\documentclass[10pt,twoside,openright,italian,english]{book}

\usepackage[twoside=true]{geometry}

\usepackage{phdthesis}

\usepackage{fancyhdr}
\usepackage{xcolor}
\usepackage{array}
\usepackage{mdwmath}
\usepackage{mdwtab}
\usepackage{amsmath,amssymb}
\usepackage[sort,nocompress]{cite}
\usepackage{graphicx}
\usepackage{glossaries}
\usepackage{listings}
\usepackage{booktabs}
\usepackage{latexsym}
\usepackage{color}
\usepackage{url}
\usepackage{bnf}
\usepackage{rotating}
\usepackage{multirow}
\usepackage{phdtitle}
\usepackage{paralist}
\usepackage{bibentry}
\usepackage[bookmarks=true,pdftex,bookmarksopen=true]{hyperref}
\usepackage{lscape}
\usepackage{longtable}
\usepackage[T1]{fontenc} 
\usepackage[latin1]{inputenc}

\usepackage[justification=centering]{subcaption}
\usepackage[normalem]{ulem}
\usepackage{epigraph}

\usepackage[toc,page,title]{appendix}

\usepackage{algpseudocode}
\usepackage[ruled,vlined,linesnumbered]{algorithm2e}
\usepackage[capitalize,noabbrev,english]{cleveref}
\usepackage{dsfont}

\usepackage{amsmath}
\usepackage{amssymb}
\usepackage{mathtools}
\usepackage{amsthm}
\usepackage{physics}
\usepackage[export]{adjustbox}
\usepackage[misc]{ifsym}
\usepackage{xspace}

\newcommand{\pisolationforest}{Preference Isolation Forest\xspace}
\newcommand{\pif}{\textsc{PIF}\xspace}
\newcommand{\spif}{Sliding-\textsc{PIF}\xspace}
\newcommand{\pspace}{Preference Space\xspace}
\newcommand{\pembedding}{Preference Embedding\xspace}
\newcommand{\pisolation}{Preference Isolation\xspace}
\newcommand{\mhash}{\textsc{MinHash}\xspace}
\newcommand{\rzhash}{\textsc{RuzHash}\xspace}

\newcommand{\vitrees}{Voronoi-\textsc{iTrees}\xspace}
\newcommand{\vitree}{Voronoi-\textsc{iTree}\xspace}
\newcommand{\rzhitree}{\rzhash-\textsc{iTree}\xspace}

\newcommand{\itree}{\textsc{iTree}\xspace}

\newcommand{\viforest}{Voronoi-\textsc{iForest}\xspace}
\newcommand{\rzhiforest}{\rzhash-\textsc{iForest}\xspace}
\newcommand{\iforest}{\textsc{iForest}\xspace}
\newcommand{\vifor}{\textsc{ViFor}\xspace}
\newcommand{\rzhifor}{\textsc{RzHiFor}\xspace}
\newcommand{\ifor}{\textsc{iFor}\xspace}
\newcommand{\eifor}{\textsc{EiFor}\xspace}
\newcommand{\lof}{\textsc{LOF}\xspace}
\newcommand{\lsh}{\textsc{LSH}\xspace}
\newcommand{\osnn}{\textsc{OSNN}\xspace}
\DeclareRobustCommand{\vect}[1]{
  \ifcat#1\relax
    \boldsymbol{#1}
  \else
    \vb*{#1}
\fi}
\newtheorem{theorem}{Theorem}
\newtheorem{corollary}{Corollary}
\newtheorem{lemma}{Lemma}
\DeclareFontShape{T1}{ptm}{m}{scit}{<->ssub*ptm/m/sc}{}

\newcommand{\onlineisolationforest}{Online Isolation Forest\xspace}
\newcommand{\onlineisolationtree}{Online Isolation Tree\xspace}
\newcommand{\oifor}{O\textsc{iFOR}\xspace}
\newcommand{\asdifor}{ASD\textsc{iFOR}\xspace}
\newcommand{\hst}{\textsc{HST}\xspace}
\newcommand{\rrcf}{\textsc{RRCF}\xspace}
\newcommand{\loda}{\textsc{LODA}\xspace}
\newcommand{\onlineiforest}{\textsc{Online}-\textsc{iForest}\xspace}
\newcommand{\onlineitree}{\textsc{Online}-\textsc{iTree}\xspace}
\newcommand{\onlineitrees}{\textsc{Online}-\textsc{iTree}s\xspace}
\newcommand{\learningprocedure}{Learning Procedure\xspace}
\newcommand{\forgettingprocedure}{Forgetting Procedure\xspace}

\newcommand{\pclustering}{Preference Clustering\xspace}
\newcommand{\mlink}{\textsc{MultiLink}\xspace}
\newcommand{\smlink}{\textsc{MLink}\xspace}
\newcommand{\gric}{\textsc{Gric}\xspace}
\newcommand{\pearl}{\textsc{Pearl}\xspace}
\newcommand{\multix}{\textsc{Multi-X}\xspace}
\newcommand{\progx}{\textsc{Prog-X}\xspace}
\newcommand{\rpa}{\textsc{RPA}\xspace}
\newcommand{\multicascadedtlinkage}{Multi-class Cascaded \tlinkage}
\newcommand{\mct}{\textsc{MCT}\xspace}
\newcommand{\tlinkage}{\textsc{T-Linkage}\xspace}
\newcommand{\stlinkage}{\textsc{TLink}\xspace}
\newcommand{\kernelfitting}{Kernel Fitting\xspace}
\newcommand{\rbf}{robust matrix factorization\xspace}
\newcommand{\biclustering}{biclustering\xspace}
\newcommand{\hoc}{higher order clustering\xspace}
\newcommand{\hp}{hypergraph partitioning\xspace}
\newcommand{\kmeans}{K-Means\xspace}
\newcommand{\matlab}{\textsc{Matlab}\xspace}

\newcommand{\Cleafy}{Cleafy\xspace}
\newcommand{\cleafy}{Cleafy\xspace}
\newcommand{\osr}{\textsc{OSR}\xspace}
\newcommand{\gboost}{\textsc{Gradient Boosting}\xspace}
\newcommand{\maxlogit}{\textsc{MaxLogit}\xspace}
\newcommand{\logit}{\textsc{Logit}\xspace}
\newcommand{\logits}{\textsc{Logits}\xspace}

\usepackage[greek,italian,main=english]{babel}

\usepackage{ulem}
\hypersetup{pdftitle={Title}, pdfauthor={Author}}
\nobibliography*

\department{Dipartimento di Elettronica Informazione e Bioingegneria}
\phdprogram{Doctoral Programme In Information Technology}

\author{Filippo Leveni}
\title{Structure-based Anomaly Detection and Clustering}
\supervisor{Giacomo Boracchi}
\cosupervisor{Luca Magri}
\titleimage{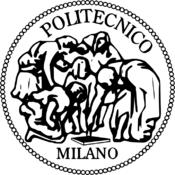}
\phdcycle{XXXV Cycle}

\begin{document}
    \pagestyle{empty}
    \maketitle
    
    \vspace*{\fill}
\emph{This thesis presents research produced during a Ph.D. funded by \cleafy.}

    \cleardoublepage
    \setlength\epigraphwidth{6.275cm}
\setlength\epigraphrule{0pt}
\epigraph{{\large \emph{``Not all those who wander are lost''\footnotemark}}}{}
\footnotetext{Gandalf, \emph{The Lord of the Rings: The Fellowship of the Ring} by J.R.R. Tolkien}

    \cleardoublepage
    \pagestyle{fancy}
    \setcounter{page}{1}
    \pagenumbering{Roman}
    \chapter*{Abstract}
    \label{sec:abstract}

    Anomaly detection is a challenging problem encountered across various application domains, including healthcare, manufacturing, and cybersecurity. Numerous statistical methods have been proposed in the literature, each relying on specific assumptions about the data being analyzed. Many of these approaches are unsupervised, where anomalies are identified as samples that deviate from the expected patterns based on the underlying assumptions.

    This thesis presents new unsupervised solutions for the anomaly detection problem in two different settings. In the first part of the thesis, we address anomaly detection from the general perspective of identifying samples that do not conform to structured patterns. In this case, we focus on the scenario where genuine data can be described by low-dimensional manifolds, while anomalous data cannot and are therefore unstructured.    
    Our main contribution to the research on structure-based anomaly detection is \pisolationforest (\pif), a novel anomaly detection framework that combines the advantages of adaptive isolation-based methods with the flexibility of \pembedding. The main intuition is to embed the data into a high-dimensional \pspace, by fitting a collection of low-dimensional manifolds, and to identify anomalies as the most isolated points within this space.
    We propose two different approaches to identify isolated points: \viforest, which leverages on suitable distances between preferences to build an ensemble of isolation trees, and \rzhiforest, which exploits a Locality Sensitive Hashing (\lsh) scheme to avoid the explicit computation of distances, hence reducing the computational complexity of the framework. Furthermore, we propose a sliding window-based extension of \pif, namely \spif, which leverages a locality prior to address situations where the global nature of the low-dimensional manifolds is unknown, enabling anomaly detection when only local information is available.
    Experiments on both synthetic and real-world datasets demonstrate that \pif successfully discriminates between structured and unstructured data, and favorably compares with state-of-the-art anomaly detection techniques.
    
    We extend our work to a scenario closely related to structure-based anomaly detection, namely structure-based clustering. In structure-based clustering, the focus is on recovering individual genuine structures rather than identifying anomalies, which are usually detected as a byproduct of the structures recovery process.
    In the literature, structure recovery is typically addressed for the single-family scenario, where individual genuine structures can be described by a specific model family, whereas the multi-family structure recovery scenario has been much less investigated.
    We propose \mlink, a novel structure-based clustering algorithm that simultaneously deals with multiple families of models in datasets contaminated by noise and outliers. In particular, \mlink considers geometric structures defined by a mixture of underlying parametric models, and tackle the structure recovery problem by means of preference analysis and clustering. \mlink combines on-the-fly model fitting and model selection in a novel linkage scheme that determines whether two clusters are to be merged. The resulting method features many practical advantages over traditional preference based methods, being faster, less sensitive to the inlier threshold, and able to compensate limitations deriving from initial models sampling.
    Experiments on several public datasets show that \mlink performs competitively with state-of-the-art alternatives, both in single-family and multi-family problems.

    In the second part of the thesis, we focus on the more traditional setting of detecting anomalies as samples that lie in low-density regions, namely density-based anomaly detection.
    We start by addressing the challenging scenario where data comes as an endless stream that may exhibit dynamic behavior, and anomalies must therefore be detected in an online way.
    Anomaly detection literature is abundant with offline methods, which require repeated access to data in memory, and impose impractical assumptions when applied to a streaming context. Existing online anomaly detection methods generally fail to address these constraints, resorting to periodic retraining to adapt to the online context. We propose \onlineiforest, a novel method explicitly designed for streaming conditions that seamlessly tracks the data generating process as it evolves over time. \onlineiforest models the data distribution using an ensemble of multi-resolution histograms that track point counts within their bins, incorporating a dynamic mechanism to adjust histograms resolution and incrementally adapt to the data generating process.
    Experimental validation on real-world datasets demonstrate that \onlineiforest is on par with online alternatives and closely rivals state-of-the-art offline anomaly detection techniques that undergo periodic retraining. Notably, \onlineiforest consistently outperforms all competitors in terms of efficiency, making it a promising solution for applications where fast identification of anomalies is of primary importance such as cybersecurity, fraud and fault detection.
    
    Finally, we address a relevant anomaly detection problem in the industrial scenario, specifically the cybersecurity challenge of identifying new malware families. Classifying a malware into its respective family is essential for building effective defence against cyber threats, enabling cybersecurity organizations to detect, respond to, and mitigate malicious activities more efficiently. However, the steady emergence of new malware families makes it difficult to acquire a comprehensive training set that encompasses all classes needed for training machine learning models. Therefore, a robust malware classification system should accurately categorize known classes while also being able to detect new ones, an anomaly detection challenge referred to in the literature as open-set recognition.
    We propose, for the first time, to combine a tree-based \gboost classifier, which is effective in classifying high-dimensional data extracted from Android manifest file permissions, to an open-set recognition technique developed within the computer vision community, namely \maxlogit.
    Our approach can be seamlessly applied to a classification pipeline based on boosted decision trees, without even affecting the classification workflow. Experiments on public and private real-world datasets demonstrate the potential of our method, which has been deployed in \cleafy's business environment and is currently part of their engine.

    \tableofcontents
    
    \cleardoublepage
    \setcounter{page}{1}
    \pagenumbering{arabic}
    \chapter{Introduction}
    \label{cha:introduction}

    \emph{Anomaly detection} is the problem of identifying data that do not conform to an expected behavior~\cite{ChandolaBanerjeeAl09}. Anomalies, by definition, are rare events, making it particularly challenging to collect a sufficient variety of them to cover all potential cases. As a result, anomaly detection is usually addressed in a \emph{semi-supervised} way, where only genuine data are known and anomalies are identified in contrast to them, or in an \emph{unsupervised} way, where anomalies are detected based on some prior assumptions.
    The significance of anomaly detection lies in the fact that anomalies in data often represent crucial, actionable insights across a range of application domains.
    A closely related concept is that of outlier, and while both terms refer to observations that deviate from the expected structure of the data, their distinction lies in perspective: outliers are typically viewed as data that hinder model performance and are usually discarded, whereas anomalies carry meaningful information and are typically analyzed and interpreted.
    In computer security, for example, anomaly detection is used in tasks such as intrusion detection, where the goal is to identify security breaches and unauthorized access in computer networks~\cite{LazarevicErtoz03}, and fraud prevention, which involves detecting fraudulent activities such as unauthorized transactions, identity theft, and account takeovers~\cite{AhmedMahmoodAl16,DalPozzoloBoracchi18}.
    In health monitoring, anomaly detection plays a crucial role in early diagnosis and intervention for conditions like arrhythmias, respiratory issues, or other medical problems. It is often implemented in wearable devices that continuously track a person's health data in real time~\cite{BanaeeAhmedAl13,UkilBandyoapdhyayAl16}.
    In industrial environments, anomaly detection is used to perform predictive maintenance~\cite{Miljkovic11,DeBenedettiLeonardiAl18,ShiWangAl18}, preventing costly breakdowns and downtime, and quality control~\cite{StojanovicDinicAl16}.

    \begin{figure}
        \centering
        \begin{subfigure}[t]{.45\linewidth}
            \centering
            \includegraphics[height=.8\linewidth]{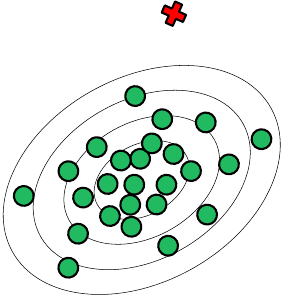}
            \caption{Density.}
            \label{subfig:density}
        \end{subfigure}
        \hfill
        \begin{subfigure}[t]{.45\linewidth}
            \centering
            \includegraphics[height=.8\linewidth]{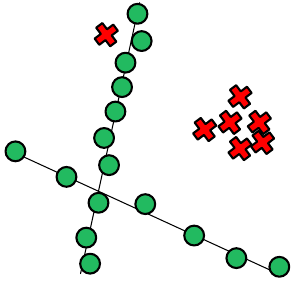}
            \caption{Structure.}
            \label{subfig:pattern}
        \end{subfigure}
        \caption{(a) An anomaly (marked as \textcolor{red}{$\times$}) is recognized as a point in a low density area. (b) Anomalies are defined with respect to their deviation from low-dimensional structures of given family $\mathcal{F}$, lines in this case.}
        \label{fig:toy_example}
    \end{figure}

    In many anomaly detection tasks, anomalies can be defined as samples deviating from structured genuine patterns that typically consist in low dimensional manifolds. We are the first to have explicitly formalized and addressed this scenario, which we have termed \emph{structure-based anomaly detection}. Structure-based anomaly detection addresses those situations where a prior regarding the parametric model family $\mathcal{F}$ describing genuine data can be leveraged to detect anomalies. Specifically, anomalies are characterized in terms of not-conformity to the low dimensional structures of family $\mathcal{F}$ describing genuine data.
    In~\cref{subfig:pattern}, we illustrate with a toy example the problem of identifying anomalous data with respect to a model family $\mathcal{F}$ describing genuine data, and this contrasts to the most traditional approach of identifying anomalies as those samples falling in low-density regions (\cref{subfig:density}). Although overly simplified, \cref{subfig:pattern} illustrates a primary task that has to be successfully addressed in several pattern recognition applications, where different model families $\mathcal{F}$ are employed, rather than lines, to identify structures or regularities in data.
    For example, images of the face of the same subject lie on a low-dimensional subspace~\cite{BasriJacobs03}, while faces of different subjects are far away from that subspace, and a density analysis would fall short in characterizing anomalies. Similarly, images of textured objects~\cite{BergmannFauserAl19,BergmannBatznerAl21} can be accurately described by few subspaces such as dictionaries~\cite{CarreraBoracchi16,CarreraManganini17}, whereas anomalous irregularities in the texture cannot. The same principle applies to 3D scans~\cite{BergmannXinAl22}, where smooth objects can be locally described by a collection of planes, while anomalous defects deviate from these structures.
    
    \medskip
    In the first part of the thesis, we address structure-based anomaly detection by proposing \emph{\pisolationforest} (\pif)~\cite{LeveniMagriAl21}, an original and general tree-based anomaly detection framework designed to identify anomalies with respect to genuine structures described by unknown instances of a parametric model family $\mathcal{F}$. \pif is composed of two main components: (i) \emph{\pembedding}, which maps data to a high-dimensional space known as \emph{\pspace}~\cite{ToldoFusiello08,MagriFusiello14} via a set of randomly fitted models from family $\mathcal{F}$, and (ii) \emph{\pisolation}, which detects anomalies as the most \emph{isolated} points in the \pspace via a forest of extremely randomized trees~\cite{GeurtsErnstAl06}. To effectively perform \pisolation in the \pspace, we proposed a novel and general tree structure we call \vitree, which partitions the space through nested Voronoi tessellations and can be employed in any metric space. Experiments demonstrate that leveraging information about model family $\mathcal{F}$ enhances anomaly detection effectiveness, and that \viforest outperforms state-of-the-art anomaly detectors when evaluated under similar conditions. We further improved \pif framework by developing a more efficient version of \viforest, specifically designed to operate in the \pspace. This new variant, we named \rzhiforest~\cite{LeveniMagriAl23}, partitions the space via \rzhash, an original Locality Sensitive Hashing (LSH)~\cite{GionisIndyk99} scheme we specifically designed for the \pspace, achieving a speed-up factor of $\times 35\%$ to $\times 70\%$ over \viforest. Finally, we extended \pif framework to scenarios where only local information about the genuine structures is available. This extension, called \spif, incorporates a sliding window mechanism to promote locality in the \pembedding step when the family $\mathcal{F}$ only approximates the genuine structures locally. Experiments show that \spif effectively leverages the local smoothness information to detect defects in real-world 3D scans, by identifying sharp irregularities in their structure.
    
    We also tackle \emph{structure-based clustering}, a problem closely related to structure-based anomaly detection. Unlike structure-based anomaly detection, which focuses solely on identifying outliers, structure-based clustering also aims at recovering genuine structures, each described by an underlying parametric model from family $\mathcal{F}$. In the literature, structure-based clustering is commonly referred to as multi-model fitting and serves as a core step in many computer vision and pattern recognition applications, including motion segmentation~\cite{WongChinAl11}, template matching~\cite{Lowe04}, primitive fitting in point clouds~\cite{HaneZachAl12}, and multi-body structure-from-motion~\cite{FitzgibbonZisserman00,SchindlerSuterAl08,OzdenSchindlerAl10}. Typically, structure-based clustering is addressed for the single-family scenario, where all genuine structures in the data are assumed to belong to the same parametric model family $\mathcal{F}$.

\begin{figure}
        \centering
        \begin{subfigure}[t]{.49\linewidth}
            \centering
            \includegraphics[width=\linewidth]{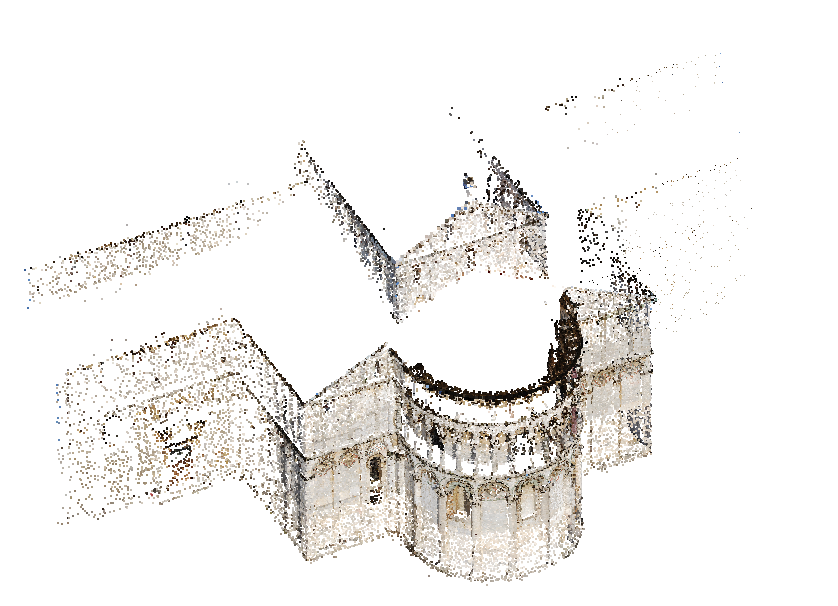}
            \caption{Input point cloud.}
            \label{subfig:point_cloud}
        \end{subfigure}
        \hfill
        \begin{subfigure}[t]{.49\linewidth}
            \centering
            \includegraphics[width=\linewidth]{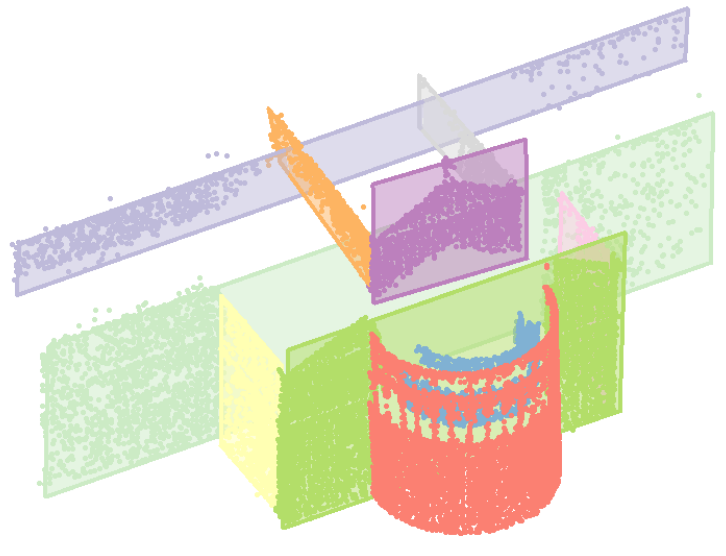}
            \caption{Recovered structures.}
            \label{subfig:structures}
        \end{subfigure}
        \caption{An example of structure-based clustering, where the goal is to recover genuine structures of multiple families $\mathcal{F}_1 \cup \dots \cup \mathcal{F}_k$, planes and cylinders in this case, from the input point cloud.}
        \label{fig:structure_based_clustering_example}
    \end{figure}

    We propose \mlink~\cite{MagriLeveniAl21}, the first preference-based clustering algorithm that is both robust to outliers and capable to deal with multiple model families $\mathcal{F}_1 \cup \dots \cup \mathcal{F}_k$ at the same time.
    In~\cref{fig:structure_based_clustering_example}, we show a real-world application of \mlink to solve a structure-based clustering problem involving multiple model families, with the objective of achieving more compressed and, therefore, more efficient information storage.
    \mlink is composed of two main components: (i) \emph{\pembedding}, and (ii) \emph{\pclustering}. The latter consists in an agglomerative single-linkage scheme in the \pspace, featuring (a) on-the-fly model sampling to reduce dependence on the initial models fitted during \pembedding, and (b) a model selection criterion to determine when structures can be effectively merged and, if so, which model family $\mathcal{F}_i$ is the most suitable.
    Experiments on both synthetic and real public datasets for structure recovery demonstrate that \mlink, thanks to the innovative combination of preference information and model selection directly into the agglomerative clustering scheme, outperforms state-of-the-art alternatives not only in multi-family problems but also in single-family scenarios.

\begin{figure}
        \centering
        \includegraphics[width=\linewidth]{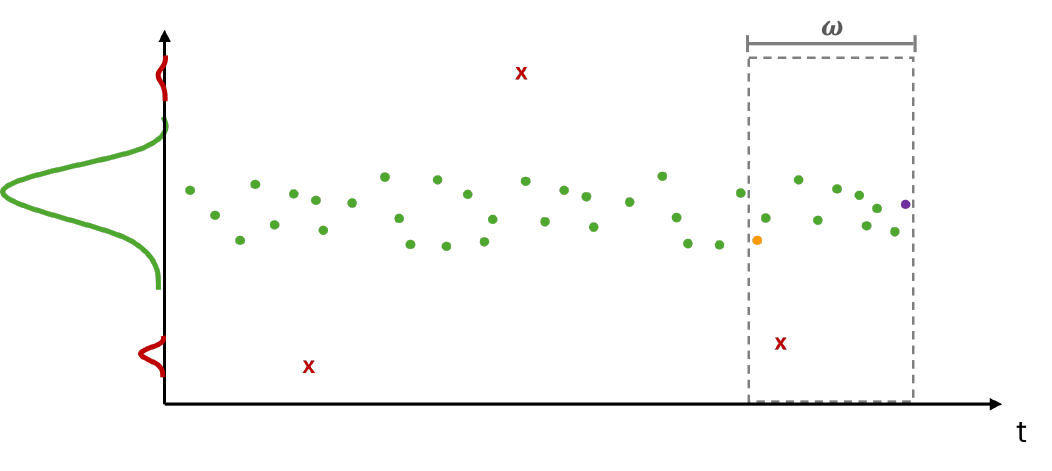}
        \caption{The goal of online anomaly detection is to identify whether a new point $\textcolor{violet}{\vect{x}_t}$ comes from the genuine $\textcolor{green}{\Phi_0}$ or anomalous $\textcolor{red}{\Phi_1}$ distribution. A common challenge in the online scenario is the strict memory limitation, which only allows to store information from a limited time horizon $\textcolor{orange}{\vect{x}_{t-\omega}}, \dots, \textcolor{violet}{\vect{x}_t}$ for each time instant.}
        \label{fig:streaming_context}
    \end{figure}

    In the second part of the thesis, we consider a more traditional setting of \emph{density-based anomaly detection}, where anomalies are identified as samples that lie in low-density regions. Our focus is on addressing the problem from an \emph{online anomaly detection} perspective, where data comes in a continuous stream and may exhibit dynamic behavior.
    In~\cref{fig:streaming_context}, we depicted the typical scenario in which online anomaly detection algorithms operate. As data are received continuously over time, the algorithm must determine, for each new instance, whether it originates from a genuine distribution or an anomalous one, both of which are usually unknown. Typical constraints in the online scenario include memory limitations, which restrict the storage of only a finite small subset of the entire data stream at each time instant, as well as the need to minimize the time interval between samples acquisition and their classification.
    
    With the vast amount of data being produced daily, online analysis has become increasingly necessary and beneficial for a wide range of applications.
    In smart city traffic management~\cite{BawanehSimon19,DjenouriBelhadiAl19}, online anomaly detection plays a key role in detecting accidents, congestion or road closures, allowing for dynamic traffic control, rerouting vehicles, and minimizing delays.
    In environmental monitoring~\cite{MiauHung20,RolloBachechiAl23}, these algorithms analyse data from IoT sensors, such as air quality or water levels, to detect hazardous conditions like pollution spikes or flood risks, enabling timely responses to protect public health and safety.
    In the finance and trading sectors~\cite{GolmohammadiZaiane15,AhmedChoudhuryAl17}, online anomaly detection is used to monitor patterns and detect irregularities that might indicate market manipulation or significant economic events.
    
    \begin{figure}
        \centering
        \includegraphics[width=.75\linewidth]{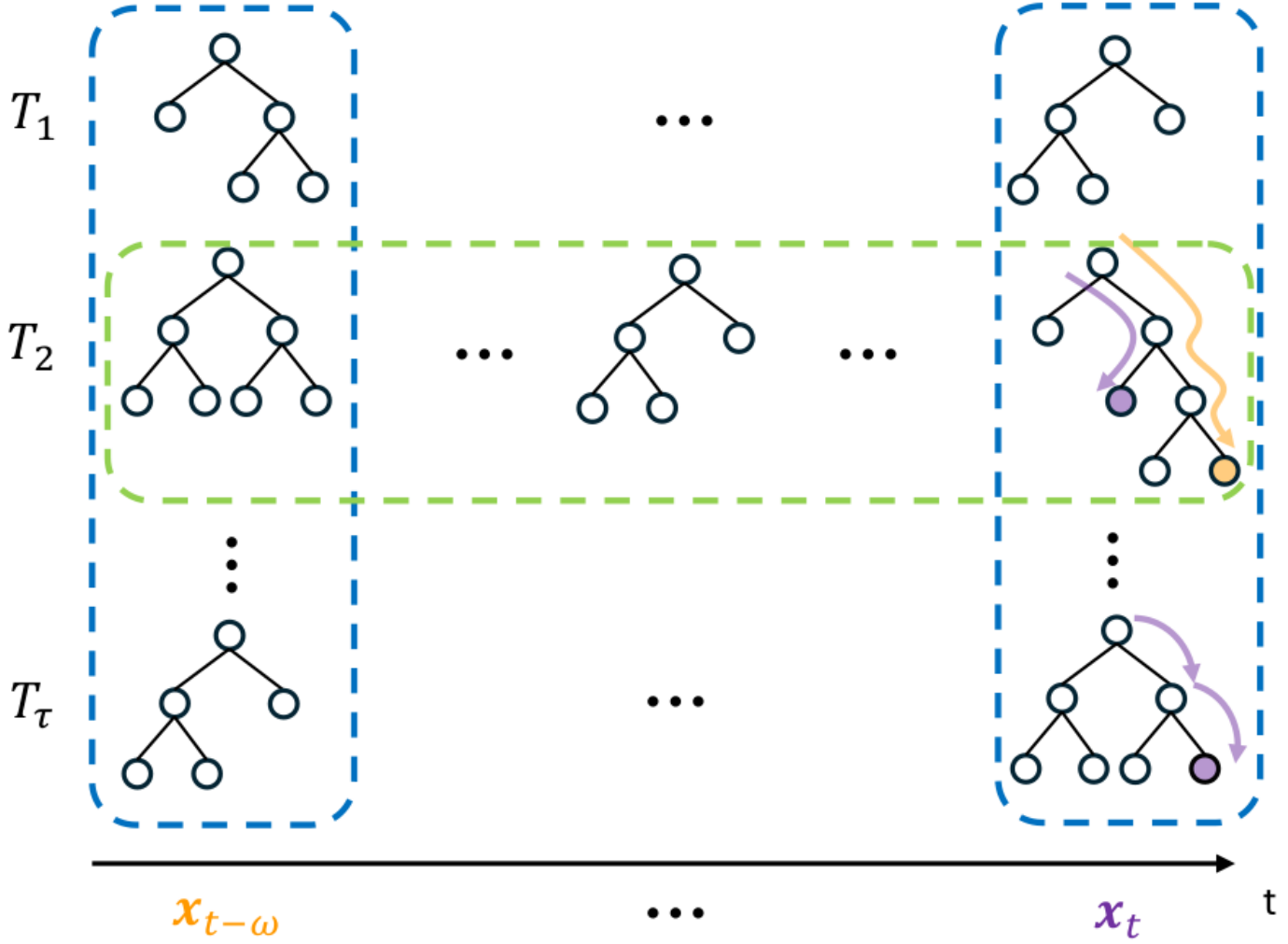}
        \caption{\onlineiforest is an ensemble of trees that continuously expand and contract their structure at each time instant $t$, by learning the new sample $\textcolor{violet}{\vect{x}_t}$, and forgetting the old sample $\textcolor{orange}{\vect{x}_{t-\omega}}$.}
        \label{fig:online_isolation_forest}
    \end{figure}
    
    The anomaly detection literature is abundant with offline methods, which require repeated access to data in memory, and impose impractical assumptions when applied to a streaming context. Many existing online methods also generally fail to address these constraints, resorting to periodic retraining to adapt to the online context.
    We address online anomaly detection by proposing \emph{\onlineisolationforest} (\onlineiforest)~\cite{LeveniCassalesAl24}, a tree-based method specifically designed for the streaming scenario, capable of effectively handle the incremental nature of data streams. \onlineiforest models the data distribution using an ensemble of multi-resolution histograms, sketched as trees in~\cref{fig:online_isolation_forest}, that track point counts within their bins.
    The fundamental component of \onlineiforest is a dynamic mechanism that enables incremental adaptation to the data distribution via two key steps: (i) the \emph{\learningprocedure}, which increases histograms resolution in the most populated regions of the space by splitting bins when they reach a certain height, and (ii) the \emph{\forgettingprocedure}, which decreases histograms resolution in the least populated areas by aggregating bins when their height drops below the threshold.
    Experiments show that \onlineiforest is characterized by an exceptionally fast operational speed, meeting the high-speed requirements of streaming contexts, while also achieving effectiveness comparable to state-of-the-art online anomaly detection methods.
    
    We also address anomaly detection from the closely related concept of \emph{Open-set recognition}~\cite{ScheirerRezende12,ScheirerJainAl14}. Both open-set recognition and anomaly detection involve identifying instances that deviate from expected patterns or originate from unknown distributions. The key difference between the two tasks is that anomaly detection can be approached either in an unsupervised or semi-supervised manner, while open-set recognition is solely tackled in a semi-supervised way. In open-set recognition, namely the classification problem in which anomalous samples (\emph{i.e.}, not belonging to any known class) must be identified, the training phase involves modeling the known data distributions, while the prediction phase aims to identify instances that come from unknown distributions, which were not seen during training.
    The close connection between open-set recognition and anomaly detection can be acknowledged by the significant overlap in their application areas. For instance, analogously to anomaly detection methods, open-set recognition techniques are employed in machine fault diagnosis~\cite{XuKovatschAl21,FrittoliCarreraAl22}, in disease detection~\cite{LiuXuAl23,XuWangAl24}, and in intrusion detection~\cite{CruzColemanAl17,HenrydossCruzAl17}.
    We leveraged on open-set recognition techniques to tackle the relevant cybersecurity problem of detecting new malware families~\cite{RuddRozsa16}. As malware developers constantly create new variants, traditional classification methods struggle to maintain optimal performance, allowing these threats to evade detection by security systems.

    We address the problem of malware family discovery by combining the tree-based \emph{Gradient Boosting} classifier~\cite{Friedman01,HastieTibshiraniAl09} with \emph{MaxLogit}~\cite{VazeHanAl21}, an open-set recognition technique widely used in computer vision. Most open-set recognition methods are designed for neural network-based classifiers and, to the best of our knowledge, they have not been extended to tree-based classifiers, which in contrast offer a higher degree of interpretability and are better able to deal with non numerical attributes. We show that MaxLogit can be seamlessly applied to a classification pipeline based on boosted decision trees without even affecting the classification workflow, and we experimentally validate its effectiveness in the cybersecurity scenario. Specifically, we validated our approach on both public and private real-world datasets comprising of malicious applications, showing that our method is promising and worthy of further investigation. Additionally, we successfully deployed our approach within \cleafy's business environment, and it is currently part of their engine.
    
    \newpage
    The vast majority of the material presented in this thesis appears in the following publications:
    \begin{itemize}
        \item \bibentry{LeveniMagriAl21}.
        \item \bibentry{MagriLeveniAl21}.
        \item \bibentry{LeveniMagriAl23}. \textbf{Best Paper Award}.
        \item \bibentry{LeveniCassalesAl24}.
    \end{itemize}
    The remaining content of this thesis is based on the following papers, which have been submitted for publication and are currently under review:
    \begin{itemize}
        \item \bibentry{LeveniMagriAl25}.
        \item \bibentry{LeveniMisturaAl25}.
    \end{itemize}
    The following publications do not fit the topic of the thesis, therefore they are not discussed here:
    \begin{itemize}
        \item \bibentry{NotarianniLeveniAl24}.
        \item \bibentry{GioriaLeveniAl25}.
        \item \bibentry{FerrarisLeveniAl25}.
    \end{itemize}

    \section{Structure of the Thesis}
        \label{sec:structure_thesis}

        The structure of this thesis is organized as follows.
        The first part focuses on the challenges of structure-based anomaly detection and clustering. In~\cref{cha:structure_based_anomaly_detection}, we begin by formalizing the problem of structure-based anomaly detection and reviewing the relevant literature, followed by an introduction to the \pif framework and its hashing-based and sliding window-based extensions, along with experimental results highlighting its benefits. In~\cref{cha:structure_based_clustering}, we formalize structure-based clustering, review the associated literature, and present \mlink, demonstrating its practical effectiveness through experimental validation.

        In the second part of the thesis, we focus on the more traditional problem of density-based anomaly detection. In~\cref{cha:online_anomaly_detection}, we begin by formalizing the online anomaly detection problem and reviewing the relevant literature. We then introduce \onlineiforest, provide a theoretical analysis of its complexity, and experimentally demonstrate its advantages over state-of-the-art methods. In~\cref{cha:open_set_recognition_malware_family_discovery}, we provide background on malware classification and malware family discovery, formalize the malware family discovery problem in our context, and review the related literature. Next, we present our approach to combine the Gradient Boosting classifier with the MaxLogit open-set recognition technique, showing its potential through experimental analysis. Finally, in~\cref{cha:conclusions}, we conclude the thesis with final remarks and suggest potential future research directions.

    \part{Structure-based Anomaly Detection and Clustering}
        \label{par:structure_based_anomaly_detection}

        \chapter{Structure-based Anomaly Detection}
    \label{cha:structure_based_anomaly_detection}

    In this chapter, we present \emph{\pisolationforest} (\pif), a tree-based unsupervised anomaly detection framework that can effectively detect anomalies with respect to genuine structures described by unknown instances of a parametric model family.
    \pif combines \emph{\pembedding}~\cite{ToldoFusiello08,MagriFusiello14}, which maps data into a high-dimensional \emph{\pspace} where unstructured data result in the most \emph{isolated} points, and \emph{\pisolation} where anomalies are identified via isolation-based techniques that use a forest of novel types of extremely randomized trees~\cite{GeurtsErnstAl06}. We propose two different tree-based algorithms to perform \pisolation: \vitree, which partitions the space using nested Voronoi tessellations and works in any metric space, and \rzhitree, which partitions the space via \rzhash, a new Locality Sensitive Hashing (LSH)~\cite{GionisIndyk99} scheme we specifically designed for the \pspace to avoid explicit computation of distances. Furthermore, we propose \spif, an extension of \pif framework which incorporates a sliding window mechanism to leverage proximity in anomaly detection problems for spatial data.
    
    In~\cref{sec:sbad_problem_formulation}, we formalize the problem of Structure-based Anomaly Detection, and in~\cref{sec:sbad_related_literature} we review the anomaly detection literature following a widely adopted taxonomy. In~\cref{sec:preference_isolation_forest}, we present the \pif framework, which consists of \pembedding (\cref{subsec:pif_preference_embedding}) and \pisolation (\cref{subsec:preference_isolation}). Specifically, in~\cref{subsec:preference_isolation} we first summarize common distances used in \pspace and then illustrate our novel \vitree algorithm. In~\cref{subsec:dealing_with_efficiency}, we address the efficiency challenges of \vitree and present \rzhash, our novel LSH scheme for \pspace, and illustrate how we integrate it into the new \rzhitree algorithm. While in~\cref{subsec:dealing_with_locality}, we introduce the locality principle and present \spif, our sliding window-based extension of \pif, highlighting also its spatial complexity benefits. Finally, in~\cref{sec:pif_experiments} we present and discuss our experimental results.

    \section{Problem Formulation}
        \label{sec:sbad_problem_formulation}
        
        \begin{figure}
            \centering
            \begin{subfigure}[t]{.275\linewidth}
                \centering
                \hspace*{-0.7cm}
                \includegraphics[height=\linewidth]{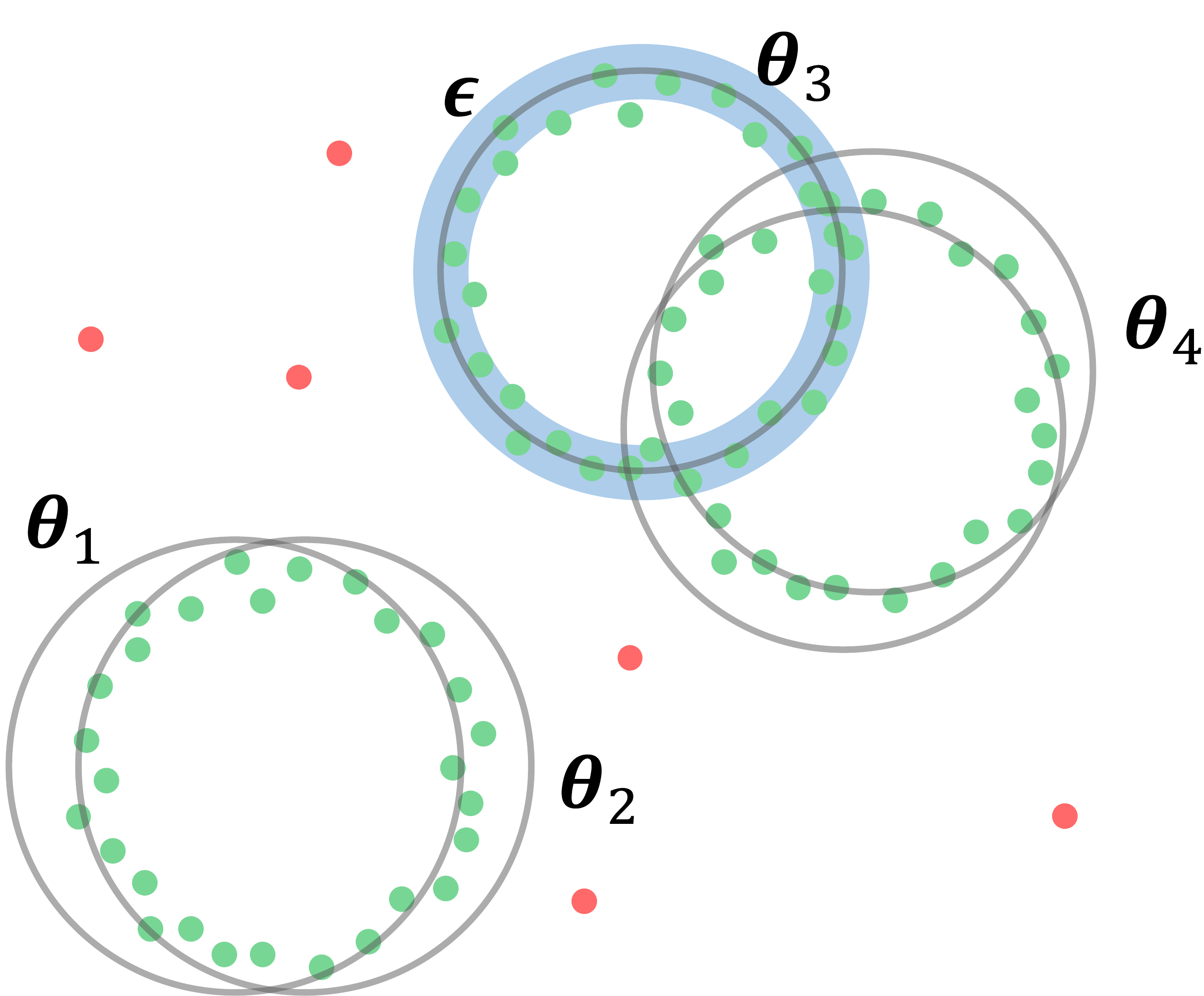}
                \caption{}
                \label{subfig:models_global_family}
            \end{subfigure}
            \hfill
            \hspace*{0.6cm}
            \begin{subfigure}[t]{.3\linewidth}
                \centering
                \hspace*{-0.6cm}
                \includegraphics[height=\linewidth]{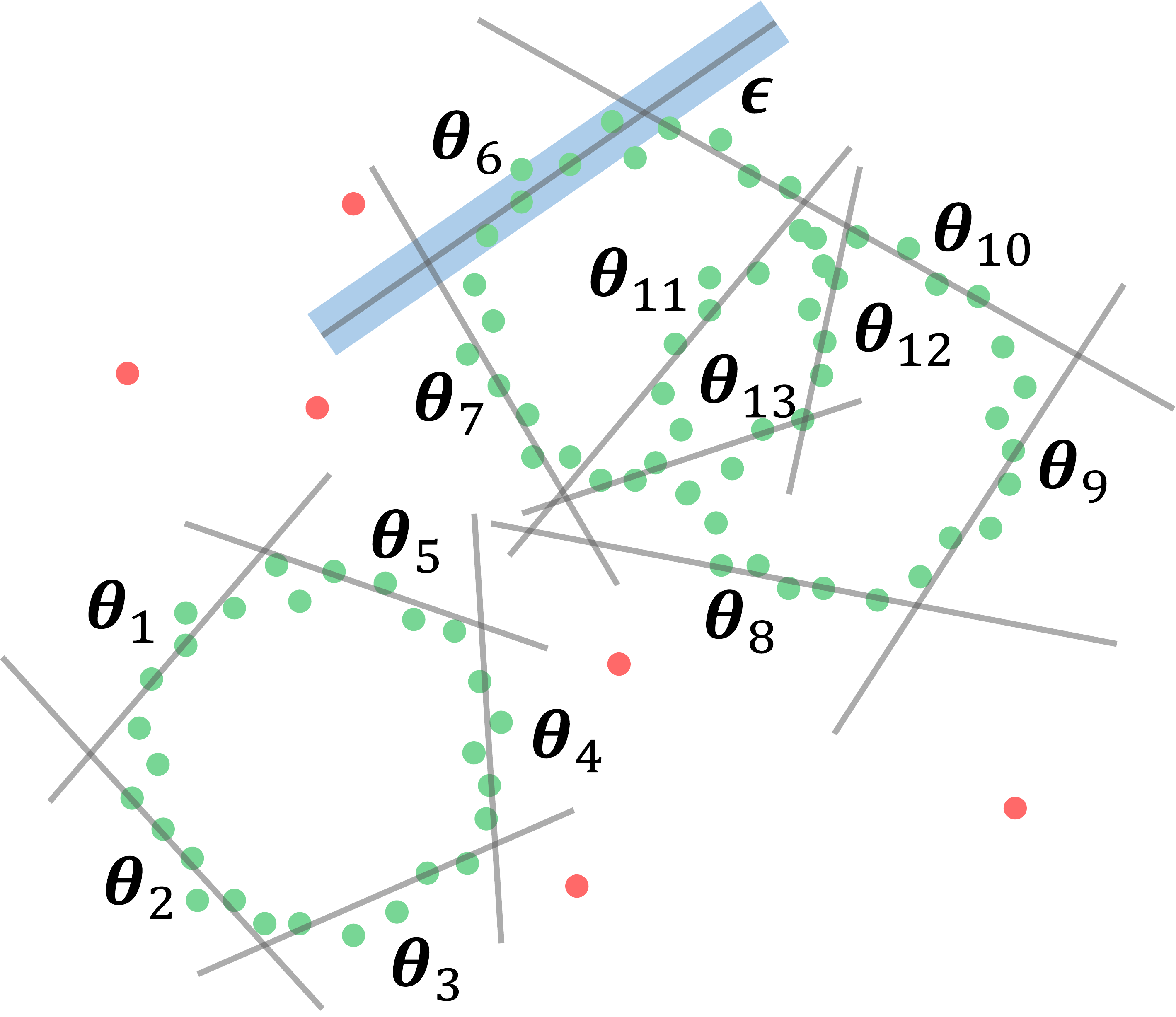}
                \caption{}
                \label{subfig:models_local_family}
            \end{subfigure}
            \hfill
            \hspace*{0.2cm}
            \begin{subfigure}[t]{.325\linewidth}
                \centering
                \includegraphics[height=\linewidth]{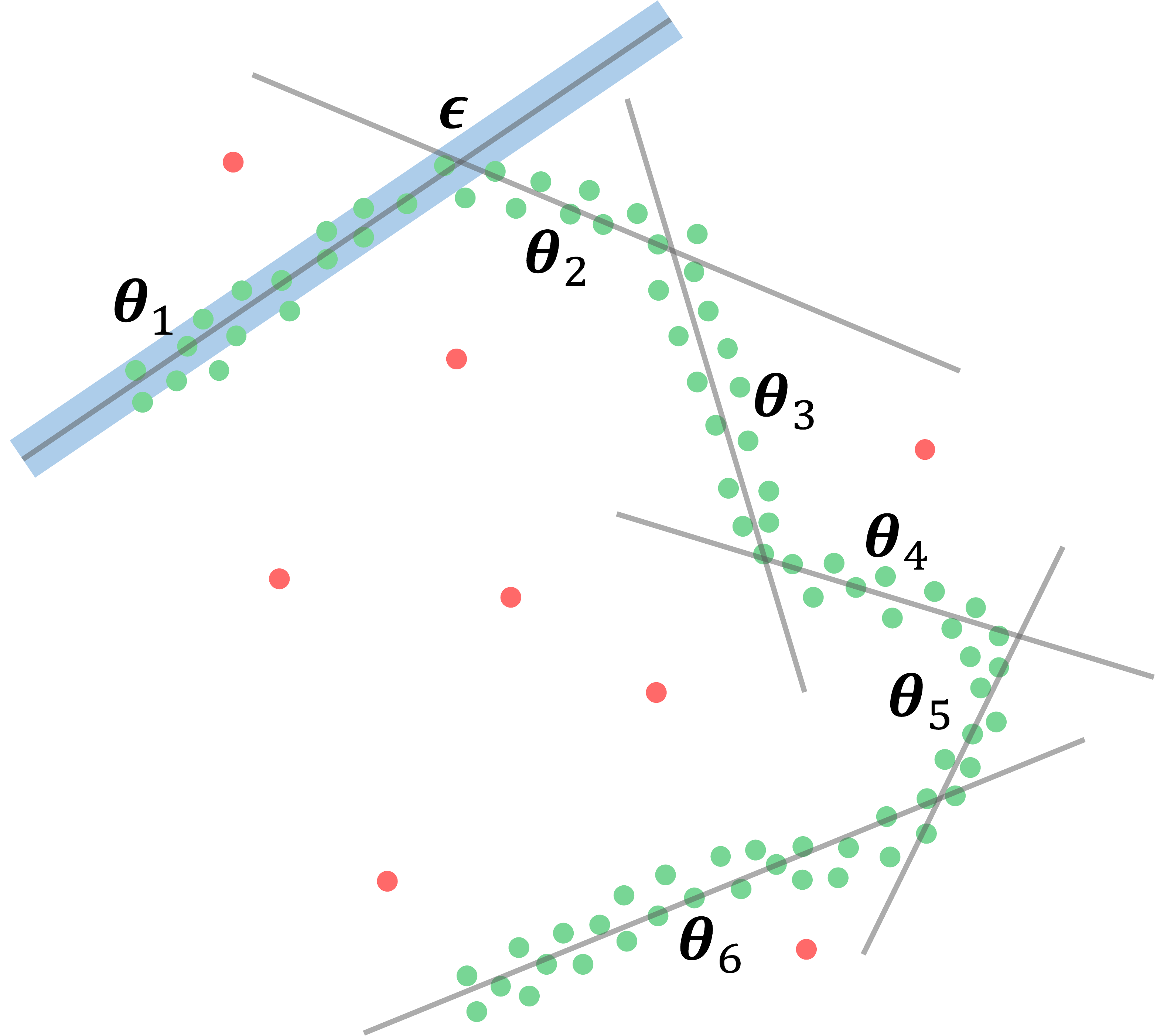}
                \caption{}
                \label{subfig:models_manifold_local_family}
            \end{subfigure}
            \caption{The same genuine data $G$ can be described by different families $\mathcal{F}$ of parametric models, circles in (a) and lines in (b). (c) Unknown complex patterns can be described by models that locally approximate the underlying manifold.}
            \label{fig:models_family}
        \end{figure}

        We consider a data set $X = G \cup A \subset \mathcal{X}$, where $G$ and $A$ are the set of genuine and anomalous data respectively, and $\mathcal{X}$ is the ambient space.
        We assume that genuine data $G$ can be described by a collection of parametric models $\{\vect{\theta}_i\}_{i = 1, \dots, k}$ whose number is unknown. In contrast, anomalous data $A$ are structure-less, and are poorly described by models $\{\vect{\theta}_i\}_{i = 1, \dots, k}$.
        Specifically, we assume that each $\vect{g} \in G$ is close to the solutions of a parametric equation $\mathcal{F}(\vect{g}, \vect{\theta}) \approx 0$, that depends on an unknown vector of parameters $\vect{\theta}$, where $\mathcal{F}$ is the model family. For example, in~\cref{subfig:models_global_family} the model family is the one of circles $\mathcal{F}(\vect{g}, \vect{\theta}) = (\theta_1 - g_1)^2 + (\theta_2 - g_2)^2 - \theta_3$ and, due to noise, genuine data satisfy the equation only up to a tolerance $\epsilon > 0$, namely $|\mathcal{F}(\vect{g}, \vect{\theta})| < \epsilon$. We assume the nature and amount of noise to be unknown. On the other hand, anomalous data $\vect{a} \in A$ poorly satisfy the parametric equation, therefore $|\mathcal{F}(\vect{a}, \vect{\theta})| \gg \epsilon$.
        
        The choice of the model family $\mathcal{F}$ is not unique, as the same data set $G$ can be described by different collections of parametric models. As an example, in~\cref{subfig:models_local_family} genuine data are described by models whose family $\mathcal{F}$ is the one of lines, thus $\mathcal{F}(\vect{g}, \vect{\theta}) = \theta_1 g_1 + \theta_2 g_2 + \theta_3$.
        The flexibility of this formulation allows to describe complex structures whose underlying manifold is unknown (\cref{subfig:models_manifold_local_family}), requiring only that the model family $\mathcal{F}$ employed is known to \textit{locally} approximate the manifold up to an appropriate tolerance $\epsilon$.
        
        The goal of Structured-based AD is to produce an anomaly scoring function $\alpha: X \rightarrow \mathbb{R}^+$ such that $\alpha(\vect{a}) \gg \alpha(\vect{g}) \ \forall \vect{a} \in A, \; \forall \vect{g} \in G$, so that it is possible to identify anomalies by setting an appropriate threshold.

    \section{Related Literature}
        \label{sec:sbad_related_literature}

        Several approaches have been proposed to identify anomalies, and for an extensive description refer to~\cite{ChandolaBanerjeeAl09, ChandolaBanerjeeAl10}. A possible taxonomy envisages four main categories: distance-based, density-based, isolation-based and model-based.
        
        In \emph{distance-based AD}~\cite{KnorrNgAl00} approaches, such as K-Nearest Neighbors (K-NN)~\cite{RamaswamyRastogiAl00}, an instance is considered anomalous when its neighborhood lacks a sufficient number of samples. Distance-based methods are independent from data distribution, and can be easily improved via data-dependent distance measures~\cite{TingZhuAl16}. However, these methods tend to be computationally intensive and to perform poorly when applied to high-dimensional data sets. Additionally, they operate in the ambient space $\mathcal{X}$, and no prior knowledge about model family $\mathcal{F}$ can be leveraged.
        We observe that distance-based methods can be framed in our formulation presented in~\cref{sec:sbad_problem_formulation}. For example, in the K-NN approach, the model family $\mathcal{F}$ consists of a set of points, while the model instance $\vect{\theta}$ is the data set itself, with the data points' coordinates as the parameters. The function $\mathcal{F}(\vect{x}, \vect{\theta})$ computes the distance between $\vect{x}$ and its $k$-th closest point within the dataset $\vect{\theta}$.
        
        In \emph{density-based AD} methods (e.g.,~\cite{BreunigKriegel00,PapadimitriouKitigawaAl03,JinTungAl06,KriegelKrogerAl09}) the key concept is that anomalous and genuine instances differ in their local density, and an important algorithm representative of this category is \lof~\cite{BreunigKriegel00}. In general, density-based detection methods carry the same computational burden of distance-based ones when dealing with large and high-dimensional data sets. Similarly to distance-based AD methods, we can frame density-based ones in our formulation. For example, in \lof the model family $\mathcal{F}$ consists of a set of points with their reachability distances, and $\mathcal{F}(\vect{x}, \vect{\theta})$ computes the local outlier factor for $\vect{x}$ with respect to $\vect{\theta}$.
        
        \emph{Isolation-based} approaches can be traced back to Isolation Forest~\cite{LiuTingAl12} (\ifor), where anomalies are identified as the most \textit{``isolated''} points, and subsequent works showed that this concept is related to the concepts of distance and density~\cite{ZhangDouAl17, LeveniMagriAl23}
        . In \ifor, they build a forest of randomized trees~\cite{GeurtsErnstAl06} 
        (\itree) that recursively partition the space, and the number of splits required to \emph{isolate} a point from the others is inversely related to its likelihood of being anomalous. The impact of \ifor can be acknowledged by the several improvements that has been subsequently proposed. Extended Isolation Forest~\cite{HaririKindAl21} and Generalized Isolation Forest~\cite{LesoupleBaudoinAl21} overcome the limitation of axis-parallel splits, while~\cite{StaermanMozharovskyiAl19} extends \ifor beyond the concept of point-anomaly, to identify functional-anomalies
        . \ifor was also extended to operate directly on pairwise distances between points~\cite{MensiTaxAl23}, or to use different criteria of aggregating paths length for the anomaly score computation~\cite{MensiBicego21}. Analogously to distance and density-based ones, isolation-based methods operate directly in the ambient space $\mathcal{X}$, and cannot leverage prior knowledge about model family $\mathcal{F}$.
        We note that even isolation-based approaches can be framed in our formulation. For example, in \ifor the model family $\mathcal{F}$ consists of an ensemble of isolation-trees, and $\mathcal{F}(\vect{x}, \vect{\theta})$ computes the normalized average depth of $\vect{x}$ in the forest $\vect{\theta}$.
        
        \emph{Model-based AD} methods first fit a model on the data, and then identify anomalies as points that do not conform to the model. Notable examples of this approach are classification-based~\cite{ScholkopfWilliamsonAl99,AbeZadroznyAl06,RuffVandermeulenAl18}, reconstruction-based~\cite{XiaCaoAl15,CarreraManganini17,ChenSatheAl17,CarreraRossi19}, and clustering-based methods~\cite{EsterKriegelAl96,JiangTsengAl01,SchubertSanderAl17}. Classification and reconstruction-based typically require a set of genuine samples to build the genuine model, making them \emph{semi-supervised} rather than \emph{unsupervised}, while clustering-based methods require to first solve the more complex problem of clustering to address AD. Model-based AD methods can be straightforwardly framed in our framework since all of them are based on an underlying model family $\mathcal{F}$. For instance, in~\cite{ScholkopfWilliamsonAl99} the family $\mathcal{F}$ consists of Support Vector Machines, while in~\cite{RuffVandermeulenAl18} it consists of Neural Networks. Autoencoders~\cite{ChenSatheAl17} and dictionaries~\cite{CarreraManganini17} are typical examples of model family $\mathcal{F}$ used in reconstruction-based methods. Again, in K-Means~\cite{JiangTsengAl01} the model family $\mathcal{F}$ consists of a set of $k$ centroids, while the function $\mathcal{F}(\vect{x}, \vect{\theta})$ computes the distance between $\vect{x}$ and the closest centroid in $\vect{\theta}$.
        
        Our proposed \emph{structure-based AD} method \pif stands out from previous approaches as it is specifically designed to straightforwardly incorporate prior knowledge about the model family $\mathcal{F}$ that describes genuine data structures. This feature makes \pif more effective than either distance-based, density-based and isolation-based methods when the model family is known. Notably, \pif operates in a truly \emph{unsupervised} manner, requiring no training data, unlike the \emph{supervised} or \emph{semi-supervised} approach typical of classification and reconstruction-based methods. Additionally, \pif directly targets AD, in contrast to clustering-based methods that address it as a byproduct of the clustering process.

    \section{\pisolationforest}
        \label{sec:preference_isolation_forest}

        \begin{figure}
            \centering
            \begin{subfigure}[t]{.325\linewidth}
                \centering
                \includegraphics[height=.8\linewidth]{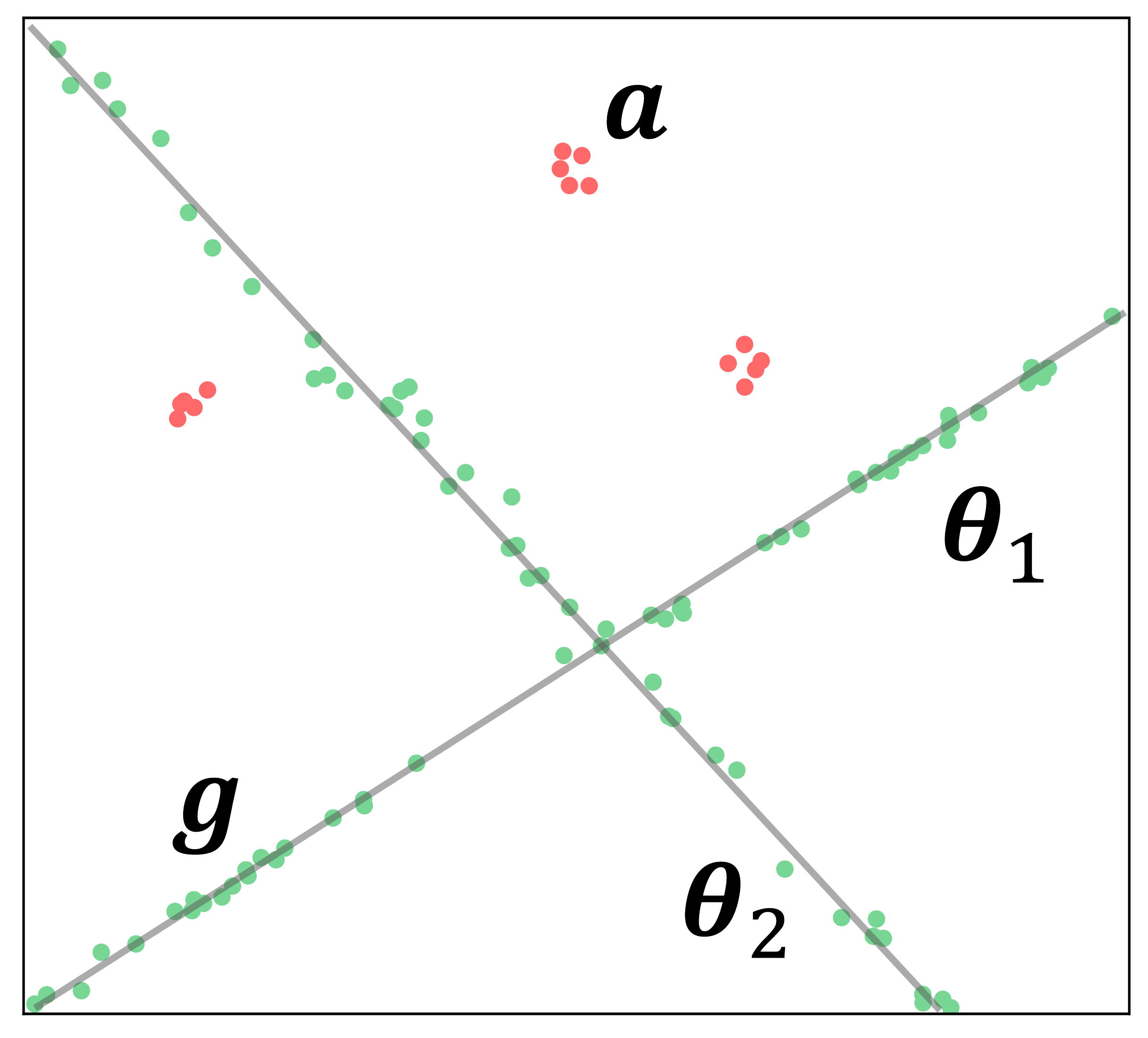}
                \caption{Input data $X = G \cup A$.}
                \label{subfig:input}
            \end{subfigure}
            \hfill
            \begin{subfigure}[t]{.325\linewidth}
                \centering
                \includegraphics[height=.8\linewidth]{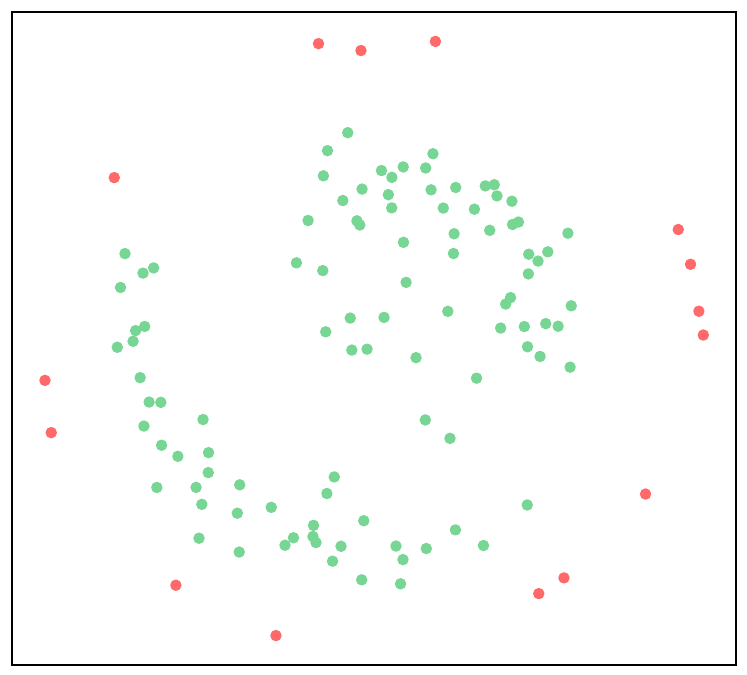}
                \caption{Preference space $\mathcal{P}$.}
                \label{subfig:embedding}
            \end{subfigure}
            \hfill
            \begin{subfigure}[t]{.325\linewidth}
                \centering
                \includegraphics[height=.8\linewidth]{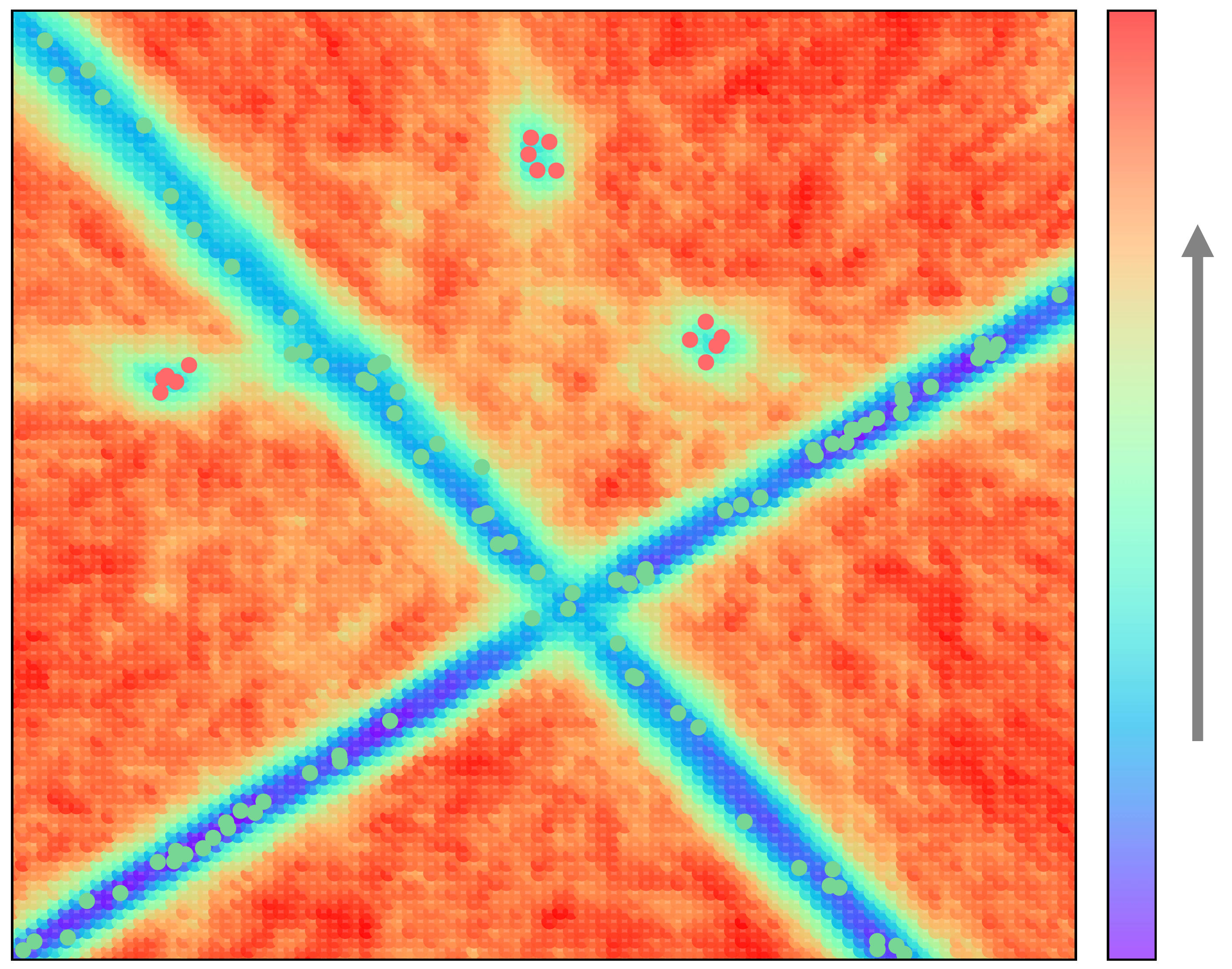}
                \caption{Anomaly scores $\alpha$.}
                \label{subfig:output}
            \end{subfigure}
            \caption{\pif detects anomalies in $X$ that do not conform to structures. (a) Genuine points $G$, in green, described by two lines with parameters $\vect{\theta}_1$ and $\vect{\theta}_2$, and anomalies $A$ in red. (b) \pembedding maps $X$ to a high-dimensional space where anomalies result in isolated points (visualized via MDS~\cite{Kruskal64a}). (c) Anomaly scores $\alpha(\cdot)$ (color coded) are computed via \pisolation.}
            \label{fig:input_output}
        \end{figure}

        The proposed Preference Isolation Forest (\pif) is composed of two main components: \pembedding and \pisolation.
        We perform \pembedding by mapping points $X$ to a high-dimensional \pspace via a set of models $\{\vect{\theta}_i\}_{i = 1, \dots, k}$ belonging to a family $\mathcal{F}$ whose parametric equation is known.
        We then carry out \pisolation to assign an anomaly score $\alpha(\vect{x})$ to each point $\vect{x}$ via \viforest, a novel isolation-based AD method we specifically designed to operate in any metric space, \pspace included.
        
        \subsection{Preference Embedding}
            \label{subsec:pif_preference_embedding}

            \begin{algorithm}[t]
                \caption{\pisolationforest (\pif) \label{alg:main}}
                \DontPrintSemicolon
                \SetNoFillComment
                \KwIn{$X$ - input data, $t$ - number of trees, $\psi$ - sub-sampling size, $b$ - branching factor}
                \KwOut{$\{\alpha_\psi(\vect{x}_j)\}_{j=1,\ldots,n}$ - anomaly scores}
                \begin{small}
                    \tcc{\pembedding\!\!\!}
                \end{small}
                Sample $m$ models $\{\vect{\theta}_i\}_{i=1,\ldots,m}$ from $X$ \label{line:begin_embedding}\\
                $P \leftarrow \{\vect{p}_j \,|\, \vect{p}_j = \mathcal{E}(\vect{x}_j)\}_{j=1,\ldots,n}$ \label{line:end_embedding}\\
                \begin{small}
                    \tcc{\viforest construction\!\!\!}
                \end{small}
                $F \leftarrow \text{\viforest}(P, t, \psi, b)$ \label{line:forest_construction}\\
                \begin{small}
                    \tcc{Anomaly scores computation}
                \end{small}
                \For{$j = 1$ \normalfont{to} $n$ \label{line:begin_detection}}
                    {$\vect{p}_j \leftarrow \text{$j$-th point in $P$}$ \label{line:point} \\
                     \For{$k = 1$ \normalfont{to} $t$}
                         {$T_k \leftarrow \text{$k$-th \vitree in $F$}$ \label{line:tree} \\
                          $d_k(\vect{p_j}) \leftarrow \text{\textsc{PathLength}}(\vect{p}_j, T_k, 0)$ \label{line:height}}
                    $\alpha_\psi(\vect{x}_j) \leftarrow 2^{-\frac{E(D_j)}{c(\psi)}}$ \label{line:anomaly_score}
                    }
                \Return $\{\alpha_\psi(\vect{x}_j)\}_{j=1,\ldots,n}$ \label{line:end_detection} \\
            \end{algorithm}

            \pembedding maps data $X \in \mathcal{X}$ to a high-dimensional \pspace $\mathcal{P}$ by leveraging prior knowledge about a suitable parametric family $\mathcal{F}$ that well describes genuine data $G$.
            
            Formally, \pembedding consists in a mapping $\mathcal{E}\colon \mathcal{X} \to \mathcal{P}$, from the ambient space $\mathcal{X}$ to the \pspace $\mathcal{P} = [0, 1]^m$.
            Such mapping is obtained by fitting a pool $\{\vect{\theta}_i\}_{i=1,\ldots,m}$ of $m$ models belonging to family $\mathcal{F}$ on the data $X$ using a RanSaC-like strategy~\cite{FischlerBolles81} (\cref{alg:main}, \cref{line:begin_embedding}): the minimal sample set -- containing the minimum number of points necessary to constrain a parametric model -- is randomly sampled from the data to fit the model parameters $\vect{\theta}_i$.
            Then (\cref{line:end_embedding}), each sample $\vect{x} \in \mathcal{X}$ is embedded to a vector $\vect{p} = \mathcal{E}(\vect{x}) \in \mathcal{P}$ whose $i$-th component is defined as:
            \begin{equation}
                \label{eq:pif_preference}
                p_i  =
                \begin{cases}
                    \phi(\delta_{i}) &\text{if $|\delta_{i}| \leq \epsilon$ }\\
                                    0 &\text{otherwise}
                \end{cases},
            \end{equation}
            where $\delta_i = \mathcal{F}(\vect{x}, \vect{\theta}_i)$, the parametric equation of model $\vect{\theta}_i$, computes the residuals of $\vect{x}$ with respect to $\vect{\theta}_i$, and $\epsilon = k\sigma$ is an inlier threshold proportional to the standard deviation $\sigma > 0$ of the noise. We assume the nature and amount of noise to be unknown, but that it can be directly estimated (e.g.,~\cite{WangSuter04}). The preference function $\phi$ is defined as:
            \begin{equation}
                \label{eq:pref}
                \phi(\delta_i) = e^{-\frac{\delta^2_i}{2 \sigma^2}}.
            \end{equation}
            We explored also the \emph{binary preference} function, that is $\phi(\delta_i) = 1$ when $|\delta_i| \leq \epsilon$ and $0$ otherwise. Hereinafter, we refer to $\mathcal{P} = [0, 1]^m$ as the \emph{continuous preference space}, unless otherwise specified.

            As illustrated in~\cref{fig:input_output}, leveraging on a model family $\mathcal{F}$ that well describes genuine data causes anomalies to became isolated points in $\mathcal{P}$, making them easily detectable in the subsequent \pisolation step of \pif.
            
            \subsubsection{Choice of the model family}
                \label{subsubsec:choice_model_family}

                The choice of the model family $\mathcal{F}$ in the \pembedding step is closely tied to the prior knowledge available about the genuine data $G$, and it plays a fundamental role in the effectiveness of the subsequent \pisolation procedure. The extent of this prior knowledge determines whether $\mathcal{F}$ provides a \emph{global} or \emph{local} description of $G$.
                
                When comprehensive prior knowledge is available, $\mathcal{F}$ can provide a \emph{global} description of $G$. As an illustrative example, in~\cref{subfig:models_global_family} we suppose to know that $G$ has been generated by a collection of circles, therefore we can explicitly choose the family $\mathcal{F}$ of circles to estimate the collection of models $\{\vect{\theta}_i\}_{i = 1, \dots, m}$. There are also scenarios where the exact model family is unknown, but we may still have some knowledge that it belongs to a broader family. In such cases, instead of assuming a specific family like circles, we can generalize to families that encompasses multiple possible shapes, such as ellipses, enabling a more flexible representation of curvilinear structures. This flexibility might be particularly beneficial when multiple model families contribute to the generation of genuine data $G$, such as either ellipses, parabolas, and hyperbolas, all of which belong to the broader family of conics, enabling the framework to handle real-world scenarios where the exact nature of the data manifold is unknown or difficult to characterize explicitly. However, general families often require larger minimal sample sets and can increase variance in the results, necessitating a careful balance between flexibility and robustness in model selection.
                
                In contrast, when only partial prior knowledge is available, we may adopt a \emph{local} description of $G$. In~\cref{subfig:models_local_family,subfig:models_manifold_local_family}, we suppose to only know that $G$ has been generated by a process whose manifold's dimensionality is one, but we lack information about the specific parametric equation governing its shape. In such cases, we can opt for any model family $\mathcal{F}$ matching the manifold dimensionality (lines in the example) to estimate $\{\vect{\theta}_i\}_{i = 1, \dots, m}$, and approximate the manifold locally. This approach provides flexibility in representing the underlying structure without requiring the exact parametric formulation. Analogously to the global case, rather than restricting $\mathcal{F}$ to lines, we might generalize to broader families of matching dimensionality such as conics or polynomials of a certain degree, and allow for more flexible local approximations.
                
        \subsection{Preference Isolation}
            \label{subsec:preference_isolation}
            
            \begin{figure}[t]
                \centering
                \hspace{-0.4cm}
                \begin{subfigure}[t]{.15\linewidth}
                    \centering
                    \hspace*{-0.6cm}
                    \includegraphics[width=2.2\linewidth, angle=90]{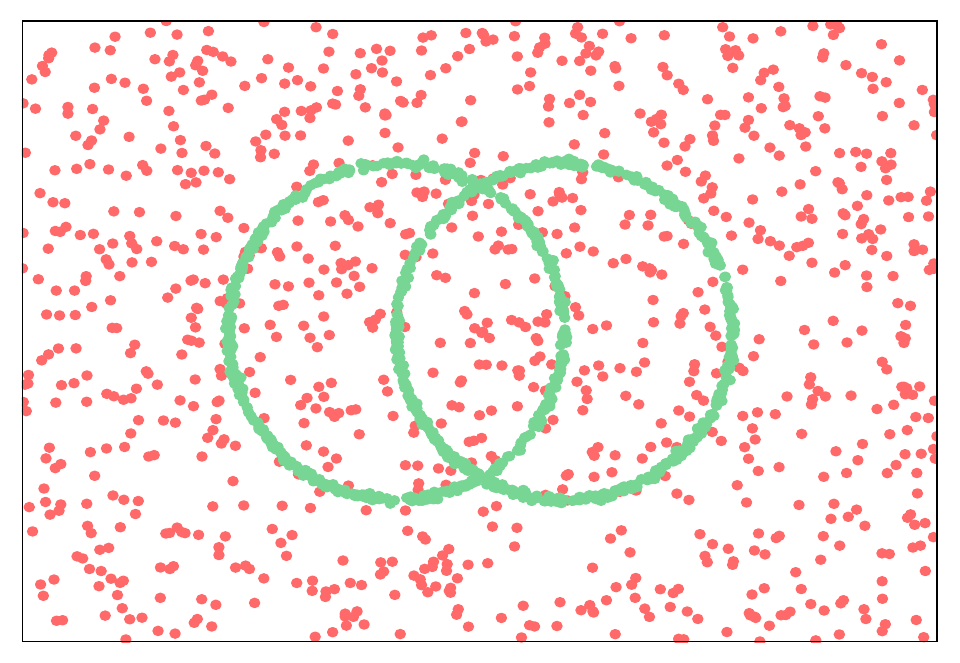}
                    \caption{Input data.}
                    \label{fig:input}
                \end{subfigure}
                \hfill
                \hspace*{0.02cm}
                \begin{subfigure}[t]{.15\linewidth}
                    \centering
                    \hspace*{-0.6cm}
                    \includegraphics[width=2.2\linewidth, angle=90]{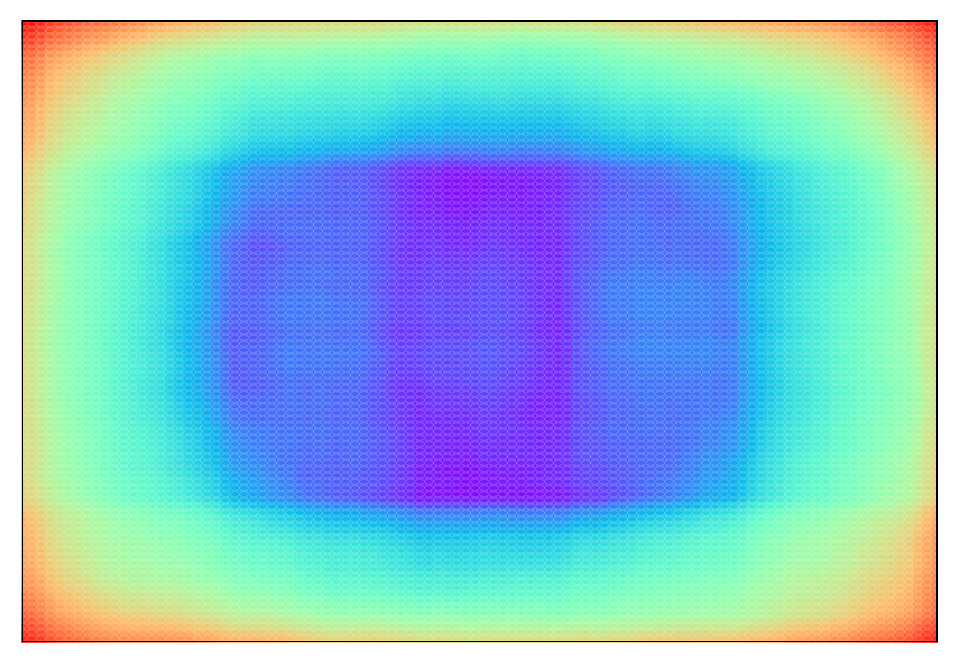}
                    \caption{\ifor~\cite{LiuTingAl12}}
                    \label{fig:ifor}
                \end{subfigure}
                \hfill
                \hspace*{-0.25cm}
                \begin{subfigure}[t]{.15\linewidth}
                    \centering
                    \hspace*{-0.35cm}
                    \includegraphics[width=2.2\linewidth, angle=90]{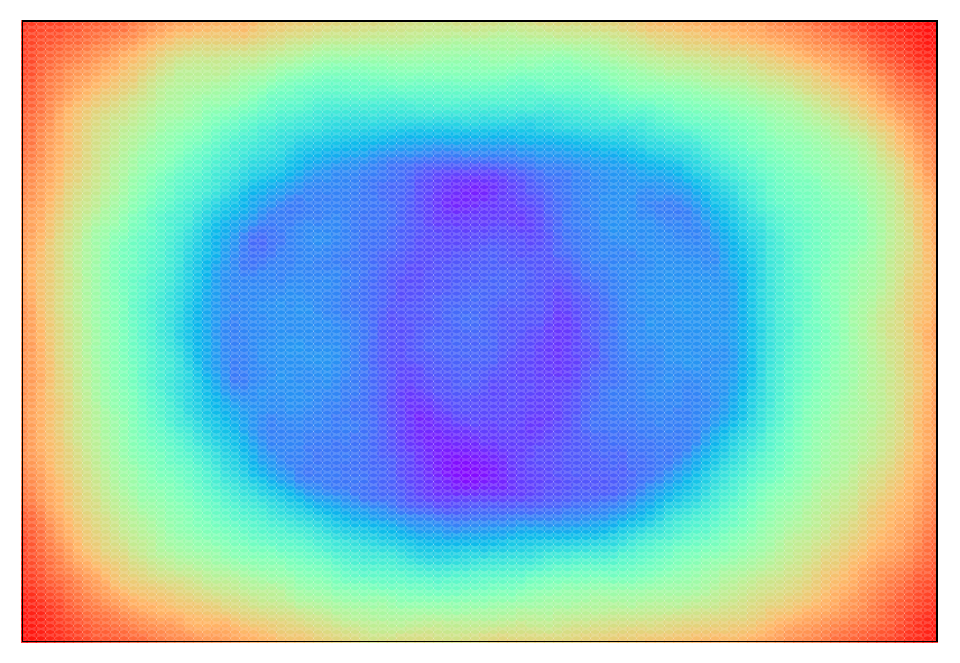}
                    \begin{minipage}{1.25\textwidth}
                        \vspace*{-0.07cm}
                        \caption{\eifor~\cite{HaririKindAl21}}
                        \label{fig:eifor}
                    \end{minipage}
                \end{subfigure}
                \hfill
                \hspace*{0.25cm}
                \begin{subfigure}[t]{.15\linewidth}
                    \centering
                    \hspace*{-0.6cm}
                    \includegraphics[width=2.2\linewidth, angle=90]{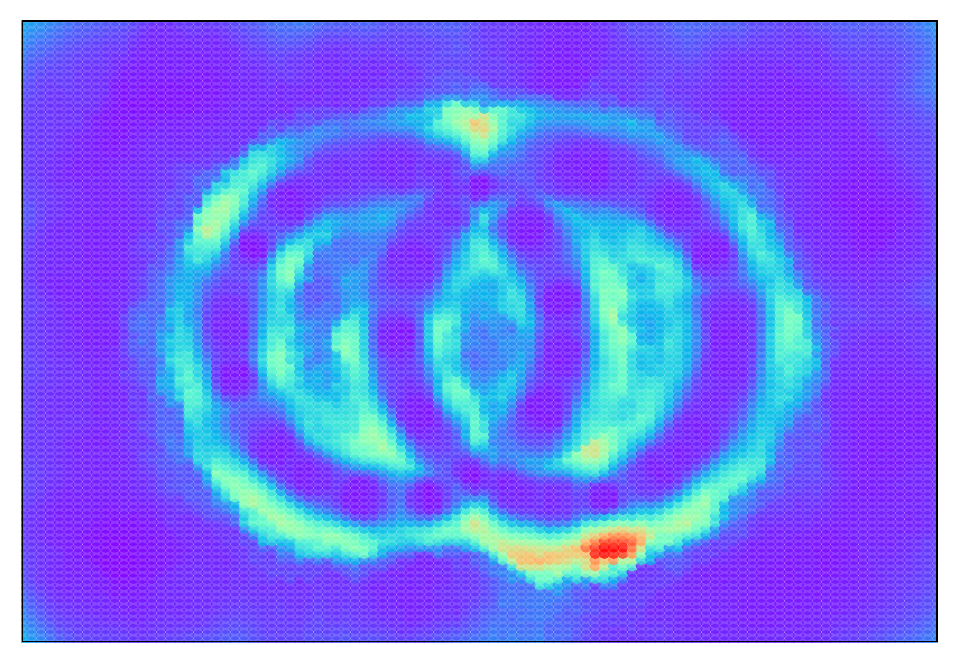}
                    \caption{\lof~\cite{BreunigKriegel00}}
                    \label{fig:lof}
                \end{subfigure}
                \hfill
                \hspace*{-0.1cm}
                \begin{subfigure}[t]{.15\linewidth}
                    \centering
                    \hspace*{-0.5cm}
                    \includegraphics[width=2.2\linewidth, angle=90]{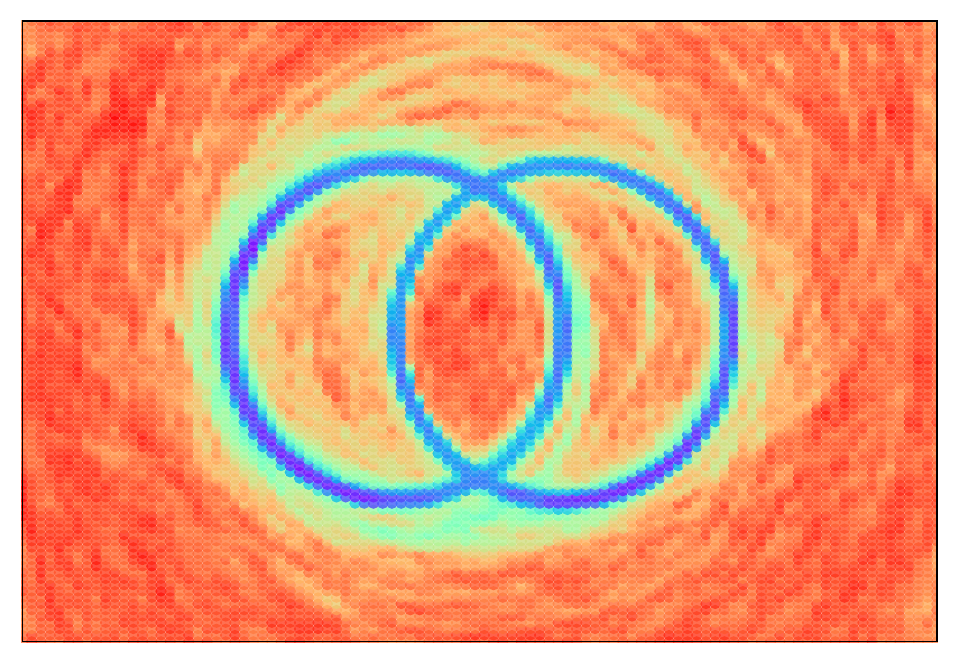}
                    \caption{\pif}
                    \label{fig:pif}
                \end{subfigure}
                \caption{Anomaly scores $\alpha(\cdot)$ computed by different algorithms. \pif effectively                
                detects anomalies
                when a suitable model family $\mathcal{F}$ (circles in this example) is chosen for \pembedding.}
                \label{fig:pif_anomaly_scores}
            \end{figure}
            
            \pisolation consists in performing isolation-based AD in the \pspace. Thanks to \pembedding, anomalous samples result in the most \emph{isolated} points in the \pspace, and our claim is that performing AD in $\mathcal{P}$ is more effective than in $\mathcal{X}$.
            As a qualitative example, in~\cref{fig:pif_anomaly_scores} we can appreciate \pif as the only approach that accurately identifies anomalies within the circles.
            Isolation-based AD is efficient, scalable, and operates without assumptions about the underlying data distribution, making it particularly suitable for high-dimensional embeddings.
            
            As we will show in~\cref{sec:pif_experiments}, state-of-the-art methods like \ifor perform better, but still worse than \pif, when directly applied in the \pspace. This is because the splitting criterion used by these techniques implicitly relies on $\ell_1/\ell_2$ norms for measuring distances. In contrast, \pif builds upon \vitrees, a novel isolation-based AD algorithm whose splitting criteria allows for any distance metric, , including the ones suited for \pspace.
            
            \subsubsection{Distances in the Preference Space}
                \label{subsubsec:distances_preference_space}

                In order to effectively perform AD in the \pspace, we identify the properties that a distance function $d(\cdot, \cdot)$ must satisfy.
                Apart from the usual properties of a distance (identity, positivity, symmetry and triangle inequality), for simplicity we require the distance to be bounded in $[0, 1]$, \emph{i.e.}, $\forall \vect{p}, \vect{q} \in \mathcal{P} \quad d(\vect{p}, \vect{q}) \leq 1$.
                Additionally, since our desiderata is to make anomalous points \emph{isolated} in $\mathcal{P}$, we require the distance to reach the maximum value when two points $\vect{p}, \vect{q}$ are orthogonal $\vect{p} \perp \vect{q}$, \emph{i.e.}, $\forall \vect{p}, \vect{q} \in \mathcal{P} \quad \lim_{\langle\vect{p}, \vect{q}\rangle \rightarrow 0} d(\vect{p}, \vect{q}) = 1$. The orthogonality property arises from the fact that anomalous data -- \emph{i.e.}, data that do not conform to the model family employed in \pembedding -- are typically poorly described by the fitted models, leading to preference vectors that are nearly orthogonal to those of genuine data.
                Several metrics satisfying the above properties have been proposed to measure distances in the \pspace~\cite{ToldoFusiello08,MagriFusiello14,LeveniMagriAl23}.
                
                \paragraph{Jaccard}
                The Jaccard~\cite{Jaccard01,Jaccard12,Kosub19}
                distance is a commonly used indicator of the dissimilarity between sets. We rely on the set interpretation of the binary \pspace $\{0, 1\}^m$~\cite{ToldoFusiello08}, and the Jaccard distance between $\vect{p}, \vect{q} \in \{0, 1\}^m$ is defined as:
                \begin{equation}
                    \label{eq:jaccard_vect}
                    d_J(\vect{p}, \vect{q}) = 1 - \frac{\sum_{i=1}^{m} (p_i \wedge q_i)}{\sum_{i=1}^{m} (p_i \vee q_i)}.
                \end{equation}

                \paragraph{Ruzicka}
                The Ruzicka~\cite{Ruzicka58} distance, also called generalized Jaccard distance~\cite{Kosub19}, extends the Jaccard distance to real-valued multisets. Analogously to the Jaccard distance, we rely on the multiset interpretation of the continuous \pspace $\mathcal{P} = [0, 1]^m$, and the Ruzicka distance between $\vect{p}, \vect{q} \in \mathcal{P}$ is defined as:
                \begin{equation}
                    \label{eq:ruzicka_vect}
                    d_R(\vect{p}, \vect{q}) = 1 - \frac{\sum_{i=1}^{m} \min\{p_i, q_i\}}{\sum_{i=1}^{m} \max\{p_i, q_i\}}.
                \end{equation}

                \paragraph{Tanimoto}
                The Tanimoto~\cite{Tanimoto57,Lipkus99}
                distance is another possible generalization of the Jaccard distance to real-valued multisets. Given $\vect{p}, \vect{q} \in \mathcal{P}$, their Tanimoto distance is:
                \begin{equation}
                    \label{eq:tanimoto_vect}
                    d_T(\vect{p},\vect{q}) = 1 - \frac{\langle\vect{p},\vect{q}\rangle}{\|\vect{p}\|^{2}+\|\vect{q}\|^{2}- \langle\vect{p},\vect{q}\rangle} = 1 - \frac{\sum_{i=1}^{m} p_i q_i}{\sum_{i=1}^{m} (p_i^{2} + q_i^{2}) - \sum_{i=1}^{m} p_i q_i}.
                \end{equation}
                
                In~\cref{subfig:embedding} we plotted the points $P$ via Multi Dimensional Scaling (MDS)~\cite{Kruskal64a}, after the Tanimoto distance has been computed between them. Genuine points $G$ generated by line $\vect{\theta}_1$ are close to each other in the \pspace, but are distant from those generated by $\vect{\theta}_2$, while anomalous points $A$ that do not belong to any structure are both distant to each other and to genuine points $G$, resulting in the most \emph{isolated} ones.
                
                \begin{figure}
                    \centering
                    \includegraphics[width = 0.85\linewidth]{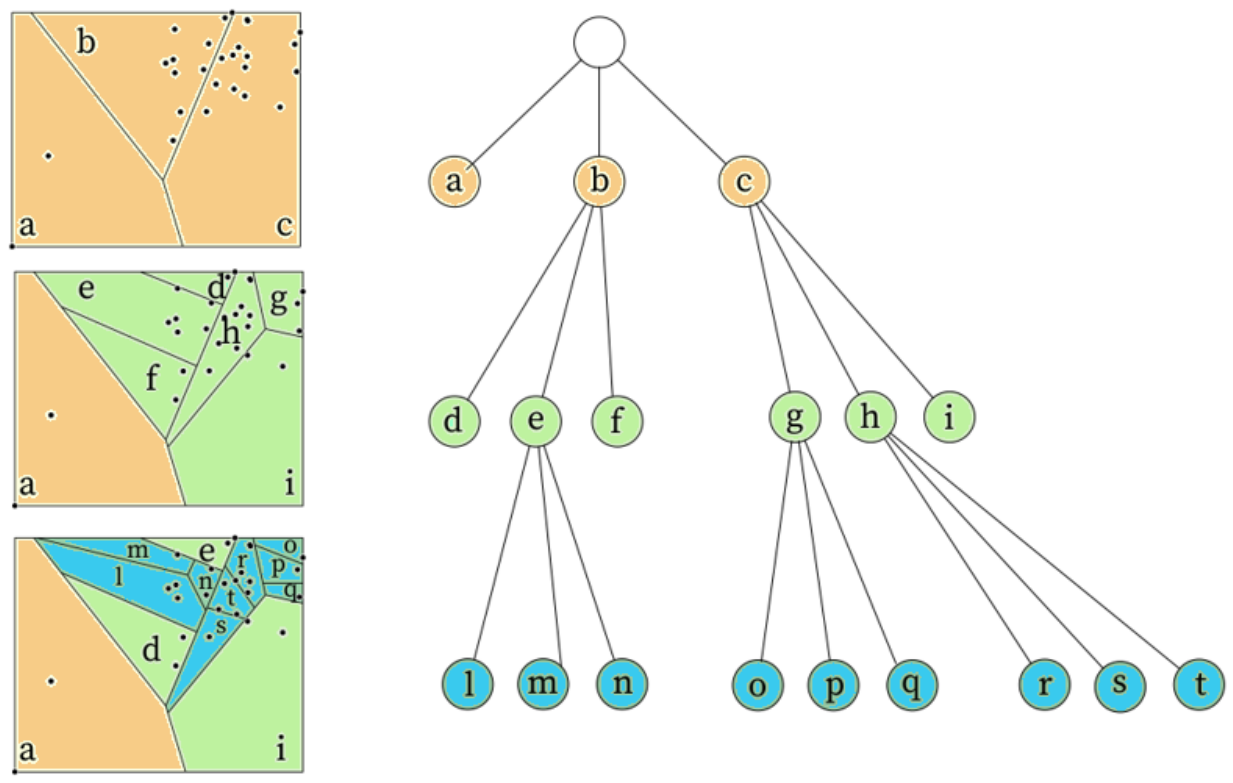}
                    \caption{A \vitree with branching factor $b=3$ and depth limit $l=3$ constructed from a set of points in $\mathbb{R}^2$. Every region is recursively split in $b$ sub-regions. The most isolated samples fall in leaves at lower depths, such as \texttt{a} and \texttt{d} cells.}
                    \label{fig:voronoi_itree}
                \end{figure}
                
                \begin{algorithm}[t]
                    \caption{\vitree \label{alg:vitree}}
                    \KwIn{$P$ - preference representations, $e$ - current tree depth, $l$ - depth limit, $b$ - branching factor}
                    \KwOut{$T$ - a \vitree}
                        \eIf{$e \geq l$ \normalfont{or} $|P| < b$\label{line:stopping_criteria}}
                            {\Return $exNode\{Size \leftarrow |P|\}$\label{line:node_creation}}
                            {$\{\vect{s}_i\}_{i=1,\ldots,b} \leftarrow \text{\textsc{Subsample}}(P, b)$\label{line:subsample_seeds}\\
                             $\{P_i\}_{i=1,\ldots, b} \leftarrow \text{\textsc{voronoiPartition}}(P, \{\vect{s}_i\}_{i=1,\ldots,b})$\label{line:vsplit}\\
                             $chNodes \leftarrow \emptyset$ \label{line:begin_recursive}\\
                             \For{$i = 1$  \normalfont{to} $b$}
                                 {$chNodes \leftarrow chNodes \cup \text{\vitree}(P_i, e + 1, l, b)$} \label{line:end_recursive}
                            \Return $inNode\{ChildNodes \leftarrow chNodes, SplitPoints \leftarrow \{\vect{s}_i\}_{i=1,\ldots,b}\}$}
                \end{algorithm}
                
            \subsubsection{\vitree}
                \label{subsubsec:voronoi_itree}
                
                \vitree is a tree-based AD technique that consists in a nested configuration of Voronoi tessellations, where each tessellation is further recursively split in $b$ sub-regions, as illustrated in~\cref{fig:voronoi_itree}.
                
                The construction of a \vitree is described in~\cref{alg:vitree}, where we first randomly sample $b$ seeds $\{\vect{s}_i\}_{i=1,\ldots,b} \subset P$ (\cref{line:subsample_seeds}) and use them to partition the set of preference representations $P$ (\cref{line:vsplit}). Specifically, the set $P$ is partitioned into $b$ subsets $\{P_i\}_{i=1,\ldots, b}$, where each $P_i \subset P$ collects those points in $P$ that have $\vect{s}_i$ as the closest seed according to the chosen distance. Each subset $P_i$ is subsequently used to build sub-trees in the recursive tree construction process (\cref{line:begin_recursive,line:end_recursive}). The partitioning process stops when the number of points in the region is less than $b$, or the tree reaches a maximum depth (\cref{line:stopping_criteria,line:node_creation}), set by default at $l = \log_{b}\psi$~\cite{Knuth98}, where $\psi$ is the number of points used to build the tree. 
                
                To decrease the variance due to randomness in \vitree realizations, we construct \viforest, a forest of \vitrees, and the \emph{average} depth of a point $\vect{x}$ is used to compute the anomaly score $\alpha(\vect{x})$~\cite{LiuTingAl12}.
                \cref{alg:viforest} details the construction of a \viforest containing
                $t$ \vitrees. Each \vitree is instantiated on a subset $P_{\psi} \subset P$ of preference representations (\cref{line:subsample}), where the subsampling factor is controlled by the parameter $\psi$.
                
                \begin{algorithm}[t]
                    \caption{\viforest \label{alg:viforest}}
                    \KwIn{$P$ - preference representations, $t$ - number of trees, $\psi$ - sub-sampling size, $b$ - branching factor}
                    \KwOut{$F = \{T_k\}_{k = 1, \dots, t}$ - set of \vitrees}
                        $F \leftarrow \emptyset$ \label{line:begin_forest_construction}\\
                        set depth limit $l = ceiling(\log_{b}\psi)$\\
                        \For{$k = 1$ \normalfont{to} $t$}
                            {$P_\psi \leftarrow \text{\textsc{Subsample}}(P, \psi)$ \label{line:subsample} \\
                             $T_k \leftarrow \text{\vitree}(P_\psi, 0, l, b)$ \label{line:tree_construction} \\
                             $F \leftarrow F \cup T_k$ \label{line:end_forest_construction}}
                        \Return $F$
                \end{algorithm}

            \subsubsection{Anomaly score computation}
                \label{subsubsec:anomaly_score_computation}

                \begin{algorithm}[t]
                    \caption{\textsc{PathLength} (\vitree) \label{alg:vitree_path_length}}
                    \KwIn{$\vect{p}$ - a sample, $T$ - a \vitree, $e$ - current path length}
                    \KwOut{$d(\vect{p})$ - path length}
                        \If{$T$ \normalfont{is a leaf}}
                          {\Return $e + c(T.size)$ \label{line:return}}
                        $childNode \leftarrow \text{\textsc{voronoiLocate}}(\vect{p}, T.splitPoints, T.childNodes)$\label{line:voronoi_locate}\\
                        \Return $\text{\textsc{PathLength}}(\vect{p}, childNode, e + 1)$ \label{line:recur}
                \end{algorithm}

                With reference to~\cref{alg:main}, we collect the depths reached by each instance $\vect{p} \in P$ in a set $D = \{d_k(\vect{p})\}_{k = 1, \dots, t}$ (\cref{line:point,line:height}).
                Then (\cref{line:anomaly_score}), analogously to~\cite{LiuTingAl12}, we compute the anomaly score as:
                \begin{equation}
                    \label{eq:anomaly_score}
                    \alpha_\psi(\vect{x}) = 2^{-\frac{E(D)}{c(\psi)}},
                \end{equation}
                where $E(D)$ is the mean value over the elements of $D$, and $c(\psi)$ is an adjustment factor.
                
                The depths are computed through the \textsc{PathLength} function described in~\cref{alg:vitree_path_length} by locating each instance $\vect{p}$ in the corresponding Voronoi partition (\cref{line:voronoi_locate}) and recursively traversing the sub-tree of the associated child node (\cref{line:recur}).
                The depth is computed as $d(\vect{p}) = e + c(T.size)$, where $e$ is the depth of the leaf and $c(n)$ is an adjustment factor to take into account the samples that fell in the considered leaf, but didn't contribute to the tree depth due to the depth limit $l$. We assume that $b=2$, and the adjustment factor $c(n)$ is computed as in~\cite{LiuTingAl12}.

        \subsection{Dealing with efficiency}
            \label{subsec:dealing_with_efficiency}

            Voronoi tessellations are effective to perform \pisolation in the \pspace, but they might be very time consuming when the number of points $n$ is large, as they require the explicit computation of distances between every point $\vect{p} \in P$ and $b$ seeds $\{\vect{s}_i\}_{i=1,\ldots,b}$, leading to a computational complexity of $O(n \, b)$ for each Voronoi tessellation. In order to overcome this bottleneck, we introduce \rzhash, a novel Locality Sensitive Hashing (LSH) scheme that splits points $P$ with complexity $O(n)$ and ensures that neighboring points according to Ruziska distance are more likely mapped to the same node, without the need to explicitly calculate distances.

            \subsubsection{\rzhash}
                \label{subsubsec:ruzhash}
                
                \rzhash is implemented as a function $h^{\vect{\pi}, \vect{\tau}}_{ruz}: [0, 1]^m \rightarrow \{1, \dots, m\}$.  The main idea, as depicted in~\cref{subfig:rzhash_partition}, is that two points $\vect{p}, \vect{q}$ close with respect to the Ruzicka distance are mapped via $h^{\vect{\pi}, \vect{\tau}}_{ruz}$ to the same node with high probability, \emph{i.e.}, $d_R(\vect{p}, \vect{q}) \approx 0 \iff Pr[h^{\vect{\pi}, \vect{\tau}}_{ruz}(\vect{p}) = h^{\vect{\pi}, \vect{\tau}}_{ruz}(\vect{q})] \approx 1$, while an isolated point $\vect{r}$ is mapped to a different node, \emph{i.e.}, $d_R(\vect{p}, \vect{r}) \approx 1 \iff Pr[h^{\vect{\pi}, \vect{\tau}}_{ruz}(\vect{p}) = h^{\vect{\pi}, \vect{\tau}}_{ruz}(\vect{r})] \approx 0$.
                Given a point $\vect{p} \in [0, 1]^m$, the \rzhash mapping is computed in  the following three steps:
                \begin{enumerate}
                    \item sample a vector of thresholds $\vect{\tau} = [\tau_1, \dots, \tau_m]$, where each $\tau_i \sim \mathcal{U}_{[0, 1)}$ is randomly drawn from the unitary interval,
                    \item sample a random permutation $\vect{\pi} = [\pi_1, \dots, \pi_m]$ of the indices $\{1, \dots, m\}$,
                    \item map $\vect{p}$ to the node having index $h^{\vect{\pi}, \vect{\tau}}_{ruz}(\vect{p}) = \min(\vect{\pi} \cdot \mathds{1}(\vect{p} > \vect{\tau}))$, where $\mathds{1}(\vect{p} > \vect{\tau}) = [\mathds{1}(p_1 > \tau_1), \dots, \mathds{1}(p_m > \tau_m)] \in \{0, 1\}^m$ is a binarization function, that is the image of the indicator function applied element-wise on $\vect{p}$ and $\vect{\tau}$ (\cref{subfig:thresholding}).
                \end{enumerate}
                
                \begin{figure}
                    \centering
                    \begin{subfigure}[t]{.65\linewidth}
                        \centering
                        \includegraphics[width=\linewidth]{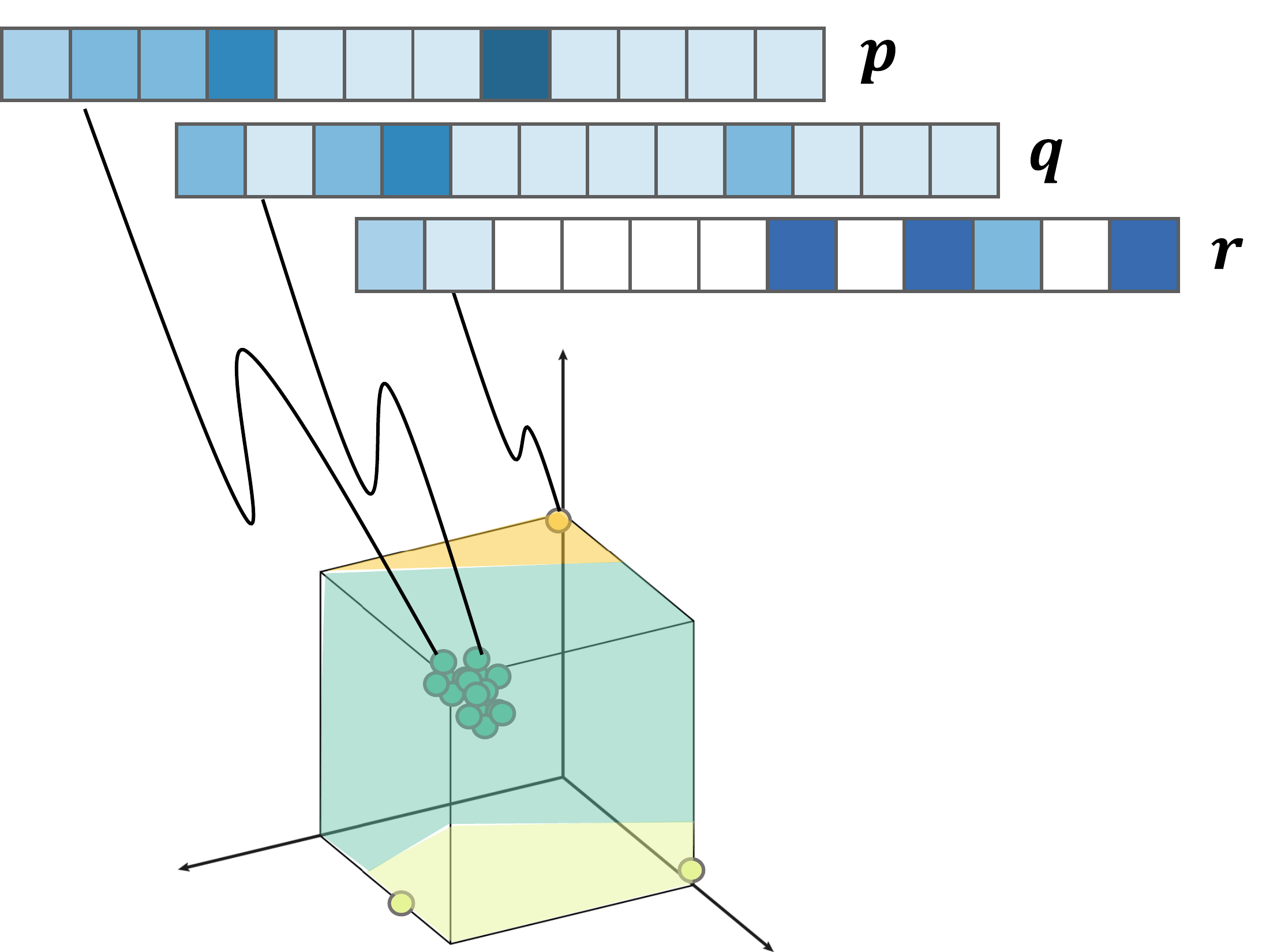}
                        \caption{}
                        \label{subfig:rzhash_partition}
                    \end{subfigure}
                    \hfill
                    \hspace{-8cm}
                    \begin{subfigure}[t]{.5\linewidth}
                        \centering
                        \includegraphics[width=\linewidth]{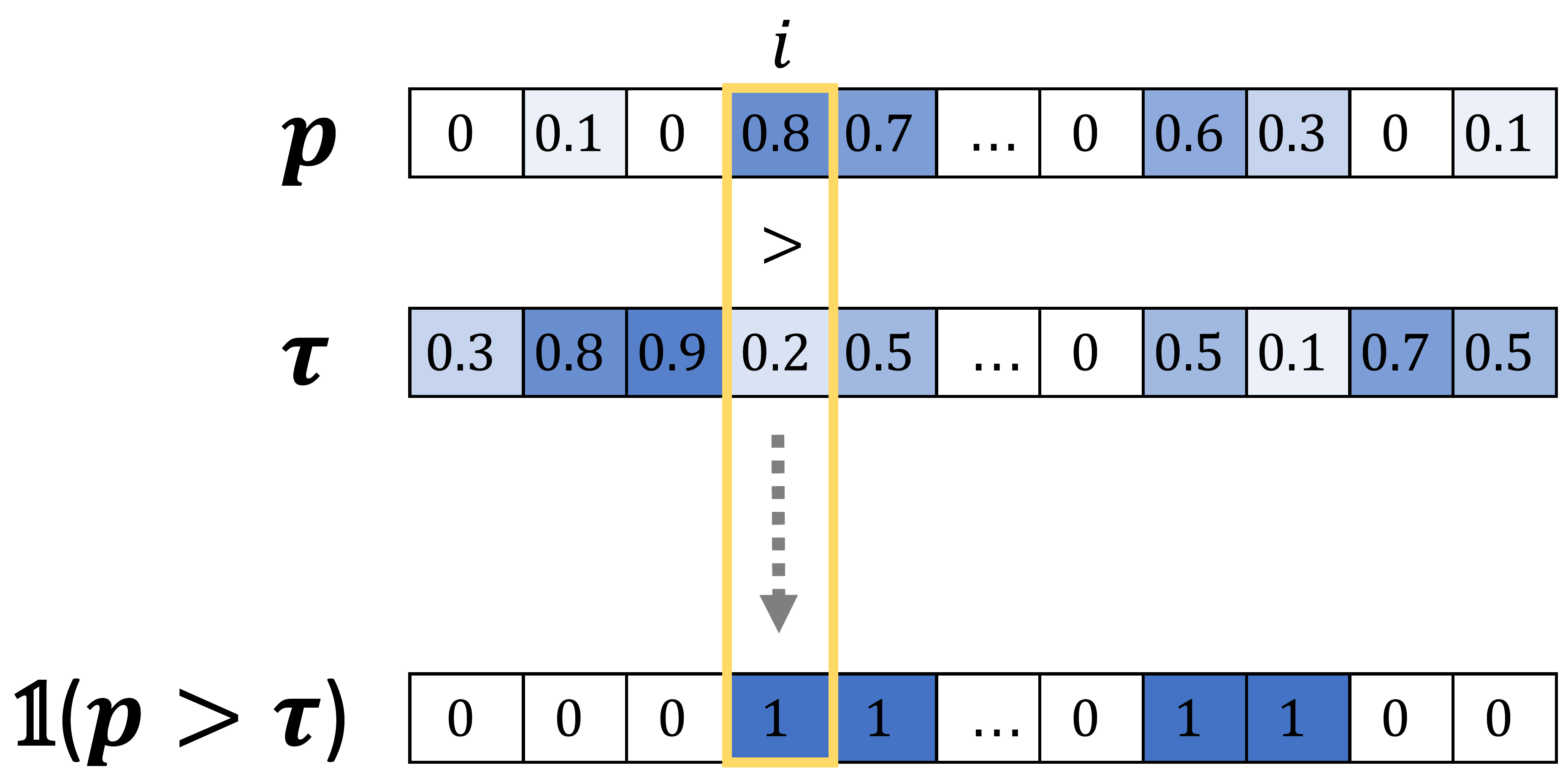}
                        \caption{}
                        \label{subfig:thresholding}
                    \end{subfigure}
                    \caption{(a) \rzhash induces a partitioning of the preference space $\mathcal{P}$, where close points are more likely mapped to the same region. We use \rzhash as splitting criteria within each node of our \rzhitree. (b) Example of binarization.}
                    \label{fig:rzhash}
                \end{figure}
                
                Our RuzHash mapping is a LSH of the Ruziska distance, as stated in the following
                \begin{theorem}
                    \label{thm:ruzhash_short}
                    Given $\vect{p}, \vect{q} \in [0, 1]^m$, then:
                    \begin{equation}
                        Pr[h^{\vect{\pi}, \vect{\tau}}_{ruz}(\vect{p}) = h^{\vect{\pi}, \vect{\tau}}_{ruz}(\vect{q})] = 1 - d_R(\vect{p}, \vect{q}),
                    \end{equation}
                \end{theorem}
                and proven in~\cref{apx:rzhash}.
                This is also illustrated in~\cref{fig:correlation}, where it can be appreciated a nearly perfect correlation between Ruzicka and its approximation via \rzhash.
                
            \subsubsection{\rzhitree}
                \label{subsubsec:ruzhash_itree}
                
                By employing the \rzhash mapping into a recursive partitioning procedure, we define a novel, more efficient version of \vitree, termed \rzhitree.
                In contrast to \vitree, which partitions points $P$ via Voronoi tessellations defined by $b$ seeds $\{\vect{s}_i\}_{i = 1, \dots, b}$, we employ the \rzhash random mapping $h^{\vect{\pi}, \vect{\tau}}_{ruz}$ defined by a vector of thresholds $\vect{\tau}$ and a permutation $\vect{\pi}$.
                Specifically, at each node of \rzhitree we split $P$ into $m$ subsets using \rzhash as depicted in~\cref{subfig:rzhash_partition}. The split is recursively performed until either: (\emph{i}) the current node contains a number of points less than $m$ or (\emph{ii}) the tree reaches a maximum depth.
                
                \begin{figure}[t]
                    \centering
                    \includegraphics[width=\linewidth]{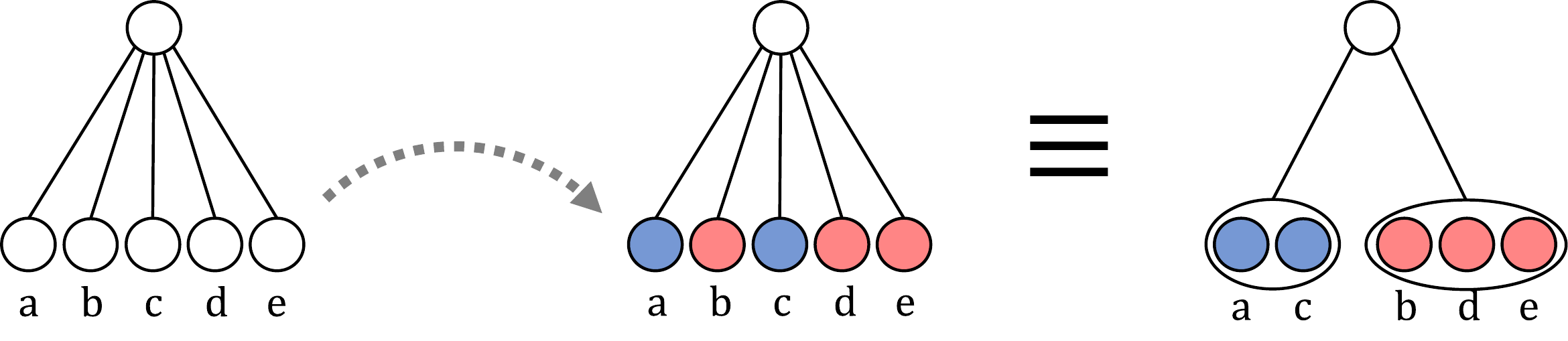}
                    \caption{On the left, split performed by \rzhash. In the middle, nodes aggregation where the groups has been color coded. On the right, resulting tree with branching factor $b = 2$.}
                    \label{fig:partitioning}
                \end{figure}
                
                \begin{algorithm}[tb]
                    \caption{\rzhitree \label{alg:rzhitree}}
                    \KwIn{$P$ - preference representations, $e$ - current tree depth, $l$ - depth limit, $b$ - branching factor}
                    \KwOut{$T$ - a \rzhitree}
                        \eIf{$e \geq l$  \normalfont{or} $|P| < b$}
                            {\Return $exNode\{Size \leftarrow |P|\}$}
                            {$\vect{\tau}, \vect{\pi} \leftarrow \text{\textsc{RuzHashSample}}(m)$\label{line:rzhsample}\\
                             $\vect{\beta} \leftarrow \text{\textsc{AggregationSample}}(m, b)$\label{line:aggrsample}\\
                             $\{P_i\}_{i=1,\ldots, b} \leftarrow \text{\textsc{RuzHashPartition}}(P, \vect{\tau}, \vect{\pi}, \vect{\beta})$\label{line:rzhsplit}\\
                             $chNodes \leftarrow \emptyset$\\
                             \For{$i = 1$  \normalfont{to} $b$}
                                 {$chNodes \leftarrow chNodes \cup \text{\rzhitree}(P_i, e + 1, l, b)$}
                            \Return $inNode\{ChildNodes \leftarrow chNodes, Thresholds \leftarrow \vect{\tau}$\\
                            \hspace{2.3cm} $Permutation \leftarrow \vect{\pi}, Aggregation \leftarrow \vect{\beta}\}$}
                \end{algorithm}
                
                \begin{table}[t]
                    \centering
                    \caption{Differences between \iforest, \viforest and \rzhiforest.}
                    \resizebox{\textwidth}{!}{
                    \begin{tabular}{c||c|c|c|c}
                                    & \multirow{2}{*}{Splitting scheme} & \multirow{2}{*}{Distance} & \multicolumn{2}{c}{Computational complexity}                                               \\ \cline{4-5}
                                    &                                   &                           & Training                                      & Testing                                    \\ \hline
                        \viforest   & Voronoi                           & Tanimoto                  & $O(\psi \: t \: b \: \log_{b} \psi)$ & $O(n \: t \: b \: \log_{b} \psi)$ \\
                        \iforest    & LSH                               & $\ell_1$                  & $O(\psi \: t \: \log_{2} \psi)$         & $O(n \: t \: \log_{2}\psi)$          \\
                        \rzhiforest & LSH                               & Ruzicka                   & $O(\psi \: t \: \log_{b} \psi)$         & $O(n \: t \: \log_{b} \psi)$         \\
                    \end{tabular}
                    }
                    \label{tab:differences}
                \end{table}
                
                \rzhash is designed to split the data into $m$ parts, where $m$ equals the dimension of $\mathcal{P} = [0, 1]^m$, however, we experienced that lower branching factors resulted in a better performance. Therefore, we accommodate for a different branching factor $b$ by randomly aggregating in $b < m$ groups the output produced by \rzhash after each split of the tree.
                \cref{fig:partitioning} shows an example of aggregation of $m = 5$ nodes produced by \rzhash into $b = 2$ nodes, resulting in a tree with branching factor $b = 2$.
                We perform this aggregation in two steps: \emph{i)} we sample a vector $\vect{\beta} = [\beta_1, \dots, \beta_m]$ where $\beta_i \sim \mathcal{U}_{\{1, \dots, b\}}$. This consists in assigning a random value in $\{1, \dots, b\}$ to each one of the $m$ nodes produced by \rzhash (in~\cref{fig:partitioning} we color-coded the different $b$ values assigned to each one of the $m$ nodes). \emph{ii)} We define a different version of \rzhash as the mapping $h^{\vect{\pi}, \vect{\tau}, \vect{\beta}}_{ruz} : [0, 1]^m \rightarrow \{1, \dots, b\}$ such that $h^{\vect{\pi}, \vect{\tau}, \vect{\beta}}_{ruz}(\vect{p}) = \beta_{\min (\vect{\pi} \cdot \mathds{1}(\vect{p} > \vect{\tau}))}$.
                In~\cref{alg:rzhitree} we detailed the construction process of a \rzhitree, where in~\cref{line:rzhsample,line:aggrsample} we sample the random vectors, while in~\cref{line:rzhsplit} we partition points $P$ via \rzhash.
                In~\cref{apx:variable_split} we theoretically studied how the choice of branching factor $b$ influences the Ruzicka distance approximation performed by \rzhash, while in~\cref{tab:differences} we summarized the main differences between \iforest, \viforest and \rzhiforest.
                
        \subsection{Dealing with locality}
            \label{subsec:dealing_with_locality}
            
            Here we present \spif, an efficient variation of \pif that exploits a \emph{locality principle}: points close in the ambient space $\mathcal{X}$ are likely to belong to the same genuine structure. This turns to be useful when the family $\mathcal{F}$ approximates the genuine data only locally (see~\cref{subfig:models_manifold_local_family}), and anomalies deviate from the regularity of smooth genuine structures.
            
            \begin{figure}[t]
                \centering
                \hspace{-1cm}
                \begin{subfigure}[t]{.4\linewidth}
                    \centering
                    \hspace*{-0.5cm}
                    \includegraphics[height=\linewidth]{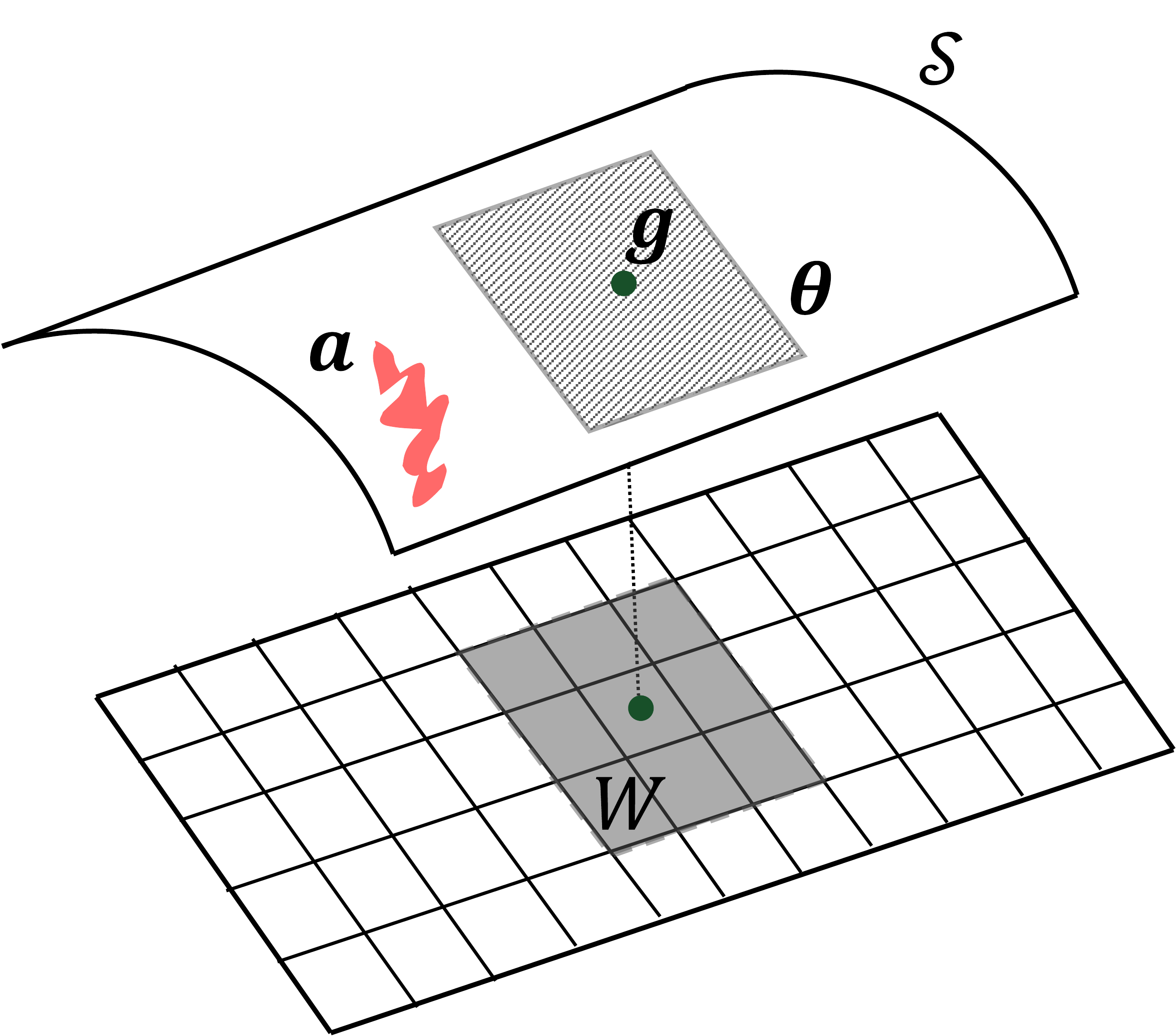}
                    \caption{}
                    \label{subfig:depthmap}
                \end{subfigure}
                \hspace{3cm}
                \begin{subfigure}[t]{.3\linewidth}
                    \centering
                    \hspace*{-1.5cm}
                    \includegraphics[height=0.8\linewidth]{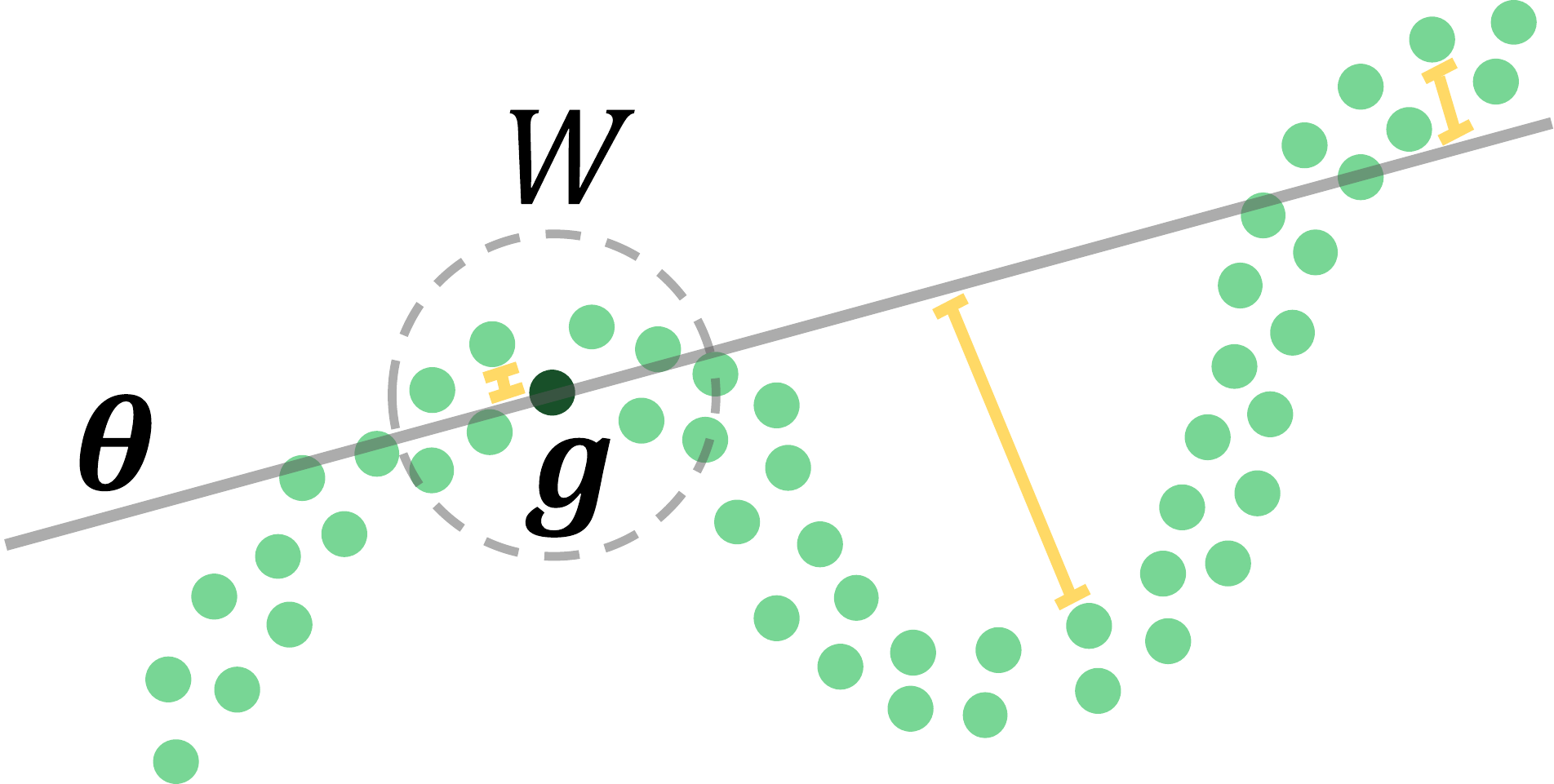}
                    \caption{}
                    \label{subfig:locality}
                \end{subfigure}
                \caption{(a) Range images connectivity enables efficient neighborhood definition. (b) Preferences are meaningful within the approximated region.}
                \label{fig:locality}
            \end{figure}
            
            \subsubsection{\spif}
                \label{subsubsec:sliding_pif}
                
                \spif leverages the locality principle by fitting models on points sampled from windows $W$ and performing structure-based AD independently within each window. To present \spif, we consider range images, which can be represented as 3D surfaces as illustrated in~\cref{subfig:depthmap}, and where we can efficiently exploit the locality principle since data are structured in a grid format.
                We assume that genuine data $G$ lie on a smooth surface $\mathcal{S}$, thus a genuine point $\vect{g}$ can be locally described by its tangent plane $\vect{\theta}$, while anomalies $A$ (defects, such as cracks) cannot.
                Rather than considering the whole dataset, we operate in a window-wise fashion by first selecting overlapping windows, and then performing \pif on the corresponding set of points $X_W$, producing an anomaly score $\alpha^W_{\psi}(\vect{x})$~\eqref{eq:anomaly_score} for each point $\vect{x} \in X_W$ and for each window $W$ it belongs to. The window-wise approach has two advantages: \emph{i)} we draw minimal sample sets from nearby points, promoting the extraction of genuine local models as in localized sampling strategies~\cite{KanazawaKawakami04}, and \emph{ii)} we compute preferences only for points within the local region of interest, yielding significant computational gain (\cref{subfig:locality}).
                Finally, we compute the overall anomaly score for each point $\vect{x}$ by averaging scores $\alpha^W_{\psi}(\vect{x})$ across all windows.
                
                The choice of the window size $\omega$ is closely tied to the size of the anomalous region, and it must be large enough to ensure anomalies remain a minority within the window. We found that windows overlapping by half of their width strike a good balance between efficiency and effectiveness.
                
                \begin{figure}[t]
                    \centering
                    \hspace{-0.5cm}
                    \begin{subfigure}[t]{.4\linewidth}
                        \centering
                        \hspace*{-0.75cm}
                        \includegraphics[height=\linewidth]{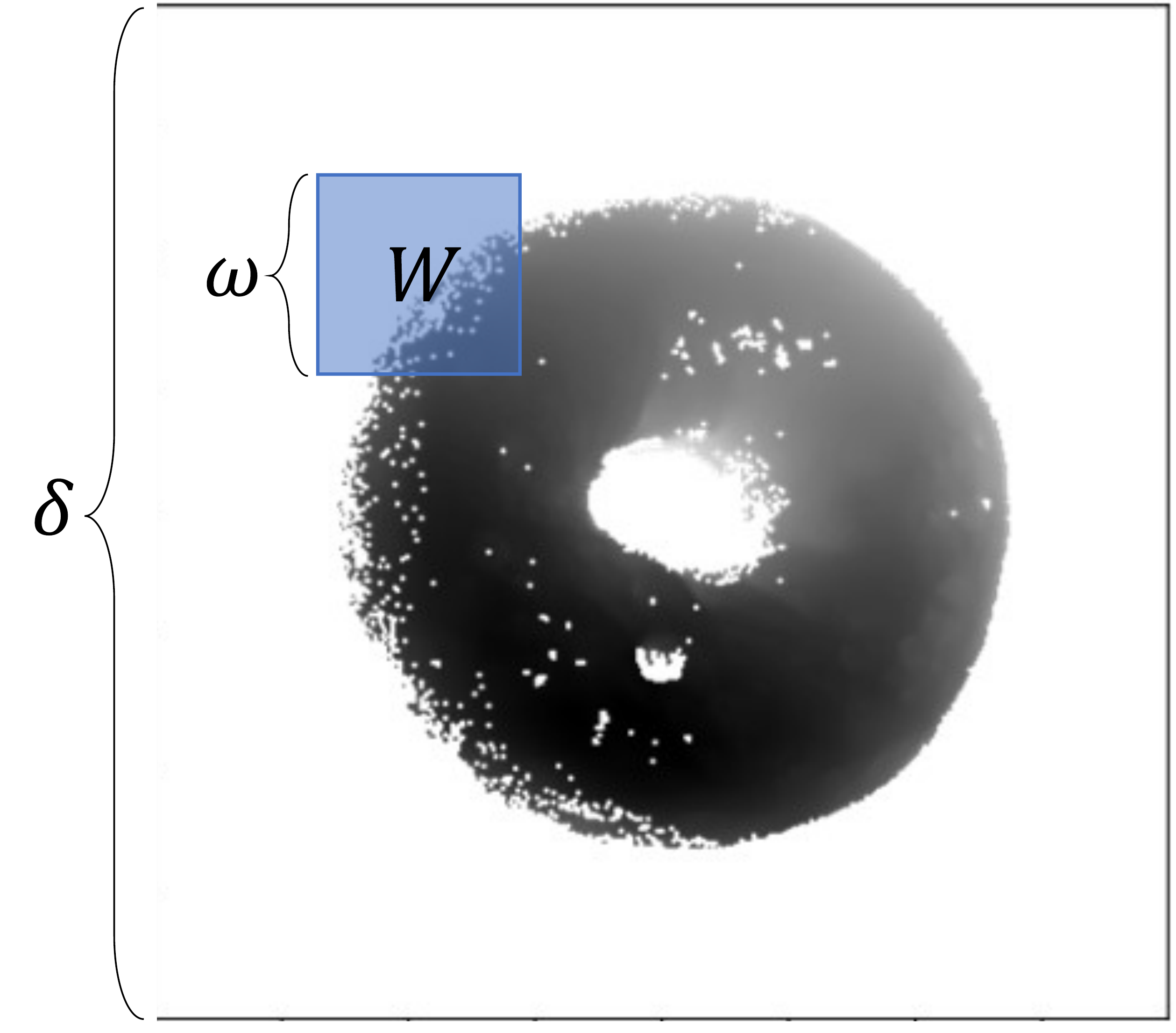}
                        \caption{}
                        \label{subfig:image_window}
                    \end{subfigure}
                    \hspace{1.25cm}
                    \begin{subfigure}[t]{.4\linewidth}
                        \centering
                        \hspace*{-0.9cm}
                        \includegraphics[height=\linewidth]{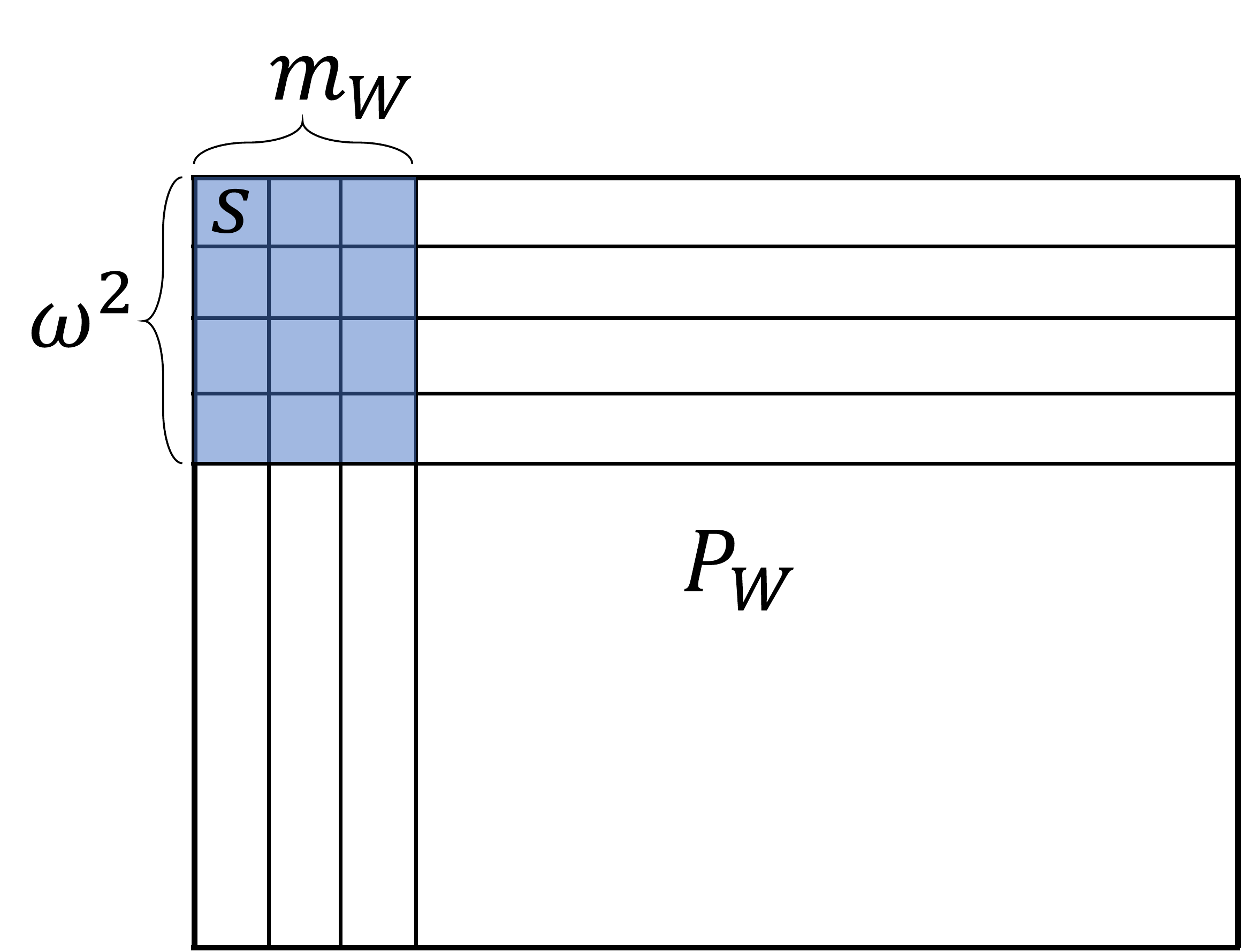}
                        \caption{}
                        \label{subfig:preference_matrix_window}
                    \end{subfigure}
                    \caption{(a) Bagel image with side length $\delta$ and superimposed window $W$ of size $\omega$, and (b) the corresponding local \pembedding matrix $P_W$.}
                    \label{fig:window}
                \end{figure}
                
                \spif benefits over \pif in terms of memory usage. \cref{subfig:image_window} shows an input image of size $\delta \times \delta$, a window $W$ of size $\omega = \frac{\delta}{k}$, and~\cref{subfig:preference_matrix_window} shows the \pembedding $P_W$ corresponding to $W$. The size of $P_W$ is proportional to the $\omega^2 = \frac{\delta^2}{k^2}$ points in $W$ and to the $m_W$ models fitted. Thus, the memory needed is $s \frac{\delta^2}{k^2} m_W$, where $s$ is the memory size of a preference value $p$.
                The total number of half-overlapping windows $W$ of size $\omega = \frac{\delta}{k}$ is $(2 \frac{\delta}{\omega} - 1)^2 = (2k - 1)^2$, resulting in a total memory need of
                \begin{equation*}
                    P_{size} = s \frac{d^2}{k^2} m_W (2k - 1)^2.
                \end{equation*}
                Thus, if we fix the total memory size to $P_{size}$, we can fit
                \begin{equation*}
                    m_W = \left\lfloor\frac{P_{size}}{s \frac{d^2}{k^2} (2k -1)^2}\right\rfloor
                \end{equation*}
                models.
                As instance, considering $s = 32$ bytes, $\delta = 800$ and $P_{size} = 1$ GB, \emph{each window} $W$ can accommodate $m_W = 419$ models for $\omega = \delta$ and $m_W = 110$ models for $\omega = \frac{\delta}{20}$, where \pif is \spif with window size $\omega = \delta$, resulting in a total number of models $m$ of $419$ and $167310$ respectively.

    \section{Experiments}
        \label{sec:pif_experiments}

        In this section, we assess the performance of \pif on both synthetic and real datasets. First, we demonstrate the superior performance of \pif compared to state-of-the-art anomaly detection methods, highlighting the importance of both leveraging a prior knowledge about genuine data and employing the suitable distances in the \pspace. Then, we show the efficiency improvement of \rzhiforest over \viforest and assess the benefits of \spif on data satisfying the locality principle.

        \subsection{Datasets}
            \label{subsec:pif_datasets}

            As regard synthetic datasets, we consider the 2D primitive fitting from~\cite{LeveniMagriAl21,LeveniMagriAl23}, consisting of points $X \in \mathbb{R}^2$, where genuine data $G$ lie on structures of a certain model family $\mathcal{F}$ (lines and circles), while anomalies are uniformly sampled within the range of $G$ such that $\frac{|A|}{|X|} = 0.5$. Each structure in the datasets contains $50$ normal points, except for stair3 and circle3, which have unbalanced structures.
            As far as real data are concerned, we consider the AdelaideRMF dataset~\cite{WongChinAl11}, consisting of stereo images with annotated matching points, where anomalies correspond to mismatches. The first $19$ sequences depict static scenes containing several planes, each giving rise to a set of matches described by the homography model family $\mathcal{F}$. The remaining $19$ sequences feature dynamic scenes with independently moving objects and give rise to a set of matches described by fundamental matrix model family $\mathcal{F}$.

            To evaluate the impact of local priors, we also considered the MVTec-3D AD~\cite{BergmannXinAl22} dataset, containing range images of $10$ different objects with various defects, such as holes and cracks. Scans were acquired using structured light, with positions $x$, $y$, and $z$ stored in three-channel tensors. We do not use the color information, and we restrict to objects that are locally smooth, namely \emph{bagel}, \emph{cookie}, \emph{peach}, \emph{potato}, and \emph{carrot}, to conform to our locality assumptions. For the same reason, we consider only the hole-defect type, as it comprises anomalies that violate the local smoothness assumption.

        \subsection{The effectiveness of structure-based anomaly detection}
            \label{subsec:effectiveness_structure_based_anomaly_detection}
            
            In the first set of experiments, we compare \pif against \ifor, \eifor and \lof. To assess the benefits of \pembedding, we performed experiments on synthetic data in the \emph{i)} ambient space $\mathbb{R}^2$, \emph{ii)} \pspace $[0, 1]^d$ and \emph{iii)} binary \pspace $\{0, 1\}^d$. Different \pisolation methods were explored for \pif, specifically \emph{i)} \vifor $\ell_2$ for the Euclidean space, \emph{i.e.}, setting $P = X$, \emph{ii)} \vifor jac for the binary \pspace, and \emph{iii)} \vifor tani, \vifor ruz and \rzhifor for the continuous \pspace. Note that in \emph{i)}, \pembedding is bypassed, and \pisolation is performed directly in the ambient space using \vifor, allowing us to assess its effectiveness as a standalone isolation-based anomaly detection algorithm. \lof was chosen as a density-based competitor because, similarly to \vifor, it can operate with arbitrary distance metrics, allowing its evaluation also in the \pspace and leading to a more comprehensive comparison.
            
            Preferences are computed with respect to a pool of $m = 10|X|$ model instances, in both circles and lines datasets. In real data experiments, performed only in the \pspace, the pool of models is determined by fitting $m = 6|X|$ instances, homographies for the first $19$ datasets, and fundamental matrices for the remaining $19$. In general, a higher $m$ yields better results, as increasing the number of fitted models improves the approximation of the true distances between points to the limit case where all possible models have been considered. In practice, we set $m$ to a sufficiently large value to ensure satisfactory performance.
            
            The parameters of \ifor, \eifor, \vifor and \rzhifor are fixed to $t = 100, \psi = 256$ and $b = 2$ throughout the experiments.
            Regarding \lof, given its sensitivity to the neighborhood size, various values of the parameter $k$ are employed, ranging from $k = 10$ to $k = 500$.
            
            Anomaly detection performance, summarized in~\cref{tab:lines_and_circles,tab:homographies_and_fundamentals}, is computed in terms of AUC, averaged over $10$ runs.
            The highest value for each dataset is underlined, and the best AUC in each embedding is also bolded if the difference is statistically significant compared to competitors (paired t-test with $\alpha = 0.05$).
            The \lof results refer to the $k$ value that maximizes the AUC or each model family and embedding (optimal $k$ parameter values are reported in~\cref{tab:neighborhood}).

            \begin{table}[t]
                \centering
                \caption{Optimal $k$ parameters of \lof.}
                \begin{tabular}{l@{\hskip 1.6cm}c@{\hskip 1.6cm}c@{\hskip 1.6cm}c}
                    \toprule
                                 & Euclidean & Preference binary & Preference \\
                    \midrule
                    circle       & $75$      & $25$              & $25$       \\
                    line         & $25$      & $150$             & $75$       \\
                    homography   & $-$       & $100$             & $100$      \\
                    fundamental  & $-$       & $75$              & $80$       \\
                    \bottomrule
                \end{tabular}
                \label{tab:neighborhood}
            \end{table}

            \cref{tab:lines_and_circles} shows that all methods improve their performance in the \pspace, highlighting the benefit of using prior knowledge about the model family $\mathcal{F}$ of genuine data. Additionally, continuous preferences are preferable over binary ones due to their higher expressiveness, which produces a space where anomalies can be identified more effectively. \lof performs better in datasets where normal data are evenly distributed, like star5, star11, circle4 and circle5, but struggles with unbalanced structures like stair3 and circle3. This is due to the fact that when normal data are evenly distributed among structures it is possible to get an optimal parameter of $k$, while in the unbalanced case not. Conversely, \vifor and \rzhifor, with fixed parameters, perform well also on unbalanced structures, with \vifor being statistically the best on average. The performance gap between \ifor, \eifor and \vifor, \rzhifor is very clear, as only the latter methods leverage the suitable distance metrics for \pspace.
            
            For real datasets, while AUC values in~\cref{tab:homographies} and~\cref{tab:fundamentals} indicate that anomaly detection on these tasks is easier than in the synthetic case, the performance gap between \ifor, \eifor and \lof, \vifor, \rzhifor remains significant.

            \begin{figure}[t]
                \hspace{-1.35cm}
                \begin{subfigure}[b]{.6\columnwidth}
                    \begin{center}
                    \centerline{\includegraphics[width=\columnwidth]{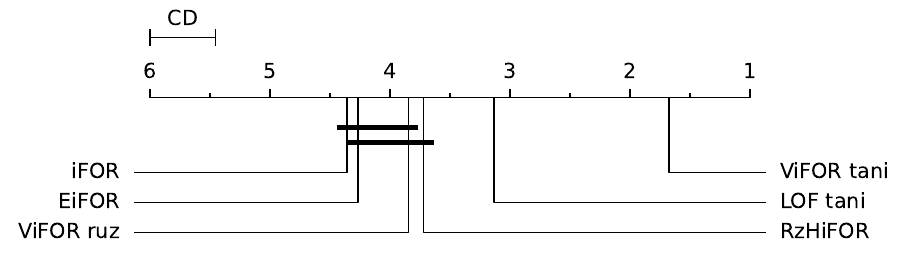}}
                    \caption{Homographies AUCs.}
                    \label{fig:cd_roc_adelH_aucs}
                    \end{center}
                \end{subfigure}
                \begin{subfigure}[b]{.6\columnwidth}
                    \begin{center}
                    \centerline{\includegraphics[width=\columnwidth]{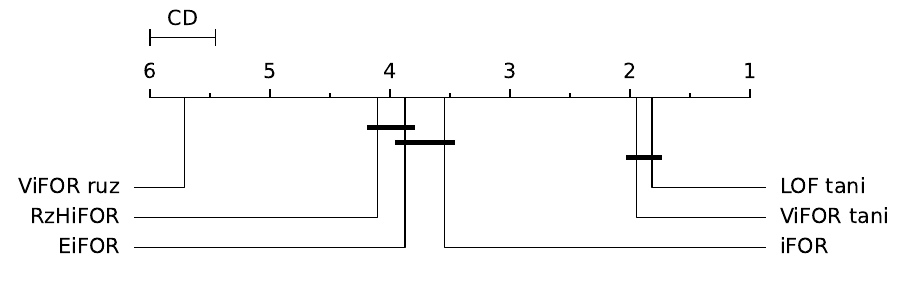}}
                    \caption{Fundamental matrices AUCs.}
                    \label{fig:cd_roc_adelFM_aucs}
                    \end{center}
                \end{subfigure}
                \hspace*{2cm}
                \begin{subfigure}[b]{.6\columnwidth}
                    \begin{center}
                    \centerline{\includegraphics[width=\columnwidth]{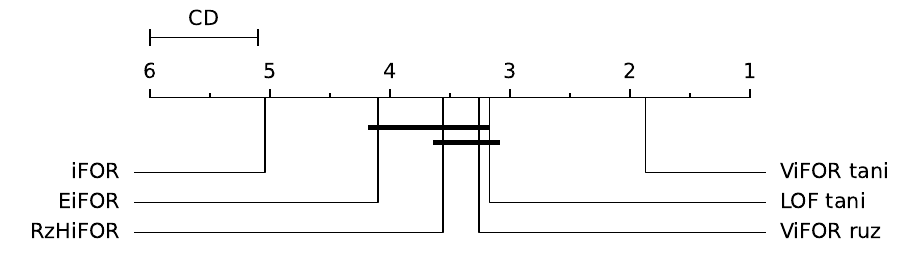}}
                    \caption{Primitive fitting AUCs.}
                    \label{fig:cd_roc_synth2D_aucs}
                    \end{center}
                \end{subfigure}
                \caption{Critical difference diagram for ROC AUCs on synthetic (bottom row) and real (top row) datasets.}
                \label{fig:sbad_cd}
            \end{figure}

            \begin{figure}[t]
                \begin{center}
                \centerline{\includegraphics[width=.6\columnwidth]{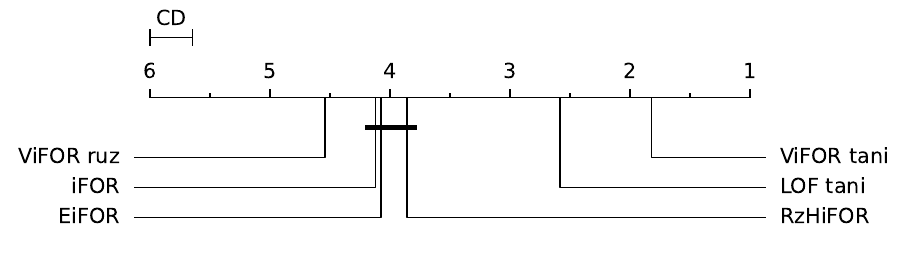}}
                \caption{Critical difference diagram for ROC AUCs on all datasets.}
                \label{fig:cd_roc_all_aucs}
                \end{center}
            \end{figure}

            In addition, the critical difference diagrams in \cref{fig:sbad_cd,fig:cd_roc_all_aucs} provide a synthesized overview of the results across multiple executions and datasets. The statistical analysis has been conducted on $6$ populations (one for each algorithm) with a total of $450$ paired ROC AUC values ($70$ for primitive fitting and $190$ each for homographies and fundamental matrices). Critical difference diagrams are based on the post-hoc Nemenyi test, and differences between populations are considered significant if their mean rank difference exceeds the critical distance $CD = 0.355$ (with specific values of $0.901$ for primitive fitting and $0.547$ for both homographies and fundamental matrices). Mean ranks and critical diagrams have been generated via the \textit{Autorank}~\cite{Herbold20} library. Critical diagrams confirm the results commented above, and emphasize the consistent advantage of approaches that incorporate appropriate distance measures, such as \lof, \vifor and \rzhifor, over those that do not, like \ifor and \eifor.
            
            Notably, \vifor consistently achieves, on average, the best performance across all the space configurations considered (\cref{tab:lines_and_circles,tab:homographies_and_fundamentals}), a trend that is almost always confirmed by the critical difference diagrams (\cref{fig:sbad_cd,fig:cd_roc_all_aucs}). This further highlights \vifor generality and suitability as a standalone isolation-based anomaly detection algorithm, with the key feature of being able to operate with arbitrary distance measures.
            
            \begin{table}[t]
                \caption{Synthetic datasets AUCs (primitive fitting).}
                \begin{subtable}{\textwidth}
                    \raggedright
                    \caption{\raggedright Euclidean ambient space $\mathcal{X}$.}
                    \begin{tabular}{l@{\hskip 1cm}cccc}
                        \toprule
                                & \lof $\ell_2$ & \ifor   & \eifor  & \vifor $\ell_2$ \\
                                \midrule
                        stair3  & $0.737$       & $0.925$ & $0.920$ & $0.918$         \\
                        stair4  & $0.814$       & $0.889$ & $0.874$ & $0.871$         \\
                        star5   & $0.771$       & $0.722$ & $0.738$ & $0.788$         \\
                        star11  & $0.671$       & $0.728$ & $0.727$ & $0.738$         \\
                        circle3 & $0.761$       & $0.698$ & $0.732$ & $0.779$         \\
                        circle4 & $0.640$       & $0.641$ & $0.665$ & $0.679$         \\
                        circle5 & $0.543$       & $0.569$ & $0.570$ & $0.633$         \\
                        \midrule
                        Mean    & $0.705$       & $0.739$ & $0.747$ & $0.772$         \\
                        \bottomrule
                    \end{tabular}
                    \label{tab:euclidean}
                \end{subtable}
                \\
                \vfill
                \begin{subtable}{\textwidth}
                    \raggedright
                    \caption{\raggedright Binary preference space $\mathcal{P} = \{0, 1\}^m$.}
                    \begin{tabular}{l@{\hskip 1cm}cccc}
                        \toprule
                                & \lof jac & \ifor   & \eifor  & \vifor jac \\
                                \midrule
                        stair3  & $0.904$  & $0.885$ & $0.864$ & $0.958$    \\
                        stair4  & $0.849$  & $0.855$ & $0.860$ & $0.941$    \\
                        star5   & $0.875$  & $0.745$ & $0.769$ & $0.872$    \\
                        star11  & $0.830$  & $0.739$ & $0.741$ & $0.771$    \\
                        circle3 & $0.719$  & $0.842$ & $0.854$ & $0.900$    \\
                        circle4 & $0.827$  & $0.686$ & $0.699$ & $0.860$    \\
                        circle5 & $0.699$  & $0.597$ & $0.617$ & $0.672$    \\
                        \midrule
                        Mean    & $0.815$  & $0.764$ & $0.772$ & $0.853$    \\
                        \bottomrule
                    \end{tabular}
                    \label{tab:jaccard}
                \end{subtable}
                \\
                \vfill
                \begin{subtable}{\textwidth}
                    \raggedright
                    \caption{\raggedright Continuous preference space $\mathcal{P} = [0, 1]^m$.}
                    \begin{tabular}{l@{\hskip 1cm}cccccc}
                        \toprule
                                & \lof tani                    & \ifor   & \eifor  & \vifor tani                  & \vifor ruz & \rzhifor                     \\
                                \midrule
                        stair3  & $0.815$                      & $0.923$ & $0.925$ & \underline{$0.971$}          & $0.969$    & $0.959$                      \\
                        stair4  & $0.881$                      & $0.912$ & $0.908$ & $0.952$                      & $0.950$    & \underline{$\mathbf{0.965}$} \\
                        star5   & \underline{$\mathbf{0.929}$} & $0.761$ & $0.822$ & $0.910$                      & $0.869$    & $0.841$                      \\
                        star11  & \underline{$\mathbf{0.900}$} & $0.738$ & $0.774$ & $0.796$                      & $0.817$    & $0.714$                      \\
                        circle3 & $0.731$                      & $0.854$ & $0.891$ & \underline{$\mathbf{0.930}$} & $0.851$    & $0.871$                      \\
                        circle4 & \underline{$0.906$}          & $0.667$ & $0.720$ & $0.897$                      & $0.777$    & $0.722$                      \\
                        circle5 & \underline{$\mathbf{0.823}$} & $0.573$ & $0.593$ & $0.780$                      & $0.565$    & $0.638$                      \\
                        \midrule
                        Mean    & $0.855$                      & $0.775$ & $0.805$ & \underline{$\mathbf{0.891}$} & $0.828$    & $0.816$                      \\
                        \bottomrule
                    \end{tabular}
                    \label{tab:tanimoto}
                \end{subtable}
                \label{tab:lines_and_circles}
            \end{table}

            \begin{table}[t]
                \caption{Real datasets AUCs.}
                \begin{subtable}{\textwidth}
                    \raggedright
                    \caption{\raggedright Homographies.}
                    \resizebox{\textwidth}{!}{
                    \begin{tabular}{l@{\hskip 1cm}cccccc}
                        \toprule
                                        & \lof tani                    & \ifor               & \eifor   & \vifor tani                  & \vifor ruz                       & \rzhifor \\
                                        \midrule
                        barrsmith       & \underline{$\mathbf{0.969}$} & $0.708$             & $0.715$  & $0.944$                      & $0.698$                          & $0.692$  \\
                        bonhall         & $0.918$                      & \underline{$0.969$} & $0.967$  & $0.949$                      & $0.960$                          & $0.951$  \\
                        bonython        & \underline{$\mathbf{0.978}$} & $0.679$             & $0.691$  & $0.954$                      & $0.911$                          & $0.857$  \\
                        elderhalla      & \underline{$0.999$}          & $0.925$             & $0.909$  & \underline{$\mathbf{0.999}$} & $0.878$                          & $0.877$  \\
                        elderhallb      & $0.986$                      & $0.924$             & $0.943$  & \underline{$\mathbf{0.999}$} & $0.966$                          & $0.976$  \\
                        hartley         & $0.963$                      & $0.749$             & $0.793$  & \underline{$\mathbf{0.989}$} & $0.911$                          & $0.882$  \\
                        johnsona        & $0.993$                      & $0.993$             & $0.993$  & \underline{$\mathbf{0.998}$} & $0.962$                          & $0.993$  \\
                        johnsonb        & $0.776$                      & \underline{$0.999$} & $0.998$  & \underline{$0.999$}          & $0.929$                          & $0.980$  \\
                        ladysymon       & $0.847$                      & $0.944$             & $0.943$  & \underline{$\mathbf{0.997}$} & $0.942$                          & $0.983$  \\
                        library         & \underline{$\mathbf{1.000}$} & $0.764$             & $0.771$  & $0.998$                      & $0.968$                          & $0.936$  \\
                        napiera         & $0.975$                      & $0.869$             & $0.879$  & \underline{$\mathbf{0.983}$} & $0.756$                          & $0.791$  \\
                        napierb         & $0.888$                      & $0.931$             & $0.936$  & \underline{$\mathbf{0.953}$} & $0.938$                          & $0.948$  \\
                        neem            & $0.985$                      & $0.896$             & $0.906$  & \underline{$\mathbf{0.996}$} & $0.928$                          & $0.969$  \\
                        nese            & \underline{$\mathbf{0.996}$} & $0.888$             & $0.892$  & $0.980$                      & $0.920$                          & $0.958$  \\
                        oldclassicswing & $0.936$                      & $0.923$             & $0.943$  & $0.987$                      & \underline{$\mathbf{0.998}$}     & $0.993$  \\
                        physics         & $0.670$                      & $0.858$             & $0.787$  & \underline{$\mathbf{1.000}$} & $0.989$                          & $0.969$  \\
                        sene            & \underline{$\mathbf{0.997}$} & $0.698$             & $0.731$  & $0.988$                      & $0.936$                          & $0.763$  \\
                        unihouse        & $0.785$                      & $0.998$             & $0.998$  & \underline{$\mathbf{0.999}$} & $0.930$                          & $0.980$  \\
                        unionhouse      & \underline{$\mathbf{0.987}$} & $0.639$             & $0.664$  & $0.968$                      & $0.877$                          & $0.840$  \\
                        \midrule
                        Mean            & $0.929$                      & $0.861$             & $0.866$  & \underline{$\mathbf{0.983}$} & $0.916$                          & $0.913$  \\
                        \bottomrule
                    \end{tabular}
                    }
                    \label{tab:homographies}
                \end{subtable}
                \\
                \vfill
                \begin{subtable}{\textwidth}
                    \raggedright
                    \caption{\raggedright Fundamental matrices.}
                    \resizebox{\textwidth}{!}{
                    \begin{tabular}{l@{\hskip 1cm}cccccc}
                        \toprule
                                          & \lof tani                    & \ifor               & \eifor              & \vifor tani                  & \vifor ruz   & \rzhifor                     \\
                                          \midrule
                        biscuit           & $0.976$                      & $0.994$             & $0.996$             & \underline{$\mathbf{1.000}$} & $0.922$      & $0.967$                      \\
                        biscuitbook       & \underline{$1.000$}          & $0.987$             & $0.988$             & \underline{$1.000$}          & $0.945$      & $0.986$                      \\
                        biscuitbookbox    & \underline{$\mathbf{1.000}$} & $0.990$             & $0.989$             & $0.996$                      & $0.895$      & $0.982$                      \\
                        boardgame         & \underline{$\mathbf{0.962}$} & $0.400$             & $0.304$             & $0.949$                      & $0.750$      & $0.883$                      \\
                        book              & $0.996$                      & \underline{$1.000$} & \underline{$1.000$} & \underline{$1.000$}          & $0.925$      & $0.982$                      \\
                        breadcartoychips  & \underline{$\mathbf{0.989}$} & $0.978$             & $0.971$             & $0.976$                      & $0.827$      & $0.975$                      \\
                        breadcube         & \underline{$\mathbf{1.000}$} & $0.998$             & $0.998$             & $0.999$                      & $0.777$      & $0.979$                      \\
                        breadcubechips    & \underline{$\mathbf{0.999}$} & $0.985$             & $0.985$             & $0.998$                      & $0.861$      & $0.976$                      \\
                        breadtoy          & $0.984$                      & \underline{$0.999$} & $0.998$             & \underline{$0.999$}          & $0.920$      & $0.984$                      \\
                        breadtoycar       & \underline{$\mathbf{0.998}$} & $0.933$             & $0.883$             & $0.991$                      & $0.827$      & $0.957$                      \\
                        carchipscube      & \underline{$\mathbf{0.993}$} & $0.981$             & $0.966$             & $0.987$                      & $0.829$      & $0.975$                      \\
                        cubetoy           & \underline{$\mathbf{1.000}$} & $0.997$             & $0.995$             & \underline{$1.000$}          & $0.960$      & $0.987$                      \\
                        cube              & \underline{$0.999$}          & $0.970$             & $0.982$             & \underline{$0.999$}          & $0.895$      & $0.952$                      \\
                        cubebreadtoychips & \underline{$0.990$}          & $0.962$             & $0.958$             & $0.989$                      & $0.752$      & $0.930$                      \\
                        cubechips         & \underline{$\mathbf{1.000}$} & $0.995$             & $0.994$             & \underline{$1.000$}          & $0.980$      & $0.996$                      \\
                        dinobooks         & $0.887$                      & $0.873$             & $0.857$             & $0.899$                      & $0.879$      & \underline{$\mathbf{0.946}$} \\
                        game              & \underline{$\mathbf{1.000}$} & $0.901$             & $0.895$             & $0.999$                      & $0.844$      & $0.898$                      \\
                        gamebiscuit       & \underline{$1.000$}          & $0.985$             & $0.988$             & \underline{$\mathbf{1.000}$} & $0.876$      & $0.981$                      \\
                        toycubecar        & \underline{$\mathbf{0.973}$} & $0.290$             & $0.192$             & $0.964$                      & $0.742$      & $0.935$                      \\
                        \midrule
                        Mean              & \underline{$0.987$}          & $0.906$             & $0.891$             & \underline{$0.987$}          & $0.864$      & $0.962$                      \\
                        \bottomrule
                    \end{tabular}
                    }
                    \label{tab:fundamentals}
                \end{subtable}
                \label{tab:homographies_and_fundamentals}
            \end{table}

        \subsection{The impact of hashing on efficiency}
            \label{subsec:impact_hashing_on_efficiency}

            \begin{figure}[t]
               \centering
               \includegraphics[width=.75\linewidth]{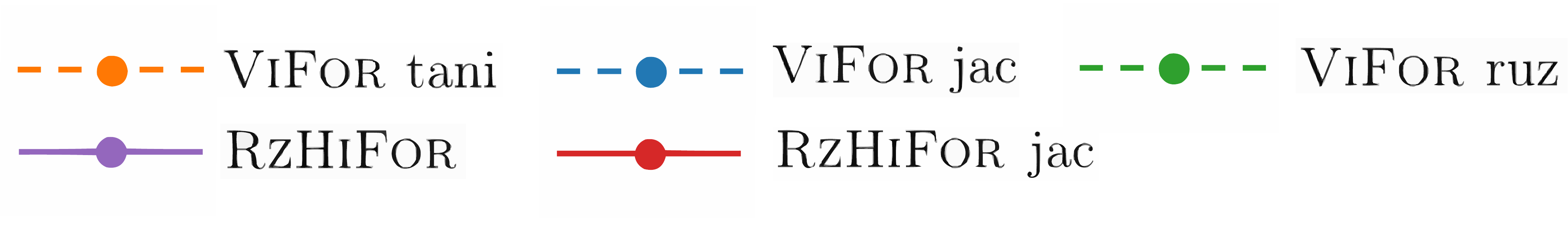}
                \\
                \hspace*{-0.85cm}
                \captionsetup[subfigure]{oneside, margin={0.75cm,0cm}}
                \subfloat[\label{fig:roc_auc}]{
                    \includegraphics[width=.35\linewidth]{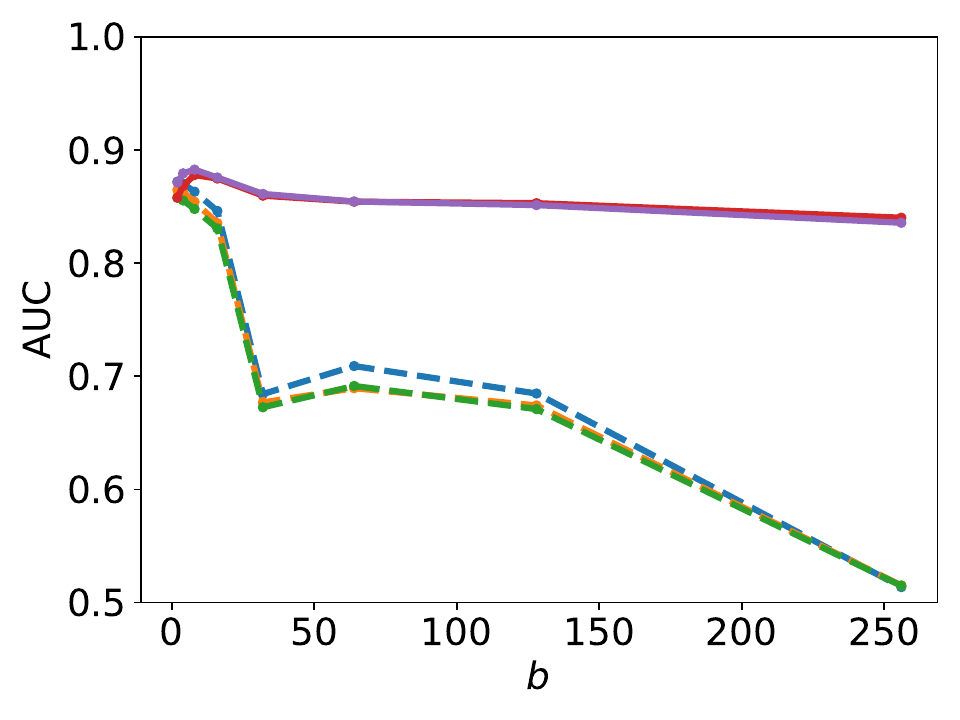}
                    \vspace{-0.25cm}
                }
                \subfloat[\label{fig:test_time}]{
                    \includegraphics[width=.35\linewidth]{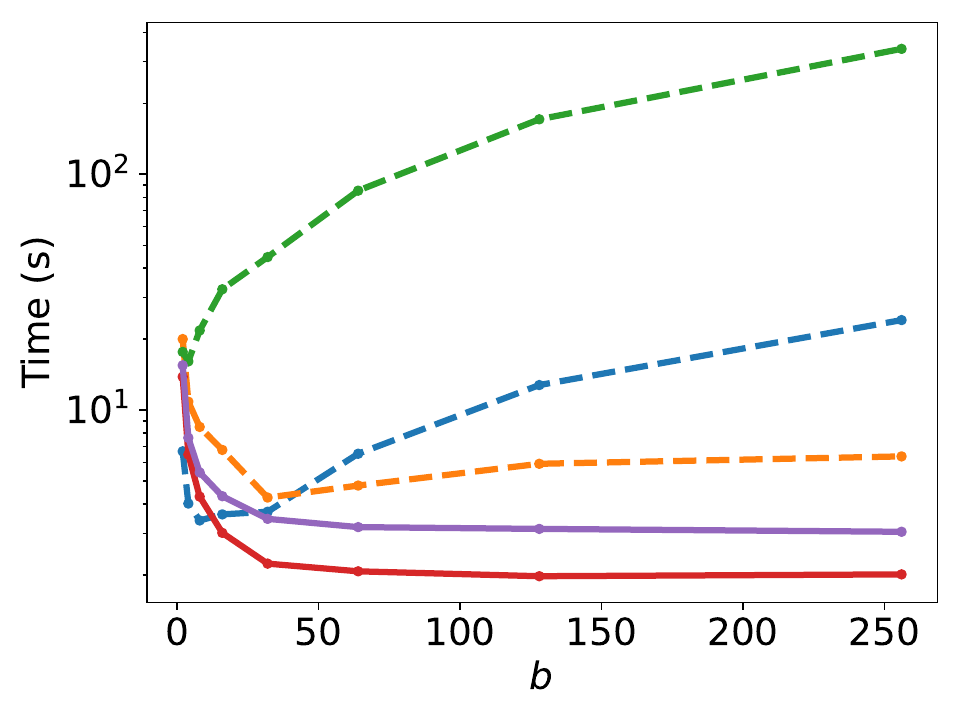}
                    \vspace{-0.25cm}
                }
                \subfloat[\label{fig:best_roc_auc_time}]{
                    \includegraphics[width=.35\linewidth]{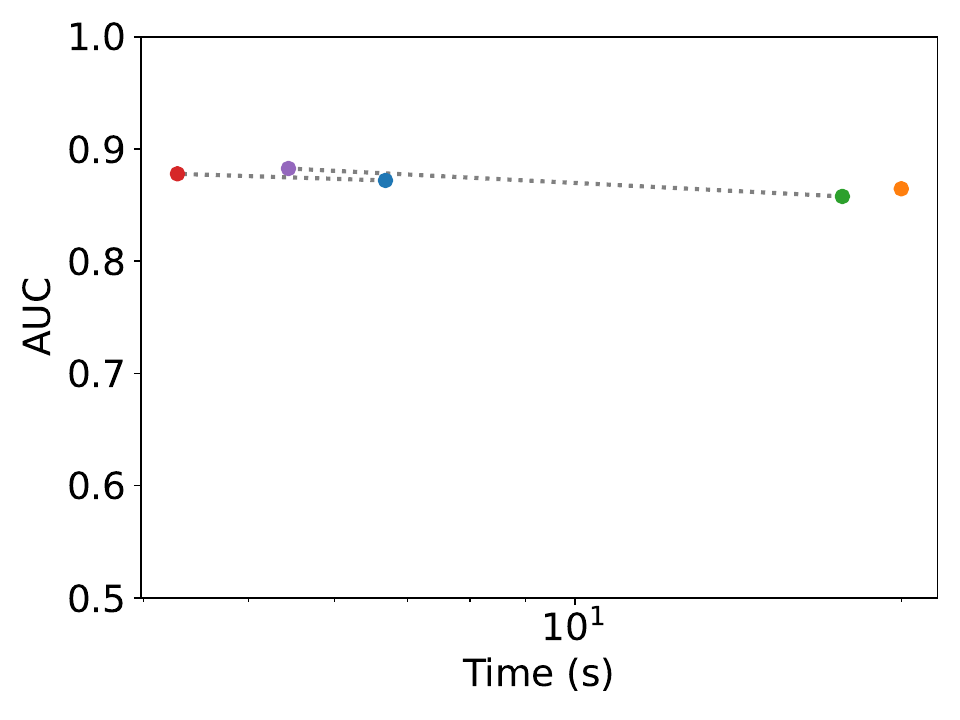}
                    \vspace{-0.25cm}
                }
                \caption{(a) Average ROC AUCs. (b) Average test times. (c) Relation between best average ROC AUC and corresponding test time.}
                \label{fig:results}
            \end{figure}

            In the second set of experiments, we compared \rzhiforest and \viforest both in the binary and in the continuous \pspace.
            Both methods were tested under the same parameters condition: $t = 100$, $\psi = 256$, with branching factor ranging varying in $b = [2, 4, 8, 16, 32, 64, 128, 256]$.

            \cref{fig:results} shows the aggregated results of our experiments. In~\cref{fig:roc_auc} and~\cref{fig:test_time}, we show the average ROC AUC and test time at different branching factors respectively. The curves represent averages from $5$ executions on both synthetic and real datasets.
            
            These experiments show that \rzhiforest is the most stable approach across different branching factors $b$ in terms of ROC AUC.
            The efficiency gain in test time is substantial, as shown in~\cref{fig:test_time}, and it is consistent with the computational complexities showed in~\cref{tab:differences}.
            
            \cref{fig:best_roc_auc_time} presents a different visualization, showing the relationship between the best ROC AUC and the corresponding test time for each method. We selected the branching factor that maximizes ROC AUC for each method in~\cref{fig:roc_auc}, and used it to identify the corresponding test time in~\cref{fig:test_time}. Dashed lines connect \rzhiforest and \viforest results that refer to the same underlying distance measure.
            It can be appreciated that \rzhiforest achieves results comparable to \viforest but with a significant reduction in execution time. Specifically, \rzhifor jac is $\times 35\%$ faster than \vifor jac, and \rzhifor is $\times 70\%$ faster than \vifor ruz.
            
        \subsection{The impact of sliding window}
            \label{subsec:impact_sliding_window}

            We assess \spif on a real-world scenario, namely the MVTec-3D AD dataset, using families of local models $\mathcal{F}$ with increasing generality (planes, spheres, and quadrics). Notably, we do not rely on any ground-truth genuine or anomalous data, since \spif is a truly unsupervised method. We ran \spif with different window sizes $\omega$ to account for various defect sizes ($\omega = [\delta, \frac{\delta}{2}, \frac{\delta}{4}, \frac{\delta}{10}, \frac{\delta}{20}]$, where $\delta$ is the image side length). For \pisolation we used \rzhiforest in the continuous \pspace with $t = 100$, $\psi = 256$ and $b = 16$ for greater efficiency.
            
            We evaluate the impact of the window size $\omega$ on qualitative experiments. \cref{tab:sliding_pif_qualitative} compares \pif ($\omega = \delta$) to \spif with smallest window size ($\omega = \frac{\delta}{20}$), alongside ground truths. \cref{tab:sliding_pif_quantitative} shows average AUC on $5$ runs achieved by \pif and \spif at different $\omega$.

            \begin{table}[!t]
                \centering
                \caption{(a) Qualitative results of \pif ($\omega = \delta$) and \spif ($\omega = \frac{\delta}{20}$), alongside ground truths. Anomaly scores $\alpha(\cdot)$ color-coded (red indicates anomalies), and AUC values displayed below objects. (b) Average AUC values on $5$ runs achieved by \pif and \spif at different $\omega$.}
                
                \hspace*{-0.5cm}
                \resizebox{\textwidth}{!}{
                \begin{subtable}[b]{.8\textwidth}
                    \centering
                    \caption{}
                    \resizebox{\textwidth}{!}{
                    \begin{tabular}{ll|ccccc}
                        $\mathcal{F}$ & $\omega$                                       & bagel & cookie & peach & potato & carrot \\
                        \midrule
                        \multirow{2}{*}{Plane}
                                      & \raisebox{0.06\textwidth}{$\delta$}            & \includegraphics[width=0.1\textwidth]{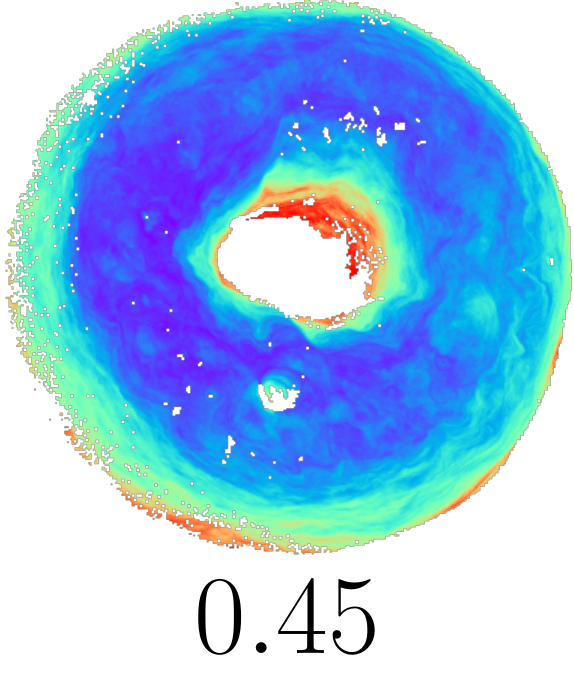}
                                                                                       & \includegraphics[width=0.1\textwidth]{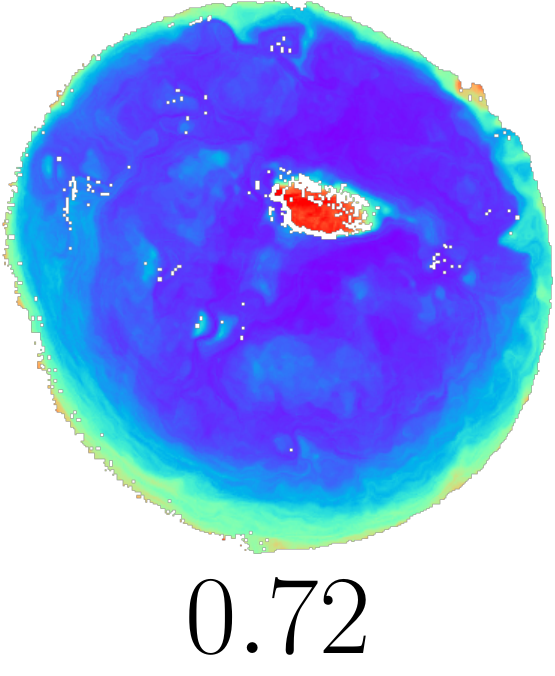}
                                                                                       & \includegraphics[width=0.1\textwidth]{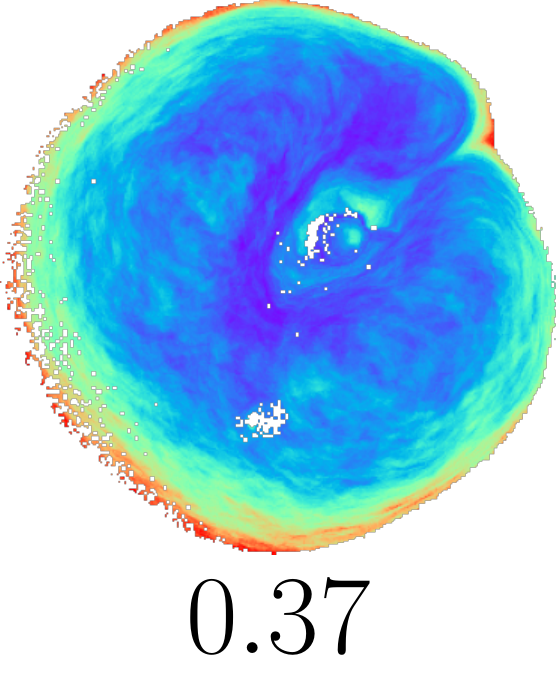}
                                                                                       & \includegraphics[width=0.16\textwidth]{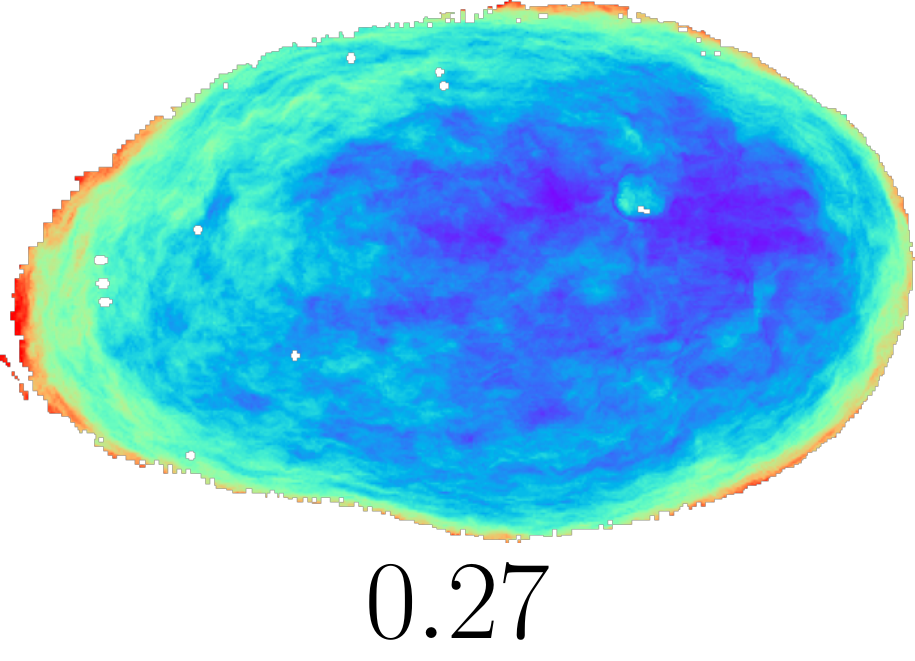}
                                                                                       & \includegraphics[width=0.16\textwidth]{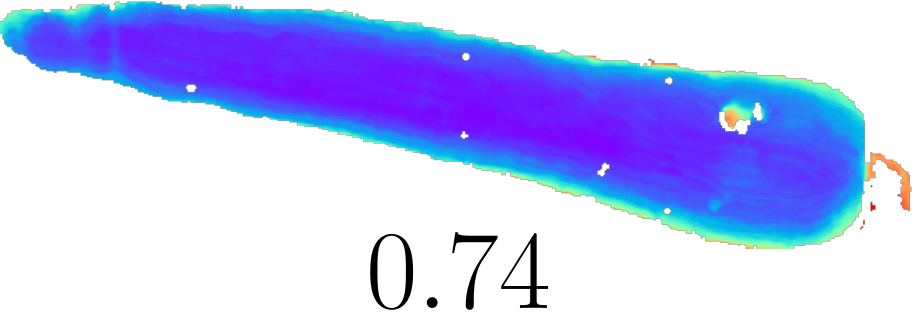} \\
                                      & \raisebox{0.06\textwidth}{$\frac{\delta}{20}$} & \includegraphics[width=0.1\textwidth]{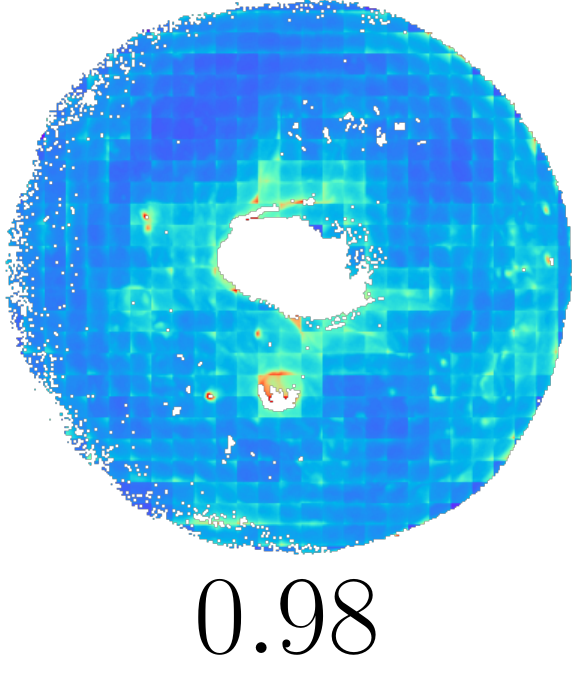}
                                                                                       & \includegraphics[width=0.1\textwidth]{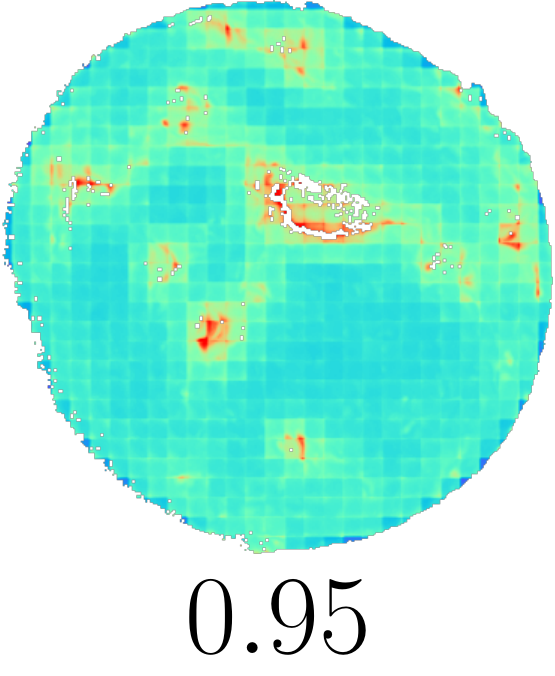}
                                                                                       & \includegraphics[width=0.1\textwidth]{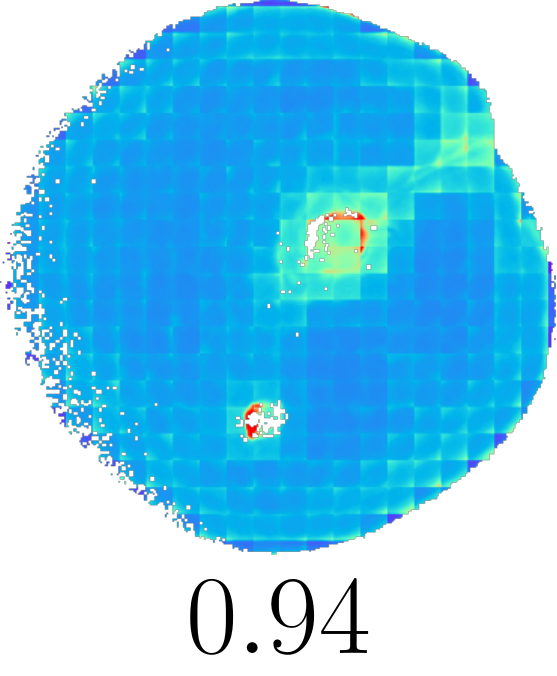}
                                                                                       & \includegraphics[width=0.16\textwidth]{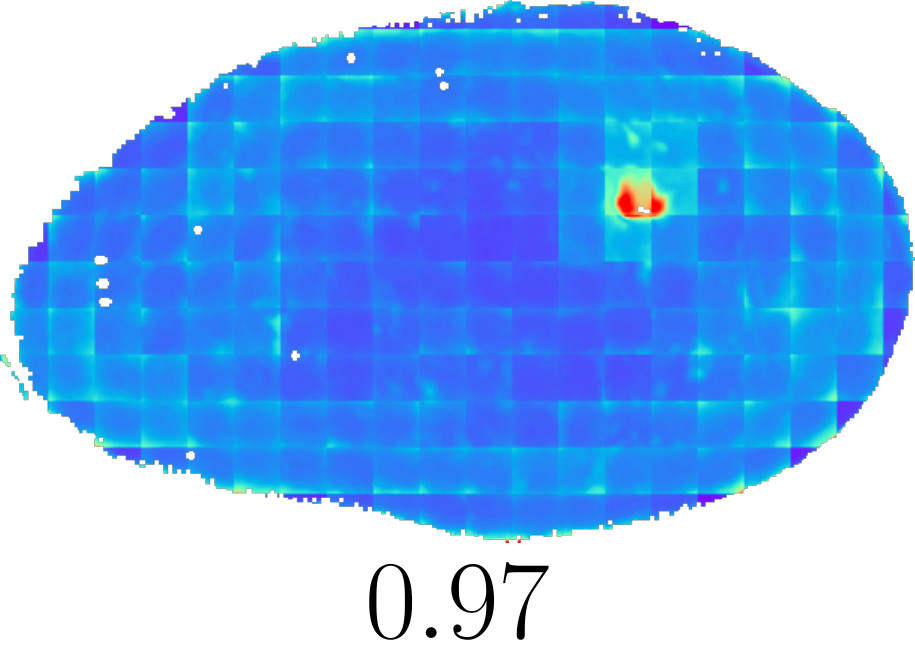}
                                                                                       & \includegraphics[width=0.16\textwidth]{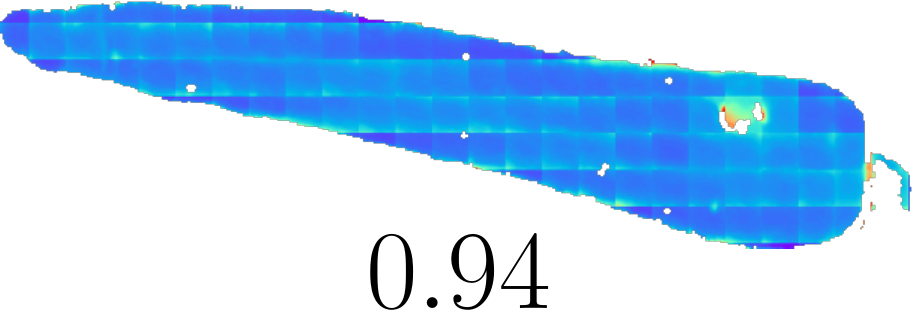} \\
                        \midrule
                        \multirow{2}{*}{Sphere}
                                      & \raisebox{0.06\textwidth}{$\delta$}            & \includegraphics[width=0.1\textwidth]{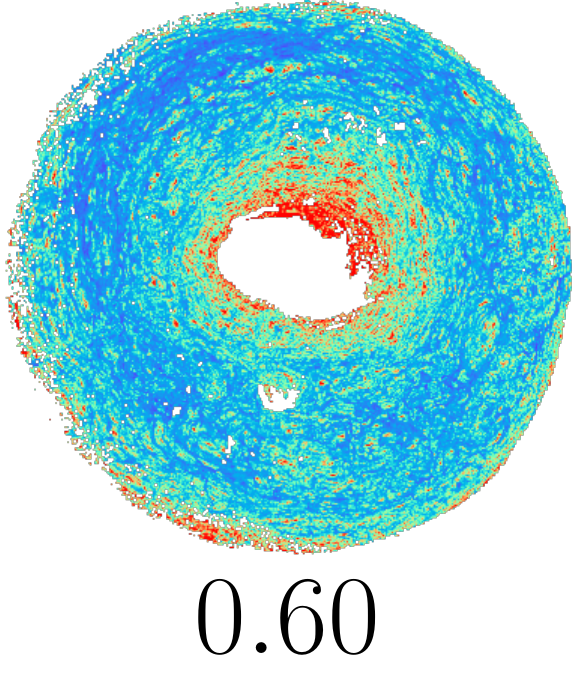}
                                                                                       & \includegraphics[width=0.1\textwidth]{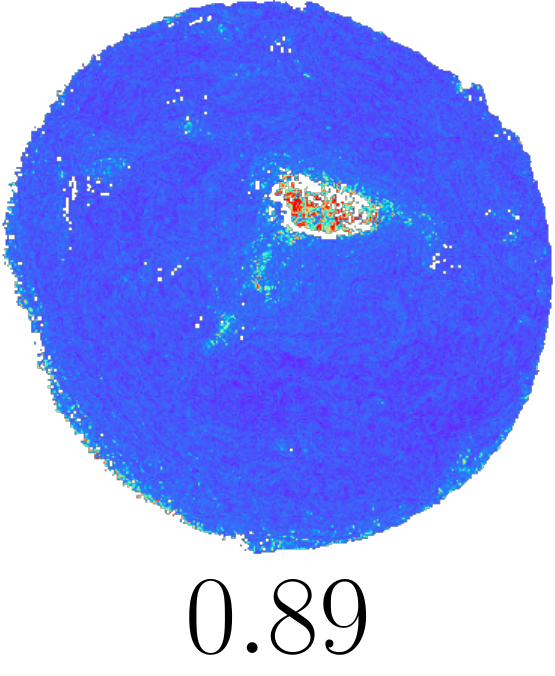}
                                                                                       & \includegraphics[width=0.1\textwidth]{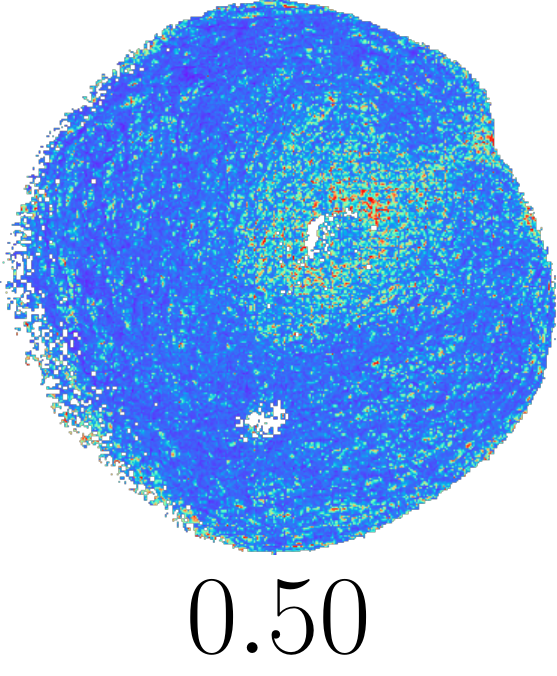}
                                                                                       & \includegraphics[width=0.16\textwidth]{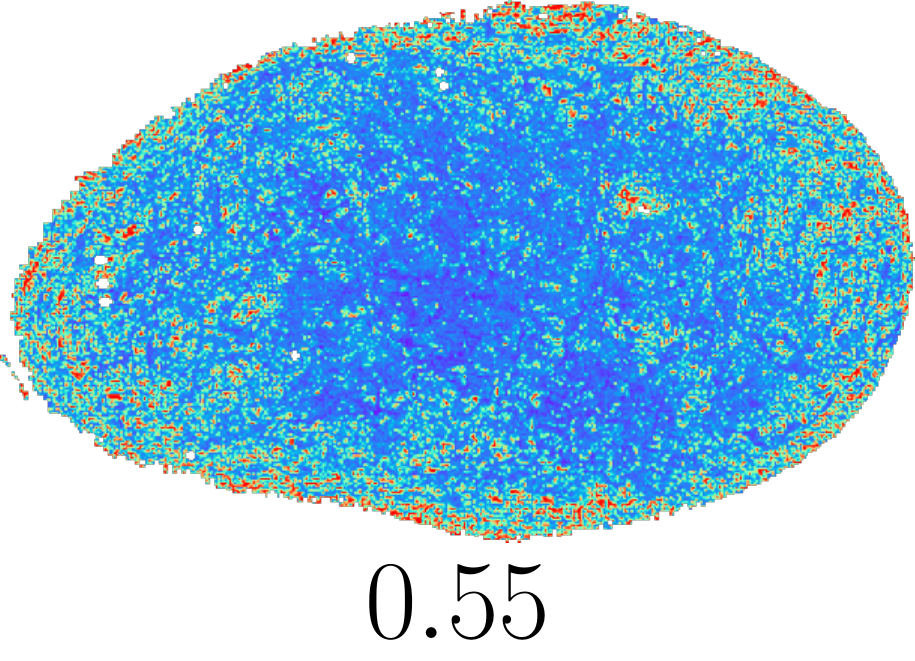}
                                                                                       & \includegraphics[width=0.16\textwidth]{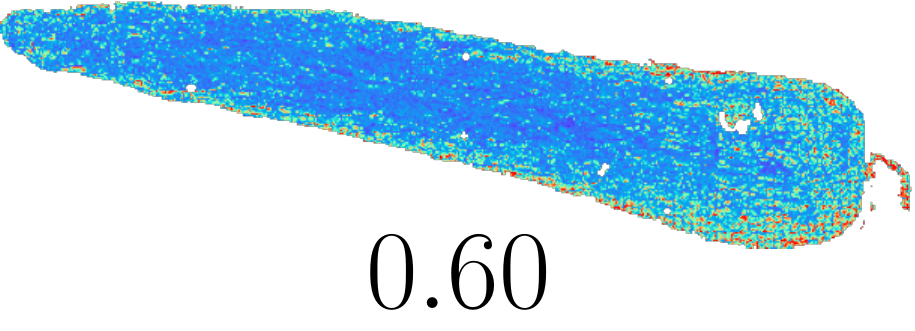} \\
                                      & \raisebox{0.06\textwidth}{$\frac{\delta}{20}$} & \includegraphics[width=0.1\textwidth]{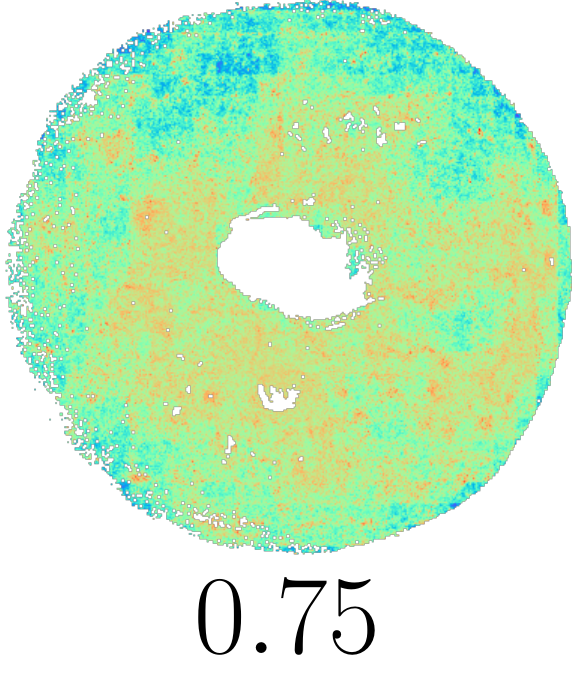}
                                                                                       & \includegraphics[width=0.1\textwidth]{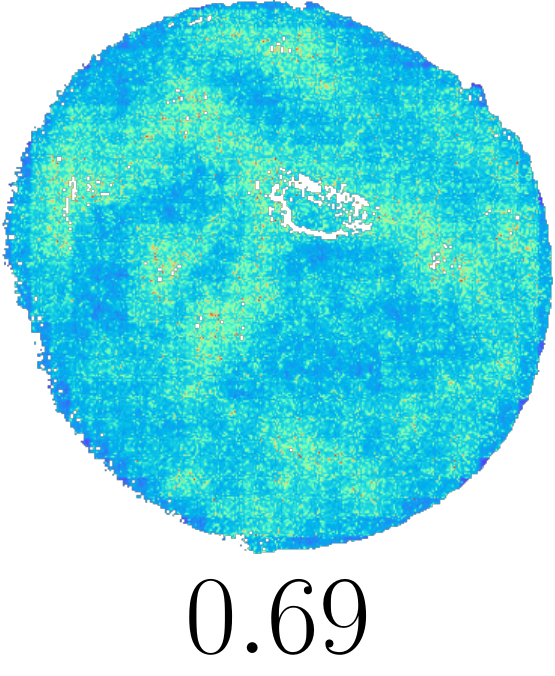}
                                                                                       & \includegraphics[width=0.1\textwidth]{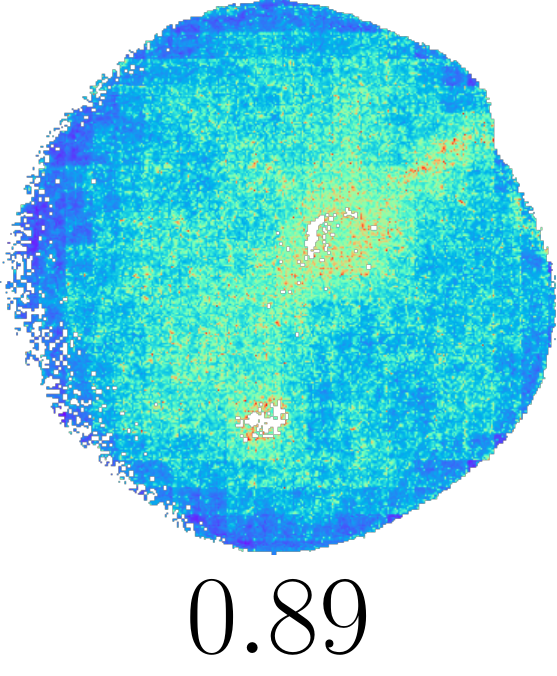}
                                                                                       & \includegraphics[width=0.16\textwidth]{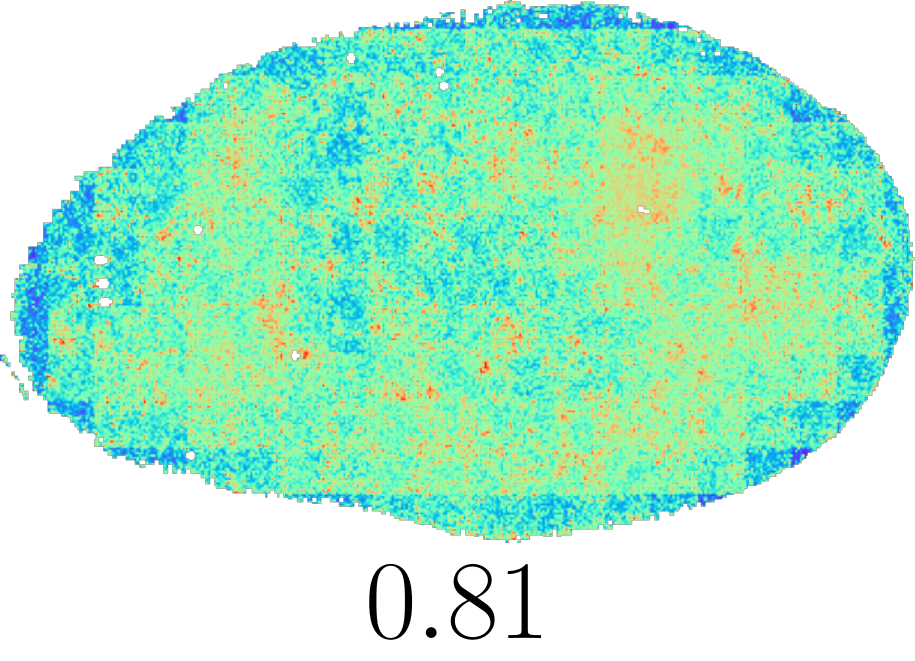}
                                                                                       & \includegraphics[width=0.16\textwidth]{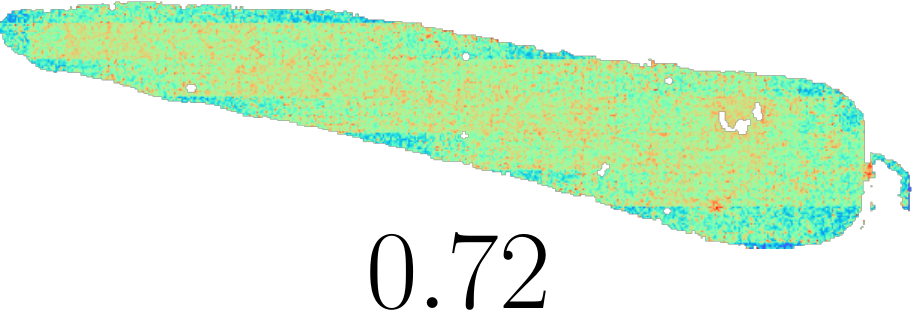} \\
                        \midrule
                        \multirow{2}{*}{Quadric}
                                      & \raisebox{0.06\textwidth}{$\delta$}            & \includegraphics[width=0.1\textwidth]{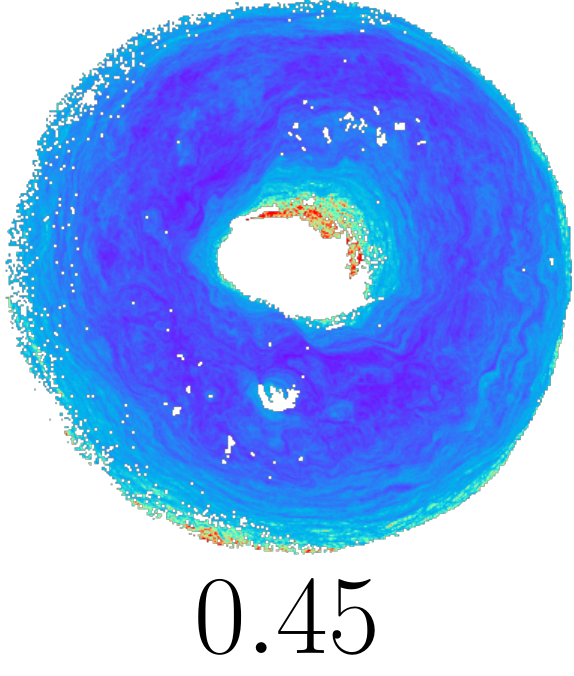}
                                                                                       & \includegraphics[width=0.1\textwidth]{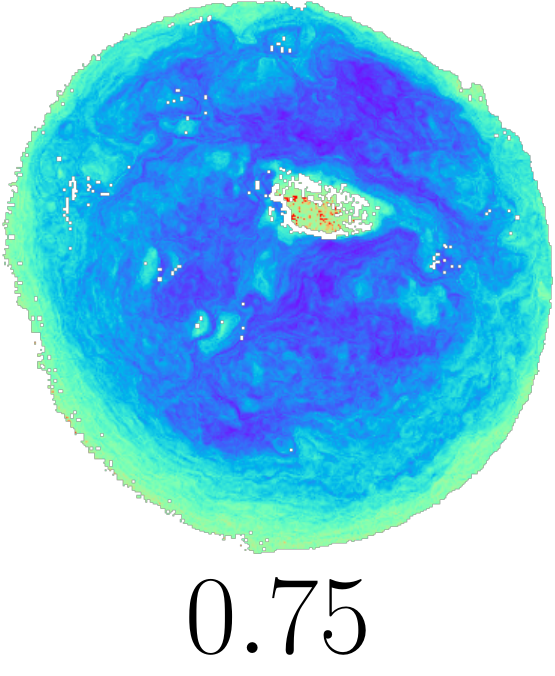}
                                                                                       & \includegraphics[width=0.1\textwidth]{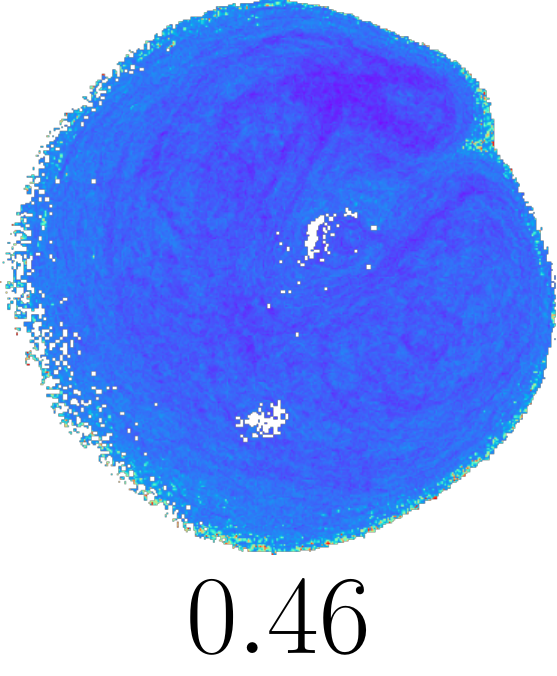}
                                                                                       & \includegraphics[width=0.16\textwidth]{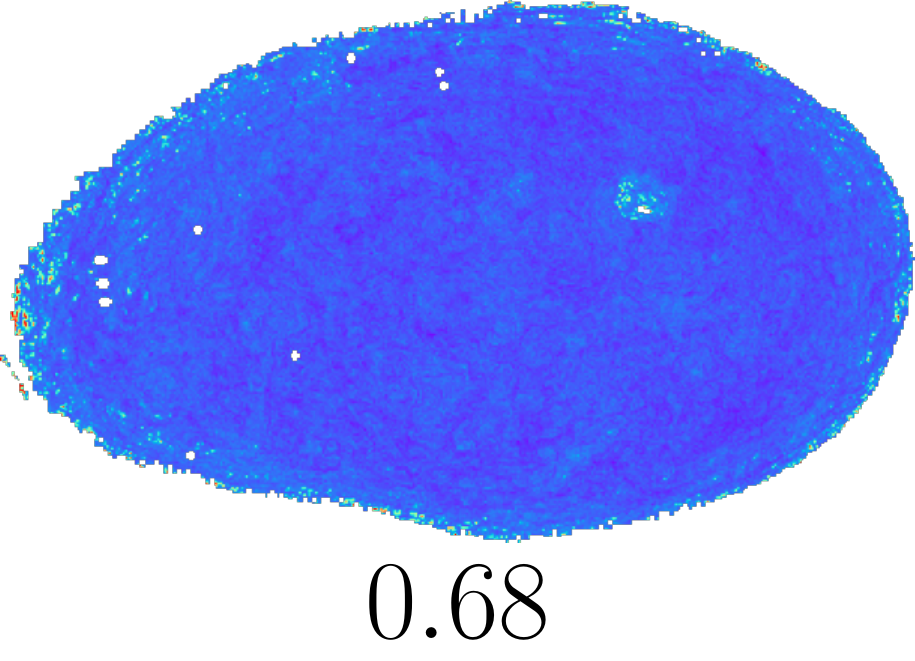}
                                                                                       & \includegraphics[width=0.16\textwidth]{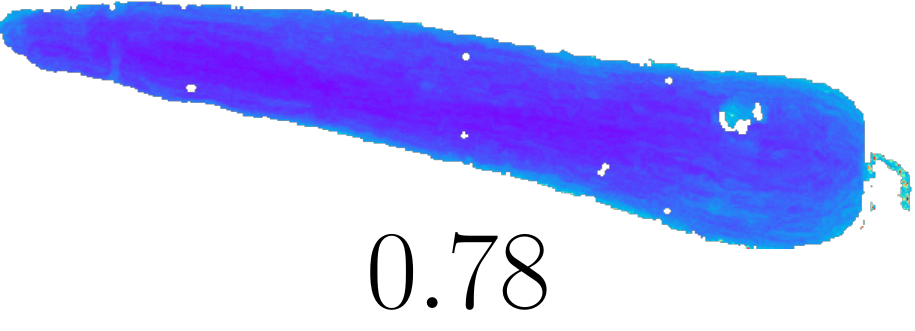} \\
                                      & \raisebox{0.06\textwidth}{$\frac{\delta}{20}$} & \includegraphics[width=0.1\textwidth]{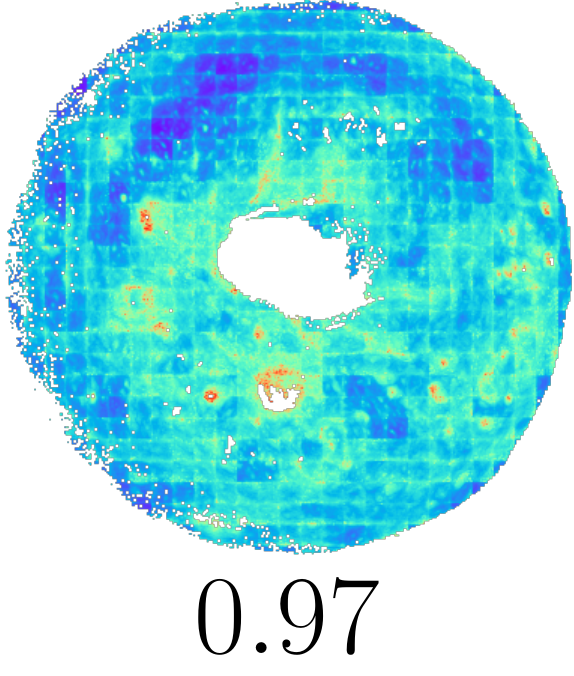}
                                                                                       & \includegraphics[width=0.1\textwidth]{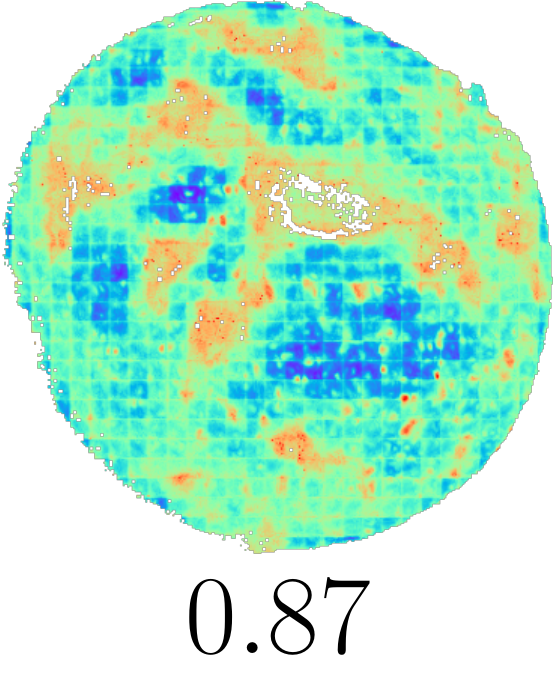}
                                                                                       & \includegraphics[width=0.1\textwidth]{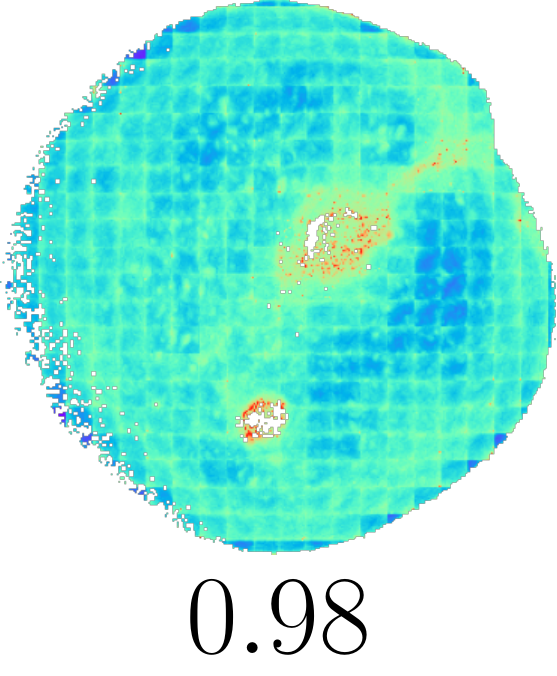}
                                                                                       & \includegraphics[width=0.16\textwidth]{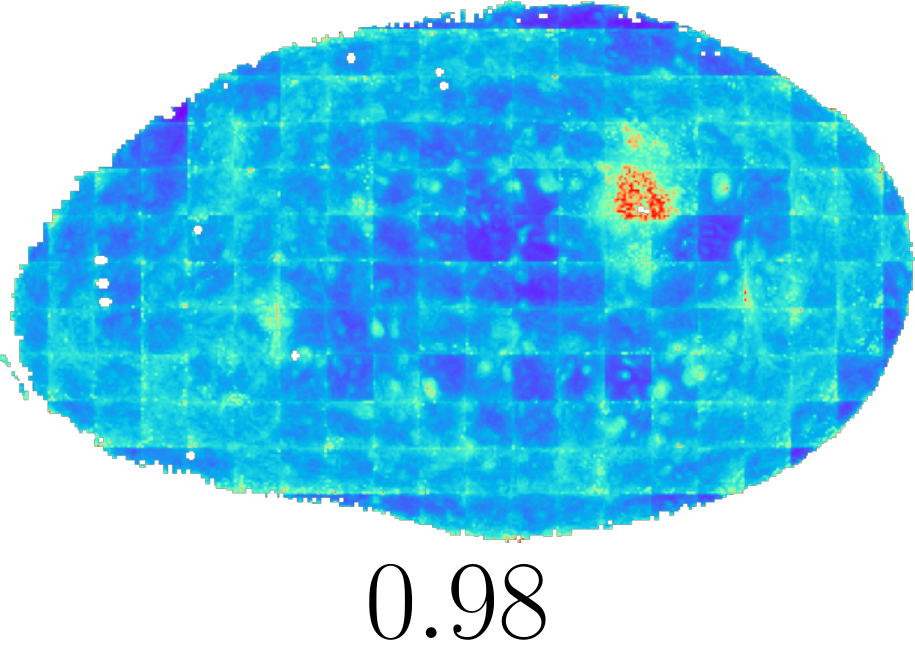}
                                                                                       & \includegraphics[width=0.16\textwidth]{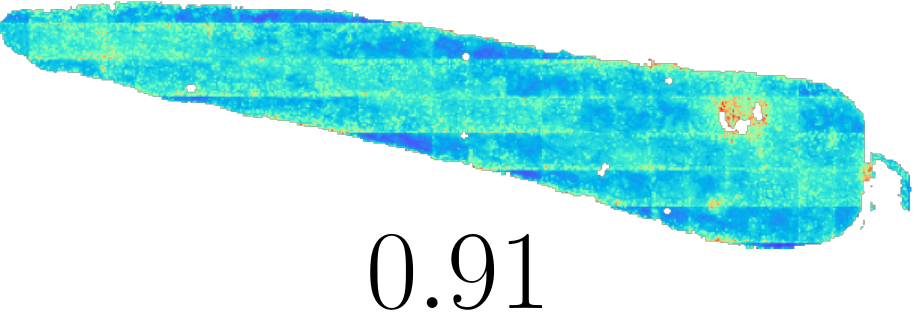} \\
                        \midrule
                        \midrule
                        \raisebox{0.05\textwidth}{\parbox{1cm}{\setlength{\baselineskip}{0.4cm}Ground\\truth}}
                                      &                                                & \includegraphics[width=0.1\textwidth]{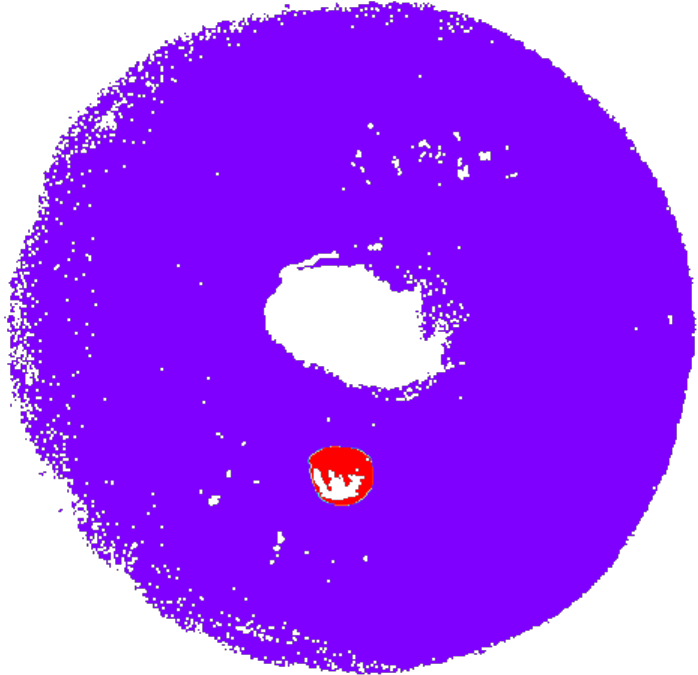}
                                                                                       & \includegraphics[width=0.1\textwidth]{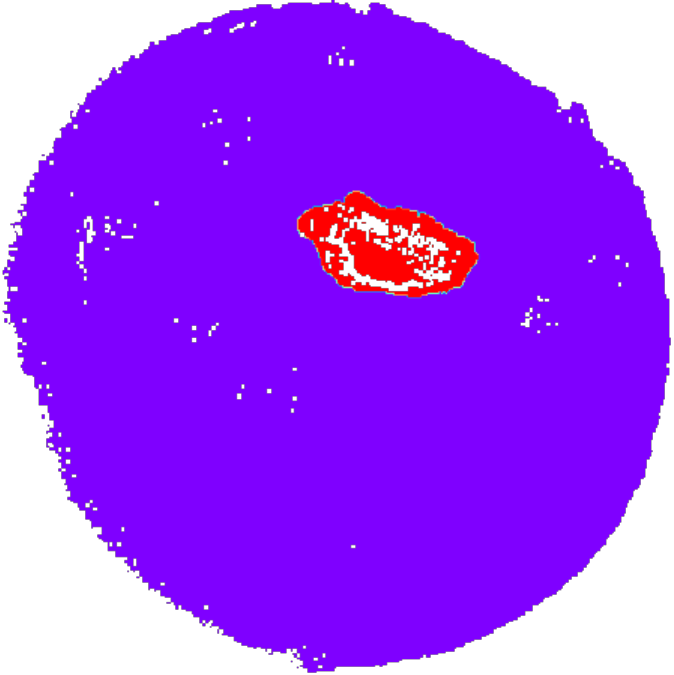}
                                                                                       & \includegraphics[width=0.1\textwidth]{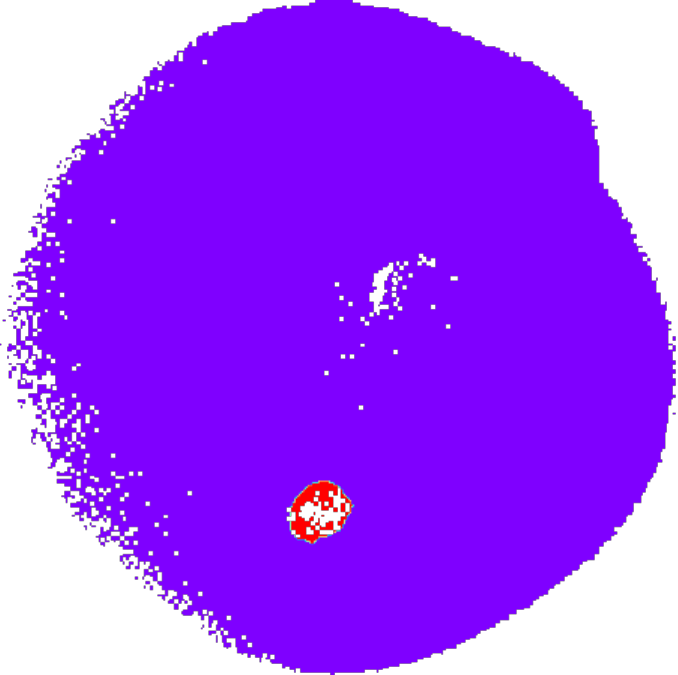}
                                                                                       & \includegraphics[width=0.16\textwidth]{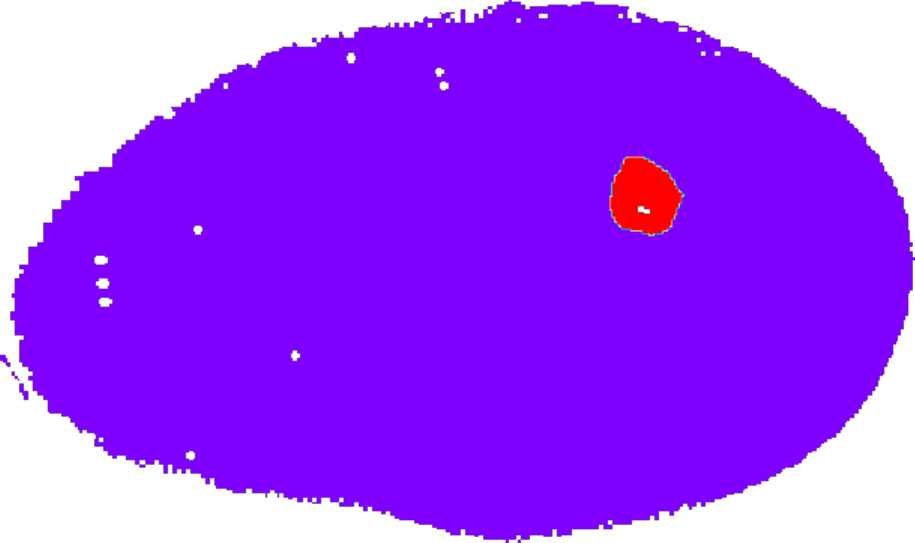}
                                                                                       & \includegraphics[width=0.16\textwidth]{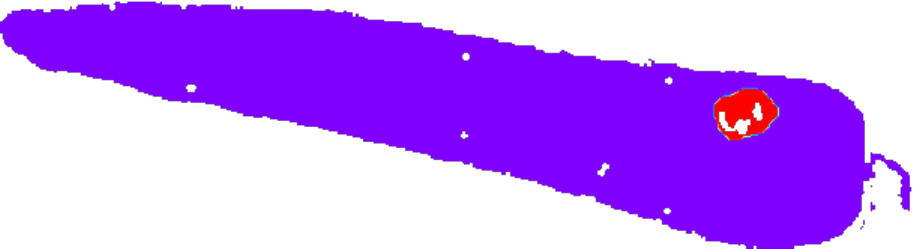} \\
                    \end{tabular}
                    }
                    \label{tab:sliding_pif_qualitative}
                \end{subtable}
                \hfill
                \begin{subtable}[b]{.275\textwidth}
                    \centering
                    \caption{}
                    \vspace{.9cm}
                    \begin{tabular}{l}
                        \includegraphics[width=\textwidth]{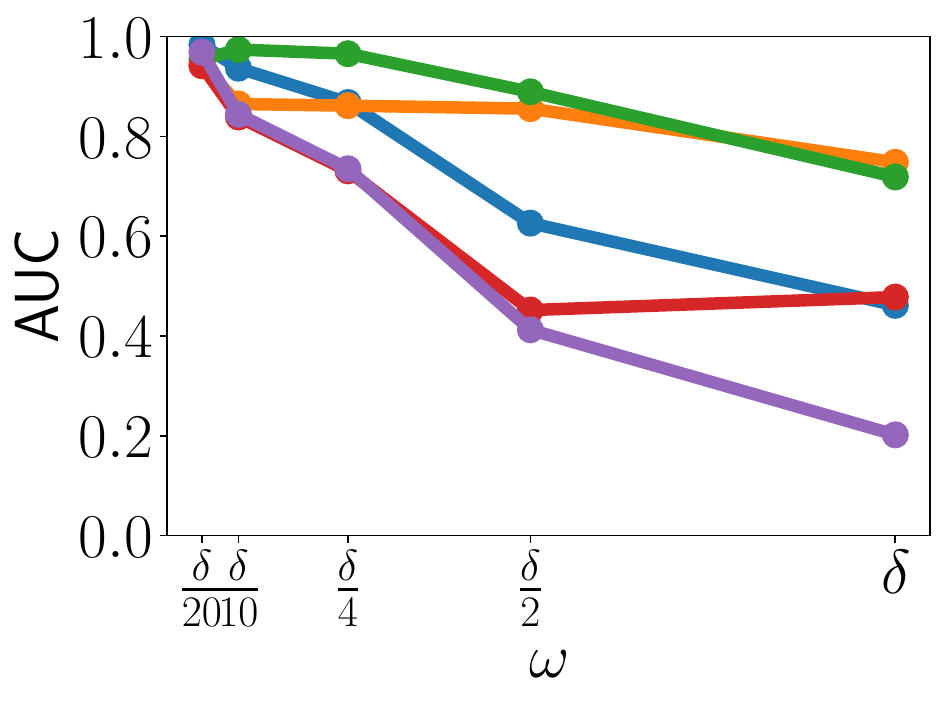}     \\
                        \vspace{.1cm}
                        \includegraphics[width=\textwidth]{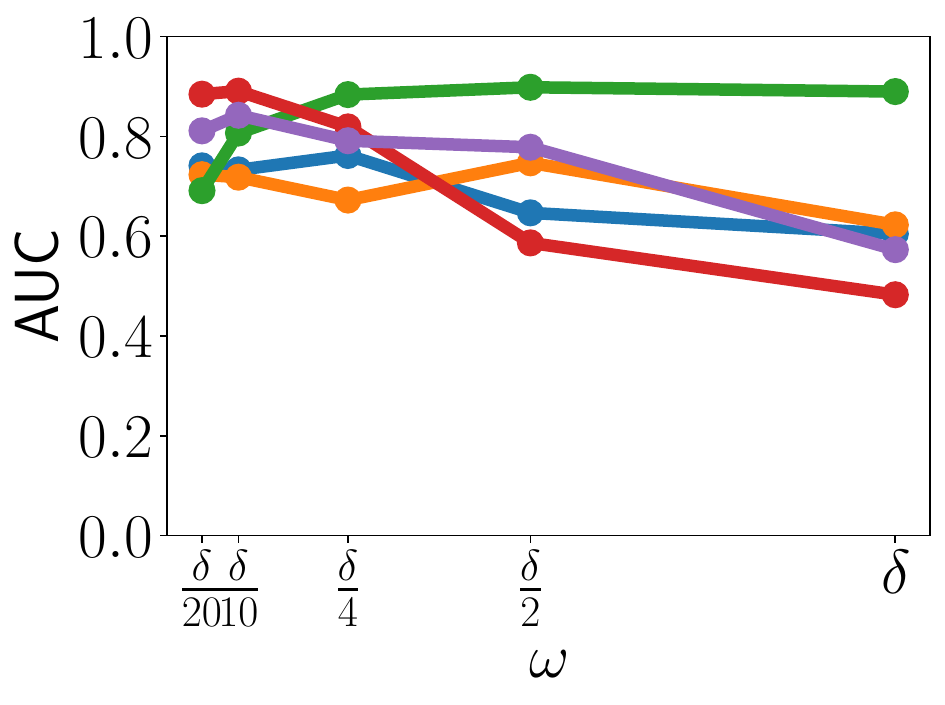}   \\
                        \vspace{.1cm}
                        \includegraphics[width=\textwidth]{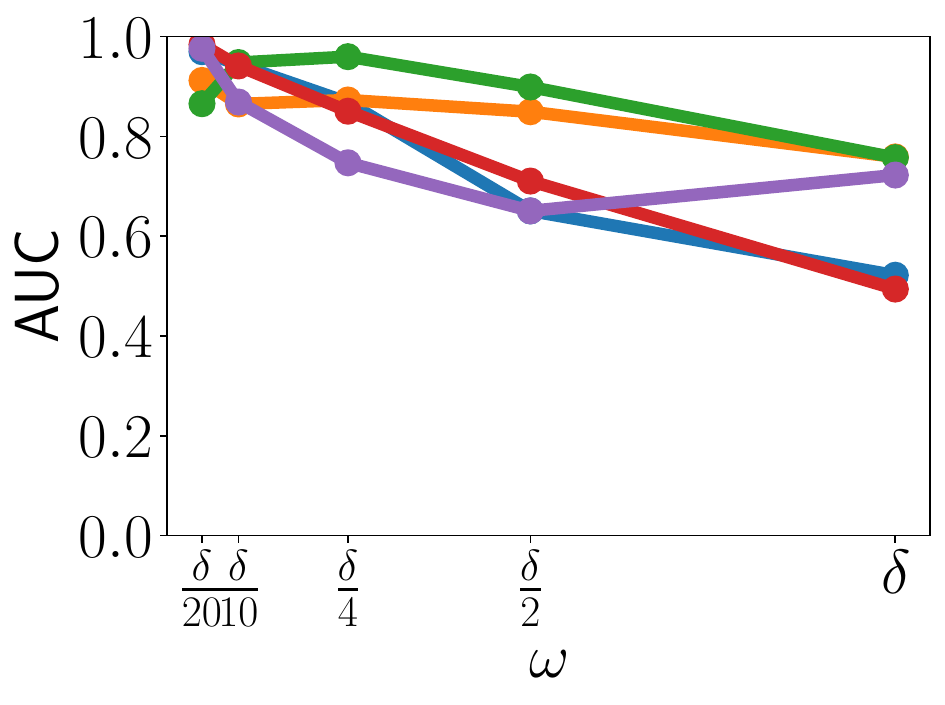} \\
                        \vspace{.1cm}
                        \parbox{\textwidth}{\vspace{-0.3cm} \includegraphics[width=1.2\textwidth]{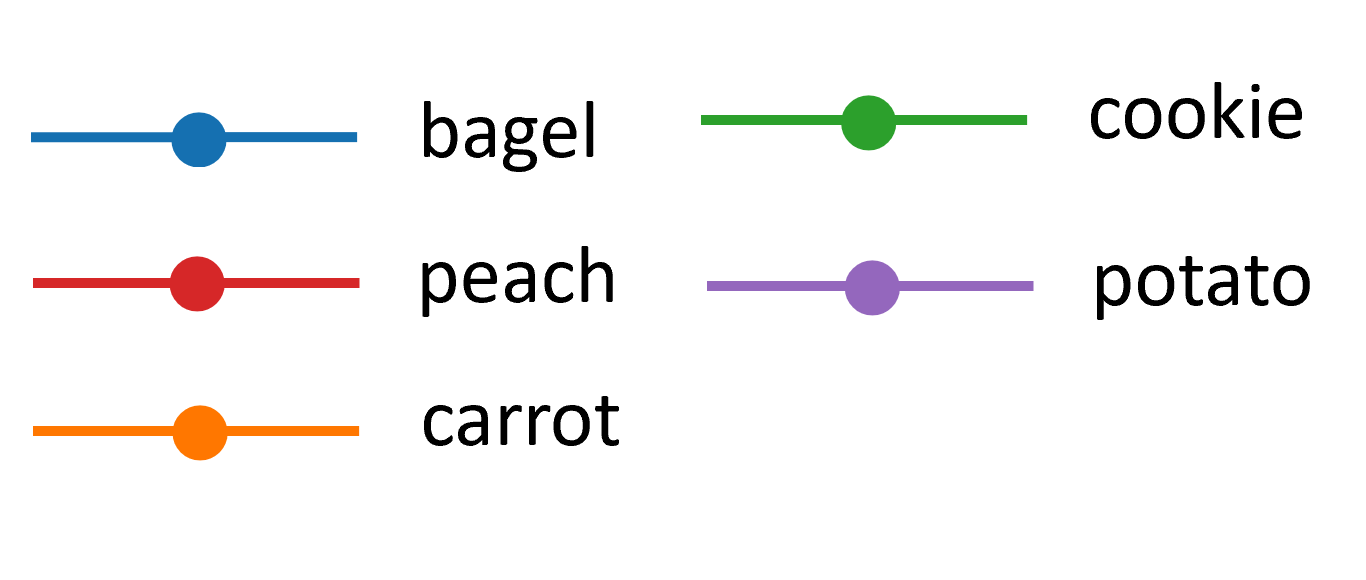}}                         \\
                        \vspace{-1.3cm}
                    \end{tabular}
                    \label{tab:sliding_pif_quantitative}
                \end{subtable}
                }
                \label{tab:sliding_pif}
            \end{table}
                
            As expected, \spif performs generally better with small $\omega$ (\cref{tab:sliding_pif_quantitative}), as local models are more accurate for smaller regions, while large $\omega$ do not satisfy the locality principle.
            An exception is cookie, which has a large anomalous area. In fact, \spif performs well when genuine data are the majority in $W$, thus the best $\omega$ for cookie is the second or third-smallest value for planes and quadrics. Spheres, on the contrary, perform best when $\omega = \delta$, since they describe accurately the entire cookie, but worst for small windows as they tend to be biased towards the center of $W$, struggling to describe borders.
            Potato has a more ellipsoidal shape, and spheres struggle to describe it for large $\omega$. With medium-sized windows, spheres perform best, showing that locally they best capture potato shape. The higher flexibility of quadrics leads to high variance in the models fitted, resulting in poor performance for medium-sized $\omega$.
            Carrots are elongated and difficult to approximate with spheres, but they are well described by quadrics and planes, since the local curvature is not very pronounced.
            Bagels and peaches are difficult for all the model families when $\omega$ is large due to their toroidal shape. However, as $\omega$ decreases, all families perform better, since the toroidal shape no longer prevails in a small neighborhood. All in all, this experiment demonstrates that when the prior on the anomalous region size $\omega$ is combined with an appropriate local prior on $\mathcal{F}$, \spif is a convenient framework to perform structure-based AD and, unlike \pif, it is a viable option for settings with a large number of points.

    \section{Conclusions}
        \label{sec:sbad_conclusions}

        In this chapter, we introduced \pif, a novel framework for detecting anomalous samples that deviate from structured patterns by means of \pembedding and \pisolation. We proposed two approaches for \pisolation: \emph{i)} \viforest, which leverages Voronoi tessellations and can be employed in any metric space, and \emph{ii)} \rzhiforest, specifically designed for \pspace, which leverages a novel \lsh criterion, \rzhash, as the splitting mechanism.

        Our empirical evaluation demonstrated that \pembedding enhances the separability between structured and unstructured data, leading to superior performance than simply performing anomaly detection directly in the original space. Notably, the proposed \pif framework outperforms all the alternatives where anomaly detection methods are straightforwardly plugged in the \pspace. Furthermore, our experiments showed that \rzhiforest not only achieves greater efficiency than \viforest, but also has stable accuracy across various branching factors, on both synthetic and real datasets.
        
        We then relaxed the assumption that genuine data must be \emph{globally} described by a parametric model. To address this, we proposed \spif, an efficient sliding window approach that leverages the locality principle to \emph{locally} approximate the manifold underlying genuine data. Experimental analysis suggests that \spif is a promising solution for real-world scenarios where the nature of the underlying manifold is typically unknown.
        
        We believe several research directions are worth exploring, particularly in employing more general and flexible model families for \pembedding. By allowing \pif to adapt to different underlying manifolds and accommodate heterogeneous generative processes, we can extend its applicability to a wide range of anomaly detection problems where some form of prior knowledge is available. The virtually unlimited variety of models that can be incorporated into the \pembedding step makes \pif an extremely adaptable anomaly detection framework. For example, in textured image analysis -- such as detecting defects in carpets, tiles or leather surfaces -- we can model local regions using subspaces. By randomly sampling small patches of the image and estimating their underlying subspaces, we can assess the reconstruction error of each region of the image to identify anomalies, such as defects disrupting the periodic texture pattern. This approach extends naturally to other periodic domains, including ECG signals and time-series data, where anomalies correspond to deviations from expected repeating patterns. To push the flexibility of \pembedding even further, we can even employ density distributions, or extremely general non-parametric families, such as clustering schemes, autoencoders or supervised regression and classification models, to address virtually any type of anomaly detection problem that can be framed in our settings.
        
        Finally, given \viforest's ability to operate with arbitrary distance metrics, it would be interesting to explore more its application as a standalone anomaly detection tool beyond \pspace, or in settings where alternative embeddings and distance measures are more appropriate. Notably, \viforest does not inherently require an explicit distance function to operate, since a distance matrix alone is sufficient for constructing a Voronoi tessellation. This means that \viforest can leverage distances derived from kernel estimations in random forests or similarities learned through neural networks, characterizing \viforest as a highly versatile anomaly detection tool that can be employed in any scenario where pairwise instance similarities can be estimated, significantly expanding its range of potential applications.

        \chapter{Structure-based Clustering}
    \label{cha:structure_based_clustering}

    In this chapter, we present \mlink, a novel structure-based clustering algorithm that efficiently and effectively recovers genuine structures present in the data. Notably, \mlink is the first among preference-based algorithms that is both robust to outliers and capable of handling general (\emph{i.e.}, non-nested) model families.
    \mlink combines \emph{\pembedding} with \emph{\pclustering}, an agglomerative linkage scheme that integrates on-the-fly sampling with a model-selection criterion to determine when structures can be effectively merged and which model, among multiple families, is the most appropriate. The \pclustering procedure reduces the severe dependencies on initial models sampling of \pembedding, leading to greater stability, and it eliminates the need to update distances at every iteration, ensuring low computational complexity.

    In~\cref{sec:sbc_problem_formulation}, we formalize the problem of Structure-based Clustering, and in~\cref{sec:sbc_related_literature} we review the literature on single and multi-family structure recovery. In~\cref{sec:multilink}, we introduce the \mlink algorithm, composed of \pembedding (\cref{subsec:ml_preference_embedding}) and \pclustering (\cref{subsec:preference_clustering}). Specifically, in~\cref{subsec:preference_clustering} we detail the algorithm by explaining the agglomerative linkage procedure, describing the on-the-fly sampling and model selection criterion via \gric cost, and highlighting the benefits of \mlink. Finally, in~\cref{sec:ml_experiments} we illustrate and discuss the experimental evaluation of \mlink.

    \section{Problem Formulation}
        \label{sec:sbc_problem_formulation}

        \begin{figure}[t]
            \centering
            \includegraphics[width=\linewidth]{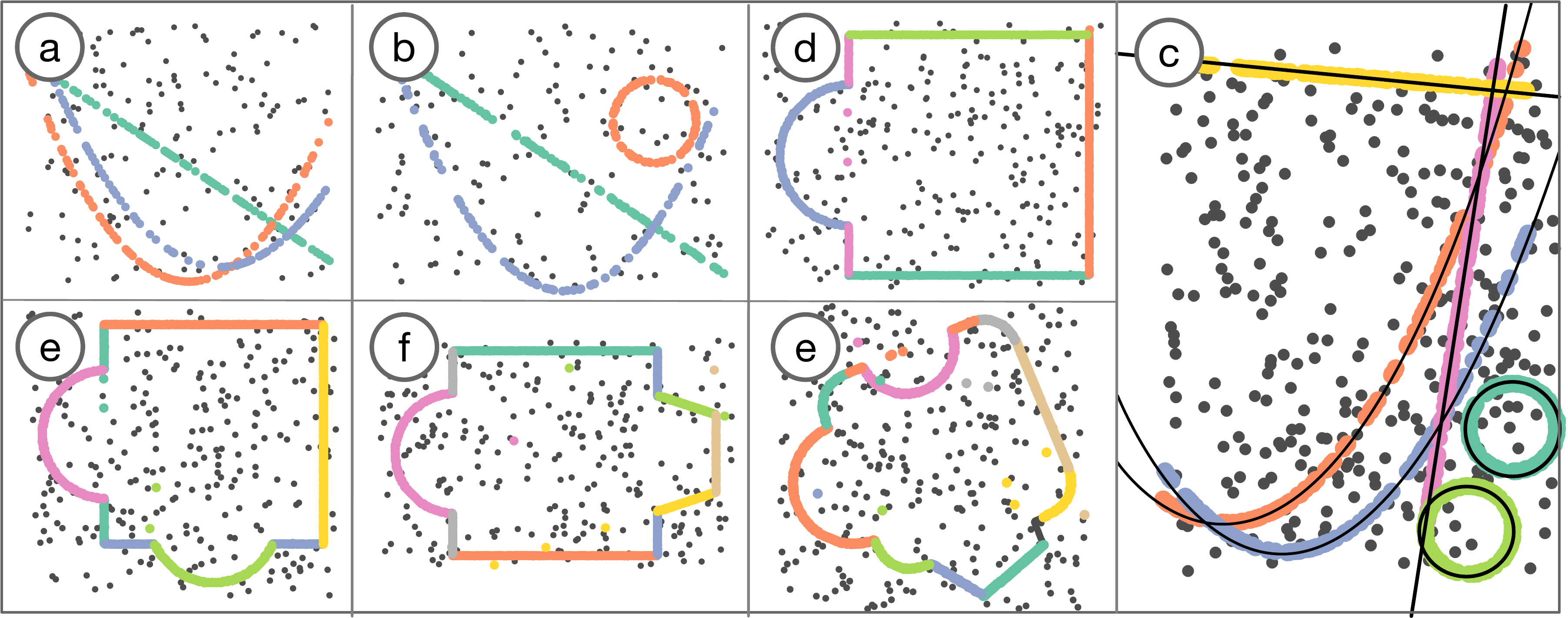}
            \caption{Worst results achieved by \mlink on conic fitting.}
            \label{fig:exp2d}
        \end{figure}

        We consider a dataset $X = G \cup A \subset \mathcal{X}$, where $G$ is the set of genuine data, $A$ represents anomalous data, and $\mathcal{X}$ is the ambient space. Unlike Structure-based Anomaly Detection (\cref{sec:sbad_problem_formulation}), our goal here is not only to detect anomalies $A$, but also to partition the genuine data $G$ into clusters.
        We assume that the genuine data $G$ can be partitioned into a union of structures, \emph{i.e.}, $G = U_1 \cup \dots \cup U_s$, where each $U_i$ is described by a parametric model $\vect{\theta}_i$ from one of the model families $\mathcal{F}_1, \dots, \mathcal{F}_k$ considered.
        Specifically, we assume that each $\vect{u} \in U_i$ is close to the solutions of a parametric equation $\mathcal{F}_i(\vect{u}, \vect{\theta}_i) = 0$, which depends on an unknown parameter vector $\vect{\theta}_i$. For instance, in~\cref{fig:exp2d} the considered model families include lines $\mathcal{F}_l$, parabolas $\mathcal{F}_p$ and circles $\mathcal{F}_c$. Due to noise, data belonging to these structures only satisfy the corresponding equations within a tolerance $\epsilon > 0$. The nature and amount of noise are assumed to be unknown. Conversely, anomalous data $\vect{a} \in A$ do not fit well the parametric equations describing the genuine data, therefore $|\mathcal{F}_i(\vect{a}, \vect{\theta}_i)| \gg \epsilon \;\; \forall i$.

        The goal of Structure-based Clustering is to produce a partition $X = U_1 \cup \dots \cup U_{\widehat{s}}$ of the input data such that the genuine and anomalous sets form disjoint groups, \emph{i.e.}, $G = U_1 \cup \dots \cup U_s$ and $A = U_{s+1} \cup \dots \cup U_{\widehat{s}}$. For each part $U_i$, it must also be possible to determine whether it belongs to $G$ or $A$, allowing for the rejection of outliers.

    \section{Related Literature}
        \label{sec:sbc_related_literature}

        While single-family structure recovery is a well established topic in Computer Vision that has attracted a lot of interest, Multi-family structure recovery in its full generality has been addressed only recently~\cite{BarathMatas18,BarathMatas19,MagriFusiello19,XuCheongAl19}, with a few earlier work that have addressed this problem only for specific applications~\cite{StrickerLeonardis95,Torr99,SugayaKanatani04,SchindlerSuterAl08}. Here we survey those methods that are most relevant for the proposed solution, while in~\cref{subsec:ml_preference_embedding} we recall some aspects of preference-based methods that are important to understand \mlink.  
        
        Both single and multi-family recovery methods can be generally divided in two main approaches: \emph{optimization-based} and \emph{preference-based}.
        Optimization-based methods were initially developed to handle the scenario where a single family $\mathcal{F}$ of models~\cite{YuChinAl11,IsackBoykov12} is present. These approaches work by minimizing an objective function that typically combines a data fidelity term, which assesses the goodness of fit, and a penalty term that accounts for model complexity. Additional terms may be added to promote spatial coherence or to incorporate specific applications related priors~\cite{PhamChinAl14}.
        These methods usually follow a two-steps \emph{hypothesize-and-verify} process. In the hypothesize step, a set of models $\{\vect{\theta}_i\}_{i = 1, \dots, m}$ is estimated via random sampling of points. During the verification step, the models that minimize the energy function are selected.
        Various techniques have been proposed, depending on how the energy function is defined. These techniques range from early approaches~\cite{Torr98} to more advanced methods that rely on graph labeling~\cite{PhamChinAl14*1}, alpha-expansion~\cite{DelongVekslerAl12, BarathMatasAl16}, convex relaxation~\cite{AmayoPinesAl18} and integer linear programming~\cite{Li07, SchindlerSuterAl08, MagriFusiello16}. 
        Particular relevant to our work is~\cite{DelongVekslerAl12}, where model-refitting is used to escape form local minima and improve convergence of energy minimization.
        
        Very recently, optimization-based methods addressed the  challenges of fitting instances from multiple families: \multix~\cite{BarathMatas18} combines alpha-expansion with a mean-shift step carried on in each model family, and \progx~\cite{BarathMatas19} further improves this approach by interleaving the hypothesize and the verify stages.
        All these methods can be considered as sophisticated implementations of a (multi-)model selection criteria, as they select the simplest models using as measure of simplicity a global energy function.
        
        Preference-based methods represent the second mainstream approach and our work falls in this category. In contrast with optimization-based methods that concentrate on models, preference-based ones put the emphasis on structures, and cast multi-structure recovery as a clustering problem, following an \emph{hypothesize-and-clusterize} scheme. During the first step, tentative models are randomly sampled and points are embedded in a \pspace $\mathcal{P}$ based on the preferences granted by the hypothesized models $\{\vect{\theta}_i\}_{i = 1, \dots, m}$. Then data are clustered in structures by inspecting the distribution of residuals with respect to the hypothesized candidate models.
        A wide variety of techniques have been proposed for segmenting preferences, including hierarchical schemes~\cite{ToldoFusiello08,ToldoFusiello13} such as \tlinkage~\cite{MagriFusiello14}, as well as \kernelfitting~\cite{ChinWangAl09}, \rbf~\cite{MagriFusiello15,TepperSapiro17}, \biclustering~\cite{TepperSapiro14,DenittoMagriAl16,BicegoFigueiredo18}, \hoc~\cite{AgarwalJongowooAl05,Govindu05,ZassShashua05,JainGovindu13} and \hp~\cite{PurkaitChinAl17,WangXiaoAl15,XiaoWangAl16,WangXiaoAl18,LinXiaoAl19}. In this work we build upon hierarchical clustering, that is robust to outliers and, in contrast to divisive alternatives, does not need to know the number of structures in advance.
        
        The preference-based approach has only recently been employed to tackle multi-family problems. One example is \multicascadedtlinkage (\mct)~\cite{MagriFusiello19}, which operates under the assumption that model families are nested, executing \tlinkage in a hierarchical manner, progressing from the most general to the simplest family.
        Then, the model selection tool~\cite{Torr99} \gric (Geometric Robust Information Criterion) is used to compare clusters derived from the general family with their corresponding nested clusters derived from simpler structures.
        However, \mct is not suited for models belonging to families that are not strictly nested within each other.
        Another example is the motion segmentation algorithm presented in~\cite{XuCheongAl18}, which can be seen as a multi-family preference-based method as well. This method focused on handling nearly-degenerate structures, which are challenging to describe accurately for real-world data. To overcome this limitation, instead of dealing with elusive model selection problems, the authors fit models from various families and combine the resulting partitions through an ad-hoc multi-view spectral clustering algorithm. Unfortunately, this method struggles when data are contaminated by outliers.
        
        It is also noteworthy that solutions for structure recovery based on deep-learning are now emerging. For instance,~\cite{XuCheongAl19*1} addresses the multi-family multi-model fitting problem by learning an embedding of the points from labeled data, which are subsequently segmented using \kmeans.
        
        In contrast, \mlink follows a different approach by combining the strengths of both optimization and preference-based methods, incorporating the clarity of model selection techniques with the flexibility of clustering.
        Specifically, we extend the preference representation to \emph{jointly} deal with multiple, potentially non-nested model families.

    \section{\mlink}
        \label{sec:multilink}

        \begin{figure}[t]
            \centering
            \hspace{-0.5cm}
            \includegraphics[width=\linewidth]{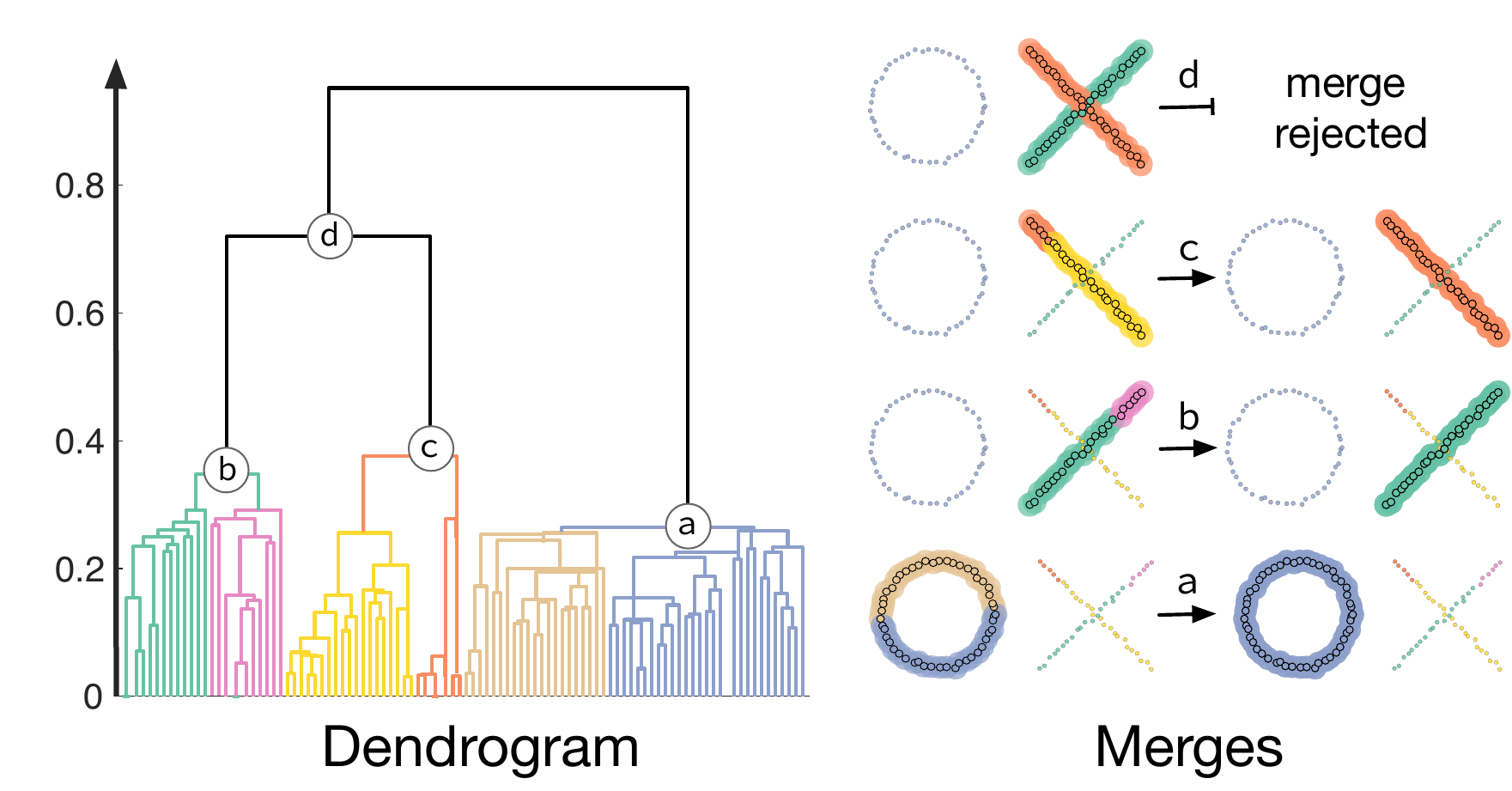}
            \caption{\mlink combines single-linkage clustering and \gric. Clusters are merged as long as the \gric score improves when fitting suitable models on-the-fly. Colors indicate how cluster aggregation proceeds in the dendrogram.}
            \label{fig:mlink}
        \end{figure}
        
        Here, we illustrate the core principles of \mlink through a structure recovery example (\cref{fig:mlink}).
        At a high level, \mlink adopts a \emph{hypothesize-and-clusterize} framework with two key distinctions compared to existing approaches.
        First, \pembedding is performed by sampling hypotheses from multiple families, in our example $\mathcal{F}_l \cup \mathcal{F}_c$, the families of lines and circles.
        Second, \pclustering is performed in the \pspace using single-linkage (see dendrogram in~\cref{fig:mlink}). The main novelty of \mlink consists in determining whether the closest clusters pairs should be merged by selecting the model family that best describes their union.
        This decision is made by fitting new models \emph{on-the-fly} to better describe the points in both clusters and then using a model selection criterion to decide whether to merge them or not.
        Specifically, we employ \gric to determine the best interpretation of the data, balancing both data fidelity and model complexity. Clusters are merged when the model fitted on their union produces a \gric score lower than the sum of their individual \gric scores. In the example shown in~\cref{fig:mlink}, clusters are first merged in \texttt{a} in a circle, then in \texttt{b} and \texttt{c} in a line, while in \texttt{d} the merge is rejected as no model can describe the two segments better than two separate lines.

        \subsection{\pembedding}
            \label{subsec:ml_preference_embedding}

            \pembedding is a mapping from the ambient space $\mathcal{X}$ to the \pspace $\mathcal{P}$, and it was used for single family of models $\mathcal{F}$.
            Let $X$ be the input data and $\epsilon > 0$ a fixed inlier threshold, the preference function of $\vect{x} \in \mathcal{X}$ with respect to a  model $\vect{\theta}_i$ is:
            \begin{equation}
                \label{eq:ml_preference}
                p_i =
                \begin{cases}
                    \phi(\delta_i) &\text{if $|\delta_i| \leq \epsilon$}\\
                    0              &\text{otherwise}
                \end{cases},
            \end{equation}
            where $\delta_i = \mathcal{F}(\vect{x}, \vect{\theta}_i)$ measures the residual between model $\vect{\theta}_i$ and point $\vect{x}$, and $\phi$ is a monotonic decreasing function in $[0,1]$ such that $\phi(0) = 1$. Intuitively, $p_i$ represents the preference that a point $\vect{x}$ grants to a model $\vect{\theta}_i$: the lower $\delta_i$, the higher the preference.
            
            In practice, a finite pool $\varTheta = \{\vect{\theta}_i\}_{i = 1, \dots, m}$ of $m$ model hypotheses is randomly sampled and used to compute the preferences as in~\eqref{eq:ml_preference}, defining an embedding $P \subset \mathcal{P} = [0,1]^m$ of data points. The rationale behind \pembedding is that points belonging to the same structure share similar preferences, thus are nearby. Several metrics have been proposed to measure distance/similarity in the \pspace, \emph{e.g.}, Ordered Residual Kernel~\cite{ChinWangAl09}, Jaccard~\cite{ToldoFusiello08} and Tanimoto~\cite{MagriFusiello14} distance.
            In this work we rely on the Tanimoto distance defined as in~\eqref{eq:tanimoto_vect}.
            
        \subsection{Preference Clustering}
            \label{subsec:preference_clustering}

            \mlink is summarized in~\cref{alg:mlink} and starts by randomly sampling a finite pool of tentative models $\mathcal{H} = \varTheta_1 \cup \dots \cup \varTheta_k$ where each set $\varTheta_i$ comes from its corresponding model family $\mathcal{F}_i$ (\cref{line:sample_hypotheses}). Hence, for each $\vect{x} \in X$, preferences are computed using~\eqref{eq:ml_preference} yielding a vector $\vect{p} \in \mathcal{P}$ (\cref{line:compute_preferences}). 
            The agglomerative step then consists in an iterative block (\cref{line:initialize_clusters,line:inhibit_merge}), where the inter-cluster distance between two clusters $U$ and $V$ is defined using the single-linkage rule:
            \begin{equation}
                \label{eq:singlelinkage}
                d(U,V) = \min_{\vect{u} \in U, \vect{v} \in V} d(\vect{u}, \vect{v})\,.
            \end{equation}
            
            Let $\vect{\theta}^U_i$ denote a model in $\varTheta_i$ fitted to the cluster $U \subseteq X$. Then, for each model family $\mathcal{F}_i$ considered, the models $\vect{\theta}^U_i,\vect{\theta}^V_i,\vect{\theta}^{U \cup V}_i$ are fitted \emph{on-the-fly} to $U, V $ and $U \cup V$, respectively. The \gric is then computed to assess the cost of these $3k$ models (\cref{line:fit_model_families,line:fit_union_model}). The \gric cost of a cluster $U \subseteq X$, with respect to a model family $\mathcal{F}_i$, is defined as~\cite{Torr99}: 
            \begin{equation}
                \label{eq:gric}
                g_i(U) = \sum_{\vect{x}_j \in U} \rho \left(\frac{\mathcal{F}_i(\vect{x}_j, \vect{\theta}^U_i)}{\sigma}\right)^2 + \lambda_1 d|U| + \lambda_2 K,
            \end{equation}
            where $\mathcal{F}(\vect{x}_j, \vect{\theta}^U_i)$ is, as in~\eqref{eq:ml_preference}, a data fidelity term that measures  the residual between $\vect{x}_j \in U$ and the fitted model $\vect{\theta}^U_i \in \varTheta_i$. Here $\sigma$ is an estimate of the residuals standard deviation, and $\rho$ is a robust function that bounds the loss at outliers. The other two terms in~\eqref{eq:gric} account for model complexity: $d$ is the dimension of the manifold of $\mathcal{F}_i$, $K$ the number of model parameters, and $|U|$ the cardinality of $U$.

            \begin{algorithm}[tb]
                \caption{\mlink}
                \label{alg:mlink}
                \DontPrintSemicolon
                \SetNoFillComment
                \KwIn{$X$ - input data, $\{\mathcal{F}_i\}_{i = 1,\dots, k}$ - model families, $\epsilon$ - inlier threshold, $\lambda_1, \lambda_2$ - \gric parameters}
                \KwOut{A partition of the data in structures $X = U_1 \cup \dots \cup U_{\widehat{s}}$}
                \begin{small}
                    \tcc{\textbf{\pembedding} \!\!\!}
                \end{small}
                Sample $m$ models $\mathcal{H} = \varTheta_1 \cup \dots \cup \varTheta_k$ \label{line:sample_hypotheses}\\
                $P \leftarrow \{\vect{p}_j \,|\, \vect{p}_j = \mathcal{E}(\vect{x}_j)\}_{j=1,\ldots,n}$ as in~\eqref{eq:ml_preference} \label{line:compute_preferences}\\
                \begin{small}
                    \tcc{\textbf{\pclustering} \!\!\!}
                \end{small}
                Put each point $\vect{x}_j$ in its own cluster $\{\vect{x}_j\}$ \label{line:initialize_clusters}\\
                Compute inter-cluster distances $d$ as in~\eqref{eq:singlelinkage} \label{line:comput_distances}\\
                \While{$\min(d) < +\infty$}
                    {Find clusters $(U,V)$ with the $\min$ distance \label{line:find_cluster}\\
                     \begin{small}
                         \tcp{\textbf{Fit models, compute \gric~\eqref{eq:gric}} \!\!\!}
                     \end{small}
                     \For{$i = 1, \ldots, k$ \label{line:fit_model_families}}
                        {Fit a model $\vect{\theta}^U_i$ to $U$ and compute $g_i(U)$ \label{line:fit_firs_model}\\
                         Fit a model $\vect{\theta}^V_i$ to $V$ and compute $g_i(V)$ \label{line:fit_second_model}\\
                         Fit a model $\vect{\theta}^{U \cup V}_i$ to $U \cup V$ and compute $g_i(U \cup V)$ \label{line:fit_union_model}\\}
                     \begin{small}
                         \tcp{\textbf{Test merge condition} \!\!\!}
                     \end{small}
                     \eIf{$\exists \widehat{i} \colon g_{\widehat{i}}(U\cup V) \leq g_{i}(U) + g_{i}(V) \enskip \forall\, i \quad i = 1, \dots, k$ \label{line:test_merge_condition}}
                     {merge $U$ and $V$, the structure is $\vect{\theta}_{\widehat{i}}(U\cup V)$ \label{line:merge_clusters}\\
                      update inter-cluster distances $d$ as in~\eqref{eq:singlelinkage} \label{line:update_distances}}
                     {$d(U,V) = +\infty$ \label{line:inhibit_merge}}}
            \end{algorithm}

            We use \gric to determine in a principled manner whether $U$ and $V$ are conveniently aggregated and, in that case, which famly of models describes $U \cup V$ at best (\cref{line:test_merge_condition,line:update_distances}). To this purpose, we compare the \gric scores of the union $g_i(U \cup V)$ with the sum of the costs for two separate fits $g_i(U) + g_i(V)$. There are two alternatives:
            \begin{enumerate}
                \item There exists $\widehat{i}$ yielding the minimum cost at $g_{\widehat{i}}(U \cup V)$:
                    \begin{equation}
                        \label{eq:merge}
                        \exists\, \widehat{i} \colon g_{\widehat{i}}(U \cup V) \leq g_{i}(U) + g_{i}(V) \quad \forall\, i
                    \end{equation}
                \item There exists $\widehat{i}$ yielding the minimum cost at $g_{\widehat{i}}(U) + g_{\widehat{i}}(V)$:
                    \begin{equation}
                        \label{eq:nonmerge}
                        \exists\, \widehat{i} \colon g_{\widehat{i}}(U) + g_{\widehat{i}}(V) < g_{i}(U \cup V) \quad \forall\, i 
                    \end{equation}
            \end{enumerate}
            In the first case, we merge $U$ and $V$ and consider $U \cup V$ as a structure of family $\mathcal{F}_{\widehat{i}}$ (\cref{line:merge_clusters}). A new cluster is created and the inter-cluster distances are updated accordingly to the single-linkage rule (\cref{line:update_distances}). Otherwise, $U$ and $V$ are considered as separate structures, and the merge is inhibited by setting $d(U,V) = +\infty$ (\cref{line:inhibit_merge}).

            \begin{figure}
                \centering
                \begin{subfigure}[b]{0.7\textwidth}
                    \centering
                    \includegraphics[width=\columnwidth]{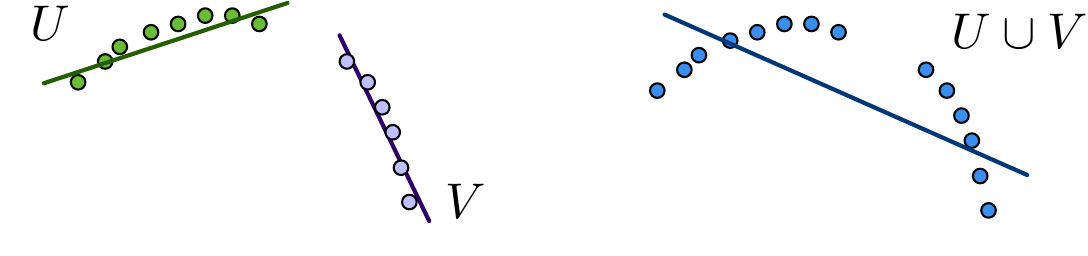}
                    \caption{Rejected with lines.}
                    \label{fig:mergeLine}
                \end{subfigure}
                \\
                \vspace{0.5cm}
                \begin{subfigure}[b]{0.7\textwidth}
                    \centering
                    \includegraphics[width=\columnwidth]{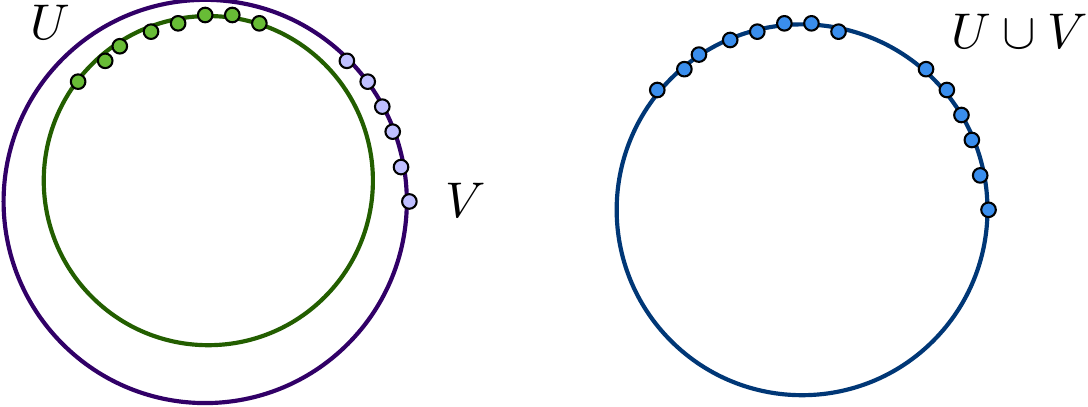}
                    \caption{Accepted  with circles.}
                    \label{fig:mergeCircle}
                \end{subfigure}
                \caption{Illustration of merge with \gric.}
                \label{fig:gric}
            \end{figure}

            Our greedy strategy guides the clustering towards the simplest explanations of the data by only accepting a merge when it results in a reduced \gric cost. In \cref{fig:gric}, when the $U$ and $V$ do not belong to the same structure, the advantage of fitting a single model over $U \cup V$ is offset by the large residuals, leading to a higher \gric cost, which prevents the merge. Conversely, when points from $U$ and $V$ belong to the same structure of a family $\mathcal{F}_{\widehat{i}}$, the cost $g_{\widehat{i}}(U \cup V)$ is lower because the regularization term penalize the complexity of fitting two separate models. Since models are fit on-the-fly (\cref{line:fit_firs_model,line:fit_union_model}) during clustering, \mlink explores new models while being driven by the linkage procedure, making these models more relevant than those in $\mathcal{H}$, the ones randomly sampled in the initial hypothesize step.
            
            \paragraph{Rationale}
            During the \emph{hypothesize} step, \mlink samples models from all the families $\mathcal{F}_1, \dots, \mathcal{F}_k$, embedding all points into a common space, regardless of the model family they refer to.
            Clusters are initially formed by aggregating points with similar preferences, even though no model is assigned to the clusters at this early stage, as they may be too small. Nonetheless, these clusters tend to group inliers from the same model, as they share strong common preferences. As clusters grow, models are associated to them during the \emph{clusterize} step, once they contain enough points to allow the fitting of models from different families. Therefore, initially \mlink leverages on reliable preference information associated to individual points, and later it leverages information from clusters. During the merging process, \mlink performs on-the-fly model fitting, allowing it to recover missed models from the initial sampling and, at the same, to recognize different model families.
            
            \paragraph{Implementation details}
            The inlier threshold $\epsilon$ should be tuned based on the noise level, which is often difficult to estimate when multiple models are present. Therefore, we perform our experiments using both a manually tuned fixed inlier threshold and a meta-heuristic approach based on the Silhouette index~\cite{ToldoFusiello09} to estimate the best $\epsilon$ within a search interval. Although this typically yields a rough estimate of the optimal $\epsilon$, it was sufficient for \mlink to achieve good performance, confirming its robustness to varying $\epsilon$ values. It's worth noting that, in principle, an ad-hoc inlier threshold $\epsilon_i$ could be set for each model family $\mathcal{F}_i$, though we used a common $\epsilon$ across all families in our experiments.
            
            \pembedding was implemented as described in~\eqref{eq:ml_preference}. As in~\cite{MagriFusiello14}, we set $\phi$ as a Gaussian function $\phi(\delta_i) = e^{-\frac{\delta^2_i}{\sigma^2}}$ where $\sigma^2 = -\frac{\epsilon^2}{\log(0.05)}$. The robust function $\rho$ in~\eqref{eq:gric} is $\rho(x) = \min(x, r-d)$, where $r$ is the dimension of the ambient space $\mathcal{X}$ of the data and $d$ is the dimension of the model manifold~\cite{Torr99}.
            
            While testing the merging conditions in~\eqref{eq:merge} and~\eqref{eq:nonmerge} on $U$ and $V$, if any of the two clusters is too small to instantiate a model from family $\mathcal{F}_i$, we apply the merging criterion from \tlinkage~\cite{MagriFusiello14}, which merges clusters if there is at least one sampled model from $\mathcal{H}$ that explains all points in both clusters.
            We also find it helpful to remove hypotheses that occurred by chance from the initial pool of models $\mathcal{H}$. To this end, we validate each hypothesis $\vect{\theta} \in \mathcal{H}$ using the preprocessing stage from~\cite{TepperSapiro14}, which exploits a Gestalt principle to assess the significance of $\vect{\theta}$.
            A model is not significant if its supporting points are uniformly distributed in space, which is verified by checking whether the number of inliers at distance $k \epsilon$ is roughly $k$ times inliers at $\epsilon$.
            
            \subsubsection{Features and benefits}
                \label{subsubsec:features_and_benefits}

                \mlink shares some similarities with hierarchical clustering-based preference methods, such as \tlinkage and \mct. However, thanks to several key differences, \mlink overcomes the main limitations of these algorithms, even in the single-family scenario. \cref{tab:diff} summarizes the key improvements of \mlink compared to \tlinkage and \mct.

                \begin{table}[t]
                    \centering
                    \caption{Differences among \tlinkage, \mct and \mlink.}
                    \begin{tabular}{l@{\hskip 0.8cm}c@{\hskip 0.8cm}c@{\hskip 0.8cm}c}
                        \toprule
                                        & \tlinkage                & \mct                    & \mlink                                        \\
                        \midrule
                        Family           & single $\mathcal{F}$ & nested $\mathcal{F}_1 \subset \mathcal{F}_2$ & multi $\mathcal{F}_1, \dots, \mathcal{F}_k$ \\
                        Models          & sampled                  & sampled                 & sampled and on-the-fly                        \\
                        Linkage         & centroid-link            & centroid-link           & single-linkage \& \gric                       \\
                        Model selection & no                       & a posteriori            & inside clustering                             \\
                        \bottomrule
                    \end{tabular}
                    \label{tab:diff}
                \end{table}

                \paragraph{\pembedding}
                \mlink implements a multi-family \pembedding that, unlike \mct, can handle multiple, non-nested model families. However, multi-family embedding alone would not be sufficient to recover multi-family structures, as in \tlinkage, where the more general models always prevail over the simpler ones, being model complexity ignored.

                \paragraph{On-the-fly sampling}
                Another essential feature of \mlink is the fitting of additional models during the clustering process, beyond those generated during the initial hypothesize step. Models fitted on-the-fly are more reliable than those in $\mathcal{H}$ as they are instantiated on emerging clusters of inliers rather than on minimal sample sets. As a result, \mlink is more robust than \tlinkage and \mct with respect to sampling imbalances, leading to more stable performance, as demonstrated in our experiments. Typically, models are fitted on-the-fly using efficient close-form solutions, while more complex non-linear refinements can be applied later, once clusters have formed.
                
                \paragraph{Agglomerative scheme}
                Both \tlinkage and \mct use a variant of centroid-linkage designed for the \pspace.
                In practice, two clusters $U$ and $V$ are merged when there is at least one model from $\mathcal{H}$ that passes through all the points of $U \cup V$ within a maximum distance of $\epsilon$, or equivalently, when the centroids of the two clusters have distance lower than $1$. Therefore, even a single outlier can significantly affect the cluster centroid, leading to over-segmentation. \mlink's single-linkage mechanism avoids this issue and also reduces the computational burden, since its distance update mechanism is more efficient as it does not involve the computation of a cluster representative at each merging step. Additionally, coupling single-linkage with the subsequent model selection step via \gric well balances its tendency to merge spatially close clusters. This creates a symbiotic mechanism where single-linkage adopts an "explorative" role, while \gric acts as a "conservative" counterbalance, preventing unnecessary merges.
                
                \paragraph{Model selection}
                While \gric was also used in \mct for model selection at the end of a stratified clustering process, \mlink integrates \gric directly into the clustering process. Additionally, tuning the parameters $\lambda_1$ and $\lambda_2$ in \mct is difficult because \gric compares a varying number of model instances. \mlink, on the other hand, always performs one-vs-a-pair comparison of models, allowing us to safely set $\lambda_1=1, \lambda_2 = 2$ in all the experiments, as shown in~\cref{tab:param}.

                \begin{table}[t]
                    \centering
                    \caption{Parameters used for \mlink.}
                    \begin{tabular}{l@{\hskip 2.75cm}c@{\hskip 2.75cm}c@{\hskip 2.75cm}c}
                        \toprule
                                      & $\epsilon$    & $\lambda_1$ & $\lambda_2$ \\
                        \midrule
                        conic (a-c)   & $0.180$       & $1$         & $2$         \\
                        conic (d-g)   & $0.900$       & $1$         & $2$         \\
                        plane seg.    & $0.070$       & $1$         & $2$         \\
                        $2$-view seg. & $0.058$       & $1$         & $2$         \\
                        video seg.    & $[0.01, 0.3]$ & $1$         & $2$         \\
                        \bottomrule
                    \end{tabular}
                    \label{tab:param}
                \end{table}

    \section{Experiments}
        \label{sec:ml_experiments}

        We test \mlink on both single-family and multi-family structure recovery problems. We first address 2D primitive fitting problems (\cref{subsec:2d_fitting}), which represent a standard benchmark for structure recovery algorithms, and demonstrate that \mlink outperforms \mct. Then, we test \mlink on real-world datasets for the estimate of two view relations from correspondences (\cref{subsec:two_views_relations}) and for video motion segmentation (\cref{subsec:video_motion_segmentation}). In all these experiments, we show that \mlink favorably compares or performs on par with recent multi-family structure recovery alternatives~\cite{BarathMatas19}. The qualitative results of \mlink on 3D primitive fitting in a sparse input point cloud~\cite{sam} has been reported in~\cref{fig:structure_based_clustering_example}.

        Performance is measured, as customarily, in terms of misclassification error (ME), \emph{i.e.}, the fraction of misclassified points with respect to the ground-truth labelling. If not stated otherwise, we always report ME averaged over $5$ runs. Parameters used to configure \mlink in each dataset are reported in~\cref{tab:param}. \matlab code of \mlink is available on-line at~\cite{blind}.

        \subsection{2D fitting}
            \label{subsec:2d_fitting}

            \begin{figure}[t]
                \centering
                \begin{subfigure}[b]{0.23\textwidth}
                    \begin{center}
                    \raisebox{0.4cm}{\includegraphics[width=\columnwidth]{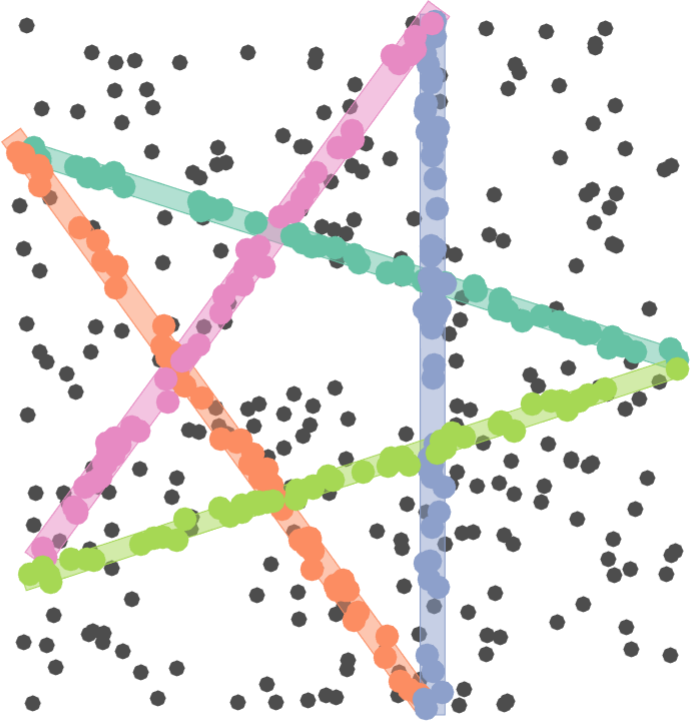}}
                    \caption{\emph{Worst} result ($3\sigma$).}
                    \label{fig:star5result}
                    \end{center}
                \end{subfigure}
                \hfill
                \begin{subfigure}[b]{0.35\textwidth}
                    \begin{center}
                    \includegraphics[width=\columnwidth]{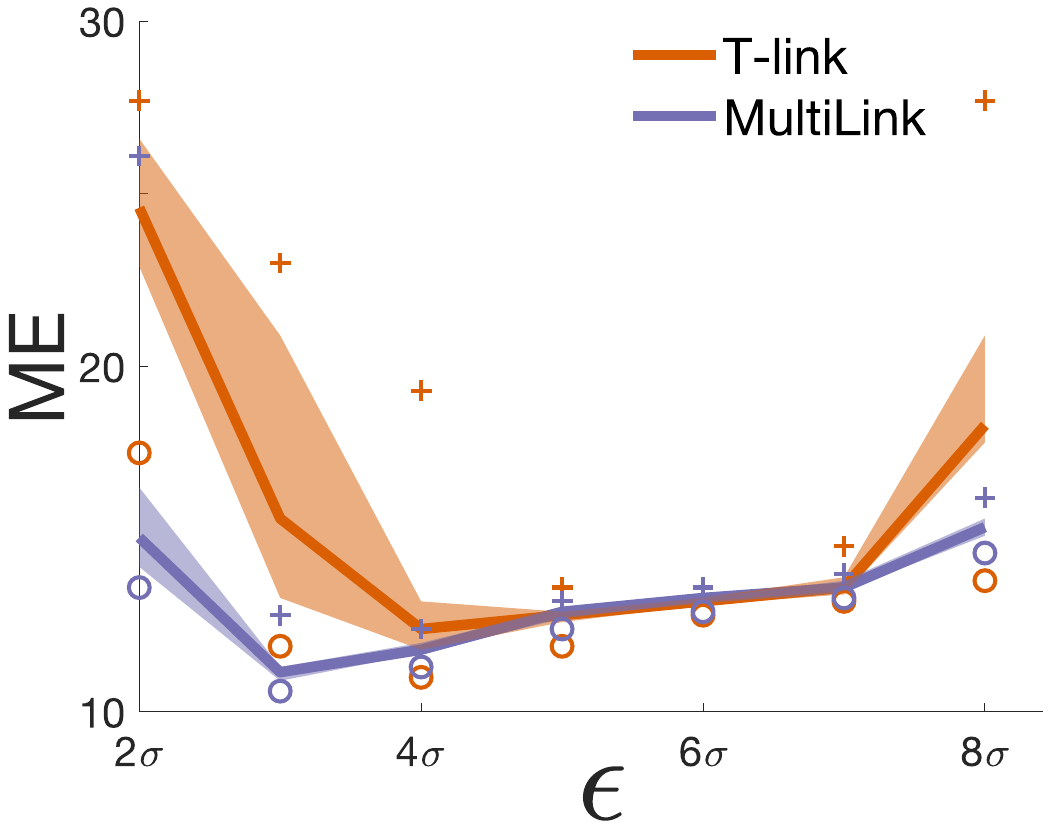}
                    \caption{ME vs. inlier threshold $\epsilon$.}
                    \label{fig:star5_me}
                    \end{center}
                \end{subfigure}
                \hfill
                \begin{subfigure}[b]{0.35\textwidth}
                    \begin{center}
                    \includegraphics[width=\columnwidth]{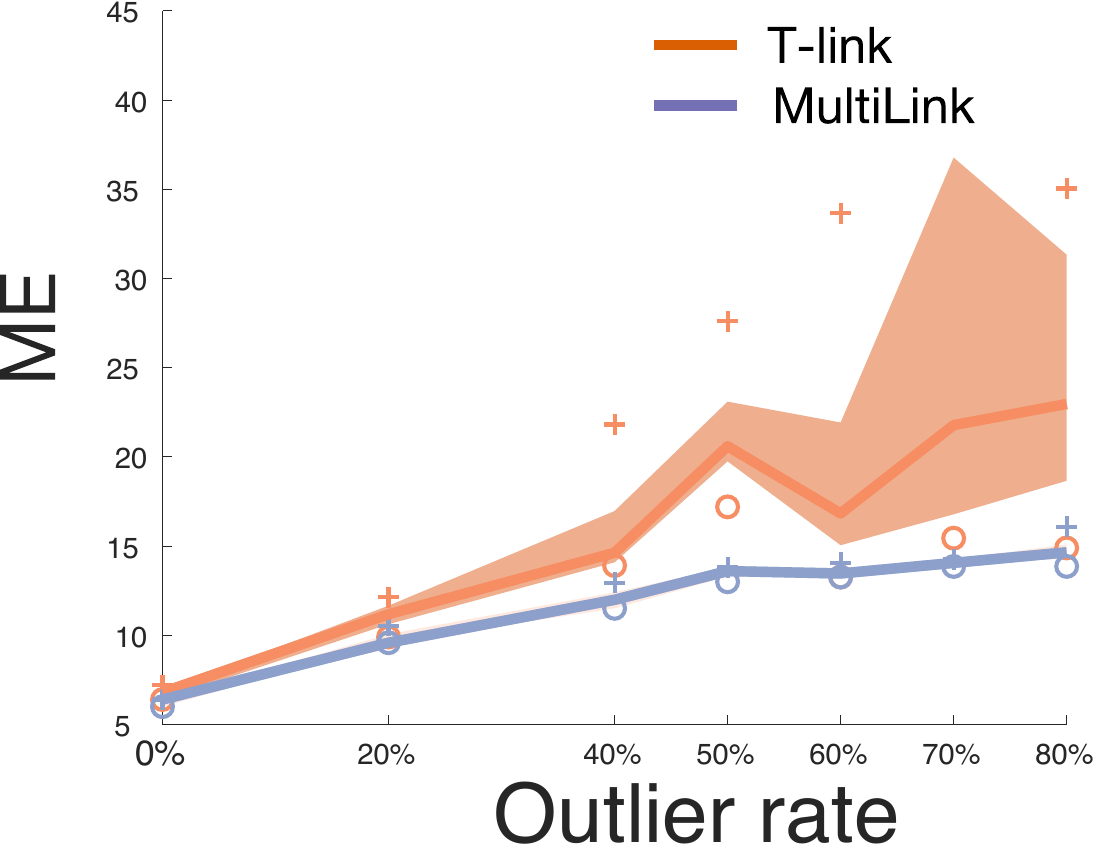}
                    \caption{ME vs. outlier rate.}
                    \label{fig:star5_outlier}
                    \end{center}
                \end{subfigure}
                \caption{Line fitting: \mlink vs. \tlinkage. (a) The \emph{worst} result of \mlink for $\epsilon = 3\sigma$ over $50$ trials. (b) and (c), the median ME (solid line), IQR (shaded area), maxima ($+$) and minima ($\circ$) as a function of $\epsilon$ and outlier ratio.}
                \label{fig:star5}
            \end{figure}

            We first consider a single-family structure recovery problem (line fitting) and show that \mlink outperforms \tlinkage in terms of accuracy, robustness to outliers and runtime. Then, we address a multi-family structure recovery problem (conic fitting) and show that \mlink outperforms \mct and \pearl.

            \paragraph{Line fitting}
            We consider \tlinkage as the closest alternative to \mlink when addressing the single-family problem illustrated in~\cref{fig:star5result}. The dataset consists of multiple lines contaminated by noise and outliers, and the \matlab implementation of \tlinkage is from~\cite{jlk}.
            
            \cref{fig:star5_me} reports the median ME over $50$ runs as a function of the inlier thresholds $\epsilon = k \sigma$, where $\sigma$ represents the noise level and $k = 2, \ldots, 8$. The plot displays the inter-quantile range (IQR) of the ME (shaded regions), along with the minimum ($\circ$) and maximum ($+$) errors. Both methods were initialized with the same hypotheses set $\mathcal{H}$, resulting in the same preference representation $P$ for points. Results indicate that \mlink consistently outperforms \tlinkage, achieving better results across the median, maximum and minimum ME metrics. Notably, except for $\epsilon = 2 \sigma$, where both methods over-segment the data, \mlink produces highly stable outputs, as evidenced by the small IQR. This confirms that fitting new models on-the-fly during clustering enhances the robustness of \mlink with respect to both $\epsilon$ and the randomly sampled hypotheses $\mathcal{H}$. In contrast, \tlinkage relies solely on the fixed pool of models in $\mathcal{H}$, and therefore it suffers from higher variability across runs, as reflected by its large IQR and maximum errors. Further analysis on the $star5$ dataset at varying outlier rates (\cref{fig:star5_outlier}) shows that \mlink consistently outperforms \tlinkage, demonstrating greater resilience to outliers.

            Even though \mlink may, in principle, have a higher worse-case complexity due to merge rejections, in practice, it does not exhibit additional computational overheads. In fact, the single-linkage scheme makes the clustering step of \mlink faster ($0.26$s) than \tlinkage ($0.76$s) on average. This experiment shows that \mlink not only outperforms its closest single-family counterpart in terms of effectiveness but also efficiency, offering more stable performance.

            \begin{table}[t]
                \centering
                \caption{Misclassification errors (ME) for conic fitting problem of~\cref{fig:exp2d} \texttt{a}.}
                \begin{tabular}{l@{\hskip 0.8cm}r@{\hskip 0.8cm}r@{\hskip 0.8cm}r@{\hskip 0.8cm}r@{\hskip 0.8cm}r@{\hskip 0.8cm}r@{\hskip 0.8cm}r}
                    \toprule
                                & (a)             & (b)             & (c)             & (d)             & (e)             & (f)             & (g)             \\
                    \midrule
                    \pearl      & $6.00$          & $16.22$         & $14.44$         & $8.05$          & $8.33$          & $17.38$         & $19.21$         \\
                    \mct        & $\mathbf{0.67}$ & $\mathbf{2.00}$ & $2.33$          & $5.23$          & $7.12$          & $5.38$          & $6.23$          \\
                    \smlink     & $2.17$          & $2.13$          & $\mathbf{1.55}$ & $\mathbf{1.83}$ & $\mathbf{0.87}$ & $\mathbf{2.46}$ & $\mathbf{4.28}$ \\
                    \bottomrule
                \end{tabular}
                \label{tab:synth}
            \end{table}

            \begin{figure}[t]
                \centering
                \includegraphics[width=.8\linewidth]{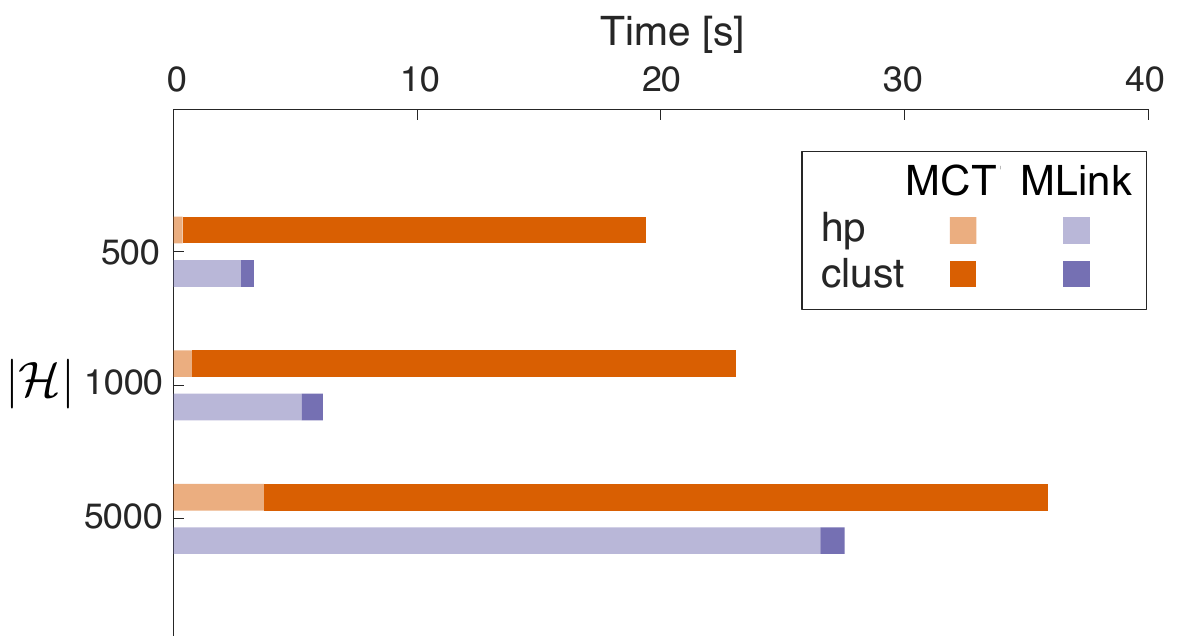}
                \caption{Execution times for conic fitting problem of~\cref{fig:exp2d} \texttt{a}.}
                \label{fig:time}
            \end{figure}
            
            \paragraph{Line and conic fitting}
            \cref{fig:exp2d} shows 2D simulated datasets from~\cite{MagriFusiello19}, used to recover lines, circles and parabolas, with \mlink worst results displayed. The datasets contain instances of lines and circles, while \texttt{a}, \texttt{b} and \texttt{c} also features parabolas. Despite the multi-family challenge, \mlink successfully recovers all geometric structures, even in its worst performing runs.
            
            \cref{tab:synth} compares the performance of \mlink with \pearl~\cite{IsackBoykov12} and \mct, as reported in~\cite{MagriFusiello19}. Results show that \mlink achieves lower ME in $5$ out of $7$ cases. Even in the two cases where \mct performs better, the ME of \mlink is rather small.
            \cref{fig:time} presents a comparison of execution times of \mct and \mlink on the dataset from~\cref{fig:exp2d}.a with reference to $|\mathcal{H}|$, the number of initial hypotheses $\varTheta_1 \cup \dots \cup \varTheta_k$. Both algorithms are implemented in \matlab, and the code of \mct is from~\cite{mct}. As expected, on-the-fly fitting makes \mlink more efficient compared to the cascaded approach of \mct. In fact, the clustering step of \tlinkage, which we show to be slower that that of \mlink in the $star5$ experiment, is repeated several times in \mct, leading to longer execution times. \mlink spends most of the time on hypotheses generation (light blue bars), while the actual clustering step takes significantly less time (dark blue bars). This is largely due to the optional pre-processing step~\cite{TepperSapiro14} for filtering out irrelevant models, whose computational burden can be substantially reduced in an optimized and parallel implementation. 
            
        \subsection{Two-views relations}
            \label{subsec:two_views_relations}

            We test \mlink on two-views segmentation over the well-known Adelaide RMF dataset~\cite{WongChinAl11}, which contains $36$ stereo images sequences with correspondences affected by noise and outliers, and annotated ground-truth matches. We first detect planar structures by fitting homographies, and then we perform motion segmentation (\cref{fig:twoviews}). This latter process was treated as a multi-family recovery task by fitting fundamental matrices, affine fundamental matrices, and homographies.

            \begin{figure}[t]
                \centering
                \begin{subfigure}[b]{0.32\textwidth}
                    \begin{center}
                    \includegraphics[width=\columnwidth]{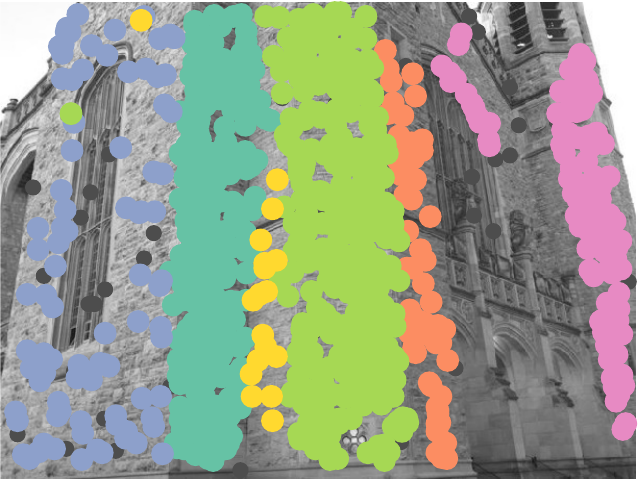}
                    \caption{}
                    \label{fig:h1}
                    \end{center}
                \end{subfigure}
                \hfill
                \begin{subfigure}[b]{0.32\textwidth}
                    \begin{center}
                    \includegraphics[width=\columnwidth]{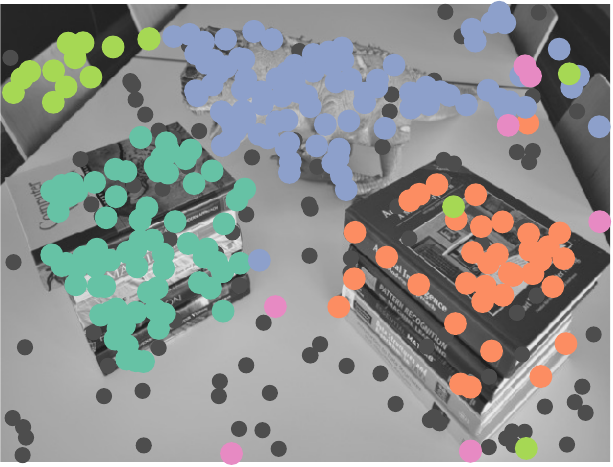}
                    \caption{}
                    \label{fig:h4}
                    \end{center}
                \end{subfigure}
                \hfill
                \begin{subfigure}[b]{0.32\textwidth}
                    \begin{center}
                    \includegraphics[width=\columnwidth]{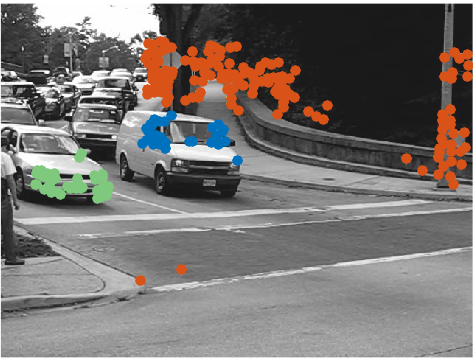}
                    \caption{}
                    \label{fig:h5}
                    \end{center}
                \end{subfigure}
                \caption{Sample results attained by \mlink on plane (a) two-view (b) and motion segmentation (c).}
                \label{fig:twoviews}
            \end{figure}
            
            \paragraph{Plane segmentation (single-family)}
            Results in~\cref{tab:adelHfixed} show that both \progx and \mlink achieve comparable best mean performance, though \mlink yields more stable results.
            To ensure a fair comparison with \mct, which was used to fit a fundamental matrix and then recover compatible nested homographies using a per-sequence tuned inlier threshold, we optimized $\epsilon$ for \mlink in the same way. The results in \cref{tab:adelHtuned} indicate that \mlink remains the best performing algorithm. Moreover, the difference in terms of ME between fixed and per-sequence tuned $\epsilon$ for \mlink is much smaller than for \tlinkage, further highlighting the robustness of \mlink with respect to the choice of $\epsilon$.

            \begin{table}[t]
                \centering
                \caption{Mean ME (in $\%$) on real datasets. Averages over $5$ runs on each sequence.}
                \begin{subtable}{\textwidth}
                    \centering
                    \caption{Plane seg. fixed parameters.}
                    \begin{tabular}{l@{\hskip 0.75cm}r@{\hskip 0.75cm}r@{\hskip 0.75cm}r@{\hskip 0.75cm}r@{\hskip 0.75cm}r@{\hskip 0.75cm}r}
                        \toprule
                             & \pearl  & \multix & \progx & \rpa    & \stlinkage & \smlink         \\
                        \midrule
                        Mean & $15.14$ & $8.71$  & $6.86$ & $23.54$ & $22.38$    & $\mathbf{6.46}$ \\
                        Std. & $6.75$  & $8.13$  & $5.91$ & $13.42$ & $7.27$     & $1.75$          \\
                        \bottomrule
                    \end{tabular}
                    \label{tab:adelHfixed}
                \end{subtable}
                \\
                \vfill
                \begin{subtable}{\textwidth}
                    \centering
                    \caption{Methods with $\epsilon$ tuned per-sequence.}
                    \begin{tabular}{l@{\hskip 2.625cm}r@{\hskip 2.625cm}r@{\hskip 2.625cm}r}
                        \toprule
                               & \stlinkage & \mct   & \smlink         \\
                        \midrule
                        Mean   & $6.60$     & $6.13$ & $\mathbf{4.10}$ \\
                        Median & $4.68$     & $4.93$ & $\mathbf{2.70}$ \\
                        \bottomrule
                    \end{tabular}
                    \label{tab:adelHtuned}
                \end{subtable}
                \\
                \vfill
                \begin{subtable}{\textwidth}
                    \centering
                    \caption{Two-view seg. fixed parameters.}
                    \begin{tabular}{l@{\hskip 0.975cm}r@{\hskip 0.975cm}r@{\hskip 0.975cm}r@{\hskip 0.975cm}r@{\hskip 0.975cm}r@{\hskip 0.975cm}r}
                        \toprule
                                   & \multicolumn{2}{c}{Fundamental} & \multicolumn{2}{c}{Affine fund.} & \multicolumn{2}{c}{Mixed} \\
                        \midrule
                                   & Mean    & Std.                  & Mean    & Std.                   & Mean            & Std.    \\
                        \midrule
                        \smlink    & $8.59$  & $4.67$                & $9.84$  & $4.09$                 & $\mathbf{7.75}$ & $4.54$  \\
                        \stlinkage & $32.20$ & $50.33$               & $41.90$ & $7.95$                 & $38.78$         & $8.21$  \\
                        \progx     & $10.73$ & $8.73$                & $-$     & $-$                    & $-$             & $-$     \\
                        \multix    & $17.13$ & $12.23$               & $10.50$ & $2.90$                 & $9.53$          & $1.43$  \\
                        \pearl     & $29.54$ & $14.80$               & $41.81$ & $15.25$                & $48.89$         & $8.16$  \\
                        \bottomrule
                    \end{tabular}
                    \label{tab:adelFMfixed}
                \end{subtable}
                \\
                \vfill
                \begin{subtable}{\textwidth}
                    \centering
                    \caption{Video seg. \textsc{s}$=$Silhouette index.}
                    \begin{tabular}{l@{\hskip 3.4cm}r@{\hskip 3.4cm}r}
                        \toprule
                                                 & Mean            & Std.    \\
                        \midrule
                        \multix                  & $12.96$         & $19.60$ \\
                        \progx                   & $8.41$          & $10.29$ \\
                        \stlinkage +s (dim. $3$) & $8.68$          & $12.23$ \\
                        \mct +s                  & $10.87$         & $12.68$ \\
                        \mlink +s (dim. $3$)     & $\mathbf{8.34}$ & $11.93$ \\
                        \mlink +s (mixed)        & $9.83$          & $13.05$ \\
                        \bottomrule
                    \end{tabular}
                    \label{tab:video}
                \end{subtable}
            \end{table}
            
            \paragraph{Two-views segmentation (multi-family)}
            We also conducted two-view motion segmentation on $19$ stereo images depicting moving objects. This dataset has been extensively used for ego motions estimation by fitting fundamental matrices, making it a benchmark for single-family multi-structure fitting~\cite{WongChinAl11,MagriFusiello14,BarathMatas18,BarathMatas19}. However, our preliminary tests suggested that some of the movements 
            can also be reliably described by affine fundamental matrices or even homographies, indicating quasi-degenerate motions.
            We tested \mlink with three different model families:
            (1) \emph{Fundamentals}: $\mathcal{F}_f$ the manifold of fundamental matrices; (2) \emph{Affine fundamentals}: $\mathcal{F}_a$ the manifold of affine fundamental matrices; (3) \emph{Mixed models}: where we consider $\mathcal{F}_f$, $\mathcal{F}_a$ and $\mathcal{F}_h$, the space of homographies.
            
            The mean ME for each configuration, along with standard deviation, is shown in~\cref{tab:adelFMfixed}. We evaluated both \mlink and \tlinkage under the three configurations mentioned above, using \emph{fixed} parameters. Additionally, we tested \progx, \multix and \pearl on the same dataset for fitting fundamental matrices, and the results from~\cite{BarathMatas19} are presented accordingly. To evaluate these methods with the affine fundamental and mixed models configurations, we adapted the source codes provided by authors of \multix and \pearl in~\cite{multix} and~\cite{pearl}, while this modification was not possible for \progx code due to its lack of flexibility for other configurations.
            Two key observations can be made: first, \mlink consistently outperforms all the other methods across the three configurations. Second, \mlink effectively handles multi-family fitting, achieving the lowest ME when the three models are used in a mixed way. These findings align with the results in~\cite{XuCheongAl18} and represents an interesting case where multi-family approaches can be successfully employed to deal with nearly degenerate data. When only affine fundamental matrices are employed, both \mlink and \tlinkage exhibit higher ME compared to using fundamental matrices alone, indicating that affine fundamental matrices may not be flexible enough to capture the full range of motion variation in the dataset.

        \subsection{Video motion segmentation}
            \label{subsec:video_motion_segmentation}
            
            Lastly, we evaluated \mlink on the video motion segmentation task using the Hopkins 155 benchmark~\cite{TronVidal07}, which contains $155$ video sequences with $2$ or $3$ moving objects whose trajectories can be approximated, under the assumption of affine projection, as a union of low dimensional subspaces. Depending on the scene considered, the dimension of the subspaces may vary, with trajectories modeled using either $2D$ or $3D$ affine subspaces~\cite{SugayaKanatani04*1}. Therefore, we configured \mlink in two ways: \emph{i}) single-family, where we fit affine subspaces of dimension $3$, and \emph{ii}) multi-family, where we fit both affine subspaces of dimension $2$ and $3$ as mixed models. The results in~\cref{tab:video} compare performance of \mlink against \multix and \progx from~\cite{BarathMatas19}, both of which were executed with fixed parameters across the entire dataset. We thus configure \mlink with all the parameters fixed, while the inlier threshold $\epsilon \in [0.01, 0.3]$ was automatically estimated for each sequence using of a variant of the Silhouette index, as described in~\cite{ToldoFusiello09}. Estimating $\epsilon$ in this way represents a very practical solution that is widely applicable in real-world scenarios. We also run \tlinkage and \mct using the same Silhouette index for threshold estimation.
            The results show that \mlink performs similarly to \progx when fitting subspaces of dimension $3$. Furthermore, the experiment demonstrates that \mlink is stable, as it can successfully compensate for inaccurate estimates of $\epsilon$. The benefits of using mixed models are not evident when looking at the average ME across the entire dataset. However, we found that \mlink with mixed family models consistently improves the results, particularly in natural video sequences with degenerate motions (\mlink with mixed families achieves a ME of $1.37\%$ on Traffic 3 and $3.14\%$ on Traffic 2, compared to the configuration with subspaces of dimension $3$ that scores $7.51\%$ and $4.18\%$ respectively). This suggests that in many sequences, $3D$ subspaces are the optimal model, and the mixed configuration actually reduces the overall performance.

    \section{Conclusions}
        \label{sec:sbc_conclusions}

        In this chapter, we presented \mlink, a simple yet effective algorithm for recovering structures from various model families in data contaminated by noise and outliers. \mlink is able to fit models from different families throughout the clustering process, and includes a novel cluster-merging scheme which leverages on-the-fly model fitting and model selection via \gric.

        Experiments on both synthetic and real-world data showed that \mlink is faster, more stable, and less sensitive to sampling and inlier thresholds compared to greedy alternatives based on preference analysis and agglomerative clustering, such as \tlinkage and \mct. Additionally, \mlink favorably compares with optimization-based methods. Moreover, \mlink is a flexible framework that can be extended by adjusting the cluster-merging conditions to meet specific constraints coming from the application at hand.

    \cleardoublepage
    \pagestyle{empty}
    \mbox{}
    \cleardoublepage
    \setlength\epigraphwidth{6.445cm}
\setlength\epigraphrule{0pt}
\epigraph{{\large \emph{``I expect they had lots of chances, like us, of turning back, only they didn't. And if they had, we shouldn't know, because they'd have been forgotten.''\footnotemark}}}{}
\footnotetext{Samvise Gamgee, \emph{The Lord of the Rings: The Two Towers} by J.R.R. Tolkien}

    \cleardoublepage
    \pagestyle{fancy}
    \part{Density-based Anomaly Detection}
        \label{par:density_based_anomaly_detection}

        \chapter{Online Anomaly Detection}
    \label{cha:online_anomaly_detection}

    In this chapter, we address the problem of online anomaly detection by proposing \emph{\onlineisolationforest} (\onlineiforest), a novel and efficient tree-based approach for identifying anomalies in data streams.
    \onlineiforest is an ensemble of multi-resolution histograms that combines a \emph{\learningprocedure}, which increases resolution in densely populated regions of the space by splitting bins, and a \emph{\forgettingprocedure}, which decreases resolution in sparsely populated areas by merging bins. These dynamic learning and forgetting mechanisms allow \onlineiforest to handle the incremental nature of data streams by continuously tracking the data distribution as it evolves over time.

    In~\cref{sec:oad_problem_formulation}, we give a formal definition for the online anomaly detection problem, and in~\cref{sec:oad_related_literature} we review the literature on online anomaly detection from a tree-based perspective. In~\cref{sec:online_isolation_forest}, we introduce the \onlineiforest algorithm, and in~\cref{subsec:online_isolation_tree} we detail \onlineitree, the multi-resolution histogram component of the \onlineiforest ensemble, along with with the \learningprocedure and \forgettingprocedure. In~\cref{subsec:complexity_analysis}, we analyse the computational complexity of \onlineiforest, and in~\cref{subsec:sliding_buffer} we briefly discuss how the sliding buffer size affects adaptation speed and model accuracy. Finally, in~\cref{sec:oif_experiments} we present and discuss the experimental validation of \onlineiforest.

    \section{Problem Formulation}
        \label{sec:oad_problem_formulation}

        \begin{figure}[t]
            \begin{subfigure}[t]{0.3\textwidth}
                \begin{center}
                \hspace*{-0.35\textwidth}
                \centerline{\includegraphics[width=0.775\columnwidth]{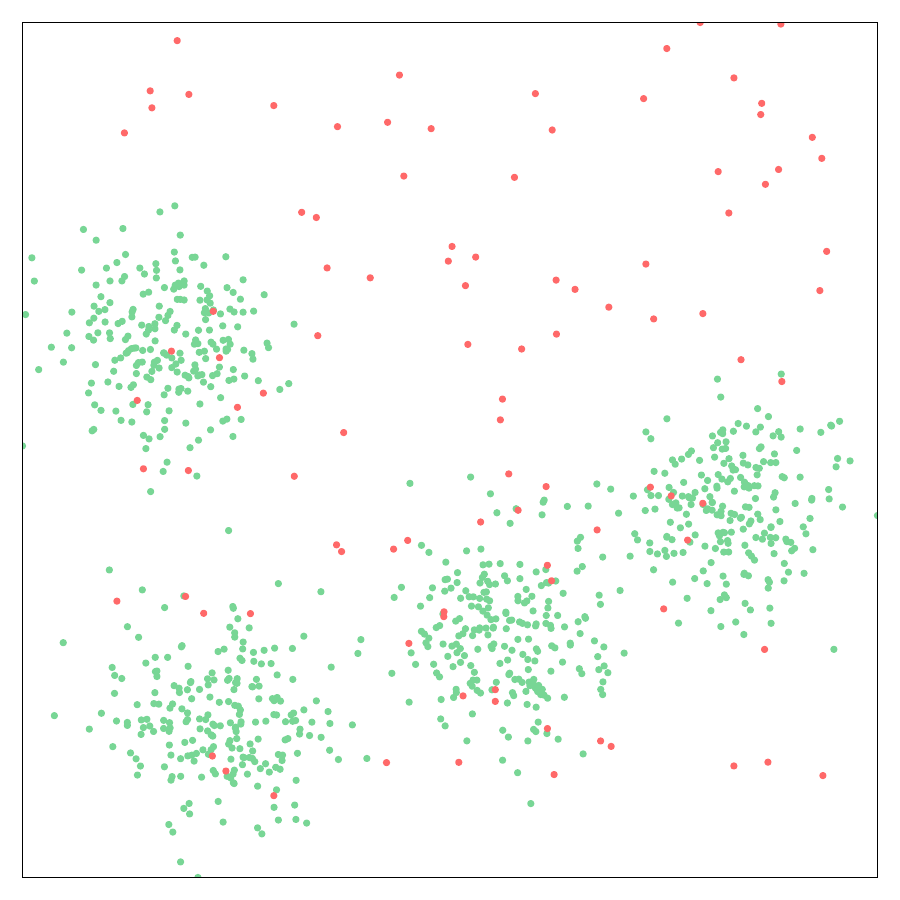}}
                \hspace*{-0.25\textwidth}
                \begin{minipage}{1.25\textwidth}
                    \caption{Data stream $\vect{x}_1, \dots, \vect{x}_t \in \mathbb{R}^d$.}
                    \label{fig:data}
                \end{minipage}
                \end{center}
            \end{subfigure}
            \hspace*{0.05\textwidth}
            \begin{subfigure}[t]{0.7\textwidth}
                \begin{center}
                \centerline{\includegraphics[width=0.325\columnwidth]{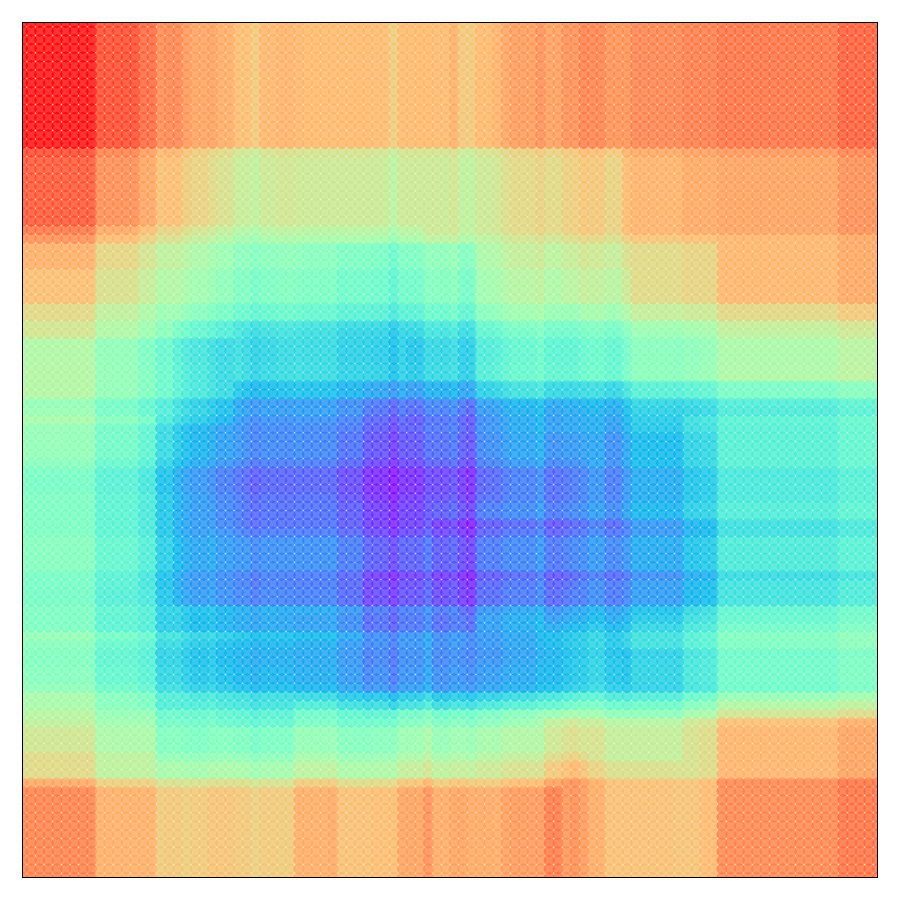} \; \includegraphics[width=0.325\columnwidth]{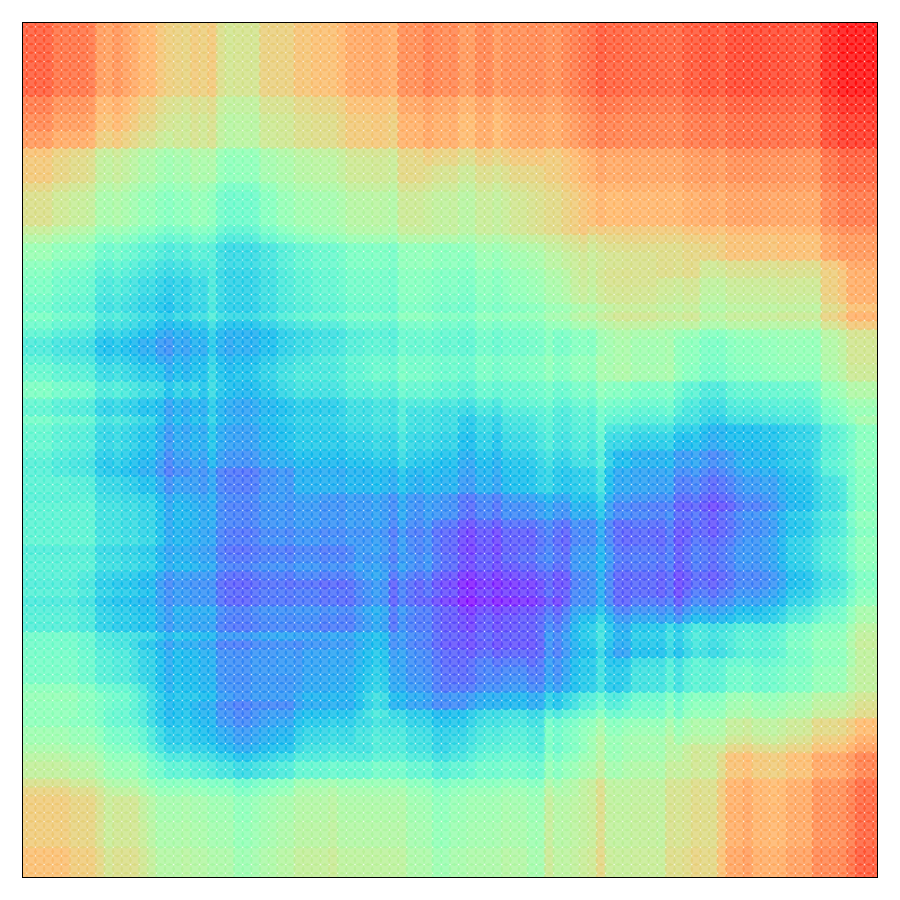} \; \includegraphics[width=0.325\columnwidth]{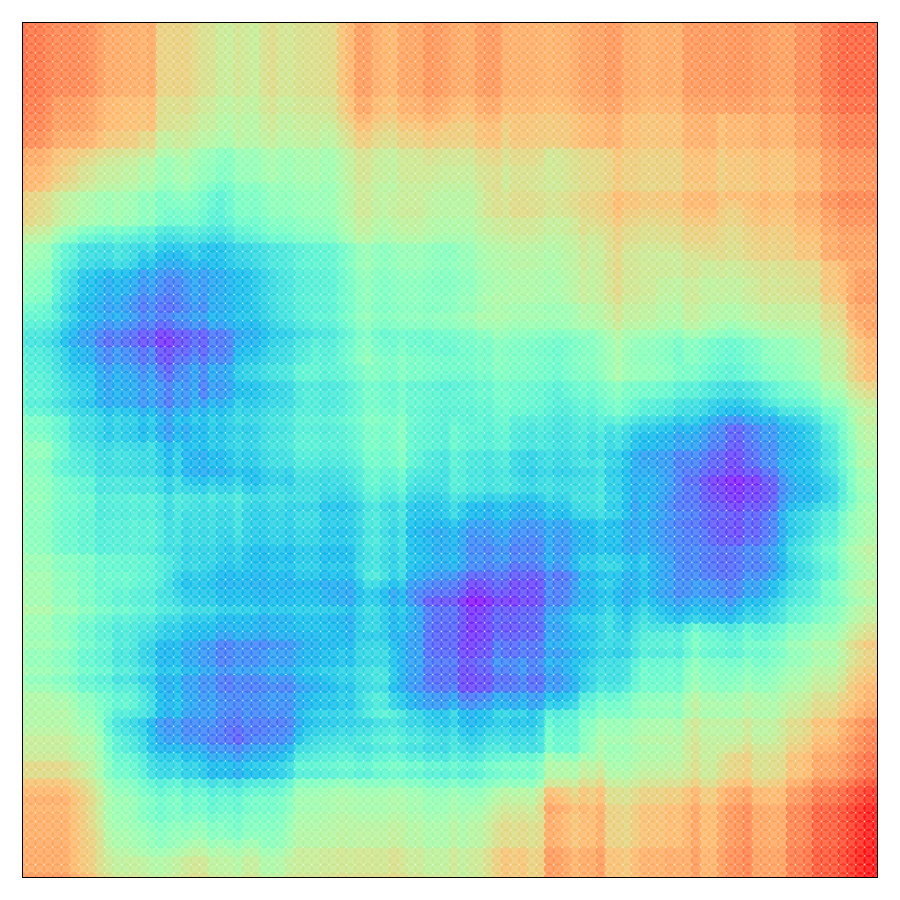} \includegraphics[width=0.04725\columnwidth]{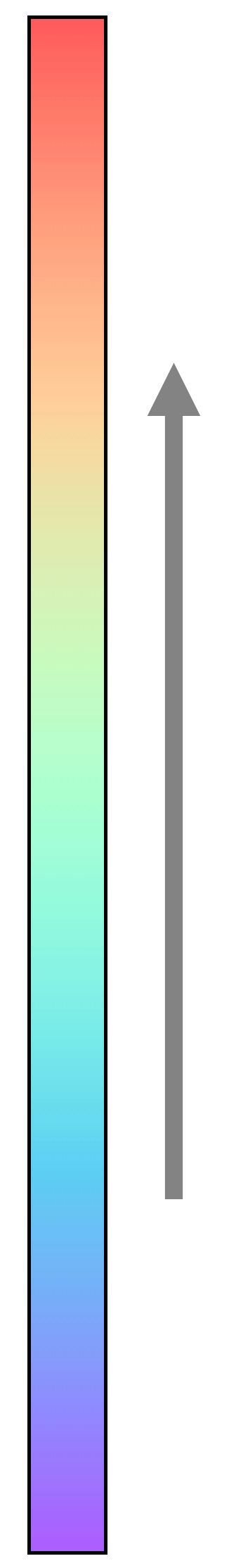}}
                \caption{Anomaly scores $s$ at different time instants $t$, from left to right.}
                \label{fig:oif_anomaly_scores}
                \end{center}
            \end{subfigure}
            \caption{\onlineiforest incrementally adapts to the data distribution of the stream and improves the anomaly scores estimate over time.}
            \label{fig:synthetic}
        \end{figure}

        We address the online anomaly detection problem in a virtually unlimited multivariate data stream $\vect{x}_1, \vect{x}_2, \dots, \vect{x}_t \in \mathbb{R}^d$, where $t \geq 1$ (\cref{fig:data}). We assume that each $\vect{x}_i$ is a realization of an independently and identically distributed (i.i.d.) random variable having unknown distribution either $\vect{X}_i \sim \Phi_0$ or $\vect{X}_i \sim \Phi_1$, where $\Phi_0$ is the distribution of genuine data and $\Phi_1$ is the distribution of anomalous data.
        Given a point $\vect{x}_t$, the goal is to identify whether $\vect{X}_t \sim \Phi_0$ or $\vect{X}_t \sim \Phi_1$ for each time instant $t$ (\cref{fig:oif_anomaly_scores}).
        
        We assume that anomalous data are \textit{``few''} and \textit{``different''}~\cite{LiuTingAl08,LiuTingAl12} and express these assumptions in the following way: (i) \textit{``few''} -- the probability that a point $\vect{x}_i$ has been generated by the distribution $\Phi_1$ of anomalous data is much lower than the probability that it has been generated by the distribution $\Phi_0$ of genuine data, \emph{i.e.}, $P(\vect{X}_i \sim \Phi_1) \ll P(\vect{X}_i \sim \Phi_0)$,
        and (ii) \textit{``different''} -- the probability that a point $\vect{x}_i$ is closer to an anomalous point $\vect{x}_j$ rather than a genuine point $\vect{x}_k$ is low.
        As a consequence, our focus is on scenarios where anomalous data do not form dense and populous clusters.

        Furthermore, we assume that we can store in memory only a finite and small subset $\vect{x}_{t-\omega}, \dots, \vect{x}_t$ of size $\omega$ from the entire data stream at each time instant $t$, were $\omega$ is small enough.
        Additionally, we require the time interval between the acquisition of a sample $\vect{x}_t$ and its classification to be as small as possible.

    \section{Related Literature}
        \label{sec:oad_related_literature}

        Among the wide literature on anomaly detection~\cite{ChandolaBanerjeeAl09}, we focus on tree-based methods, as they achieve state-of-the-art performance at low computational and memory requirements.
        \ifor~\cite{LiuTingAl08,LiuTingAl12} introduced the concept of \textit{``isolation''} as a criteria to categorize anomaly detection algorithms, and subsequent works showed that it is strongly related to the concepts of both distance and density~\cite{ZhangDouAl17,LeveniMagriAl23}.
        
        The core component of \ifor is an ensemble of random trees constructed through an iterative branching process. Each individual tree is built by randomly selecting a data dimension and a split value within the bounding box containing data points in that dimension. Anomalous data are identified via an anomaly score computed on the basis of the average path length from the root node to the leaf node, under the assumption that anomalies are easier to isolate.

        The most straightforward extension of \ifor to the streaming scenario is \textit{Isolation Forest ASD}~\cite{DingFei13} (\asdifor), which periodically trains from scratch a new \ifor ensemble on the most recent data and discards the old ensemble. The periodic retraining delays the adaptation to new data and, depending on how frequently it is performed, slows down the execution. In contrast, \onlineiforest seamlessly updates its internal structure at each new sample processed, enabling fast and truly online anomaly detection.
        \textit{Robust Random Cut Forest}~\cite{GuhaMishraAl16} (\rrcf) represents the first attempt to adapt \ifor to the streaming context. \rrcf dynamically manages tree structures, and introduces a novel anomaly score based on the discrepancy in tree complexity when a data point is removed from the stream. The anomaly score grounds on the assumption that anomalies' impact on tree structures is more evident compared to genuine data points. However, tree modifications performed by \rrcf tend to be resource-intensive compared to the fast bin splitting and aggregation of \onlineiforest.
        \textit{LODA}~\cite{Pevny16} leverages on the Johnson-Lindenstrauss~\cite{JohnsonLindenstrauss84} lemma to project data onto $1$-dimensional spaces and subsequently model data distributions via an ensemble of $1$-dimensional histograms. \loda makes use of fixed resolution histograms, which make it ineffective in describing multi-modal and complex data arrangements. Conversely \onlineiforest histograms, thanks to their ability to increase resolution, successfully adapt to various data configurations.
        \textit{Half Space Trees}~\cite{TanTingAl11} (\hst), in contrast to \loda, employs an ensemble of $d$-dimensional multi resolution histograms. Specifically, \hst builds an ensemble of complete binary trees by picking a random dimension and using the mid-point to bisect the space. Under the assumption that anomalous data points lie in sparse regions of the space, authors of \hst propose a score based on node masses. Similarly to \asdifor, \hst suffers of periodic retraining. Additionally, the bins in \hst histograms are generated without data, leading to high resolution in empty regions of the space and subsequent memory inefficiency.
        On the other hand, the data-dependent histograms of \onlineiforest allows for a more detailed description of data distribution in the most populous regions of the space.

    \section{\onlineisolationforest}
        \label{sec:online_isolation_forest}
        
        \onlineiforest, denoted as $\mathcal{F}$, is an ensemble of \onlineitrees $\{T_1,\\ \dots, T_\tau\}$ that we specifically designed to continuously and efficiently learn in streaming manner and adapt to the evolving data distribution inherent in streaming contexts.
        Each \onlineitree is a $d$-dimensional histogram that evolves its bins, both in terms of structure and the associated height, as it collects more information about the unknown distributions $\Phi_0$ and $\Phi_1$ over time.
        We rely on a sliding buffer $W = [\vect{x}_{t-\omega}, \dots, \vect{x}_t]$ containing the $\omega$ most recent points and, at each time instant $t$, we use the most and least recent points $\vect{x}_t$ and $\vect{x}_{t-\omega}$ from the buffer $W$ to respectively expand and contract the tree accordingly.
        
        \onlineiforest, detailed in~\cref{alg:online_iforest}, initializes the buffer $W$ as an empty list, and each tree $T \in \mathcal{F}$ is initially composed of the root node only. At each time step $t$, we get a new point $\vect{x}_t$ from the data stream (\cref{line:get_point}), store it in the sliding buffer $W$, and use it to update every base model $T$ within our ensemble (\cref{line:append_window,line:insert_point}). A tree $T$ learns the new point $\vect{x}_t$ by updating each bin along the path from the root to the corresponding leaf, and potentially expanding its tree structure as described by the learning procedure in~\cref{alg:online_itree_insert}. Subsequently, we remove the oldest point $\vect{x}_{t-\omega}$ from the buffer $W$ and force every tree $T$ to forget it (\cref{line:check_window,line:remove_point}) via the forgetting procedure that entails updating the bins along the leaf path. The forgetting step, in contrast to the learning one, involves a potential contraction of the tree structure instead of an expansion, as outlined in~\cref{alg:online_itree_remove}. We formalize and detail the learning and forgetting procedures in ~\cref{subsec:online_isolation_tree}.
        
        \begin{algorithm}[t]
            \caption{\onlineiforest}
            \label{alg:online_iforest}
            \DontPrintSemicolon
            \SetNoFillComment
            \KwIn{$\omega$  window size, $\tau$ - number of trees, $\eta$ - max leaf samples}
            initialize $W$ as empty list \label{line:initialize_window}\\
            initialize $\mathcal{F}$ as set of $\tau$ empty trees $\{T_1, \dots, T_{\tau}\}$ \label{line:initialize_forest}\\
            \While{true}
                {$\vect{x}_t \leftarrow$ get point from stream \label{line:get_point}\\
                 \begin{small}
                    \tcc{\textbf{Update forest} \!\!\!}
                 \end{small}
                 append $\vect{x}_t$ to $W$ \label{line:append_window}\\
                 \For{$i = 1$ \normalfont{to} $\tau$ \label{line:insert_loop}}
                     {\texttt{learn\_point}$(\vect{x}_t, T_i.rootN, \eta, \texttt{c}(\omega, \eta))$ \label{line:insert_point}}
                 \If{\normalfont{length of} $W$ \normalfont{greater than} $\omega$ \label{line:check_window}}
                    {$\vect{x}_{t-\omega} \leftarrow$ pop oldest point from $W$ \label{line:remove_window}\\
                     \For{$i = 1$ \normalfont{to} $\tau$ \label{line:remove_loop}}
                         {\texttt{forget\_point}$(\vect{x}_{t-\omega}, T_i.rootN, \eta)$ \label{line:remove_point}}}
                \begin{small}
                    \tcc{\textbf{Score point} \!\!\!}
                \end{small}
                $\mathcal{D} \leftarrow \emptyset$ \label{line:depth_init}\\
                \For{$i = 1$ \normalfont{to} $\tau$ \label{line:score_loop}}
                    {$\mathcal{D} \leftarrow \mathcal{D}$ $\cup$ \texttt{point\_depth}$(\vect{x}_t, T_i.rootN)$  \label{line:compute_depth}}
                $s_t \leftarrow 2^{-\frac{E(\mathcal{D})}{\texttt{c}(\omega, \eta)}}$ \label{line:compute_score}}
        \end{algorithm}
        
        We compute the anomaly score $s_t \in [0, 1]$ for the point $\vect{x}_t$ (\cref{line:depth_init,line:compute_score}) according to the principles behind \ifor~\cite{LiuTingAl08,LiuTingAl12}. In particular, anomalous points are easier to isolate than genuine ones, and therefore they are more likely to be separated early in the recursive splitting process of a random tree.
        Therefore, we determine the anomaly score by computing the depth of all the leaves where $\vect{x}_t$ falls in each tree $T \in \mathcal{F}$, and consequently mapping the depths to the corresponding $s_t$ via the following normalization function
        \begin{equation}
            s_t \leftarrow 2^{-\frac{E(\mathcal{D})}{\texttt{c}(\omega, \eta)}},
        \end{equation}
        where $E(\mathcal{D})$ is the average depth along all the $\tau$ trees of the ensemble, and $\texttt{c}(\omega, \eta) = \log_2\frac{\omega}{\eta}$ is an adjustment factor as a function of the window size $\omega$ and the number $\eta$ of points required to perform a bin split. The adjustment factor represents the average depth of an \onlineitree and we derive it in~\cref{subsec:complexity_analysis}.
        The computation of the leaf depth where $\vect{x}_t$ falls into is detailed in~\cref{alg:online_itree_depth}.

        \subsection{\onlineisolationtree}
            \label{subsec:online_isolation_tree}

            \begin{figure}[t]
                \begin{subfigure}[b]{0.225\textwidth}
                    \begin{center}
                    \centerline{\includegraphics[width=\columnwidth]{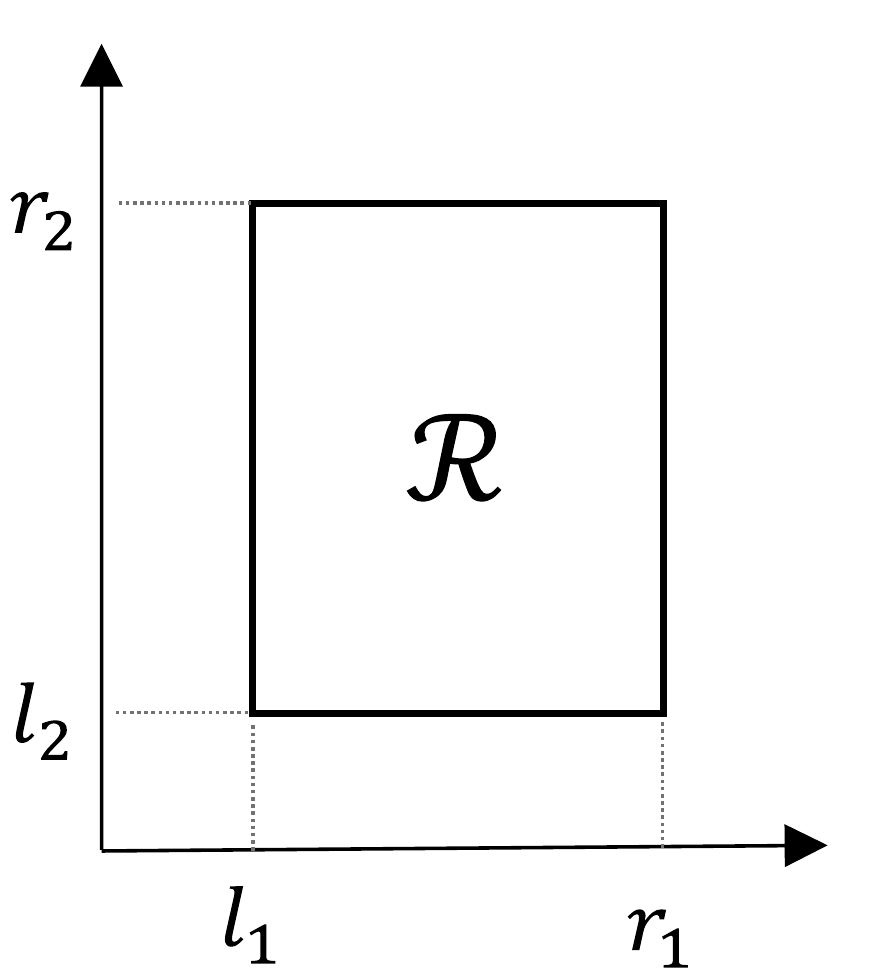}}
                    \caption{Bin support $\mathcal{R}$ and its boundaries $[l_i, r_i]$.}
                    \label{fig:support}
                    \end{center}
                \end{subfigure}
                \hfill
                \begin{subfigure}[b]{0.245\textwidth}
                    \begin{center}
                    \centerline{\vspace*{0.26cm} \includegraphics[width=\columnwidth]{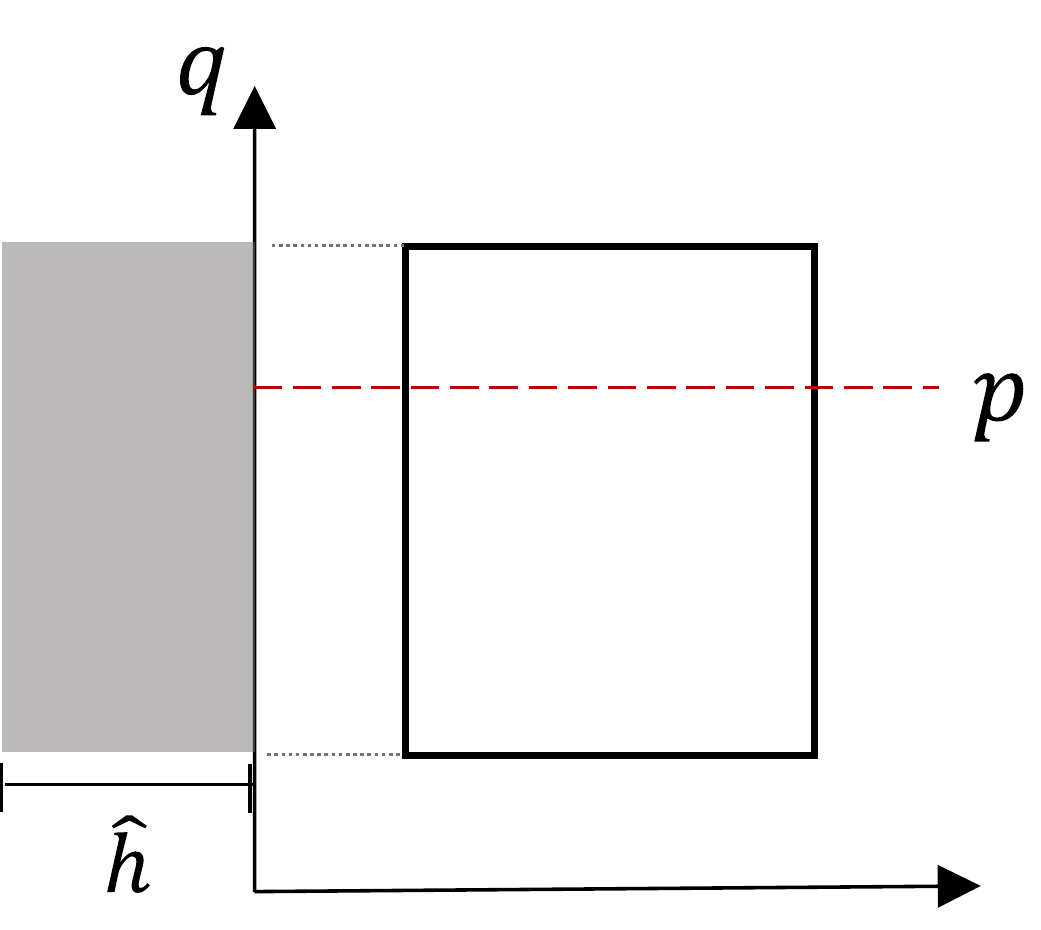}}
                    \caption{Maximum bin height $\widehat{h}$ and split information $q$ and $p$.}
                    \label{fig:split}
                    \end{center}
                \end{subfigure}
                \hfill
                \begin{subfigure}[b]{0.215\textwidth}
                    \begin{center}
                    \centerline{\vspace*{0.35cm} \includegraphics[width=\columnwidth]{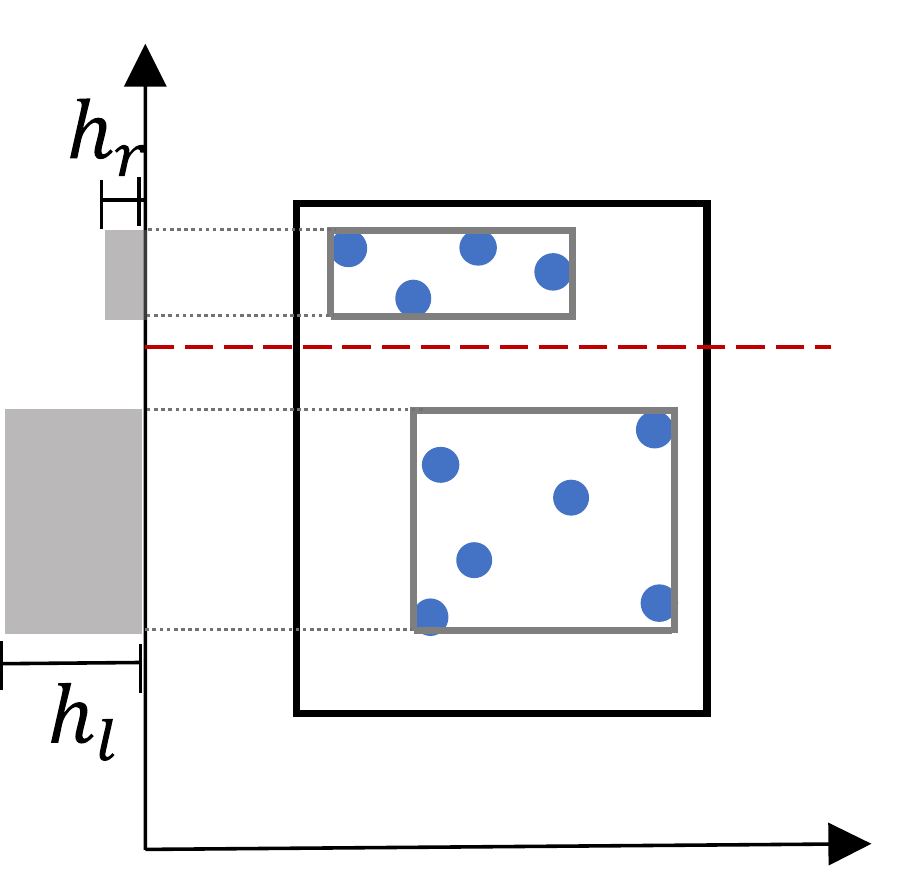}}
                    \caption{Sampled points $\mathcal{X}$ and new bins height $h_l$ and $h_r$.}
                    \label{fig:points}
                    \end{center}
                \end{subfigure}
                \hfill
                \begin{subfigure}[b]{0.185\textwidth}
                    \begin{center}
                    \centerline{\vspace*{0.35cm} \includegraphics[width=\columnwidth]{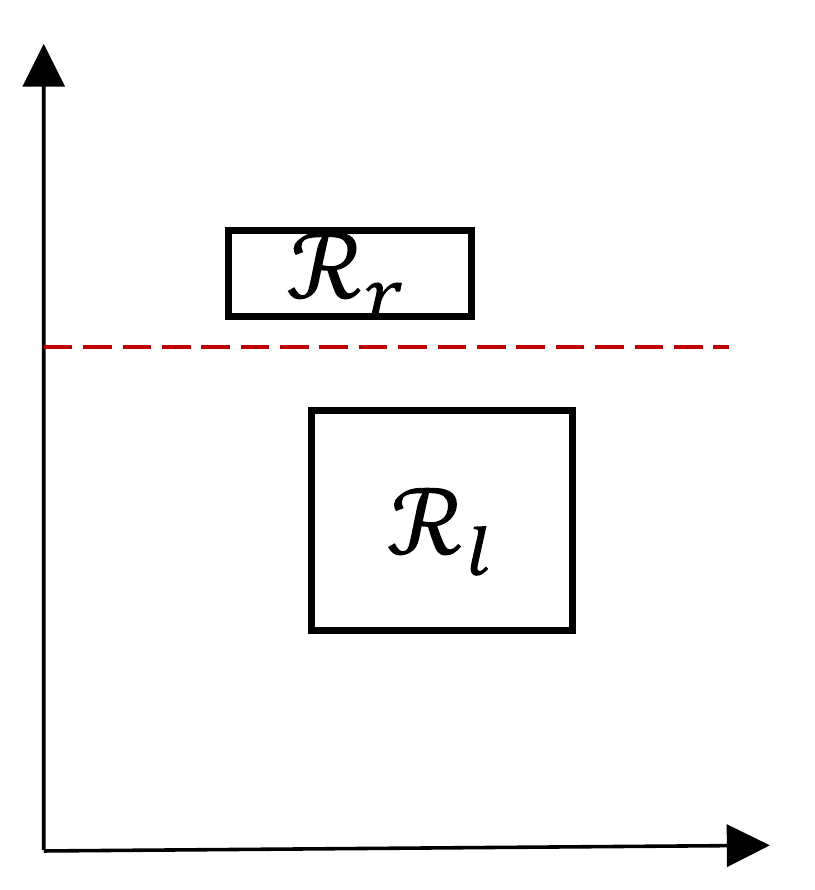}}
                    \caption{New bins support $\mathcal{R}_l$ and $\mathcal{R}_r$.}
                    \label{fig:new_support}
                    \end{center}
                \end{subfigure}
                \caption{Split procedure of a leaf node $N = (h, \mathcal{R})$ in a tree $T$. (b) As soon as a node $N$ reaches the maximum bin height $\widehat{h}$, we randomly select a dimension $q$ and a split value $p$. (c) We randomly sample a set $\mathcal{X}$ of $\widehat{h}$ points from the support $\mathcal{R}$ and use them to initialize bins of newborn child nodes $N_l = (h_l, \mathcal{R}_l)$ and $N_r = (h_r, \mathcal{R}_r)$.}
                \label{fig:split_procedure}
            \end{figure}

            The fundamental component of our solution is the \onlineitree structure. Every \onlineitree is a $d$-dimensional histogram constructed by recursively splitting the input space $\mathbb{R}^d$ into bins, such that each bin stores the number of points that fell in the corresponding region of the space. We define \onlineitree as a dynamic collection of nodes $T = \{N_j\}_{j = 1, \dots, m}$ that is continuously updated as new points are learned and old points are forgotten by the tree. We characterize the $j$-th node as $N_j = (h_j, \mathcal{R}_j)$, where $h_j$ is the number of points that crossed it in their path to the leaf, that is the bin height, and $\mathcal{R}_j = \bigtimes_{i = 1}^d [l_i, r_i]$ is the minimal $d$-dimensional hyperrectangle that encloses them, that is the support of the bin, where $\bigtimes$ denotes the Cartesian product.
            It is worth noting that $h_j$ and $\mathcal{R}_j$ are sufficient for achieving an efficient online adaptation of \ifor.

            When a new sample $\vect{x}_t$ is received from the data stream we run, independently on each \onlineitree, a learning procedure to update the tree. The learning procedure involves sending the incoming sample $\vect{x}_t$ to the corresponding leaf, and updating the heights $h$ and supports $\mathcal{R}$ of all the bins along the path accordingly.
            When a leaf reaches the maximum height $\widehat{h}$, we split the corresponding bin in two according to the procedure illustrated in~\cref{fig:split_procedure} and described next.
            This is repeated until the window $W$ gets full, then, together with the learning procedure for the new incoming sample $\vect{x}_t$, we include a forgetting procedure for the oldest sample $\vect{x}_{t-\omega}$ in $W$. The forgetting procedure might involve aggregating nearby bins in a single one as illustrated in~\cref{fig:forget_procedure}. The two fold updating mechanism enables \onlineitree to (i) incrementally learn when the stream starts and (ii) track possible evolution of the stream.
            
            \onlineitree bins associated to more populated regions of the space undergo frequent splits, whereas bins associated with sparsely populated regions undergo less frequent splits. Since each split increases the depth of the tree's leaf nodes, we can distinguish between anomalous and genuine points based on the depth $k$ of the leaf nodes they fall into. In~\cref{fig:oif_anomaly_scores} we see that an ensemble of \onlineitrees assigns different anomaly scores to regions with high and low population.

            \subsubsection{Learning procedure}
                \label{subsubsec:learning_procedure}

                \begin{figure}[t]
                    \begin{subfigure}[b]{0.49\textwidth}
                        \begin{center}
                        \centerline{\includegraphics[width=\columnwidth]{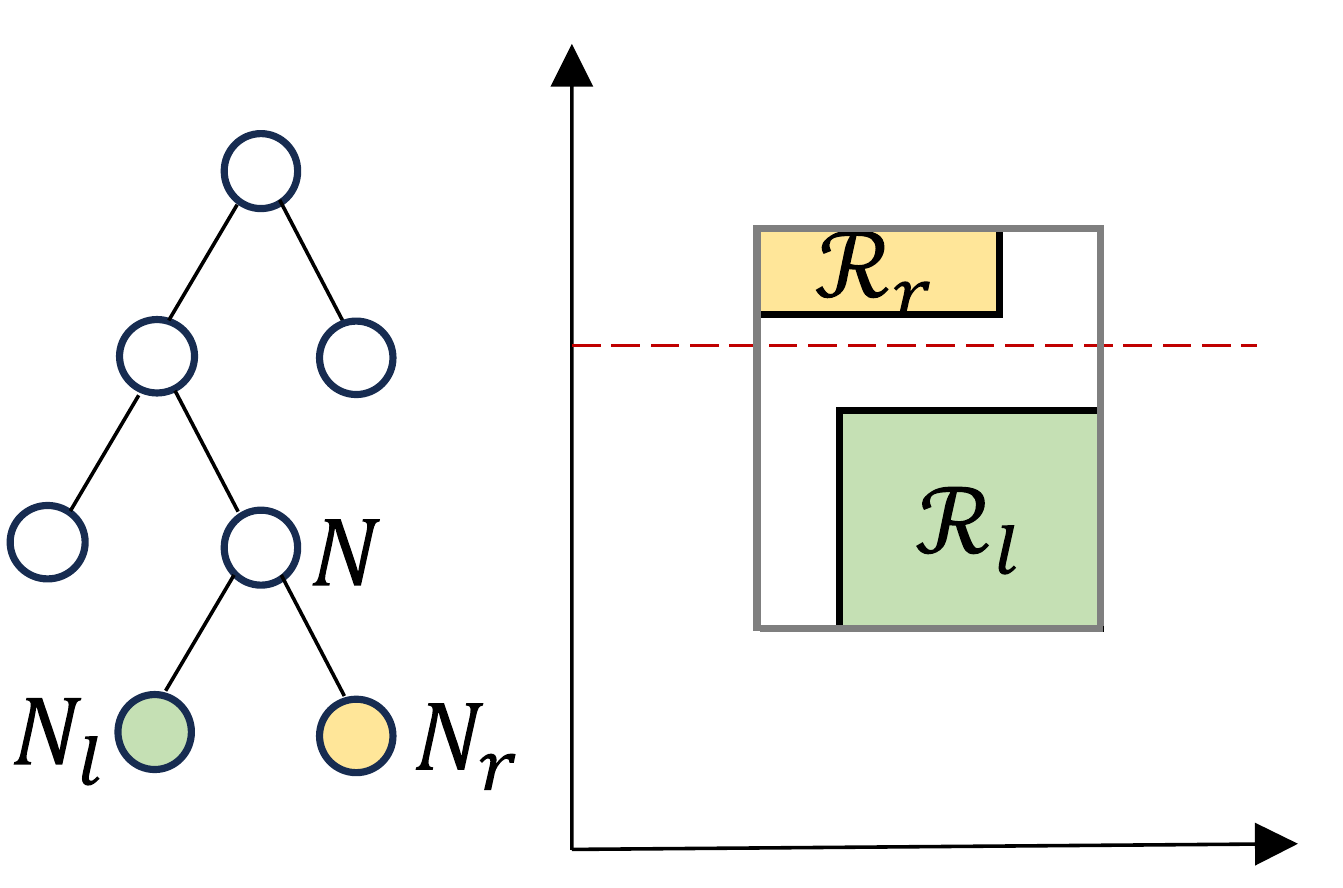}}
                        \caption{Before}
                        \label{fig:before_forget}
                        \end{center}
                    \end{subfigure}
                    \hfill
                    \begin{subfigure}[b]{0.49\textwidth}
                        \begin{center}
                        \centerline{\includegraphics[width=\columnwidth]{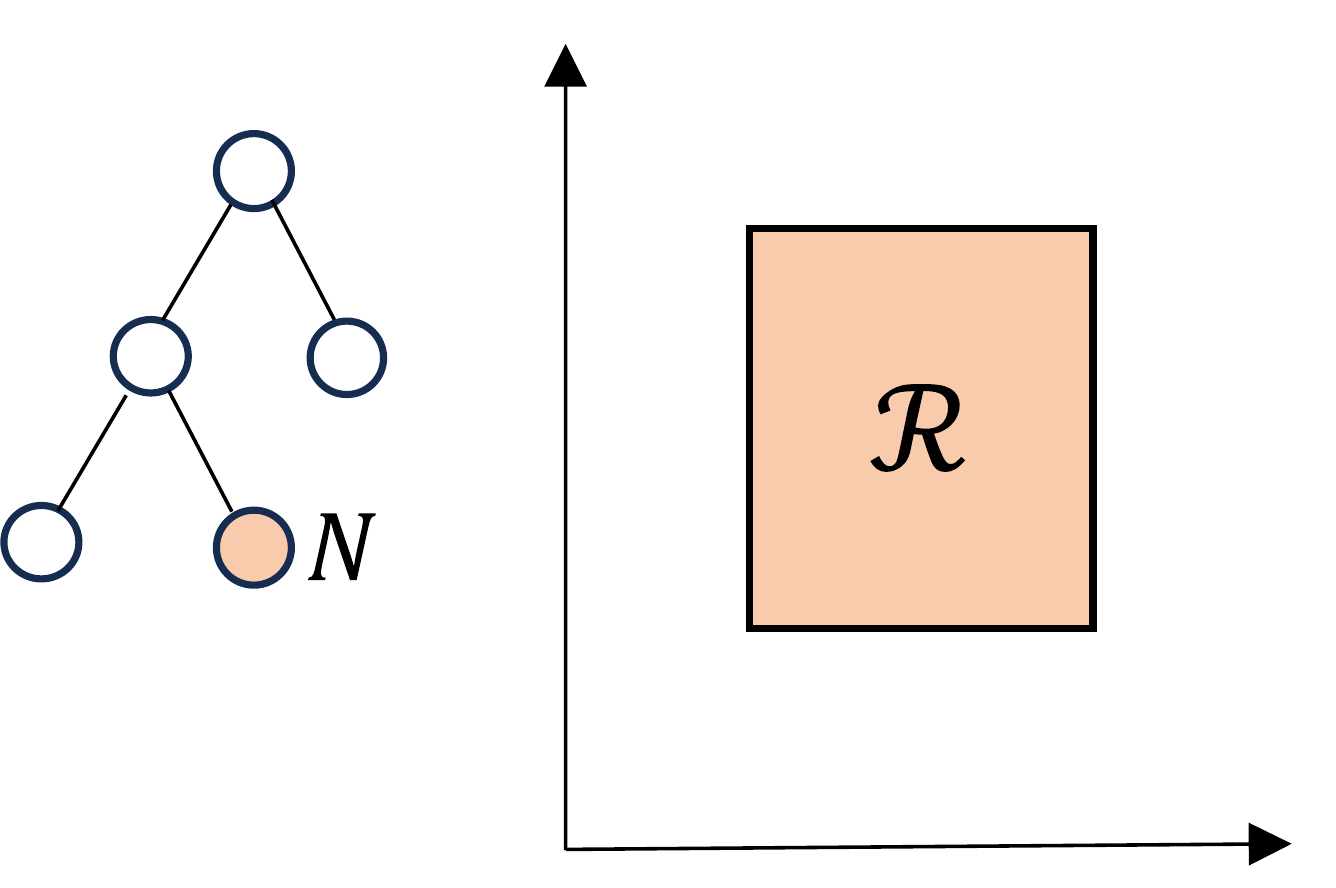}}
                        \caption{After}
                        \label{fig:after_forget}
                        \end{center}
                    \end{subfigure}
                    \caption{Following the forgetting procedure, support $\mathcal{R}$ of node $N$ is the minimal hyperrectangle that encloses supports $\mathcal{R}_l, \mathcal{R}_r$ of child nodes $N_l, N_r$.}
                    \label{fig:forget_procedure}
                \end{figure}

                \begin{algorithm}[t]
                    \caption{\onlineitree{} -- learn point}
                    \label{alg:online_itree_insert}
                    \DontPrintSemicolon
                    \SetNoFillComment
                    \KwIn{$\vect{x}$ - input data, $N$ - a tree node, $\eta$ - max leaf samples, $\delta$ - depth limit}
                    \SetKwFunction{FMain}{learn\_point}
                    \SetKwProg{Fn}{Function}{:}{}
                    \Fn{\FMain{$\vect{x}, N, \eta, \delta$}}{
                    update bin height $h$ and support $\mathcal{R}$ \label{line:update_bin} \\
                    \If{$N$ \normalfont{is a leaf}}
                        {\If{$h \geq \eta \: 2^{k}$ \normalfont{\textbf{and}} $k < \delta$ \label{line:split_condition}}
                            {$q \leftarrow$ sample from $\mathcal{U}_{\{1, \dots, d\}}$ \label{line:sample_dimension} \\
                             $p \leftarrow$ sample from $\mathcal{U}_{[l_q, r_q]}$ \label{line:sample_value} \\
                             $\mathcal{X} \leftarrow$ sample from $\mathcal{U}_{\mathcal{R}}$ \label{line:sample_points} \\
                             partition $\mathcal{X}$ into $\mathcal{X}_l$, $\mathcal{X}_r$ \label{line:split_points} \\
                             compute $h_l$, $h_r$ and $\mathcal{R}_l$, $\mathcal{R}_r$ \\
                             initialize child nodes $N_l, N_r$ \label{line:initialize_children} 
                         }
                       \Else
                        {\If{$x_q < p$}
                            {\texttt{learn\_point}$(\vect{x}, N_l, \eta, \delta)$}
                         \Else
                            {\texttt{learn\_point}$(\vect{x}, N_r, \eta, \delta)$}
                        }}}
                \end{algorithm}

                \begin{algorithm}[t]
                     \caption{\onlineitree{} -- forget point}
                     \label{alg:online_itree_remove}
                     \DontPrintSemicolon
                     \SetNoFillComment
                     \KwIn{$\vect{x}$ - input data, $N$ - a tree node, $\eta$ - max leaf samples}
                     \SetKwFunction{FMain}{forget\_point}
                     \SetKwProg{Fn}{Function}{:}{}
                     \Fn{\FMain{$\vect{x}, N, \eta$}}{
                     decrease bin height $h$ \label{line:update_bin_remove} \\
                     \If{$N$ \normalfont{is NOT a leaf}}
                             {\If{$h < \eta \: 2^{k}$ \label{line:forget_condition}}
                                 {update bin support $\mathcal{R}$ from $\mathcal{R}_l, \mathcal{R}_r$\label{line:update_support} \\
                                  forget split $q, p$ and child nodes $N_l, N_r$ \label{line:forget_split}
                                  }
                                  
                             \Else
                                 {\If{$x_q < p$}
                                     {\texttt{forget\_point}$(\vect{x}, N_l, \eta)$}
                                  \Else
                                     {\texttt{forget\_point}$(\vect{x}, N_r, \eta)$}
                                 }
                             }
                     }
                \end{algorithm}

                Every time we feed a point $\vect{x}_t$ to the tree via the learning procedure, we update both the height $h$ and the support $\mathcal{R}$ of all the nodes crossed by $\vect{x}_t$ along the path from the root to the leaf (\cref{line:update_bin} of \cref{alg:online_itree_insert}). When the bin height $h$ of a leaf node $N$ reaches a maximum value $\widehat{h}$, we increase the resolution of the histogram in the corresponding region of the space by splitting the associated bin in two, and generating two child nodes $N_l$ and $N_r$.
                
                We define the maximum height of a bin as $\widehat{h} = \eta \: 2^k$, where $\eta$ is a user-defined parameter, and $k$ is the depth of the corresponding node $N \in T$. Since the maximum height $\widehat{h}$ grows exponentially as a function of the depth $k$ of the considered node, we increase the number of points required to perform a split in deeper nodes of the tree. This design choice results in compact trees, leading to a substantial efficiency gain.
                We perform the split procedure only if the depth $k$ of the leaf node $N$ is less than a maximum value $\delta = \log_2\frac{\omega}{\eta}$, as a function of the window size $\omega$ and the number $\eta$ of points required to split histogram bins. Trees with limited depth had already been discussed in \ifor as a solution to the swamping and masking effects~\cite{Murphy51} in anomaly detection.
    
                The split procedure, invoked when the bin height $h$ reaches the maximum value $\widehat{h}$, is as follows. First, we randomly sample a split dimension $q \in \{1, \dots, d\}$ and split value $p \in [l_q, r_q]$ from the random variables $Q \sim \mathcal{U}_{\{1, \dots, d\}}$ and $P \sim \mathcal{U}_{[l_q, r_q]}$ respectively, where $\mathcal{U}$ denotes the uniform distribution (\cref{line:sample_dimension,line:sample_value} of \cref{alg:online_itree_insert}).
                Second, we sample a set $\mathcal{X}$ of $\widehat{h}$ points from the support $\mathcal{R} = \bigtimes_{i = 1}^d [l_i, r_i]$ such that each element $\vect{x} \in \mathcal{X}$ is distributed according to $\vect{X} \sim \mathcal{U}_{\mathcal{R}}$ (\cref{line:sample_points}). The uniform sampling of $\mathcal{X}$ grounds on the approximation of the data distribution by a piece wise uniform distribution represented by the union of histogram leaves' bins.
                Finally, we partition the elements of $\mathcal{X}$ into $\mathcal{X}_l = \{\vect{x} \in \mathcal{X} | x_q < p\}$ and $\mathcal{X}_r = \{\vect{x} \in \mathcal{X} | x_q \geq p\}$, and use them to initialize heights $h_l$, $h_r$ and supports $\mathcal{R}_l$, $\mathcal{R}_r$ of the newborn left and right child nodes $N_l$ and $N_r$ respectively (\cref{line:split_points,line:initialize_children}).
                For illustration purposes, we depicted the split procedure in~\cref{fig:split_procedure} when $d = 2$.

            \subsubsection{Forgetting procedure}
                \label{subsubsec:forgetting_procedure}

                \begin{algorithm}[t]
                    \caption{\onlineitree{} -- point depth}
                    \label{alg:online_itree_depth}
                    \DontPrintSemicolon
                    \SetNoFillComment
                    \KwIn{$\vect{x}$ - input data, $N$ - a tree node, $\eta$ - max leaf samples}
                    \KwOut{depth of point $\vect{x}$}
                    \SetKwFunction{FMain}{point\_depth}
                    \SetKwProg{Fn}{Function}{:}{}
                    \Fn{\FMain{$\vect{x}, N$}}{
                    \If{$N$ \normalfont{is a leaf}}
                       {\Return $k + \texttt{c}(h, \eta)$}
                    \Else
                        {\If{$x_q < p$}
                            {\Return \texttt{point\_depth}$(\vect{x}, N_l)$}
                         \Else
                            {\Return \texttt{point\_depth}$(\vect{x}, N_r)$}}
                    }
                \end{algorithm}

                In contrast to the learning procedure, which involves creating new nodes and thereby enhancing the histogram resolution in that area, in the forgetting procedure we aggregate nodes and merge the associated bins, ultimately reducing the number of bins and, hence, the histogram resolution in the corresponding region of the space. Specifically, every time we feed a point $\vect{x}_{t-\omega}$ to the tree via the forgetting procedure, we decrease the bin height $h$ of all the nodes $N$ crossed by it along the path from the root to the leaf (\cref{line:update_bin_remove} of~\cref{alg:online_itree_remove}). When the height $h$ of an internal node (\emph{i.e.}, of a node that experienced a split) drops below the threshold $\widehat{h}$, we forget the split in $N$ by merging its two child nodes $N_l$ and $N_r$.
                The forget procedure (illustrated in~\cref{fig:forget_procedure}) consists in first updating the bin support $\mathcal{R}$ of the node $N$ as the minimal hyperrectangle that encloses bin supports $\mathcal{R}_l$ and $\mathcal{R}_r$ (\cref{line:update_support} of \cref{alg:online_itree_remove}) of $N_l$ and $N_r$ respectively. Then split information $q, p$ and child nodes $N_l, N_r$ are discarded (\cref{line:forget_split}).

        \subsection{Complexity analysis}
            \label{subsec:complexity_analysis}

            The computational complexity is a crucial aspect in the online context, where data streams must be processed at high speed with low memory requirements. Time and space complexities of \onlineiforest are closely tied to the depth of the \onlineitrees within the ensemble. Therefore, we first derive \onlineitree depth in both the average and worst case scenarios, and then express time and space complexities of \onlineiforest as functions of these depths (\cref{tab:complexity}).
            
            \paragraph{Average case}
            To determine the average depth $\bar{k}$ of an \onlineitree, we first note that a perfectly balanced binary tree constructed with $\omega$ points has exactly $\frac{\omega}{2^{k}}$ points at each node at depth $k$~\cite{Knuth23}. Since a node at depth $k$ requires $\eta \: 2^{k}$ points to undergo a split in \onlineitree, we can state that depth $k$ exists if and only if
            \begin{equation}
                \frac{\omega}{2^{k-1}} \geq \eta \: 2^{k-1},
            \end{equation}
            \emph{i.e.}, if there were enough points at depth $k - 1$ to perform the split.
            Making explicit the inequality with respect to $k$ we have
            \begin{equation}
                k \leq \frac{1}{2} \: \log_2\frac{\omega}{\eta} + 1,
            \end{equation}
            from which it follows that the average depth of an \onlineitree is
            \begin{equation}
                \bar{k} = \lfloor \frac{1}{2} \: \log_2\frac{\omega}{\eta} + 1 \rfloor.
            \end{equation}
            Hence, the average \textit{time} complexity of \onlineiforest (\emph{i.e.}, the computational complexity of traversing each tree from the root to a leaf), is $O(n \: \tau \: \log_2\frac{\omega}{\eta})$, where $\tau$ it the number \onlineitrees in the ensemble and $n$ is the number of samples in the data stream.
            We express the average \textit{space} complexity of \onlineiforest (\emph{i.e.}, the amount of memory space required) as a function of the number of nodes in an \onlineitree, that is in turn tied to the average depth $\bar{k}$, plus the buffer size $\omega$. Specifically, we note that the number of nodes in a perfectly balanced binary tree with depth $k$ is $2^{k+1} - 1$~\cite{Knuth23}. Therefore, the average space complexity of \onlineiforest is $O(\tau \: 2^{\frac{1}{2}\log_2\frac{\omega}{\eta}} + \omega) = O(\tau \sqrt{\frac{\omega}{\eta}} + \omega)$, and it is independent from the number $n$ of samples in the data stream.

            \begin{table}[t]
                \caption{Average and worst case complexity of \onlineiforest.}
                \begin{center}
                    \begin{sc}
                    \begin{tabular}{l@{\hskip 2.75cm}c@{\hskip 2.75cm}c}
                    \toprule
                    Complexity & Average case                                  & Worst case                             \\
                    \midrule
                    Time       & $O(n \: \tau \: \log_2\frac{\omega}{\eta})$   & $O(n \: \tau \: \log_2\frac{w}{\eta})$ \\
                    Space      & $O(\tau \sqrt{\frac{\omega}{\eta}} + \omega)$ & $O(\tau \frac{\omega}{\eta} + \omega)$ \\
                    \bottomrule
                    \end{tabular}
                    \end{sc}
                \end{center}
                \label{tab:complexity}
            \end{table}

            \paragraph{Worst case}
            We note that a binary tree degenerated into a linked list, constructed with $\omega$ points, has $\omega - k$ points at depth $k$. Therefore, similarly to the average case, depth $k$ exists if and only if
            \begin{equation}
                \omega-(k-1) \geq \eta \: 2^{k-1}.
            \end{equation}
            By making the depth $k$ explicit, and placing an upper bound on it, we have
            \begin{equation}
                k \leq \log_2\frac{w+1-k}{\eta} + 1 \leq \log_2\frac{w+1}{\eta} + 1
            \end{equation}
            from which it follows that the worst case depth of an \onlineitree is
            \begin{equation}
                 \tilde{k} \leq \lfloor \log_2\frac{w+1}{\eta} + 1 \rfloor.
            \end{equation}
            Thus, the worst case \textit{time} and \textit{space} complexities of \onlineiforest are $O(n \: \tau \: \log_2\frac{w}{\eta})$ and $O(\tau \: 2^{\log_2\frac{w+1}{\eta}} + \omega) = O(\tau \frac{\omega}{\eta} + \omega)$ respectively.

        \subsection{Adaptation Speed vs. Modeling Accuracy}
            \label{subsec:sliding_buffer}

            The length $\omega$ of the sliding buffer $W$ plays a crucial role in controlling the trade-off between adaptation speed and modeling accuracy in \onlineiforest. Specifically, adopting a small $\omega$ allows \onlineiforest to quickly adapt to changes, but it results in a coarse modeling of the underlying data distribution. To this regard, we can observe~\cref{fig:oif_anomaly_scores}, illustrating the anomaly scores as \onlineiforest processes an increasing number of points. Moving from left to right, the anomaly scores describe the learned data distribution after processing $100$, $300$ and $1000$ points, and this is equivalent to what \onlineiforest would learn over sliding buffers of corresponding lengths. Notably, after processing $100$ points, the anomaly scores are coarse, whereas they become more fine-grained after $1000$ points.

    \section{Experiments}
        \label{sec:oif_experiments}

        In this section we first compare the performance of \onlineiforest and state-of-the-art methods on a large anomaly detection benchmark where anomalous and genuine distributions $\Phi_0$ and $\Phi_1$ are stationary. Then, we resort to a dataset exhibiting concept drift to assess their capability to adapt to distribution changes.

        \subsection{Datasets}
            \label{subsec:oif_datasets}

            \begin{table}[t]
                \caption{Stationary datasets properties.}
                \begin{center}
                    \begin{sc}
                    \begin{tabular}{l@{\hskip 1.25cm}r@{\hskip 1.25cm}r@{\hskip 1.25cm}r@{\hskip 1.25cm}r}
                    \toprule
                    Dataset              & $n$      & $d$  & $\%$ of anomalies \\
                    \midrule
                    Donors               & $619326$ & $10$ & $5.90$            \\
                    Http                 & $567497$ & $3$  & $0.40$            \\
                    ForestCover          & $286048$ & $10$ & $0.90$            \\
                    fraud                & $284807$ & $29$ & $0.17$            \\
                    Mulcross             & $262144$ & $4$  & $10.00$           \\
                    Smtp                 & $95156$  & $3$  & $0.03$            \\
                    Shuttle              & $49097$  & $9$  & $7.00$            \\
                    Mammography          & $11183$  & $6$  & $2.00$            \\
                    nyc\_taxi\_shingle   & $10273$  & $48$ & $5.20$            \\
                    Annthyroid           & $6832$   & $6$  & $7.00$            \\
                    Satellite            & $6435$   & $36$ & $32.00$           \\
                    \bottomrule
                    \end{tabular}
                    \end{sc}
                \end{center}
                \label{tab:datasets}
            \end{table}

            \paragraph{Stationary}
            We run our experiments on the eight largest datasets used in~\cite{LiuTingAl08,LiuTingAl12} (Http, Smtp~\cite{YamanishiTakeuchiAl04}, Annthyroid, Forest Cover Type, Satellite, Shuttle~\cite{AsunctionNewman07}, Mammography and Mulcross~\cite{RockeWoodruff96}), two datasets from Kaggle competitions (Donors and Fraud~\cite{PangShenAl19}), and the shingled version of NYC Taxicab dataset used in~\cite{GuhaMishraAl16}.
            We chose these datasets as they contain real data where the genuine and anomalous distributions $\Phi_0$ and $\Phi_1$ are unknown, and contain labels about anomalous data to perform performance evaluation.
            Information on the cardinality $n$, dimensionality $d$, and $\%$ of anomalies for the datasets is outlined in~\cref{tab:datasets}.

            \paragraph{Non-stationary}
            We use the INSECTS dataset~\cite{SouzaDosReisAl20} previously used for change detection~\cite{FrittoliCarreraAl22,StucchiMagriAl23,StucchiRizzoAl23}. INSECTS contains feature vectors ($d = 33$) describing the wing-beat frequency of six (annotated) species of flying insects. This dataset contains $5$ real changes caused by temperature modifications that affect the insects' flying behavior. For our purposes, we selected the two most populous classes as genuine (`ae-aegypti-female', `cx-quinq-male'), and the least populous as the anomalous one (`ae-albopictus-male'). This results in a total of $n = 212514$ points with $5.50\%$ of anomalies.

        \subsection{Competing methods and methodology}
            \label{subsec:competing_methods_and_methodology}

            In our experiments we compared \onlineiforest (\oifor) to state-of-the-art methods in the online anomaly detection literature described in~\cref{sec:oad_related_literature}. In particular, we compared to \textit{iForestASD} (\asdifor), \textit{Half Space Trees} (\hst), \textit{Robust Random Cut Forest} (\rrcf) and \textit{LODA} (\loda) using their \textit{PySAD}~\cite{YilmazKozat20} implementation.
            
            For comparison purposes, we set the number of trees $\tau = 32$ for all the algorithms, and considered the number of random cuts in \loda equivalent to the number of trees. The low number of trees is a reasonable choice given the computational constrains of online methods. In fact, while this choice may affect result accuracy and stability~\cite{ChabchoubTogbeAl22,AzzariBicego24}, the memory and time costs grow with the number of trees $\tau$, making it necessary to balance performance and efficiency in online scenarios where both resources are strictly limited.
            
            We set window size $\omega = 2048$ for both \oifor and \asdifor, and used the default value $\omega = 250$ for \hst. We set the subsampling size used to build trees in \asdifor to the default value $\psi = 256$, while the number of bins for each random projection in \loda to $b = 100$. The trees maximum depth $\delta$ depends on the subsamping size $\psi$ in \asdifor, on the window size $\omega$ and number $\eta$ of points required to split histogram bins in \oifor, while it is fixed to the default value $\delta = 15$ in \hst. The parameters configuration for all the algorithms is illustrated in~\cref{tab:parameters}.

            Each algorithm was executed $30$ times on both stationary and non-stationary datasets. We randomly shuffled every stationary dataset before each execution, then used the same shuffled version to test all the algorithms.
            Processing each data point individually is prohibitive in terms of time due to the data stream size. To solve this problem we divided every dataset in batches of $100$ points each and passed one chunk at a time to the algorithms in an online manner.

            We use the ROC AUC and execution time (in seconds) to evaluate the effectiveness and efficiency of the considered algorithms, respectively. In addition, we employ the critical difference diagram for both metrics to synthesize the results across multiple executions on datasets with diverse characteristics.

            \begin{table}[t]
                \vskip 0.15in
                \caption{Algorithms execution parameters.}
                \begin{center}
                    \begin{sc}
                    \begin{tabular}{l@{\hskip 1.1cm}c@{\hskip 1.1cm}c@{\hskip 1.1cm}c@{\hskip 1.1cm}c@{\hskip 1.1cm}c@{\hskip 1.1cm}c}
                    \toprule
                    Algorithm     & $\tau$ & $\omega$ & $\eta$ & $\psi$ & $\delta$       & $b$   \\
                    \midrule
                    \oifor        & $32$   & $2048$   & $32$   & --     & $\log_2\frac{\omega}{\eta}$ & --    \\
                    \asdifor      & $32$   & $2048$   & --     & $256$  & $\log_2\psi$   & --    \\
                    \hst          & $32$   & $250$    & --     & --     & $15$         & --    \\
                    \rrcf         & $32$   & --       & --     & $256$  & --             & --    \\
                    \loda         & $32$   & --       & --     & --     & --             & $100$ \\
                    \bottomrule
                    \end{tabular}
                    \end{sc}
                \end{center}
                \vskip -0.1in
                \label{tab:parameters}
            \end{table}

        \subsection{Results and Discussion}
            \label{subsec:pif_results}
            
            \begin{table}[t]
                \caption{ROC AUCs (a) and total execution times (b).}
                \begin{subtable}{\textwidth}
                    \raggedright
                    \caption{}
                    \begin{sc}
                    \tabcolsep=0.15cm
                    \begin{tabular}{l@{\hskip 0.5cm}r@{\hskip 0.5cm}r@{\hskip 0.5cm}r@{\hskip 0.85cm}r@{\hskip 0.7cm}r@{\hskip 0.625cm}r}
                                                 && \multicolumn{5}{c}{AUC $(\uparrow)$}                                         \\
                                                    \cmidrule{3-7}                                                               
                                                 && oIFOR            & asdIFOR          & HST       & RRCF             & LODA    \\
                        \midrule
                        Donors                   && $\mathbf{0.795}$ & $0.769$          & $0.715$   & $0.637$          & $0.554$ \\
                        Http                     && $0.998$          & $\mathbf{0.999}$ & $0.992$   & $0.996$          & $0.632$ \\
                        ForestCover              && $0.887$          & $0.861$          & $0.722$   & $\mathbf{0.917}$ & $0.500$ \\
                        fraud                    && $0.936$          & $0.946$          & $0.910$   & $\mathbf{0.951}$ & $0.722$ \\
                        Mulcross                 && $\mathbf{0.995}$ & $0.952$          & $0.011$   & $0.800$          & $0.506$ \\
                        Smtp                     && $0.861$          & $\mathbf{0.905}$ & $0.851$   & $0.894$          & $0.731$ \\
                        Shuttle                  && $0.992$          & $\mathbf{0.996}$ & $0.981$   & $0.957$          & $0.528$ \\
                        Mammography              && $0.854$          & $\mathbf{0.855}$ & $0.831$   & $0.824$          & $0.622$ \\
                        nyc\_taxi\_shingle       && $0.572$          & $0.709$          & $0.342$   & $\mathbf{0.725}$ & $0.499$ \\
                        Annthyroid               && $0.685$          & $\mathbf{0.810}$ & $0.636$   & $0.740$          & $0.589$ \\
                        Satellite                && $0.651$          & $\mathbf{0.709}$ & $0.531$   & $0.662$          & $0.501$ \\
                        \midrule
                        median                   && $\mathbf{0.866}$ & $0.863$          & $0.739$   & $0.832$          & $0.541$ \\
                        mean rank                && $2.167$          & $\mathbf{1.583}$ & $3.917$   & $2.500$          & $4.833$ \\
                        \bottomrule
                    \end{tabular}
                    \end{sc}
                \end{subtable}
                \\
                \vfill
                \begin{subtable}{\textwidth}
                    \raggedright
                    \caption{}
                    \begin{sc}
                    \tabcolsep=0.15cm
                    \begin{tabular}{l@{\hskip 0.5cm}r@{\hskip 0.5cm}r@{\hskip 0.5cm}r@{\hskip 0.5cm}r@{\hskip 0.5cm}r@{\hskip 0.5cm}r}
                                                 && \multicolumn{5}{c}{Time $(\downarrow)$}                          \\
                                                    \cmidrule{3-7}
                                                 && oIFOR             & asdIFOR  & HST       & RRCF      & LODA      \\
                        \midrule
                        Donors                   && $\mathbf{252.36}$ & $551.85$ & $2145.85$ & $4924.46$ & $2111.09$ \\
                        Http                     && $\mathbf{179.36}$ & $509.40$ & $2016.00$ & $8367.16$ & $2017.85$ \\
                        ForestCover              && $\mathbf{107.65}$ & $197.82$ & $1045.39$ & $2997.86$ & $1009.92$ \\
                        fraud                    && $\mathbf{100.09}$ & $285.69$ & $973.91$  & $4936.03$ & $931.93$  \\
                        Mulcross                 && $\mathbf{90.33}$  & $270.79$ & $936.01$  & $3244.96$ & $848.12$  \\
                        Smtp                     && $\mathbf{29.95}$  & $142.65$ & $325.77$  & $1273.98$ & $254.69$  \\
                        Shuttle                  && $\mathbf{16.35}$  & $108.28$ & $167.48$  & $770.61$  & $130.61$  \\
                        Mammography              && $\mathbf{3.32}$   & $80.01$  & $37.92$   & $118.22$  & $29.55$   \\
                        nyc\_taxi\_shingle       && $\mathbf{8.03}$   & $83.15$  & $36.70$   & $151.67$  & $36.82$   \\
                        Annthyroid               && $\mathbf{2.00}$   & $77.06$  & $24.26$   & $93.40$   & $19.79$   \\
                        Satellite                && $\mathbf{3.74}$   & $78.90$  & $21.78$   & $93.77$   & $17.55$   \\
                        \midrule
                        median                   && $\mathbf{29.95}$  & $142.57$ & $323.29$  & $1274.45$ & $254.64$  \\
                        mean rank                && $\mathbf{1.000}$  & $2.667$  & $3.500$   & $5.000$   & $2.833$   \\
                        \bottomrule
                    \end{tabular}
                    \end{sc}
                \end{subtable}
                \label{tab:performance}
            \end{table}

            \begin{figure}[t]
                \hspace{-1.35cm}
                \begin{subfigure}[b]{.6\columnwidth}
                    \begin{center}
                    \centerline{\includegraphics[width=\columnwidth]{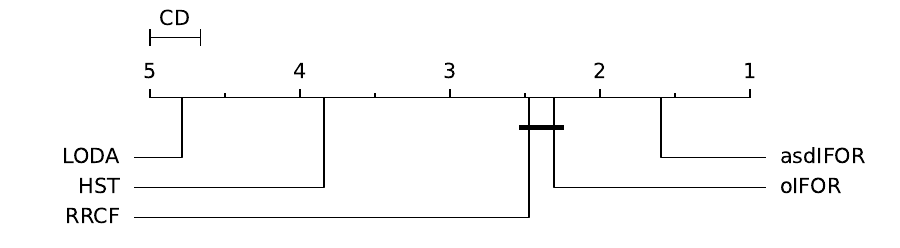}}
                    \caption{ROC AUCs.}
                    \label{fig:cd_roc_aucs}
                    \end{center}
                \end{subfigure}
                \begin{subfigure}[b]{.6\columnwidth}
                    \begin{center}
                    \centerline{\includegraphics[width=\columnwidth]{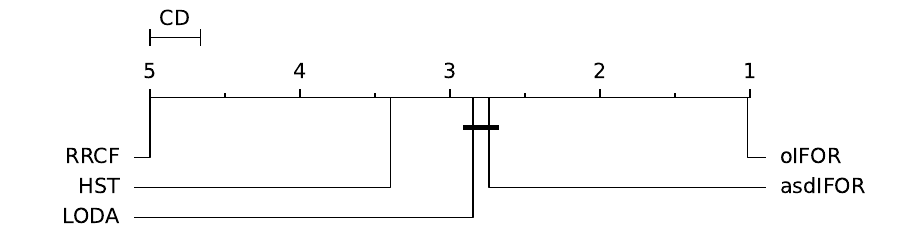}}
                    \caption{Times.}
                    \label{fig:cd_times}
                    \end{center}
                \end{subfigure}
                \caption{Critical difference diagram for ROC AUCs and total execution times.}
                \label{fig:cd}
            \end{figure}

            \begin{figure}[t]
                \begin{center}
                \centerline{\includegraphics[width=.9\columnwidth]{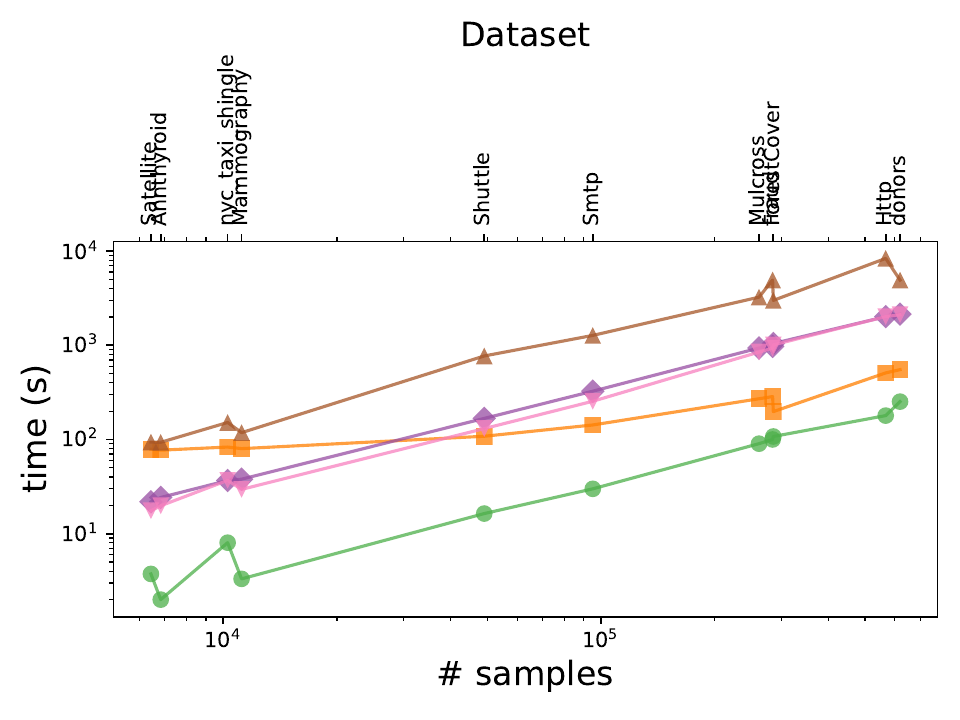}}
                \centerline{\hspace{0.95cm} \includegraphics[width=.7\linewidth]{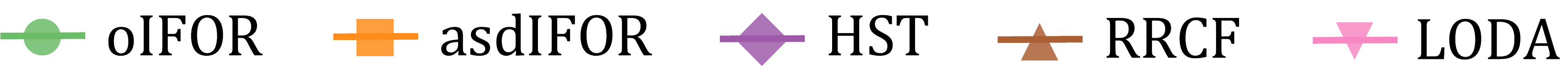}}
                \caption{Total execution times ordered by dataset size.}
                \label{fig:times}
                \end{center}
            \end{figure}

            \subsubsection{Anomaly detection in stationary data streams}
                \label{subsubsec:stationary}
                
                In~\cref{tab:performance} we show the median ROC AUC for each algorithm after processing each dataset, as well as the median total execution time for each algorithm to process the entire data stream at hand.
                The last two rows represent the median value and the mean rank among the total $330$ executions. We highlight the best row-wise result in bold.
                
                \oifor exhibits, by far, the lowest time complexity with respect to all the competitors at hand. In particular, when we compare \oifor to the second best method (\asdifor), we can notice that \oifor execution time is less than half in $4$ out of $5$ biggest datasets, and it is reduced by more than an order of magnitude in $4$ out of $6$ smallest ones. This result is confirmed by the critical difference diagram presented in~\cref{fig:cd_times}, which shows that \oifor is statistically better than all the others, while \loda and \asdifor are statistically equivalent. The statistical analysis has been conducted for $5$ populations (one for each algorithm) with $330$ paired total execution times. Critical difference diagrams are based on the post-hoc Nemenyi test, and differences between populations are significant when the difference of their mean rank is greater than the critical distance $CD = 0.336$. Mean ranks and critical diagrams have been generated via \textit{Autorank}~\cite{Herbold20} library.
                In~\cref{fig:times} we sorted the median total execution times listed in~\cref{tab:performance} by dataset size, and we can appreciate the linear trend exhibited by all the algorithms.
    
                \cref{tab:performance} shows that \oifor, \asdifor and \rrcf exhibit the best detection performance over different datasets. While \oifor exhibits a slightly higher overall median ROC AUC, the Nemenyi test highlights that \asdifor is the most effective algorithm, and that \oifor and \rrcf are statistically equivalent (\cref{fig:cd_roc_aucs}).
                The exceptionally low performance of \hst on the Mulcross dataset in~\cref{tab:performance} is due to the fact that the size of anomaly clusters is large and that anomaly clusters have an equal or higher density compared to genuine ones in that dataset. This scenario, combined with the high default maximum depth value $\delta$, makes this situation particularly difficult for \hst to handle, as it is based on the opposite assumptions.
                
                \paragraph{Learning in the early stages of a data stream}
                In addition to the conventional evaluation of online anomaly detection algorithms based on their final anomaly detection capability, we highlight the initial learning speed demonstrated by various methods. We focused to the early stages of the data stream, aiming to comprehend how the various methods learn in a critical phase such as the initial one, and analyzed their performance within the first $1000$ samples of the stream.
                Solid curves in~\cref{fig:roc_auc_initial} show the median ROC AUC of each algorithm over all the $30$ executions and the $11$ datasets, for a total of $330$ runs. We computed the ROC AUCs at each time instant $\dot{t}$ using the scores from $t=1$ to $t=\dot{t}$. All the algorithms show a fast learning speed, since within the first $1000$ samples all of them get very close to the final median performance showed in~\cref{tab:performance}.
                The corresponding efficiency is shown in~\cref{fig:times_initial}, where all the algorithms exhibit similar trends, with \oifor being the fastest to adapt.

            \subsubsection{Anomaly detection in non-stationary data streams}
                \label{subsubsec:non_stationary}

                \begin{figure}[t]
                    \centerline{\includegraphics[width=\linewidth]{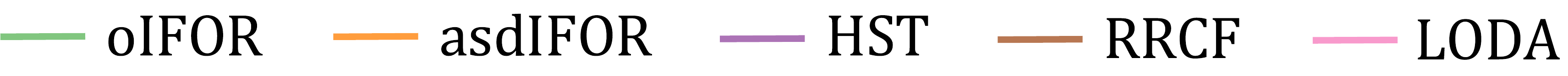}}
                    \begin{subfigure}[t]{\columnwidth}
                        \begin{center}
                        \centerline{\includegraphics[width=.8\columnwidth]{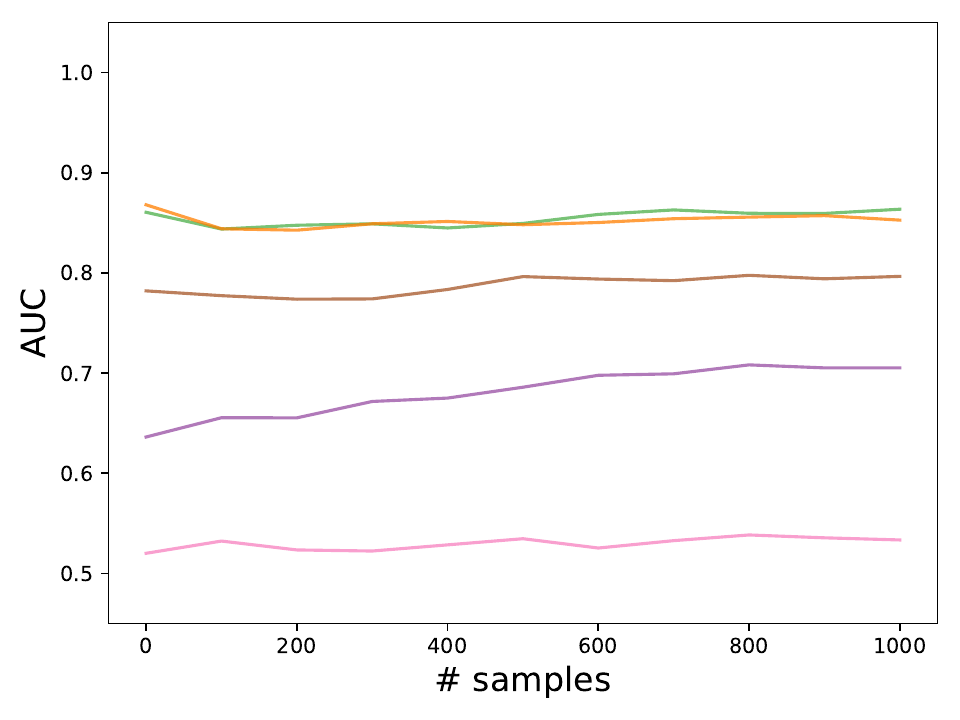}}
                        \caption{ROC AUCs.}
                        \label{fig:roc_auc_initial}
                        \end{center}
                    \end{subfigure}
                    \\
                    \vfill
                    \begin{subfigure}[t]{\columnwidth}
                        \begin{center}
                        \centerline{\includegraphics[width=.8\columnwidth]{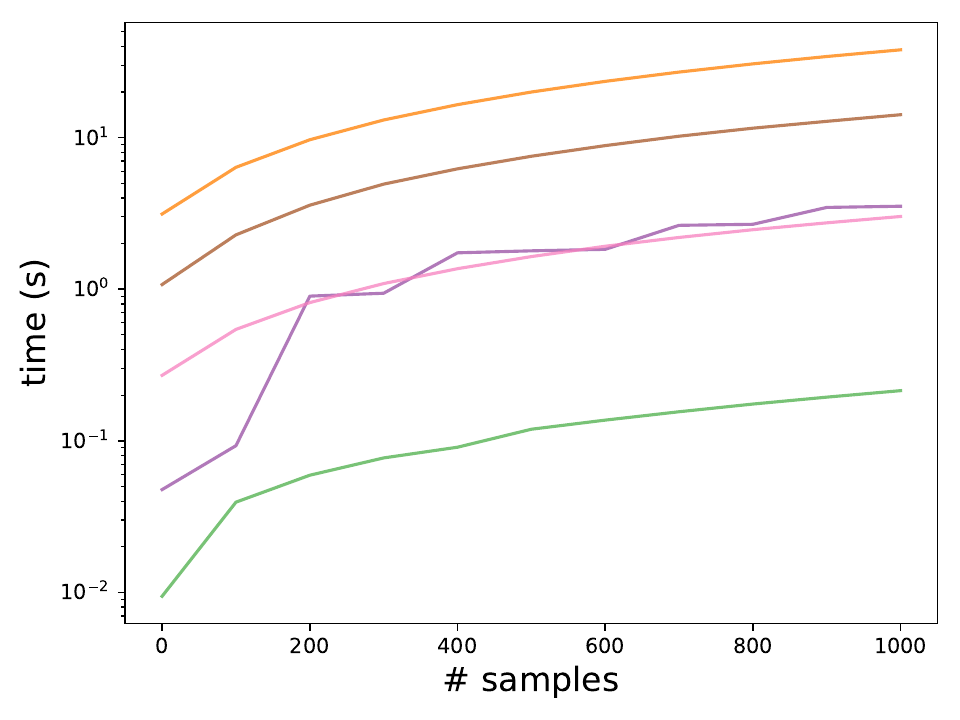}}
                        \caption{Times.}
                        \label{fig:times_initial}
                        \end{center}
                    \end{subfigure}
                    \caption{Evolution of the median ROC AUCs and total execution times within the first $1000$ samples of the stream.}
                    \label{fig:performance_initial}
                \end{figure}

                \begin{figure}[t]
                    \begin{center}
                    \centerline{\includegraphics[width=\linewidth, center]{pictures/online_anomaly_detection/online_isolation_forest/experiments/real/legend.pdf}}
                    \centerline{\includegraphics[width=\linewidth]{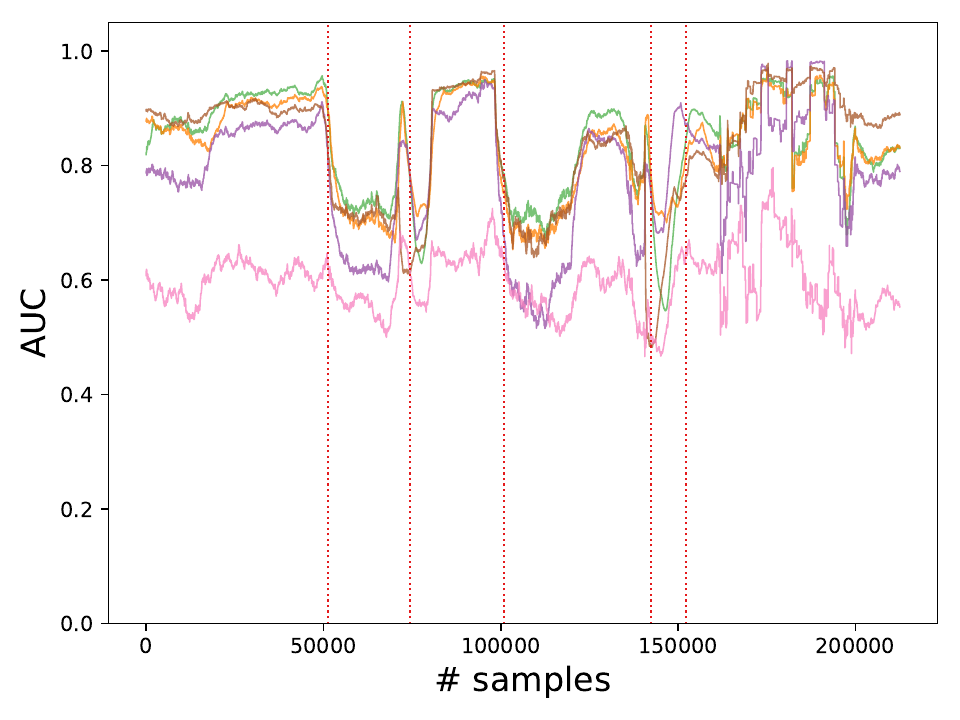}}
                    \caption{The performance of online anomaly detection methods is significantly influenced by changes in data distributions $\Phi_0$ and $\Phi_1$.}
                    \label{fig:convoluted_aucs}
                    \end{center}
                \end{figure}

                In~\cref{fig:convoluted_aucs} we show the median ROC AUC of each algorithm over $30$ executions on the INSECTS dataset. We are interested in investigating the instantaneous anomaly detection performance of all the algorithms, therefore we computed the ROC AUCs at each time instant $\dot{t}$ using the scores within a window of size $5000$ centered in $\dot{t}$, \emph{i.e.}, from $t=\dot{t}-2500$ to $t=\dot{t}+2500$. The choice of $5000$ for window size was made to guarantee that the resulting curves exhibit a satisfactory degree of smoothness.
                Vertical dotted lines represent the time instants when the change in distributions occur.
                \cref{fig:convoluted_aucs} shows that all the tested algorithms are affected in a similar way by sudden changes in the genuine and anomalous distributions $\Phi_0$ and $\Phi_1$, and we cannot identify a method that consistently maintains performance after a change. Although \loda is less affected by changes compared to the others, the overall low performance indicates that \loda struggles in learning the underlying distributions.
                The execution times of the algorithms are not influenced by distribution changes, and \onlineiforest remains the fastest option.

    \section{Conclusions}
        \label{sec:oad_conclusions}

        In this chapter, we presented \onlineiforest, an anomaly detection algorithm specifically designed for the streaming scenario. \onlineiforest is an ensemble of histograms that dynamically adapt to the data distribution keeping only statistics about data points. Thanks to a sliding window, \onlineiforest is able to selectively forget old data points and update histograms accordingly.
        Extensive experiments showed that \onlineiforest features an extremely fast processing and learning speed while maintaining effectiveness comparable to that of state-of-the-art methods.
        The intuitive operational approach, coupled with its high speed, positions \onlineiforest as a good candidate for addressing real-world streaming anomaly detection challenges.

        As future work, we aim to remove the sliding window $W$ while retaining the forgetting capabilities of \onlineiforest. Additionally, we seek to automate the selection of the number $\eta$ of points required to split histogram bins.

        \chapter{Open-Set Recognition for Malware Family Discovery}
    \label{cha:open_set_recognition_malware_family_discovery}

    In this chapter, we tackle the anomaly detection problem of identifying new malware families in Android applications by analyzing the permissions they request. While many malware classes have already been identified and categorized, the widespread popularity of the Android operating system motivates malware developers to continuously craft new malicious applications to gain access to sensitive data or system resources.
    Therefore, we frame malware family discovery as an \emph{Open-set recognition} (\osr) problem, and address it by combining the tree-based \gboost classifier~\cite{Friedman01,HastieTibshiraniAl09} with \maxlogit~\cite{VazeHanAl21}, a popular \osr technique. Most \osr methods are designed for neural network-based classifiers and, to the best of our knowledge, they have not been extended to tree-based classifiers. We show that \maxlogit can be seamlessly integrated into a classification pipeline based on boosted decision trees without affecting the workflow. Experiments on both a public and a private dataset, provided by \cleafy, show that our approach is promising and worthy of further investigation. Additionally, we have successfully deployed our approach in \cleafy's business environment, where it is currently part of their engine.

    In~\cref{sec:mfd_background}, we provide an overview of malware detection, in~\cref{sec:mfd_problem_formulation} we frame the problem of malware family discovery in the \osr settings, and in~\cref{sec:mfd_related_literature} we review the related literature. In~\cref{sec:open_set_recognition_with_gradient_boosting}, we introduce our framework, detailing the \logits extraction process (\cref{subsec:logit_extraction}) and the \maxlogit computation (\cref{subsec:maxlogit_computation}). Finally, in~\cref{sec:osr_experiments} we describe our experimental setup and present our classification and \osr results.

    \section{Background}
        \label{sec:mfd_background}

        Malware detection approaches are typically based on heuristic rules, and can be divided in static, such as signature based and permission based, and dynamic~\cite{OdusamiAbayomi-AlliAl18}. Dynamic analysis involves observing application behavior at run time within a sandbox environment, and monitoring system calls to construct a function call graph which is consequently analyzed to identify malicious behaviour.
        Some dynamic approaches employ machine learning algorithms to classify a malware, usually trained using altered function call graphs as input~\cite{HassenChan17}. Depending on the classification model, this approach can easily integrate with open-set recognition techniques to discover new malware families~\cite{HassenChan20,JiaChan22}. Despite the high potential of dynamic analysis, the runtime testing required to construct the function call graph makes it inefficient and unsuitable for high throughput scenarios.
        
        In signature based approaches, unique identifiers of a known malware is generated and stored in a database as a model, and any application to be tested is compared to it and eventually flagged as malware~\cite{FarukiLaxmiAl15}. Although signature based analysis proves efficient, it allows the identification of malware families that are already stored in the database as models, making it a not appropriate method to identify new ones.
        
        Permission analysis, on the other hand, consists in flagging applications as potentially harmful based solely on the permissions they request from the operating system~\cite{RovelliVigfusson14}. In this work we focus on permission analysis as it is more flexible compared to signature based analysis and more efficient compared to dynamic analysis.

        \medskip
        The permissions required by Android applications are outlined within a file named AndroidManifest. This file is mandatory for every Android application and contains essential information about the application, including permissions.
        Permissions can be classified into two categories: custom and system. Custom permissions don't require access to sensitive data such as contacts or filesystem. Conversely, system permissions, totaling more than $1500$, encompass all permissions exposed by the system, with only a subset of these (the most sensitive ones) requiring explicit user approval. While system permissions are largely standardized, certain manufacturers introduce specific ones based on device peculiarities. Our study focuses solely on the latter, as they have the potential to result in adverse consequences for the user.

        The permission extraction process entails the following steps: extracting permissions from the manifest file, filtering out custom permissions, and applying one-hot encoding to system permissions enabling malware analysis through a machine learning model. One-hot encoding is a common preprocessing method, where each permission is represented by a binary value ($1$ for requested, $0$ for not requested). This results in each application being represented by a high-dimensional binary vector of length $P$, where $P$ is the total number of permissions. Despite this preprocessing, the limited number of samples typically available in real-world malware classification scenarios (small $n$), coupled with the high-dimensional feature space (large $P$) poses challenges for effectively training machine learning classification models.

    \section{Problem Formulation}
        \label{sec:mfd_problem_formulation}

        A malicious application is represented by a vector of permissions $\vect{p} \in \{0, 1\}^P$, where $P$ is the total number of permissions considered, and $p_i = 1$ if the $i$-th permission is listed in the application manifest.
        These malicious applications might either belong to a known class $\ell \in \mathcal{L}$ indicating a known malware family, or to a novel family that had never been observed before.

        Our goal is to train an open-set classifier $\mathcal{K}$ that associates to each malicious application $\vect{p}$ either a known class label or the \emph{Novel} label, \emph{i.e.}:
        \begin{equation}
            \label{eq:osr_classifier}
            \mathcal{K}(\vect{p}) =
            \begin{cases}
                \textit{Novel}\\
                \widehat{\ell}(\vect{p}) \in \mathcal{L}
            \end{cases}.
        \end{equation}
        We assume that we are provided with a training set $\mathcal{TR} = \{(\vect{p}_i, \ell_i) \;|\; \ell_i \in \mathcal{L}\}_{i = 1, \dots, n}$ of annotated malicious applications belonging to known families and with a test set $\mathcal{TS} = \{(\vect{p}_i, \ell_i) \;|\; \ell_i \in \mathcal{L} \cup \{Novel\}\}_{i = 1, \dots, m}$ of annotated malicious applications belonging to both known and novel families.

    \section{Related Literature}
        \label{sec:mfd_related_literature}

        Using the manifest file permissions alongside machine learning models for malware classification is a well-established practice. For instance, in~\cite{RovelliVigfusson14,HerronGlissonAl21}, the permission vector serves as behavioral marker and it is fed to machine learning models to classify potentially harmful applications. The machine learning algorithms used range from Support Vector Machines and Gaussian Naive Bayes to Random Forests~\cite{Aggarwal15,HanPeiAl22,Breiman01}. Their findings showed that, compared to existing anti-virus engines, all machine learning models trained on the manifest file permissions alone yielded significant improvements in classification rates, with Random Forest having the highest performance.
        In~\cite{TurnipSitumorangAl20} they propose to use a tree-based \gboost~\cite{ChenGuestrin16} model to classify six classes of malware based on three permission categories, highlighting that boosting is particularly suited for Android malware classification.
        All these methods operate as closed-set classifiers, namely they are restricted to classify a malware into already known classes, and are therefore unable to discover new families.

        \medskip
        An effective machine learning malware detection system should both accurately classify a known malware and identify novel, unknown malware families. This challenge, known as \osr~\cite{BendaleBoult16}, is typically addressed by assessing the classifier's degree of uncertainty in its classifications.
        \osr systems can be categorized in two types. The first type distinguishes between instances of known classes and unknown ones, but does not differentiate within known classes~\cite{ScheirerRochaAl12,BodesheimFreytagAl13,BodesheimFreytag15}. This kind of \osr approach, also referred to as anomaly detection, does not address known malware classification which is a primary requirement in our setting. Conversely, the second type of \osr systems can both discriminate known classes as well as identify instances of unknown ones~\cite{JainScheirerAl14,BendaleBoult15,BendaleBoult16,DaYuAl14,GeDemyanovAl17}. An example of the latter type is Open-Set Nearest-Neighbor (\osnn)~\cite{MendesDeSouzaAl17}, which extends a $1$-NN classifier to address the \osr task. \osnn computes the ratio of the distances between a sample and its two nearest neighbors from different families and classifies the sample as unknown if the ratio falls below a specified threshold. Although straightforward, \osnn relies on Euclidean distances, which are not meaningful when dealing with high-dimensional binary data, resulting in suboptimal performance for the \osr task.
        
        The \osr problem has been extensively studied within the computer vision community~\cite{BendaleBoult16,GeDemyanovAl17,LeeLeeAl18,NealOlson18,ChenQiaoAl20,SunYang20,ZhangLi20,VazeHanAl21,CardosoFrancaAl15}, and most of the proposed solutions build on top of a convolutional neural network classifier.
        Neural networks have also been used for \osr in malware detection. In~\cite{HassenChan20,JiaChan22}, they first run applications in a sandbox environment to extract function call graphs~\cite{Ryder79} and rearrange them into adjacency matrices. Subsequently, they train a convolutional neural network on those matrices using different loss functions to learn a discriminative representation of malware families within a latent space. During testing, they identify an instance as unknown if the distance to the closest centroid in the latent space exceeds a precomputed threshold.
        Despite the effectiveness of these methods, the extraction of function call graphs is time consuming, making them impractical for high throughput scenarios.
        
        Alternative input representations other than function call graphs, as well as diverse ways to extend existing closed-set intrusion detection systems for \osr has been proposed~\cite{RieckTriniusAl11,RuddRozsa16}.
        In~\cite{ZhouWang12} they use a permission-based footprinting scheme to detect samples of known Android malware families and then apply a heuristics-based filtering scheme to detect specific behaviors exhibited by unknown malicious families. If an application is classified as malicious and does not appear in the database, they consider it as a novel malware and generate the corresponding permission-based footprint in a feedback loop. However, their heuristic-based approach targets only specific Android features that may be exploited to load new code, either Java binary code or native machine code, thus limiting the ability to detect different types of an unknown malware.
        
        Our proposed approach combines the effectiveness of \gboost in classifying Android malware programs based on manifest file permissions, with the \maxlogit \osr technique, originally developed within the computer vision community. Our approach classifies malware instances into known families while detecting unknown ones by relying exclusively on permission analysis, making it faster than methods that require function call graph extraction and manipulation. Additionally, it is entirely data-driven, eliminating reliance on human-crafted heuristics that may introduce biases and limit its classification potential.

    \section{Open-set Recognition with Gradient Boosting}
        \label{sec:open_set_recognition_with_gradient_boosting}

        Our proposed open-set classifier $\mathcal{K}$, depicted in~\cref{fig:open_set_classifier}, consists of a closed-set classifier $\mathcal{C}$ and an open-set recognition module $\mathcal{O}$. The closed-set classifier assigns to each malicious application $\vect{p}$ a known class label, expressed as $\mathcal{C}(\vect{p}) = \widehat{\ell}(\vect{p}) \in \mathcal{L}$. The open-set recognition module $\mathcal{O}(\vect{z}_{\vect{p}}) \in \{Novel, Not\;novel\}$ determines whether $\vect{p}$ belongs to a $novel$ (unknown) or $not$ $novel$ (known) class based on the raw output values $\vect{z}_{\vect{p}}$ assigned to $\vect{p}$ by the classifier $\mathcal{C}$. Therefore, we can restate~\eqref{eq:osr_classifier} in the following way:
        \begin{equation}
            \label{eq:osr_classifier_new}
            \mathcal{K}(\vect{p}) =
            \begin{cases}
                \textit{Novel} \quad \textnormal{if} \; \mathcal{O}(\vect{z}_{\vect{p}}) = Novel\\
                \widehat{\ell}(\vect{p}) \quad\;\;\textnormal{otherwise}
            \end{cases}.
        \end{equation}
        In our approach we employ a tree-based \gboost~\cite{HastieTibshiraniAl09} as closed-set classifier, that is a set of boosted classification trees $\mathcal{C} = \{T_i\}_{i = 1, \dots, t}$, and \maxlogit~\cite{VazeHanAl21} as open-set scoring module $\mathcal{O}$, that is a threshold $\tau \in \mathbb{R}$ set on the classifier's raw output values $\vect{z}_{\vect{p}}$ to ensure a certain false alarm rate.

        \subsection{\logits extraction}
            \label{subsec:logit_extraction}

            \begin{figure}[t]
                \centering
                \includegraphics[width=\linewidth]{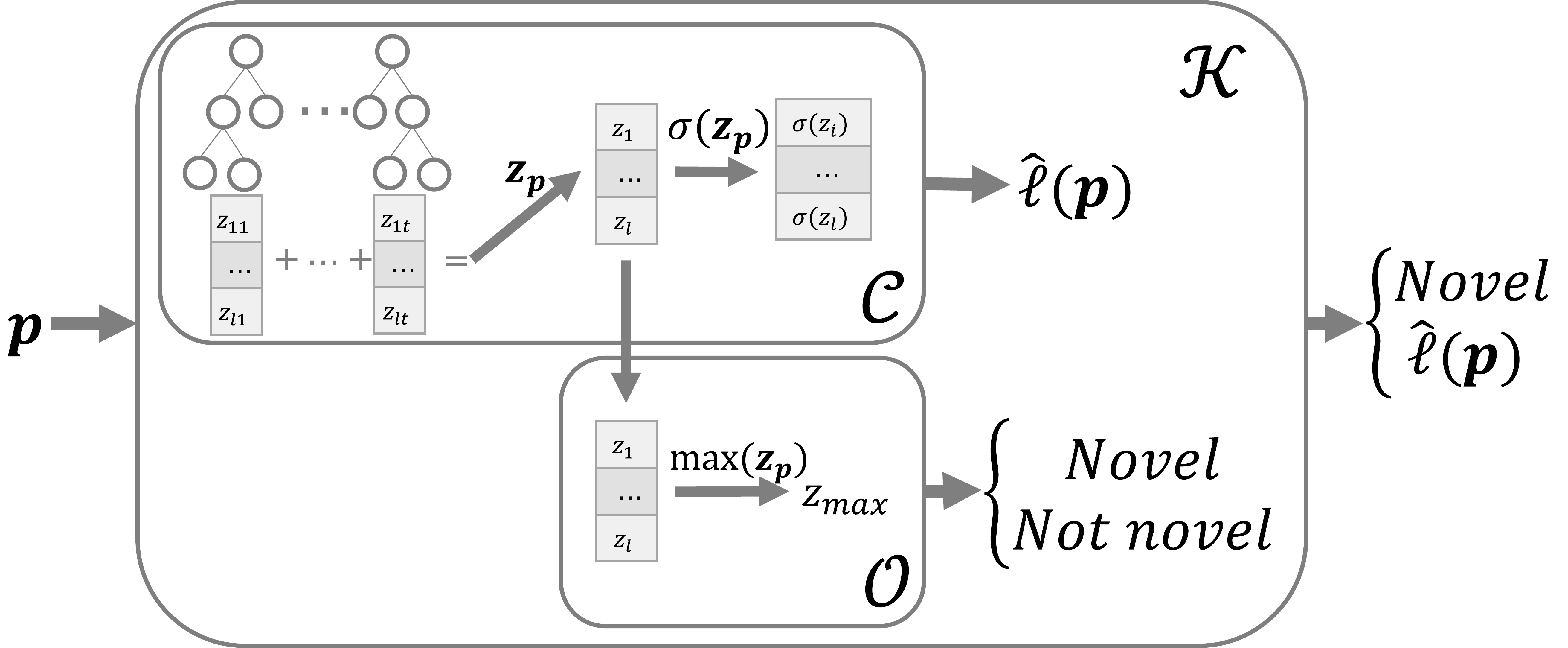}
                \caption{Depiction of our proposed open-set classifier $\mathcal{K}$.}
                \label{fig:open_set_classifier}
            \end{figure}
    
            \begin{figure}[t]
                \centering
                \begin{subfigure}[t]{\linewidth}
                    \centering
                    \includegraphics[width=\linewidth]{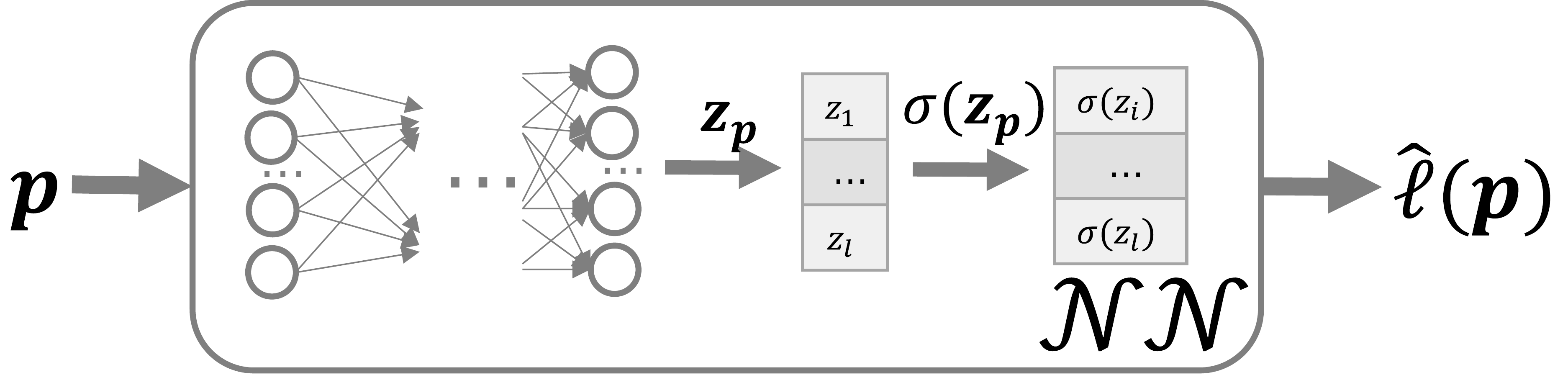}
                    \caption{}
                    \label{fig:nn_maxlogit}
                \end{subfigure}
                \hfill
                \begin{subfigure}[t]{\linewidth}
                    \centering
                    \includegraphics[width=\linewidth]{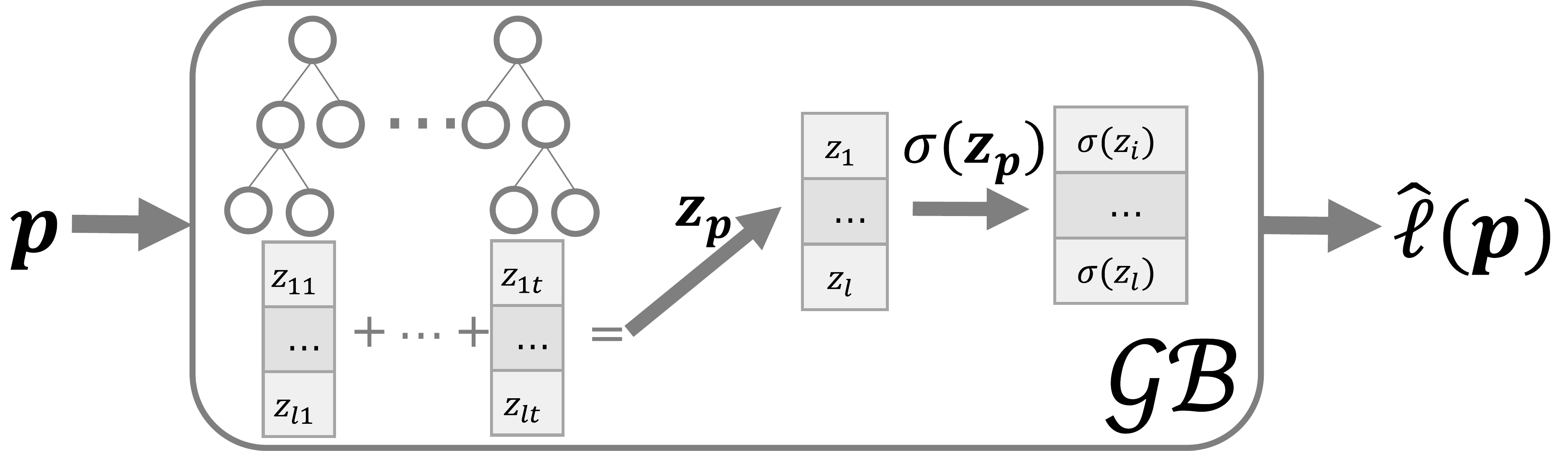}
                    \caption{}
                    \label{fig:gb_maxlogit}
                \end{subfigure}
                \caption{Analogy between the \logits vector in a neural network-based classifier (a) and the decision values in a tree-based \gboost classifier (b). In both cases, the vector $\vect{z}_{\vect{p}}$ represents the raw output values before applying the softmax function $\sigma(\vect{z}_{\vect{p}})$.}
                \label{fig:maxlogit}
            \end{figure}

            \maxlogit has been originally developed within the computer vision community and has traditionally been coupled with neural network-based classifiers. In this context, \maxlogit operates on the raw output scores $\vect{z}_{\vect{p}} = [z_1, \dots, z_l]$ generated by the last layer of the neural network, known as the \logits, before the softmax function. In~\cref{fig:nn_maxlogit} we illustrated a neural network classifier, where $\vect{p}$ is the input point and $\widehat{\ell}(\vect{p})$ is the predicted label.
            Since the softmax does not change the order of the scores in $\vect{z}_{\vect{p}}$, the predicted class $\widehat{\ell}(\vect{p})$ corresponds to the entry having the maximum value within the probability vector $\sigma(\vect{z}_{\vect{p}}) = [\sigma(z_1), \dots, \sigma(z_l)]$, which denotes the output of the softmax function to each element $z_i$ of the \logits vector $\vect{z}_{\vect{p}}$, \emph{i.e.}:
            \begin{equation}
                \label{eq:softmax}
                \sigma(z_i) = \frac{e^{z_i}}{\sum_{j = 1}^{l}e^{z_j}}.
            \end{equation}
            \maxlogit approach operates directly on the raw output values $\vect{z}_{\vect{p}}$ because \logits display clearer separation between known and unknown classes when compared to normalized probability values $\sigma(\vect{z}_{\vect{p}})$~\cite{VazeHanAl21}.
            Unfortunately, neural networks are not well suited for handling high-dimensional sparse binary data. Hence, we opted for tree-based \gboost as the classification algorithm due to its effectiveness in managing such data. To the best of our knowledge, this is the first time that the \maxlogit approach is used in conjunction with classifiers other than those based on neural networks.
            
            \gboost classifier doesn't have proper \logits, but it produces decision values, which are essentially the raw, unnormalized scores assigned to each class by the ensemble model before being transformed into probabilities (\cref{fig:gb_maxlogit}). Our intuition is to use \gboost decision values as the \logits vector produced by neural networks.
            In \gboost, in order to obtain the final prediction $\widehat{\ell}(\vect{p})$ for a given input sample $\vect{p}$, decision values are aggregated across all the trees in the ensemble, resulting in a final vector $\vect{z}_{\vect{p}} = [z_1, \dots, z_l]$. Each $z_i = \sum_{j = 1}^{t} z_{ij}$ is obtained by summing the decision values for the $i$-th label along all the trees, as illustrated in~\cref{fig:gb_maxlogit}. Subsequently, similarly to neural network classifiers, the softmax~\eqref{eq:softmax} is applied to $\vect{z}_{\vect{p}}$, and the class corresponding to the highest probability value in $\sigma(\vect{z}_{\vect{p}})$ is assigned as the predicted class for the input sample $\vect{p}$.

        \subsection{\maxlogit computation}
            \label{subsec:maxlogit_computation}

            The \maxlogit open-set recognition module $\mathcal{O}$ operates as a binary classifier, employing the maximum \logit value obtained from $\mathcal{C}$ to distinguish between $novel$ and $not$ $novel$ classes.
            When a sample $\vect{p}$ is put into the \gboost classifier $\mathcal{C}$, we extract the \logits vector $\vect{z}_{\vect{p}} = [z_1, \dots, z_l]$ and identify the maximum value $\max(\vect{z}_{\vect{p}}) = \max(z_1, \dots, z_l)$. To decide whether a sample belongs to a novel class, we apply a threshold $\tau$ to $\max(\vect{z}_{\vect{p}})$ and, if $\max(\vect{z}_{\vect{p}}) < \tau$, it indicates that the classifier is not confident in its classification, thereby we classify $\vect{p}$ as $novel$:
            \begin{equation}
                \label{eq:osr_module_new}
                \mathcal{O}(\vect{z}_{\vect{p}}) =
                \begin{cases}
                    \textit{Novel} \qquad\quad \textnormal{if} \; \max(\vect{z}_{\vect{p}}) < \tau\\
                    \textit{Not novel} \quad\;\textnormal{otherwise}
                \end{cases}.
            \end{equation}

            The value of the threshold $\tau$ is fundamental to control false alarms raised by the \osr module, \emph{i.e.}, the amount of samples belonging to $not$ $novel$ malware families classified as instances of a $novel$ family. Since the classification of a sample $\vect{p}$ as belonging to a new family leads to a subsequent manual inspection by an human expert, the tuning of threshold $\tau$ is essential to avoid waste in human time resources.
            Threshold $\tau$ is typically tuned using an external training set $\mathcal{TT} = \{(\vect{p}_i, \ell_i) \;|\; \ell_i \in \mathcal{L}\}_{i = 1, \dots, k}$, which is composed only of samples belonging to known classes. This is due to the fact that we focus on false positive rate, which is solely impacted by misclassifications of known samples as novel.
            In particular, given a trained closed-set classifier $\mathcal{C}$, an external training set $\mathcal{TT}$ and desired false positive rate $FPR \in [0, 1]$, we set the threshold $\tau$ to ensure that:
            \begin{equation}
                P(\max(\vect{z}_{\vect{p}}) < \tau \;|\; \vect{p} \in \mathcal{TT}) \leq FPR
            \end{equation}
            where $P$ is the probability symbol. In practice, it is enough to set the threshold $\tau$ equal to the $FPR$-quantile of the empirical distribution of $\max(\vect{z}_{\vect{p}})$ computed from all the elements of $\mathcal{TT}$.

            \medskip
            We remark that our method does not add any complexity to the \gboost closed-set classifier $\mathcal{C}$. The open-set recognition module $\mathcal{O}$ is seamlessly integrated into the classification workflow, preserving an inference time complexity of $O(t \: d)$ per sample, where $t$ is the number of trees in the ensemble and $d$ is the maximum tree depth.

    \section{Experiments}
        \label{sec:osr_experiments}

        In this section, we assess the benefits of our open-set recognition solution in identifying new malware families on both a publicly available dataset and a private dataset provided by \cleafy. We first introduce the dataset in~\cref{subsec:osr_dataset}, and detail the experimental methodology in~\cref{subsec:osr_methodology}. Subsequently, we validate our approach for both closed-set and open-set recognition on the considered datasets, and discuss its performance within \cleafy's real-world deployment settings in~\cref{subsec:osr_results}.

        \subsection{Dataset}
            \label{subsec:osr_dataset}

            \begin{figure}[t]
                \centering
                \includegraphics[width=\linewidth]{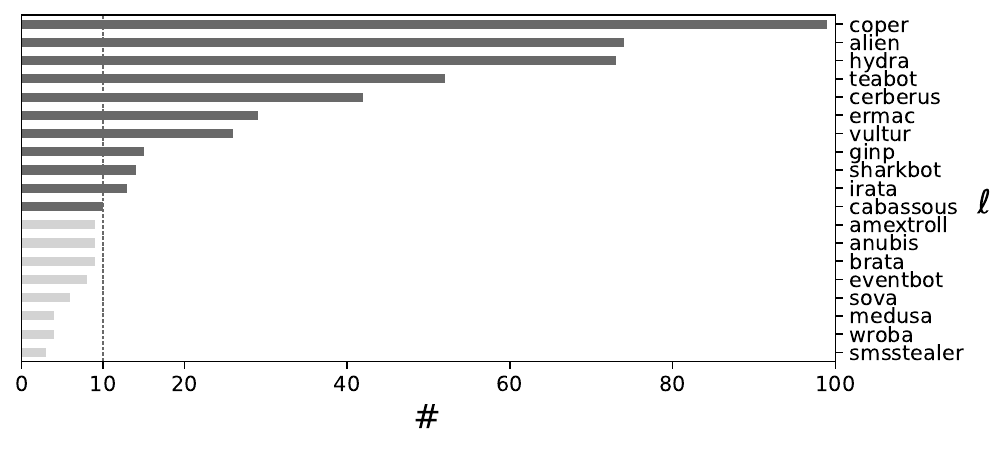}
                \caption{Imbalance in the number of instances $\#$ associated with each malware family $\ell$ in our dataset $\mathcal{D}$.}
                \label{fig:families_cardinality}
            \end{figure}

            The public dataset used is Drebin~\cite{ArpSpreitzenbarthAl14}, which consists of $n = 5560$ applications from $179$ different malware families, collected in the period from August 2010 to October 2012. Each application is characterized by several features grouped into the following categories: \emph{hardware components}, \emph{required permission}, \emph{app components}, \emph{filtered intents}, \emph{restricted API calls}, \emph{used permission}, \emph{suspicious API calls} and \emph{network addresses}. Since this study focuses on system permissions, we considered only features in the \emph{required permission} group explicitly referring to the Android operating system.
            
            The private dataset, kindly provided by \cleafy, comprises records of $n = 499$ malicious applications, identified as a subset of the threat intelligence telemetry provided by an online cybersecurity software. The considered telemetry contains applications identified as potentially malicious on user devices worldwide, with a particular focus on Europe, in the period from February 2022 to January 2023. The malware labels were assigned through manual analysis of malware programs conducted by a highly specialized threat intelligence team with strong domain expertise. \cref{fig:families_cardinality} illustrates the various malware families present in the private dataset and their cardinalities, highlighting the imbalanced nature of the classification problem.
            
            Each application is described by a binary vector $\vect{p}_i \in \{0, 1\}^P$, where $P$ is $154$ and $1800$ for the public and private dataset respectively, containing the one-hot encoding of the requested Android permissions, along with the corresponding label $\ell_i \in \mathcal{L}$ denoting the malware family. The resulting datasets are in the form of $\mathcal{D} = \{(\vect{p}_i, \ell_i) \;|\; \ell_i \in \mathcal{L}\}_{i = 1, \dots, n}$.
            
        \subsection{Methodology}
            \label{subsec:osr_methodology}

            We evaluate the closed-set recognition performance of \gboost, on both public and private datasets, through stratified $10$-fold cross-validation and average results over the $10$ test folds. Malware families with less than $10$ samples were excluded from the $10$-fold validation procedure, therefore they are not considered in closed-set recognition performance assessment. We grouped these samples in a dummy class labeled as $others$, which we then subsequently employed as the $novel$ class for evaluating the open-set recognition performance of our solution. This results in an open-set recognition problem with $54$ and $11$ known malware families for Drebin and private dataset respectively, while the $others$ class includes $125$ and $8$ families respectively.
            We perform an additional experiment on the public dataset, where we grouped all the malware families other than the top $10$ most populous ones into the $others$ class, and we refer to this configuration as Drebin\textsubscript{10}. This setup allows us to investigate a different scenario, where the closed-set classifier $\mathcal{C}$ has to deal with a low number of populous families, while the open-set recognition module $\mathcal{O}$ has to identify a larger, more heterogeneous set of samples as $novel$.

            We perform an additional experiment following a leave-one-class-out approach, by training our model on all the populous classes of Drebin\textsubscript{10} except a single malware class, which we consider $novel$ at test time. By doing so, we evaluate the effectiveness of our model in recognizing each malware class as novel when trained on the other classes. We do not consider leave-$k$-class-out procedures with $k > 1$, as increasing $k$ reduces the number of known classes, thereby simplifying both the closed-set and open-set recognition tasks. This is further supported by the comparison of results between Drebin and Drebin\textsubscript{10}, which differ significantly in the number of known classes. We employ stratified $10$-fold cross-validation within each leave-one-class-out iteration, and average the results over the $10$ test folds. To further assess the capability of our solution in identifying each individual malware class as $novel$, we plot the ROC curves for each novel class detection problem, treating the novel malware class as the positive class and merging all the classes used for training into the negative class.
            
            \medskip
            Unfortunately, we have no external training set to estimate the threshold $\tau$, nor can we use a portion of our datasets $\mathcal{D}$ solely for this purpose due to their limited size. Therefore, we set a desired false positive rate $FPR = 0.005$ and tuned the threshold $\tau$ on \emph{training} data, being aware that by doing so we are underestimating the real false positive rate at test time. A very low $FPR$ is mandatory in our settings, as classifying a sample $\vect{p}$ as belonging to a new family triggers manual inspection, which incurs a significant human resource cost.
            
        \subsection{Results and Discussion}
            \label{subsec:osr_results}
            
            \begin{figure*}[t]
                \centering
                \begin{subfigure}[t]{.45\linewidth}
                    \centering
                    \includegraphics[width=\linewidth]{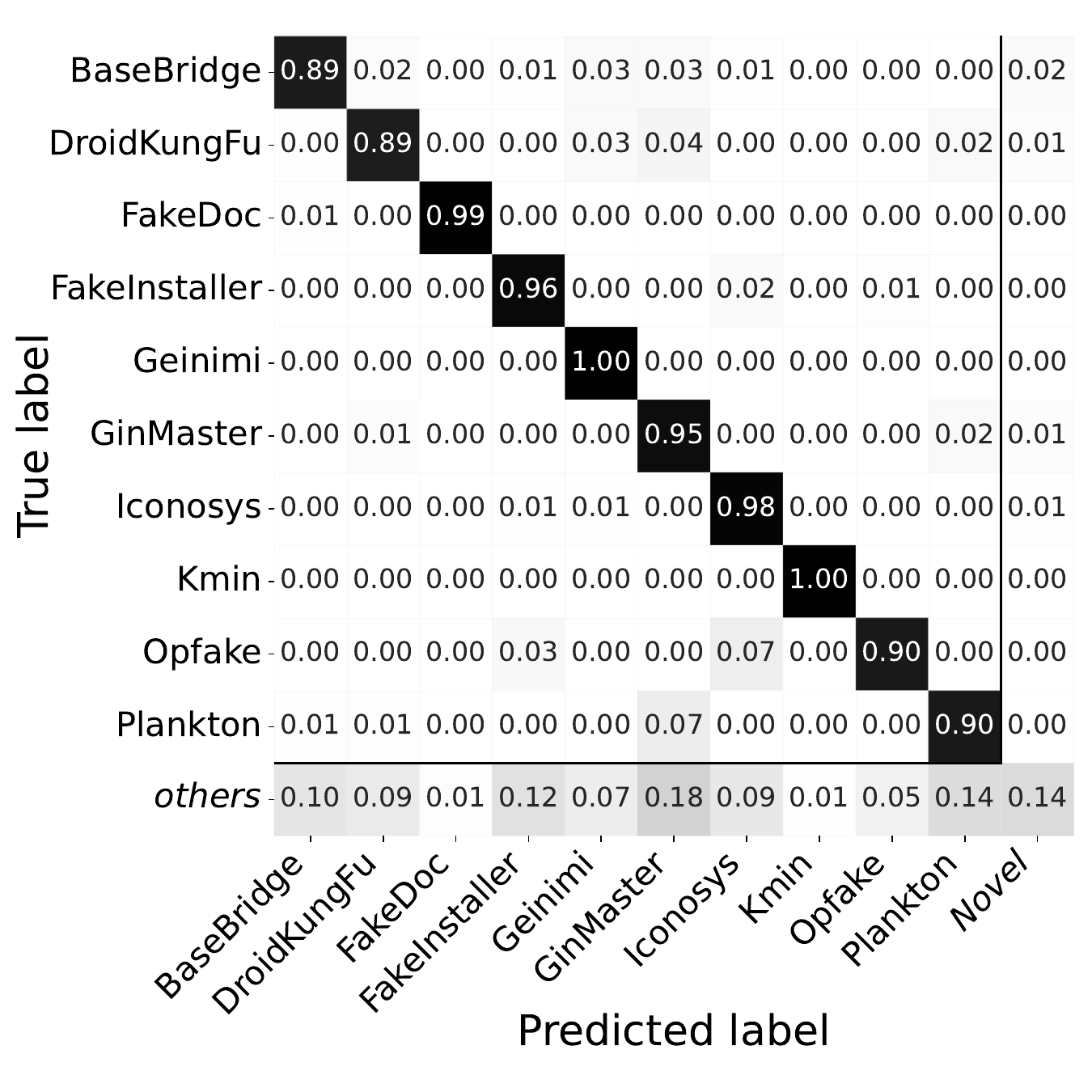}
                    \vspace*{-0.6cm}
                    \caption{}
                    \label{fig:kfold_osr_recall_matrix_public}
                \end{subfigure}
                \hfill
                \begin{subfigure}[t]{.45\linewidth}
                    \centering
                    \includegraphics[width=\linewidth]{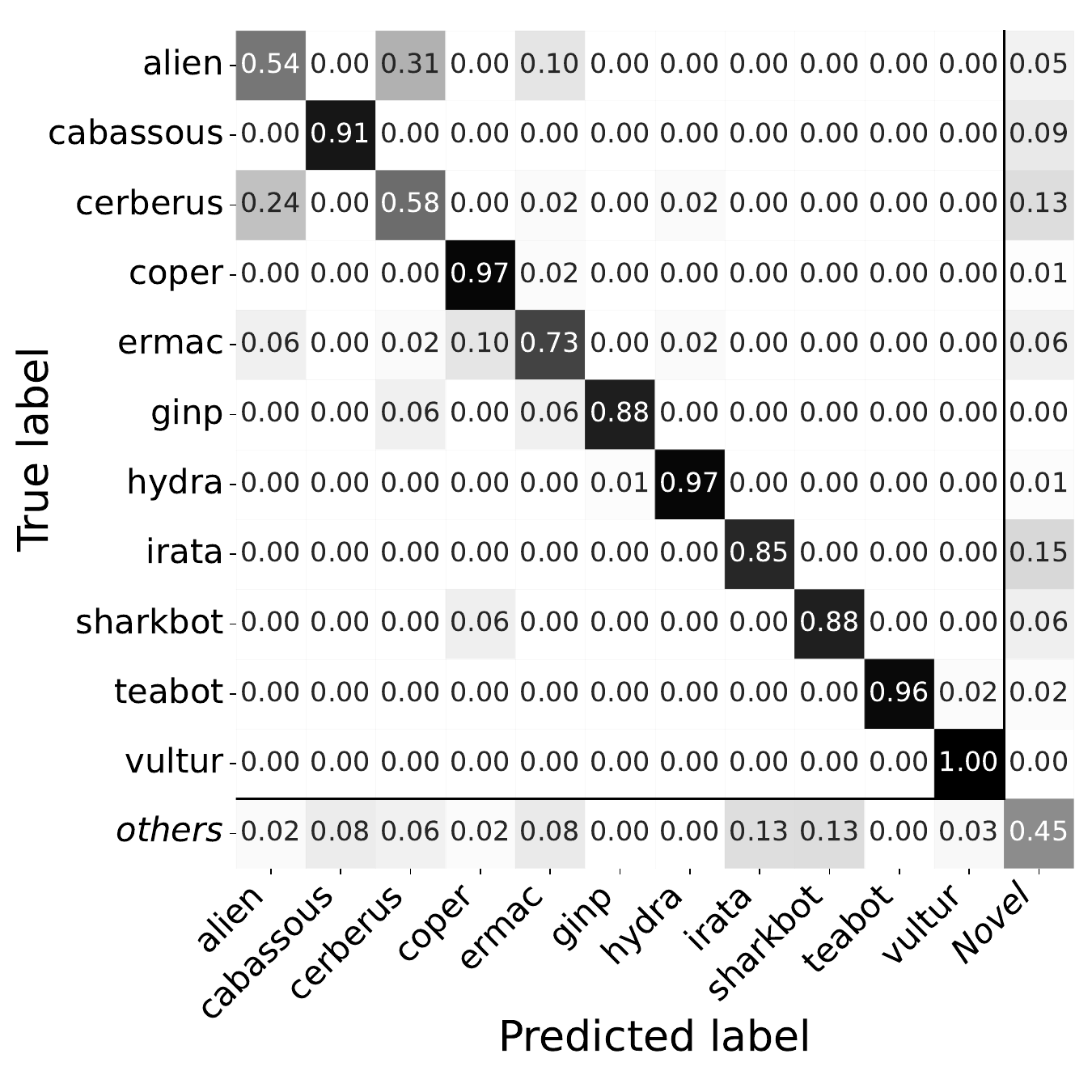}
                    \vspace*{-0.6cm}
                    \caption{}
                    \label{fig:kfold_osr_recall_matrix_private}
                \end{subfigure}

                \begin{subfigure}[t]{.45\linewidth}
                    \centering
                    \includegraphics[width=\linewidth]{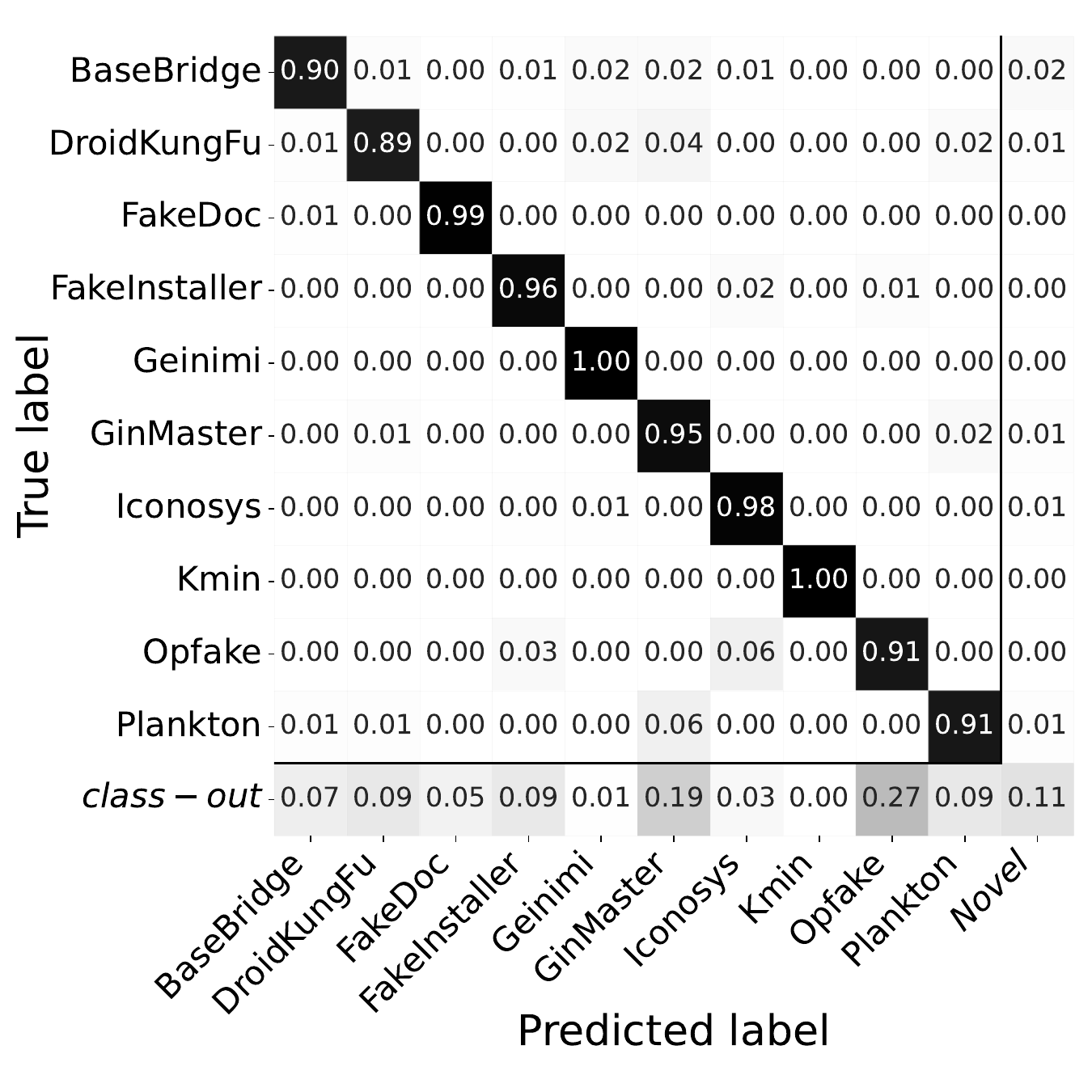}
                    \vspace*{-0.6cm}
                    \caption{}
                    \label{fig:loco_kfold_osr_recall_matrix_public}
                \end{subfigure}
                \hfill
                \begin{subfigure}[t]{.45\linewidth}
                    \centering
                    \includegraphics[width=\linewidth]{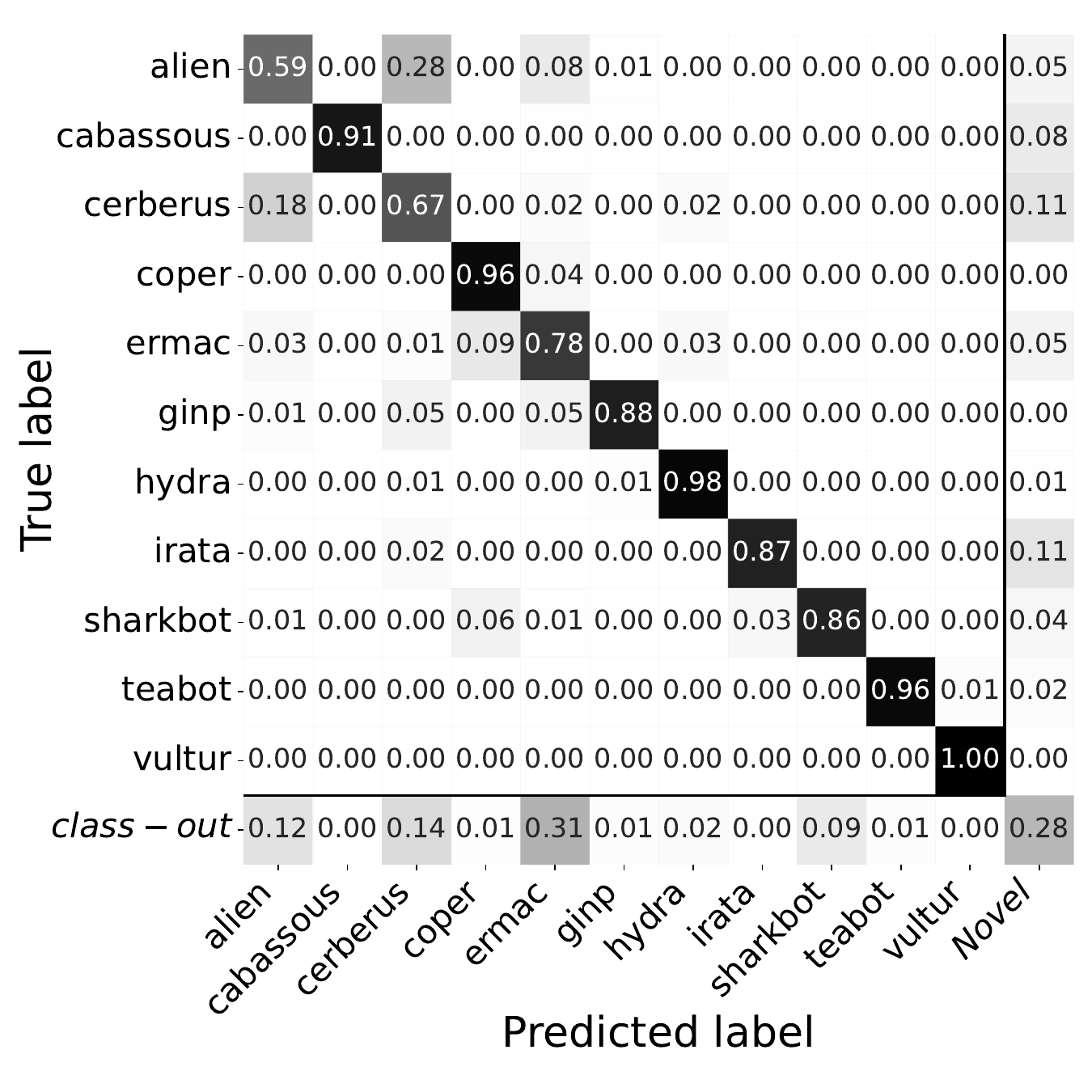}
                    \vspace*{-0.6cm}
                    \caption{}
                    \label{fig:loco_kfold_osr_recall_matrix_private}
                \end{subfigure}

                \hspace{0.7cm}
                \begin{subfigure}[t]{.4\linewidth}
                    \centering
                    \includegraphics[width=\linewidth]{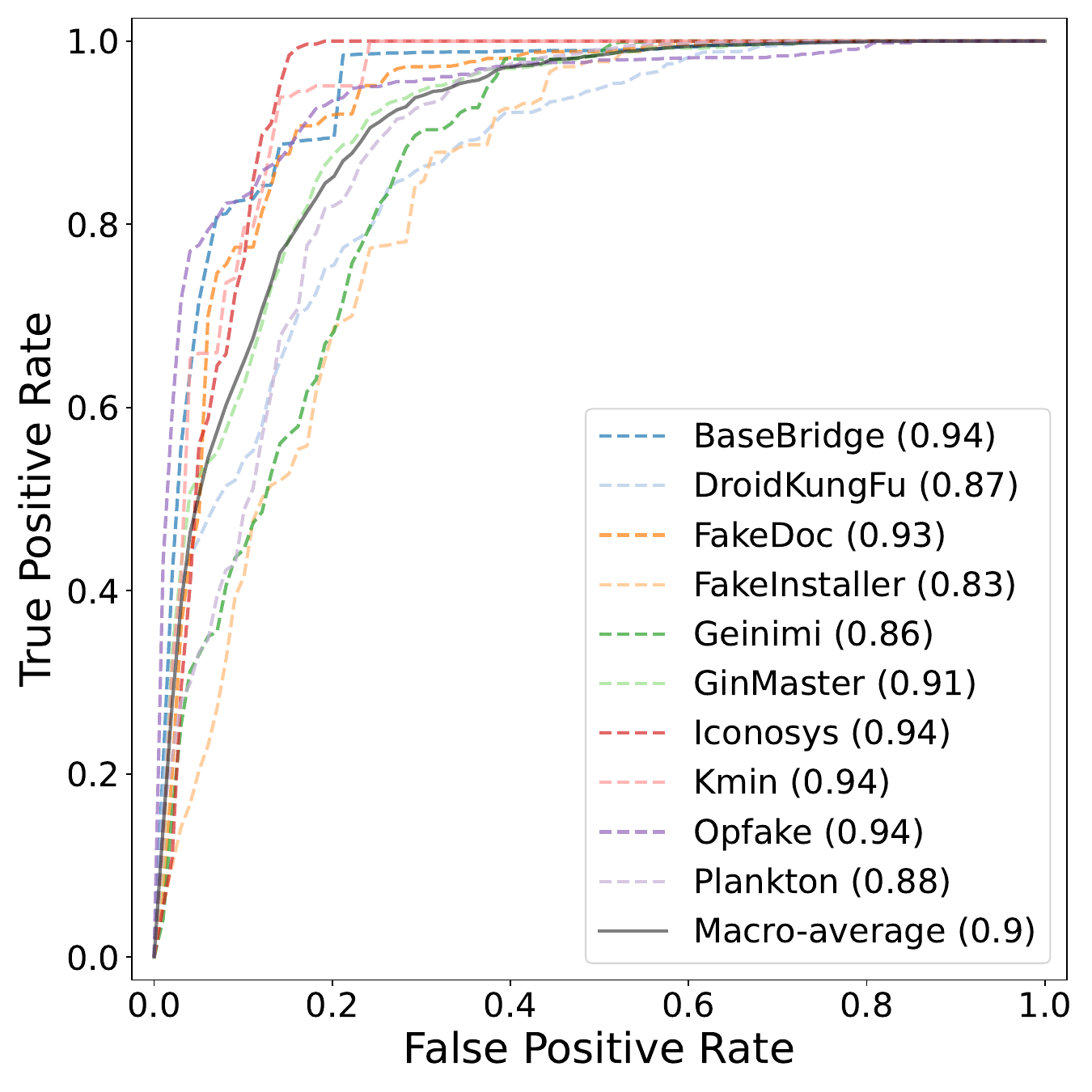}
                    \vspace*{-0.6cm}
                    \caption{}
                    \label{fig:loco_kfold_osr_binary_roc_curves_public}
                \end{subfigure}
                \hfill
                \hspace{-3cm}
                \begin{subfigure}[t]{.4\linewidth}
                    \centering
                    \includegraphics[width=\linewidth]{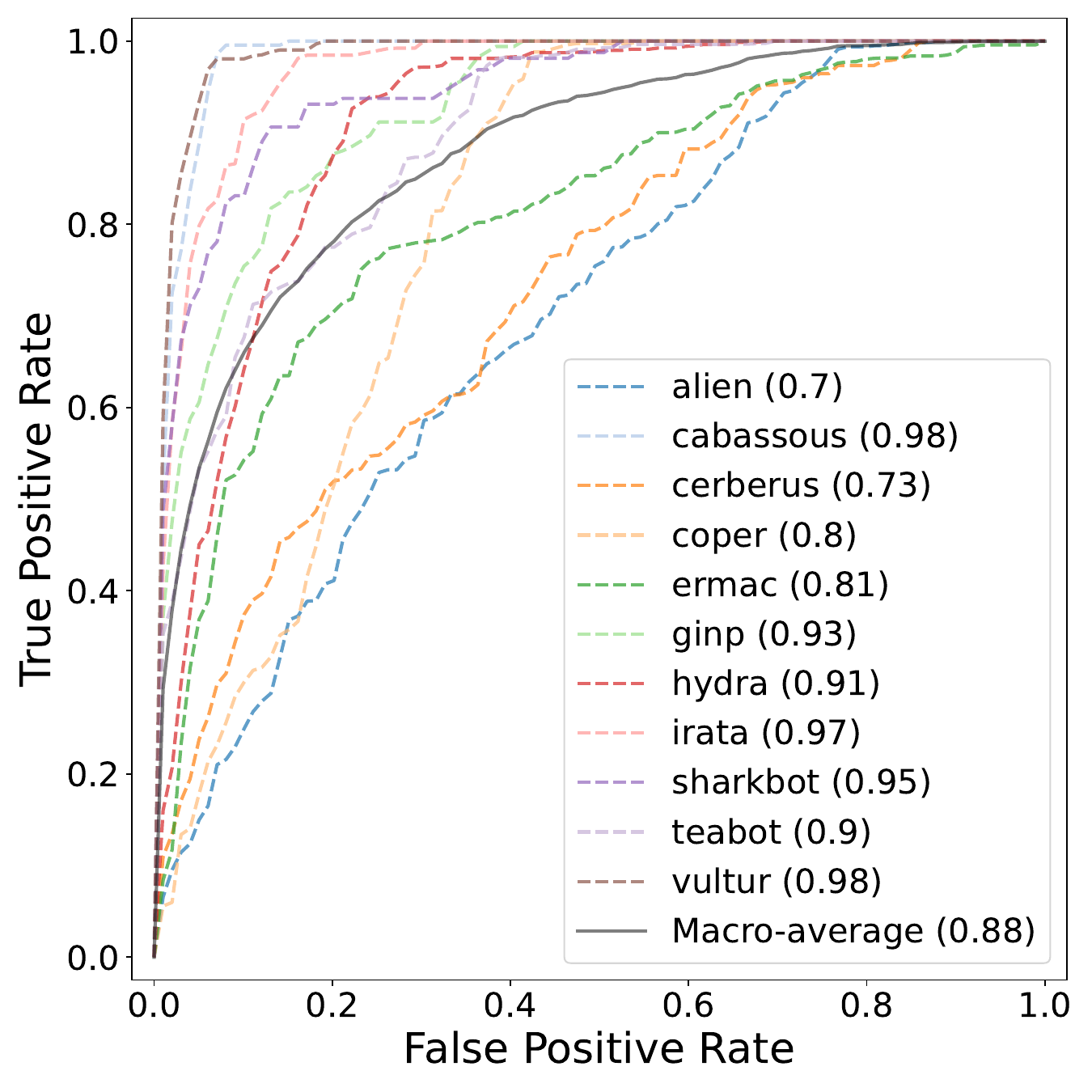}
                    \vspace*{-0.6cm}
                    \caption{}
                    \label{fig:loco_kfold_osr_binary_roc_curves_private}
                \end{subfigure}
                \caption{Recall confusion matrix of the open-set classifier $\mathcal{K}$, on the public (first column) and private (last column) datasets, when (a),(b) instances of the less populous classes are grouped in the dummy class $others$ and treated as $novel$, and (c),(d) each class is sequentially treated as $novel$ in the leave-one-class-out process, along with the associated novelty detection ROC curves (e),(f).}
                \label{fig:kfold_loco_kfolr_results_public}
            \end{figure*}
            
            We first assess the effectiveness of the tree-based \gboost classifier $\mathcal{C}$ for closed-set recognition on the $10$-fold experiment on both public and private datasets. A good performance in closed-set recognition is closely tied to the reliability of the subsequent open-set recognition procedure~\cite{VazeHanAl21}. Subsequently, we assess the effectiveness of our open-set recognition classifier $\mathcal{K}$ by comparing it against \osnn. In the latter experiment, we further investigate the relation between misclassified samples in the closed-set recognition task and the false positive rate in the open-set recognition task. Ultimately, we discuss the performance of our solution within \cleafy's anti-fraud business environment since its deployment on their engine.

            \begin{table}[t]
                \centering
                \caption{Micro-average (accuracy) and macro-average recall for closed-set $\mathcal{C}$ and open-set $\mathcal{K}$ classifiers on both public and private datasets. The results are shown for the two tested settings (a) and (b).}
                \begin{subtable}{\textwidth}
                    \centering
                    \caption{Less popolous malware families grouped into a dummy class $others$ and considered as the $novel$ class.}
                    \begin{tabular}{l@{\hskip 0.825cm}||@{\hskip 0.825cm}c@{\hskip 0.825cm}|@{\hskip 0.825cm}c@{\hskip 0.825cm}||@{\hskip 0.825cm}c@{\hskip 0.825cm}|@{\hskip 0.825cm}c@{\hskip 0.825cm}}
                                                  & \multicolumn{2}{c||}{$\mathcal{C}$} & \multicolumn{2}{c}{$\mathcal{K}$} \\ \cline{2-5}
                                                  & Micro   & Macro                    & Micro   & Macro                   \\ \hline
                        Drebin\textsubscript{10}  & $0.931$ & $0.949$                  & $0.711$ & $0.874$                 \\
                        Drebin                    & $0.823$ & $0.856$                  & $0.772$ & $0.839$                 \\
                        Private                   & $0.863$ & $0.869$                  & $0.842$ & $0.844$                 \\
                    \end{tabular}
                    \label{tab:kfold}
                \end{subtable}
                \\
                \vfill
                \begin{subtable}{\textwidth}
                    \centering
                    \caption{Each populous malware family, in turn, designated as the $novel$ class in a leave-one-class-out fashion.}
                    \begin{tabular}{l@{\hskip 0.825cm}||@{\hskip 0.825cm}c@{\hskip 0.825cm}|@{\hskip 0.825cm}c@{\hskip 0.825cm}||@{\hskip 0.825cm}c@{\hskip 0.825cm}|@{\hskip 0.825cm}c@{\hskip 0.825cm}}
                                                  & \multicolumn{2}{c||}{$\mathcal{C}$} & \multicolumn{2}{c}{$\mathcal{K}$} \\ \cline{2-5}
                                                  & Micro   & Macro                    & Micro   & Macro                   \\ \hline
                        Drebin\textsubscript{10}  & $0.935$ & $0.953$                  & $0.858$ & $0.874$                 \\
                        Drebin                    & $0.825$ & $0.850$                  & $0.810$ & $0.835$                 \\
                        Private                   & $0.871$ & $0.881$                  & $0.858$ & $0.860$                 \\
                    \end{tabular}
                    \label{tab:loco_kfold}
                \end{subtable}
                \label{tab:kfold_loco_kfold}
            \end{table}
            
            \medskip
            \subsubsection{Malware classification}
                \cref{tab:kfold_loco_kfold} summarizes the aggregated classification performance of the closed-set classifier $\mathcal{C}$. We observe similar performance levels on the public and private dataset in both scenarios (a) and (b), with slightly better results on the private dataset, and this could be attributed to the smaller number of families to classify ($11$ compared to $54$). This is confirmed by the scenario where only the top $10$ classes of the public dataset are retained, resulting in significantly higher micro-average and macro-average recall. We observe also that dealing with a higher-dimensional space ($1800$ compared to $154$) does not necessarily make it easier to separate different classes, as in the Drebin\textsubscript{10} scenario $\mathcal{C}$ achieves higher performance despite dealing with a comparable number of classes with respect to the private dataset ($10$ compared to $11$). The slightly lower micro-average recall, compared to the macro-average, is due to some misclassifications within the most populous classes, as the micro-average accounts for different class sizes. Overall, \gboost classifier proves effective in classifying malware families based on high-dimensional binary permission vectors.

            \medskip
            \subsubsection{Malware family discovery}

                \begin{figure*}[t]
                    \centering
                    \begin{minipage}[t]{\textwidth}
                        \centering
                        \includegraphics[width=.4\linewidth]{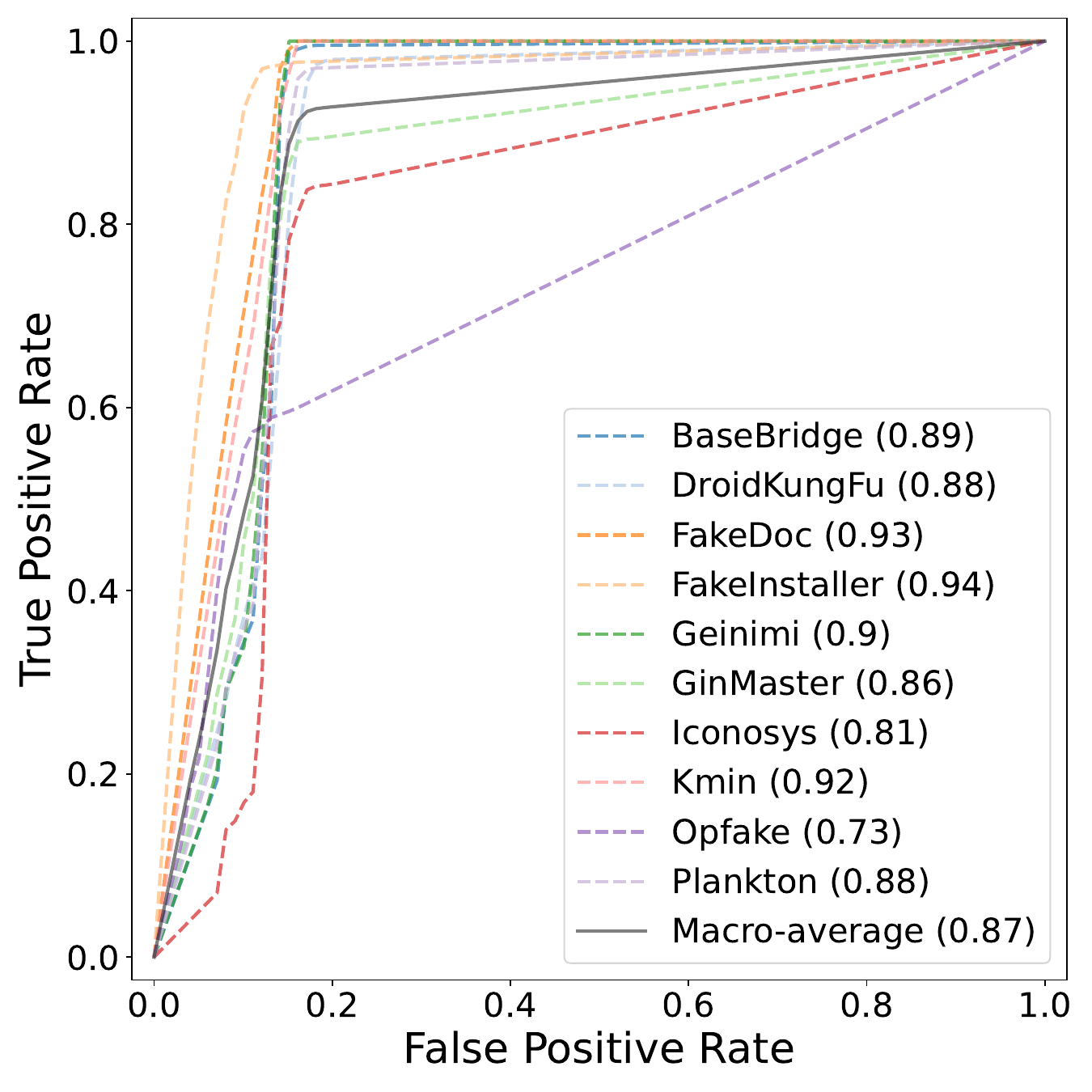}
                        \caption{Novelty detection ROC curves for \osnn on the public dataset, with each class treated as $novel$ in the leave-one-class-out process.}
                        \label{fig:loco_kfold_osr_binary_roc_curves_public_osnn}
                    \end{minipage}
                    \\
                    \begin{minipage}{\textwidth}
                        \centering
                        \nextfloat
                        \begin{subfigure}[t]{.37\linewidth}
                            \centering
                            \raisebox{1cm}{\includegraphics[width=\linewidth]{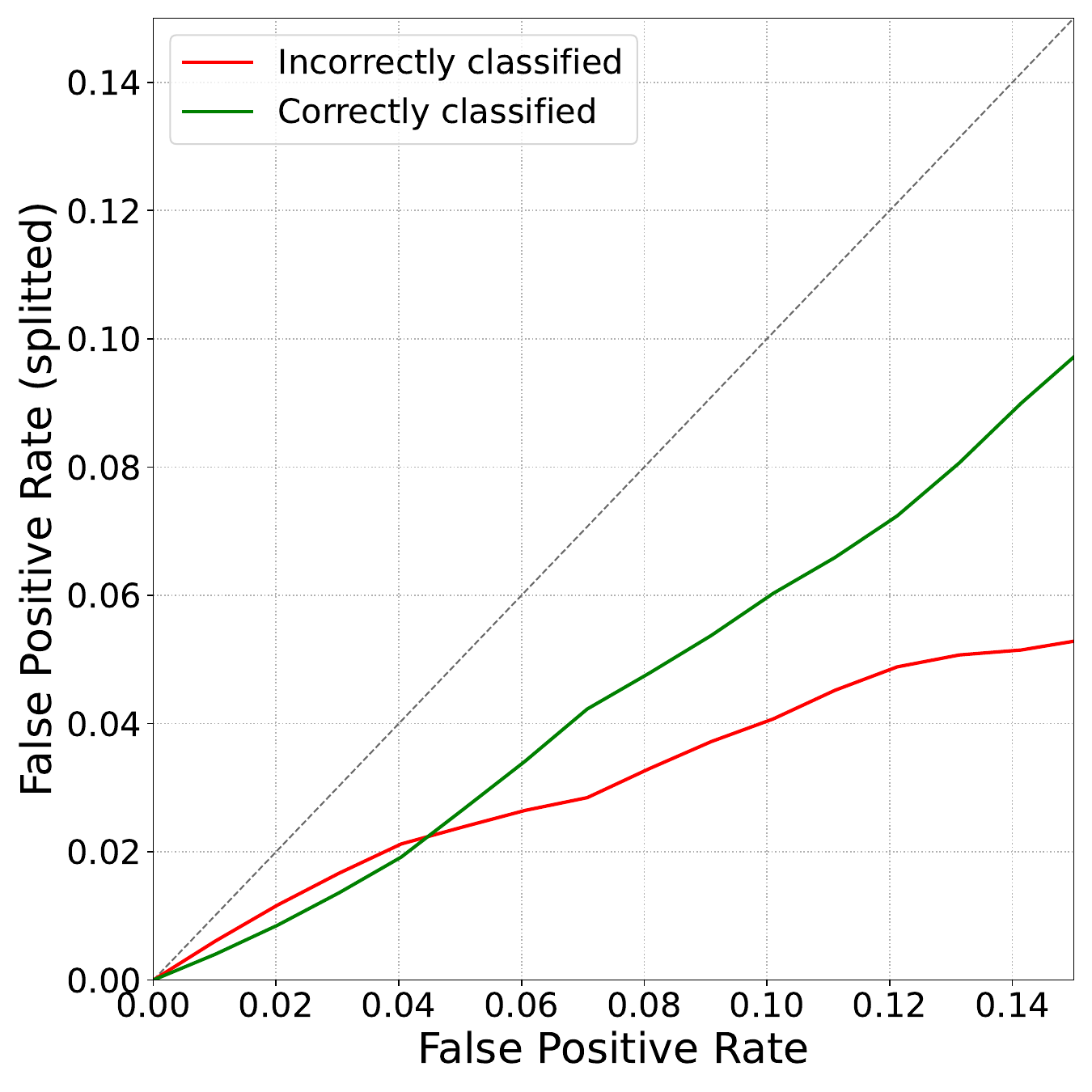}}
                            \caption{}
                            \label{fig:loco_kfold_osr_fpr_decomposition_public}
                        \end{subfigure}
                        \hfill
                        \begin{subfigure}[t]{.47\linewidth}
                            \centering
                            \includegraphics[width=\linewidth]{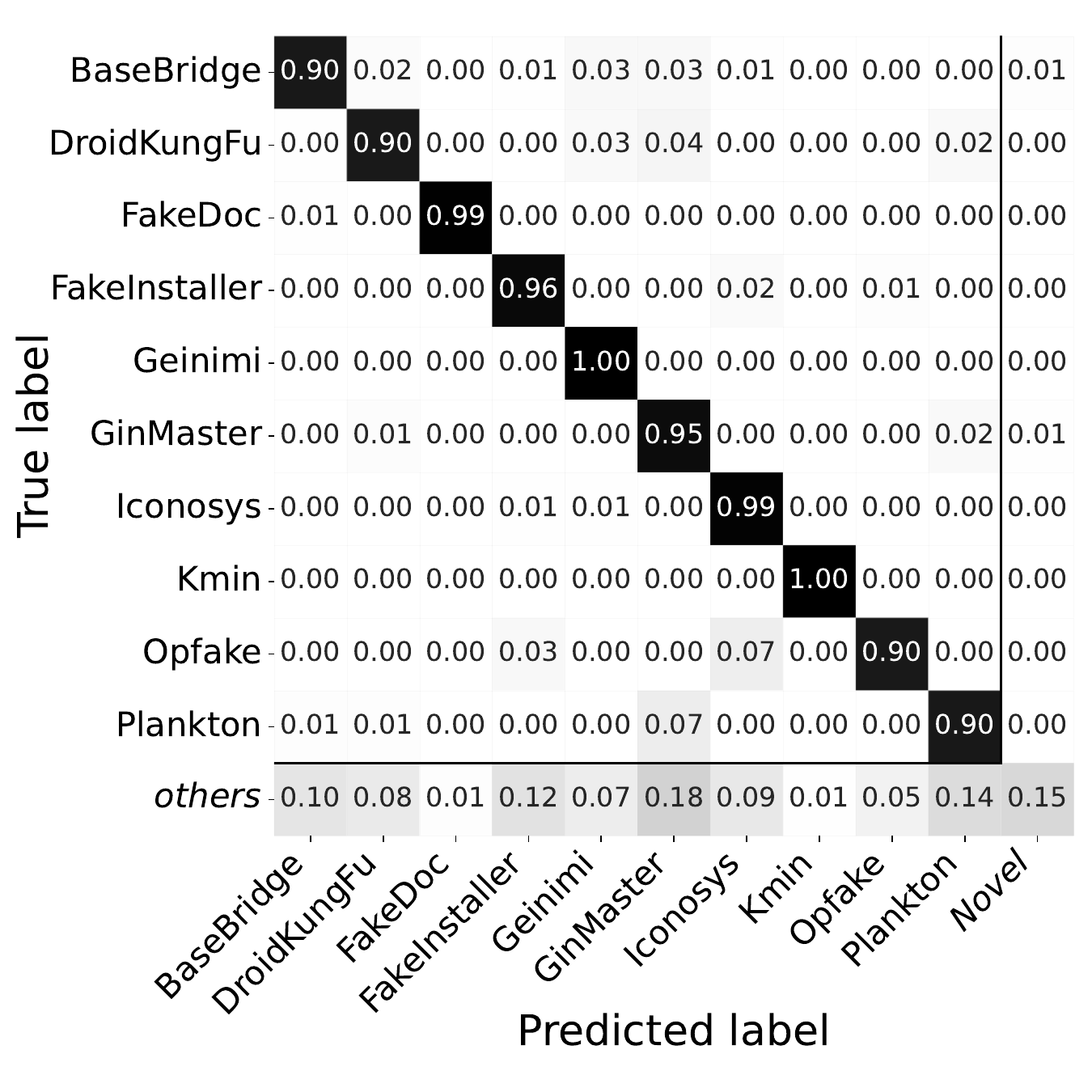}
                            \caption{}
                            \label{fig:loco_kfold_osr_recall_matrix_decomposition_public}
                        \end{subfigure}
                        \caption{(a) Different contributions made by samples correctly (green curve) and incorrectly (red curve) classified by $\mathcal{C}$ in composing the total false positive rate of the open-set recognition module $\mathcal{O}$ on the public dataset. (b) Treating the misclassified instances as $novel$ results in better real performance for the open-set classifier $\mathcal{K}$.}
                        \label{fig:loco_kfold_osr_fpr_analysis_public}
                    \end{minipage}
                \end{figure*}
            
                \cref{tab:kfold_loco_kfold} also reports the performance of the open-set classifier $\mathcal{K}$, which is, as expected, consistently lower than the closed-set classifier $\mathcal{C}$ due to the additional challenge of recognizing novel malware families and the subsequent impact of false positives.
                
                In~\cref{fig:kfold_osr_recall_matrix_public,fig:kfold_osr_recall_matrix_private}, we show the recall confusion matrices of the open-set classifier $\mathcal{K}$ when the instances of the $10$ and $11$ underrepresented classes, respectively for the public and private datasets, are grouped in the $others$ class and considered as $novel$. The recall confusion matrices reveal some false positives, where known malware families are misclassified as $novel$, especially for the private dataset, with $cerberus$ and $irata$ families being the most affected. This suggests that these are classes where the closed-set classifier exhibits great uncertainty, resulting in low confidence levels in their classification. We also observe that, while the families of the public dataset are easily separable, as shown by the almost clean outer diagonal recall matrix, this does not hold for the private dataset. Specifically, we observe a tendency to misclassify instances belonging to $alien$ and $cerberus$ classes, as evidenced by the prominent errors highlighted. We explain these results as a consequence of the fact that $alien$ is a more recent version of $cerberus$, hence, we expect a substantial overlap in the permissions they require to the system.
                
                \cref{fig:loco_kfold_osr_recall_matrix_public,fig:loco_kfold_osr_recall_matrix_private} present the recall confusion matrices obtained when each individual class is sequentially designated as $novel$ in a leave-one-class-out procedure, where similar observations to those previously discussed hold.
                \cref{fig:loco_kfold_osr_binary_roc_curves_private,fig:loco_kfold_osr_binary_roc_curves_public} show the ROC curves for the novelty detection task, together with respective AUC values in round brackets, when the specified class was considered as the positive $novel$ class. A low AUC value highlights the tendency of the open-set classifier to incorrectly classify, with a high confidence, the unknown class as belonging to one of the known classes used in the training process. This happens in the private dataset for the $alien$ and $cerberus$ families due to their similarity, consistent with the classification errors observed before, while in the public dataset, values are overall higher and more stable, aligning with the results shown in the confusion matrices.

                \medskip
                We conducted the same latter experiment on \osnn, and we show the resulting ROC curves and AUC values in~\cref{fig:loco_kfold_osr_binary_roc_curves_public_osnn}. \osnn exhibits very low $TPR$ values at low $FPR$, making it unsuitable for our requirements. In fact, a low $FPR$ is mandatory in our settings, as classifying a sample $\vect{p}$ as belonging to a new family triggers manual inspection, which involves a significant human resource cost. When comparing \osnn to our approach in~\cref{fig:loco_kfold_osr_binary_roc_curves_public}, we observe that at similarly low $FPR$ values, our method achieves significantly higher $TPR$, further highlighting its suitability for the malware family discovery task. The piecewise linear shape of \osnn's ROC curve reflects its limitations in handling high-dimensional binary data. Specifically, \osnn classifies a malware sample as belonging to a new family if the ratio of the distances to its two nearest neighbors from different families falls below a given threshold. However, when dealing with binary data, the number of distinct distance values is limited and highly redundant, leading to a discrete and heavily skewed distribution of distance ratios, and ultimately causing the piecewise progression in $TPR$ values we observe in~\cref{fig:loco_kfold_osr_binary_roc_curves_public_osnn}.
                
                A comparison of computational time between our method and \osnn highlights the greater efficiency of our approach, with inference times of $1.359 \cdot 10^{-5}$ and $1.620 \cdot 10^{-4}$ seconds respectively, showing a difference of more than an order of magnitude. This disparity stems from \osnn's higher computational complexity, which is dominated by the $O(n \: logn)$ sorting of distances between the query sample and the training dataset. This step is required to compute the ratio of distances to the two nearest neighbors, which is necessary to determine whether the query sample should be classified as novel.

            \medskip
            \subsubsection{False alarms analysis}
                \label{subsubsec:false_alarms_analysis}
                
                In this paragraph, we focus on examining the composition of false alarms on the public dataset Drebin\textsubscript{10}. To this end, we decomposed the false positives of the open-set recognition module into two groups: those originating from instances that were correctly classified by $\mathcal{C}$ before introducing the open-set recognition module $\mathcal{O}$, and those that would have been misclassified regardless of $\mathcal{O}$.
                
                \cref{fig:loco_kfold_osr_fpr_decomposition_public} shows the false positive rate computed over these two groups and we observe that most of the false positives of $\mathcal{O}$, in particular at low $FPR$ values, are samples that would have been misclassified by $\mathcal{C}$ if the open-set module were not in place. This was expected, as instances misclassified by $\mathcal{C}$ are typically associated with low confidence values. We argue that those samples are worth being checked by the threat analysts, since incorrect malware classification results in ineffective solutions against the threat. Therefore, if we do not consider misclassified samples as false positives, thus we compute the false positive rate only from samples belonging to known families that would have been correctly classified if there was no \osr module $\mathcal{O}$, the false positive rate drops considerably. \cref{fig:loco_kfold_osr_recall_matrix_decomposition_public} displays the corresponding recall confusion matrix when these instances are treated as $novel$, where the model shows attains a micro-average recall of $0.714$ alongside a macro-average recall of $0.877$.

            \medskip
            \subsubsection{Real-world deployment performance}
                \label{subsubsec:real_world_deployment_performance}
                
                Our solution is currently used by analysts at \cleafy as a complementary tool to enhance malware classification and discover new families. Testing our solution in their operational environment began in the latter half of 2023, and is currently in use in monitoring telemetry gathered by the threat intelligence division, for the analysis of data from worldwide sources.
                
                The closed-set classification performance of our deployed solution resembles closely those reported in experiments on the private dataset, achieving an accuracy of $83\%$ when tested on about $300$ telemetry applications. Unfortunately, the portion of the telemetry data used for testing did not allow us to obtain a final judgment on the model's performance in detecting new families. The ability to discover new real malware families cannot be assessed on the deployed system by means of strategies like the leave-one-class-out we adopted before, and recently there have been no cases of completely different malware families within the Android landscape. However, there have been a couple of noteworthy cases. In the first case, the \osr system reported as \textit{novel} a new version of a known malware. In this case, the malware detected as \textit{novel} had a few significant differences with respect to the other, but not enough to consider this to belong to a new family (contrary to $alien$ and $cerberus$). In the second case, a variation of a known malware was labeled as $novel$ due to different permissions requirements compared to the known one. Additionally, a manual review revealed that hundreds of known malware instances classified as $novel$ actually belonged to families not represented in the training set, indicating they were correctly identified as novel. In fact, most of the false alarms concerned malware programs other than banking malware programs (the primary interest for \cleafy's system), and therefore correctly classified as $novel$.

                \Cleafy prefers not to disclose further information on deployment performance for strategic reasons.

    \section{Conclusions}
        \label{sec:mfd_conclusions}

        In this chapter, we addressed the problem of detecting new malware families by combining, for the first time, a tree-based \gboost classifier with the \maxlogit \osr method. We conducted experiments on both a public and a proprietary dataset to validate the suitability of our approach and discussed its deployment performance in a real-world environment. Additionally, our analysis of false alarms and the impact of misclassified instances emphasized the practical advantages of applying an \osr approach to malware classification.

        We envisage several potential future directions. These include investigating advanced feature engineering techniques to enhance the performance of \osr further, as well as exploring alternative or ensemble approaches to integrate different \osr methods. Another interesting direction involves developing a dynamic thresholding system that adjusts the $FPR$ threshold $\tau$ adaptively, taking into account contextual information such as the current threat landscape to achieve the desired sensibility of the \osr module.

    \chapter{Concluding Remarks}
    \label{cha:conclusions}
    
    In this thesis, we addressed the problem of anomaly detection, which involves identifying data that deviate from expected or genuine behavior. Anomaly detection has relevant applications in various fields, each presenting unique challenges and constraints based on how genuine behavior is modeled and how data are processed. This thesis introduced new solutions for addressing anomaly detection under two distinct modeling settings.
    
    In the first part of the thesis, we modeled genuine data as samples that conform to structured patterns represented by low-dimensional manifolds, while anomalous data were defined as those that deviate from these patterns. We refer to this scenario as structure-based anomaly detection.
    We proposed \pisolationforest (\pif), an anomaly detection framework designed to identify anomalous samples that do not conform to structured patterns. This is achieved by combining two key components: (i) \pembedding, which maps data do a high-dimensional \pspace, and (ii) \pisolation, carried out by isolation-based algorithms. We developed two versions of \pisolation: the first one, namely \viforest, employs an ensemble of nested Voronoi tessellations and can be applied to any metric space. The second one, namely \rzhiforest, is specifically designed for the \pspace and uses a novel Locality Sensitive Hashing (\lsh) called \rzhash to avoid explicit distances computation, thereby improving \pif's efficiency.
    Our empirical evaluation demonstrated that \pembedding enhances the separability between genuine (structured) and anomalous (unstructured) data, resulting in better performance than performing anomaly detection directly in the ambient space. Notably, the proposed \pif framework outperforms all alternative approaches when they are straightforwardly plugged in the \pspace, highlighting the greater effectiveness of both \viforest and \rzhiforest. Furthermore, we showed that \rzhiforest not only achieves greater efficiency than \viforest, but it also exhibits more stable performance across different branching factors, both on synthetic and real-world data.
    We then introduced \spif, a sliding window-based extension of \pif framework, designed for scenarios where only local information about genuine structures is available. By leveraging the locality principle, \spif is particularly useful in real-world cases where the global nature of the underlying low-dimensional manifold is typically unknown, but we can provide local approximations. By employing a collection of local models, complex manifolds can be effectively described in a way akin to how a plane curve is represented, in projective geometry, by the set of lines tangent to it (\emph{i.e.}, its dual). Experiments on 3D scanned objects demonstrate that \spif successfully identifies defects by detecting sharp irregularities in otherwise smooth surfaces.
    
    We also addressed a problem closely related to structure-based anomaly detection, namely structure-based clustering, where the primary focus is on recovering individual genuine structures rather than identifying anomalies, which are typically detected as a byproduct of the structures recovery process.
    We proposed \mlink, a simple and effective algorithm for recovering genuine structures from different model families in data contaminated by noise and outliers. \mlink operates by fitting models from various families within a hierarchical clustering linkage scheme, and includes a novel cluster merging strategy based on on-the-fly model fitting and selection through \gric (Geometric Robust Information Criterion).
    Experiments on both synthetic and real-world data demonstrated that \mlink is faster, more stable, and less sensitive to both the initial models sampling and the inlier threshold compared to preference-based competitors, and at the same time favorably compares with optimization-based methods. Furthermore, \mlink can be further extended by modifying cluster merging conditions to account for specific constraints coming from the application at hand.
    
    In the second part of the thesis, we focused on the more traditional setting where genuine data are modeled as samples that lie in high-density regions, while anomalies fall in low-density regions. This scenario is commonly referred to as density-based anomaly detection.
    We started by addressing it from the online anomaly detection perspective, where data is processed as an endless stream and may exhibit dynamic behavior.
    To this end, we proposed \onlineiforest, an anomaly detection algorithm specifically designed for the streaming scenario. \onlineiforest consists of an ensemble of multi-resolution histograms that incrementally adapt to the data distribution by keeping only statistics about data points. Thanks to a sliding window, \onlineiforest selectively forgets older data points and update histograms accordingly.
    Extensive experiments showed that \onlineiforest features an extremely fast processing and learning speed while maintaining effectiveness comparable to that of state-of-the-art methods, positioning \onlineiforest as a good candidate for addressing real-world streaming anomaly detection challenges.
    
    Finally, we addressed the relevant industrial anomaly detection problem of identifying new malware families.
    We proposed to combine the tree-based \gboost classifier with the \maxlogit open-set recognition technique to tackle the problem of malware family discovery. Our approach can be seamlessly applied to a classification pipeline based on boosted decision trees without even affecting the classification workflow.
    Experiments conducted on both a public and a proprietary dataset validated the suitability of our approach, which has been deployed in a business environment and is currently part of their engine.
    
It can also incorporate various model families within the \pembedding process, including density distributions, subspaces, clustering schemes, autoencoders, and even supervised regression and classification models such as support vector machines and neural networks. This flexibility allows \pif to tackle a broader range of challenges and applications where some form of prior knowledge is available
    \section{Future Work}
        \label{sec:future_work}

        The research presented in this thesis can be expanded following different directions.
        First, the generality of the \pif anomaly detection framework is not limited to parametric models. It can also utilize model families such as density distributions, subspaces, clustering schemes, autoencoders, and even supervised regression and classification models like support vector machines and neural networks, enabling \pif to address a much wider range of challenges and applications where some form of prior knowledge is available. The virtually unlimited variety of model families that can be incorporated into the \pembedding step makes \pif an extremely adaptable framework, able to address any type of anomaly detection problem that can be framed in the \emph{structure-based anomaly detection} settings.
        
        Furthermore, since \viforest can operate with arbitrary distance metrics, it could be interesting to apply \viforest as a standalone anomaly detection tool in spaces other than the \pspace, where different distances may be more appropriate for distinguishing between anomalies and genuine data. The inherent ability to work directly with distance matrices allows \viforest to be applied even in complex scenarios where no explicit distance function is available, such as when distances are derived from kernel estimations in random forests or similarities are learned through neural networks. This makes \viforest a highly versatile standalone anomaly detection tool, able to handle any scenario where pairwise instance similarities can be estimated, significantly expanding its range of potential applications.
        
        Finally, our work on online anomaly detection, namely \onlineiforest, can be improved by eliminating the need for a sliding window of points, while preserving the ability to forget information about the past. All the information about the generative process seen so far by \onlineiforest is already embedded within the density distribution learned by the ensemble of histograms. Therefore, we can remove the sliding window by sampling candidate points to forget directly from the histogram bins according to the estimated density. This approach would reduce memory requirements and would make \onlineiforest more self-contained and theoretically sound, allowing to analytically characterize the learning and forgetting procedures as stochastic processes.

    \cleardoublepage
    \pagestyle{empty}
    \mbox{}
    \cleardoublepage
    \setlength\epigraphwidth{8.74855cm}
\epigraph{{\large \emph{``I have thought of a nice ending for [my book]: and he lived happily ever after to the end of his days.''\footnotemark}}}{}
\footnotetext{Bilbo Baggins, \emph{The Lord of the Rings: The Fellowship of the Ring} by J.R.R. Tolkien}

    \cleardoublepage
    \pagestyle{fancy}
    \appendix
    \chapter{Theoretical results on RuzHash}
    \label{cha:theoretical_results_ruzhash}

    \section{\rzhash is a LSH for Ruzicka distance}
            \label{apx:rzhash}
            
            We present the proof of~\cref{thm:ruzhash_short}, introduced in~\cref{subsubsec:ruzhash}, stating that \rzhash is a Locality Sensitive Hashing (LSH) for the Ruzicka distance. The proof builds on~\cref{lem:minhash}, a well-known result stating that \mhash is an LSH for the Jaccard distance. Additionally, \cref{fig:correlation} empirically illustrates the correlation between \rzhash and Ruzicka.
            
            \begin{lemma}[\mhash]
                \label{lem:minhash}
                Given $\vect{p}, \vect{q} \in \{0, 1\}^m$, let $h^{\vect{\pi}}_{min} : \{0, 1\}^m \rightarrow \{1, \dots, m\}$ the hashing performed by \mhash, where $\vect{\pi} = [\pi_1, \dots, \pi_m]$ is a random permutation of the $m$ dimensions $\{1, \dots, m\}$. In is known that~\cite{BroderCharikarAl00}
                
                \begin{equation}
                    \begin{split}
                        Pr[h^{\vect{\pi}}_{min}(\vect{p}) = h^{\vect{\pi}}_{min}(\vect{q})] &= \frac{\# positive\ outcomes}{\# total\ outcomes} =\\
                                                                              &= \frac{\sum_{\vect{\pi}} \mathds{1}(h^{\vect{\pi}}_{min}(\vect{p}) = h^{\vect{\pi}}_{min}(\vect{q}))}{m!} =\\
                                                                              &= \frac{\sum_{i=1}^{m} (p_i \wedge q_i)}{\sum_{i=1}^{m} (p_i \vee q_i)} =\\
                                                                              &= \frac{\sum_{i=1}^{m} \min\{p_i, q_i\}}{\sum_{i=1}^{m} \max\{p_i, q_i\}} = 1 - d_J(\vect{p},\vect{q}),
                    \end{split}
                    \label{eq:mhash}
                \end{equation}
                where $\mathds{1}$ is the indicator function, $m!$ is the total number of permutations of $m$ dimensions, and with $\sum_{\vect{\pi}}$ we summed over all the possible realizations of the permutation $\vect{\pi}$.
            \end{lemma}

            \setcounter{theorem}{0}
            \begin{theorem}[\rzhash]
                \label{thm:ruzhash}
                Given $\vect{p}, \vect{q} \in [0, 1]^m$, let $\vect{\tau} = [\tau_1, \dots, \tau_m]$ be a vector of thresholds where each $\tau_i \sim \mathcal{U}_{[0, 1)}$ is randomly drawn from the unitary interval, let $\mathds{1}(\vect{p} > \vect{\tau}) = [\mathds{1}(p_1 > \tau_1), \dots, \mathds{1}(p_m > \tau_m)] \in \{0, 1\}^m$ be the indicator function applied element-wise on $\vect{p}$ and $\vect{\tau}$, and let $h^{\vect{\pi}}_{min} : \{0, 1\}^m \rightarrow \{1, \dots, m\}$ be the hashing performed by \mhash (\cref{lem:minhash}). Let \rzhash be the function $h^{\vect{\pi}, \vect{\tau}}_{ruz} = h^{\vect{\pi}}_{min} \circ \mathds{1}(\vect{p} > \vect{\tau}) :  [0, 1]^m \rightarrow \{1, \dots, m\}$. Then,
                \begin{equation}
                    Pr[h^{\vect{\pi}, \vect{\tau}}_{ruz}(\vect{p}) = h^{\vect{\pi}, \vect{\tau}}_{ruz}(\vect{q})] = \frac{\sum_{i=1}^m \min\{p_i, q_i\}}{\sum_{i=1}^m \max\{p_i, q_i\}} = 1 - d_R(\vect{p}, \vect{q}).
                    \label{eq:rzhash}
                \end{equation}
            \end{theorem}
            \begin{proof}
                Relying on the definition of \rzhash as $h^{\vect{\pi}, \vect{\tau}}_{ruz} = h^{\vect{\pi}}_{min} \circ \mathds{1}(\vect{p} > \vect{\tau})$, it holds that
                \begin{equation}
                    \begin{split}
                        P[h^{\vect{\pi}}_{min}&(\mathds{1}(\vect{p} > \vect{\tau})) = h^{\vect{\pi}}_{min}(\mathds{1}(\vect{q} > \vect{\tau}))] =\\
                                                                                         &= \frac{\# positive\ outcomes}{\# total\ outcomes} =\\
                                                                                         &= \frac{\int_{\vect{\tau}} \sum_{\vect{\pi}} \mathds{1}(h^{\vect{\pi}}_{min}(\mathds{1}(\vect{p} > \vect{\tau})) = h^{\vect{\pi}}_{min}(\mathds{1}(\vect{q} > \vect{\tau}))) d\tau}{m!} =\\
                                                                                         &= \frac{\int_{\vect{\tau}} \sum_{i=1}^{m} (\mathds{1}(p_i > \tau_i) \wedge \mathds{1}(q_i > \tau_i)) d\vect{\tau}}{\int_{\vect{\tau}} \sum_{i=1}^{m} (\mathds{1}(p_i > \tau_i) \vee \mathds{1}(q_i > \tau_i)) d\vect{\tau}} =\\
                                                                                         &= \frac{\int_{\vect{\tau}} \sum_{i=1}^m \min\{\mathds{1}(p_i > \tau_i), \mathds{1}(q_i > \tau_i)\} d\vect{\tau}}{\int_{\vect{\tau}} \sum_{i=1}^m \max\{\mathds{1}(p_i > \tau_i), \mathds{1}(q_i > \tau_i)\} d\vect{\tau}},
                    \end{split}
                \end{equation}
                where with $\int_{\vect{\tau}}$ we integrated over all the possible realizations of the vector $\vect{\tau}$ and the rest follows from~\cref{lem:minhash} because $\mathds{1}(\vect{p} > \vect{\tau}), \mathds{1}(\vect{q} > \vect{\tau}) \in \{0, 1\}^m$.
                Since each $\tau_i \sim \mathcal{U}_{[0, 1)}$ is an \textit{independent} realization of the uniform distribution, we can write the last term as
                \begin{equation}
                    \label{eq:independent}
                    \text{\hspace{-0.75cm}}
                    \frac{\int_{\vect{\tau}} \sum_{i=1}^m \min\{\mathds{1}(p_i > \tau_i), \mathds{1}(q_i > \tau_i)\} d\vect{\tau}}{\int_{\vect{\tau}} \sum_{i=1}^m \max\{\mathds{1}(p_i > \tau_i), \mathds{1}(q_i > \tau_i)\} d\vect{\tau}} = \frac{\sum_{i=1}^m \int_0^1 \min\{\mathds{1}(p_i > \tau_i), \mathds{1}(q_i > \tau_i)\} d\tau_i}{\sum_{i=1}^m \int_0^1 \max\{\mathds{1}(p_i > \tau_i), \mathds{1}(q_i > \tau_i)\} d\tau_i}.
                \end{equation}
                We have that
                \begin{equation}
                    \min\{\mathds{1}(p_i > \tau_i), \mathds{1}(q_i > \tau_i)\} =
                    \begin{cases}
                        1 & \text{if $\min\{p_i, q_i\} > \tau_i$}\\
                        0 & \text{otherwise},
                    \end{cases}
                \end{equation}
                and analogously for $\max\{\mathds{1}(p_i > \tau_i), \mathds{1}(q_i > \tau_i)\}$.
                Therefore, we split the integrals in~\eqref{eq:independent} as
                \begin{equation}
                    \text{\hspace{-1cm}}
                    \frac{\sum_{i=1}^m \int_0^1 \min\{\mathds{1}(p_i > \tau_i), \mathds{1}(q_i > \tau_i)\} d\tau_i}{\sum_{i=1}^m \int_0^1 \max\{\mathds{1}(p_i > \tau_i), \mathds{1}(q_i > \tau_i)\} d\tau_i} = \frac{\sum_{i=1}^m (\int_0^{\min\{p_i, q_i\}} 1 \,d\tau_i + \int_{\min\{p_i, q_i\}}^1 0 \,d\tau_i)}{\sum_{i=1}^m (\int_0^{\max\{p_i, q_i\}} 1 \,d\tau_i + \int_{\max\{p_i, q_i\}}^1 0 \,d\tau_i)}
                \end{equation}
                then
                \begin{equation}
                    \text{\hspace{-1.25cm}}
                    \frac{\sum_{i=1}^m (\int_0^{\min\{p_i, q_i\}} 1 \,d\tau_i + \int_{\min\{p_i, q_i\}}^1 0 \,d\tau_i)}{\sum_{i=1}^m (\int_0^{\max\{p_i, q_i\}} 1 \,d\tau_i + \int_{\max\{p_i, q_i\}}^1 0 \,d\tau_i)} = \frac{\sum_{i=1}^m \int_0^{\min\{p_i, q_i\}} 1 \,d\tau_i}{\sum_{i=1}^m \int_0^{\max\{p_i, q_i\}} 1 \,d\tau_i} = \frac{\sum_{i=1}^m \min\{p_i, q_i\}}{\sum_{i=1}^m \max\{p_i, q_i\}},
                \end{equation}
                and this proves~\eqref{eq:rzhash}.
            \end{proof}
            
            In~\cref{fig:correlation} we show the exact correlation between $1 - d_R(\vect{p}, \vect{q})$ and \rzhash for $n = 1000$ randomly generated couples of points $\vect{p}, \vect{q} \in [0, 1]^m$, when $m = 10000$ and \rzhash has been executed $k = 100$ times for each couple, compared to the exact correlation between $1 - d_J(\vect{p}, \vect{q})$ and \mhash in the same settings.

    \section{\rzhash linearly correlates with Ruzicka distance}
            \label{apx:variable_split}

            We prove that, when the \rzhash mapping is modified as described in~\cref{subsubsec:ruzhash_itree}, it maintains a correlation with the Ruzicka distance, with the slope being proportional to the new branching factor $b$. In~\cref{fig:ruzicka_variable}, we empirically illustrate the linear correlation for different branching factors.

            \begin{theorem}[$d_R(\vect{p}, \vect{q})$ linearly correlates with $b$]
                \label{thm:correlation}
                Given $\vect{p}, \vect{q} \in [0, 1]^m$, let $h^{\vect{\pi}, \vect{\tau}}_{ruz}(\vect{p}), h^{\vect{\pi}, \vect{\tau}}_{ruz}(\vect{q}) \in \{1, \dots, m\}$ the corresponding values computed by \rzhash. Let $\vect{\beta} = [\beta_1, \dots, \beta_m]$ be a vector in $\{1, \dots, b\}^m$, where each $\beta_i \sim \mathcal{U}_{\{1, \dots, b\}}$.
                Then,
                \begin{equation}
                    \label{eq:correlation}
                        P[\beta_{h^{\vect{\pi}, \vect{\tau}}_{ruz}(\vect{p})} = \beta_{h^{\vect{\pi}, \vect{\tau}}_{ruz}(\vect{q})}] = 1 + \frac{1 - b}{b}d_R(\vect{p}, \vect{q}).
                \end{equation}
            \end{theorem}
            \begin{proof}
                We split $P[\beta_{h^{\vect{\pi}, \vect{\tau}}_{ruz}(\vect{p})} = \beta_{h^{\vect{\pi}, \vect{\tau}}_{ruz}(\vect{q})}]$ in two terms
                \begin{equation}
                    \label{eq:split}
                    \text{\hspace{-2.5cm}}
                    \begin{split}
                        P[\beta_{h^{\vect{\pi}, \vect{\tau}}_{ruz}(\vect{p})} = \beta_{h^{\vect{\pi}, \vect{\tau}}_{ruz}(\vect{q})}] &= P[\beta_{h^{\vect{\pi}, \vect{\tau}}_{ruz}(\vect{p})} = \beta_{h^{\vect{\pi}, \vect{\tau}}_{ruz}(\vect{q})} | h^{\vect{\pi}, \vect{\tau}}_{ruz}(\vect{p}) = h^{\vect{\pi}, \vect{\tau}}_{ruz}(\vect{q})] \, P[h^{\vect{\pi}, \vect{\tau}}_{ruz}(\vect{p}) = h^{\pi, \tau}_{ruz}(\vect{q})] +\\
                                                                                                                                     &+ P[\beta_{h^{\vect{\pi}, \vect{\tau}}_{ruz}(\vect{p})} = \beta_{h^{\vect{\pi}, \vect{\tau}}_{ruz}(\vect{q})} | h^{\vect{\pi}, \vect{\tau}}_{ruz}(\vect{p}) \neq h^{\vect{\pi}, \vect{\tau}}_{ruz}(\vect{q})] \, P[h^{\vect{\pi}, \vect{\tau}}_{ruz}(\vect{p}) \neq h^{\vect{\pi}, \vect{\tau}}_{ruz}(\vect{q})].
                    \end{split}
                \end{equation}
                We know from~\cref{thm:ruzhash} that $P[h^{\vect{\pi}, \vect{\tau}}_{ruz}(\vect{p}) = h^{\vect{\pi}, \vect{\tau}}_{ruz}(\vect{q})] = 1 - d_R(\vect{p}, \vect{q})$, thus we rewrite~\eqref{eq:split} as
                \begin{equation}
                    \text{\hspace{-0.75cm}}
                    \begin{split}
                        P[\beta_{h^{\vect{\pi}, \vect{\tau}}_{ruz}(\vect{p})} = \beta_{h^{\vect{\pi}, \vect{\tau}}_{ruz}(\vect{q})}] &= P[\beta_{h^{\vect{\pi}, \vect{\tau}}_{ruz}(\vect{p})} = \beta_{h^{\vect{\pi}, \vect{\tau}}_{ruz}(\vect{q})} | h^{\vect{\pi}, \vect{\tau}}_{ruz}(\vect{p}) = h^{\vect{\pi}, \vect{\tau}}_{ruz}(\vect{q})] \, (1 - d_R(\vect{p}, \vect{q})) +\\
                                                                                                                                     &+ P[\beta_{h^{\vect{\pi}, \vect{\tau}}_{ruz}(\vect{p})} = \beta_{h^{\vect{\pi}, \vect{\tau}}_{ruz}(\vect{q})} | h^{\vect{\pi}, \vect{\tau}}_{ruz}(\vect{p}) \neq h^{\vect{\pi}, \vect{\tau}}_{ruz}(\vect{q})] \, d_R(\vect{p}, \vect{q}).
                    \end{split}
                \end{equation}
                The probability that $\beta_{h^{\vect{\pi}, \vect{\tau}}_{ruz}(\vect{p})}$ equals $\beta_{h^{\vect{\pi}, \vect{\tau}}_{ruz}(\vect{q})}$ is exactly $1$ when $h^{\vect{\pi}, \vect{\tau}}_{ruz}(\vect{p}) = h^{\vect{\pi}, \vect{\tau}}_{ruz}(\vect{q})$, and $\frac{1}{b}$ when $h^{\vect{\pi}, \vect{\tau}}_{ruz}(\vect{p}) \neq h^{\vect{\pi}, \vect{\tau}}_{ruz}(\vect{q})$. The last follows from the fact that each $\beta_i \sim \mathcal{U}_{\{1, \dots, b\}}$ is a realization of a uniform random variable in $\{1, \dots, b\}$.
                Therefore
                \begin{equation}
                    \begin{split}
                        P[\beta_{h^{\vect{\pi}, \vect{\tau}}_{ruz}(\vect{p})} = \beta_{h^{\vect{\pi}, \vect{\tau}}_{ruz}(\vect{q})}] &= 1 - d_R(\vect{p}, \vect{q}) + \frac{1}{b} \, d_R(\vect{p}, \vect{q}) =\\
                                                                                                       &= 1 + \frac{1 - b}{b}d_R(\vect{p}, \vect{q}),
                    \end{split}
                \end{equation}
                and this proves~\eqref{eq:correlation}.
            \end{proof}
            
            \cref{fig:ruzicka_variable} shows the linear correlation existing between $1 - d_R(\vect{p}, \vect{q})$ and $b$, when different $b$ values are considered, for $n = 1000$ randomly generated couples of points $\vect{p}, \vect{q} \in [0, 1]^m$, when $m = 10000$ and $k = 100$.
            
            \begin{corollary}[$d_J(\vect{p}, \vect{q})$ linearly correlates with $b$]
                We observe that substituting \rzhash $h^{\vect{\pi}, \vect{\tau}}_{ruz}(\vect{p}), h^{\vect{\pi}, \vect{\tau}}_{ruz}(\vect{q})$ with \mhash $h^{\vect{\pi}}_{min}(\vect{p}), h^{\vect{\pi}}_{min}(\vect{p})$ in~\cref{thm:correlation}, and given~\cref{lem:minhash}, the proof is equivalent. It follows that
                \begin{equation}
                        P[\beta_{h^{\vect{\pi}}_{min}(\vect{p})} = \beta_{h^{\vect{\pi}}_{min}(\vect{q})}] = 1 + \frac{1 - b}{b}d_J(\vect{p}, \vect{q}).
                \end{equation}
            \end{corollary}
            
            \begin{figure}[tb]
                \centering
                \begin{subfigure}[t]{.45\linewidth}
                    \centering
                    \includegraphics[width=\linewidth]{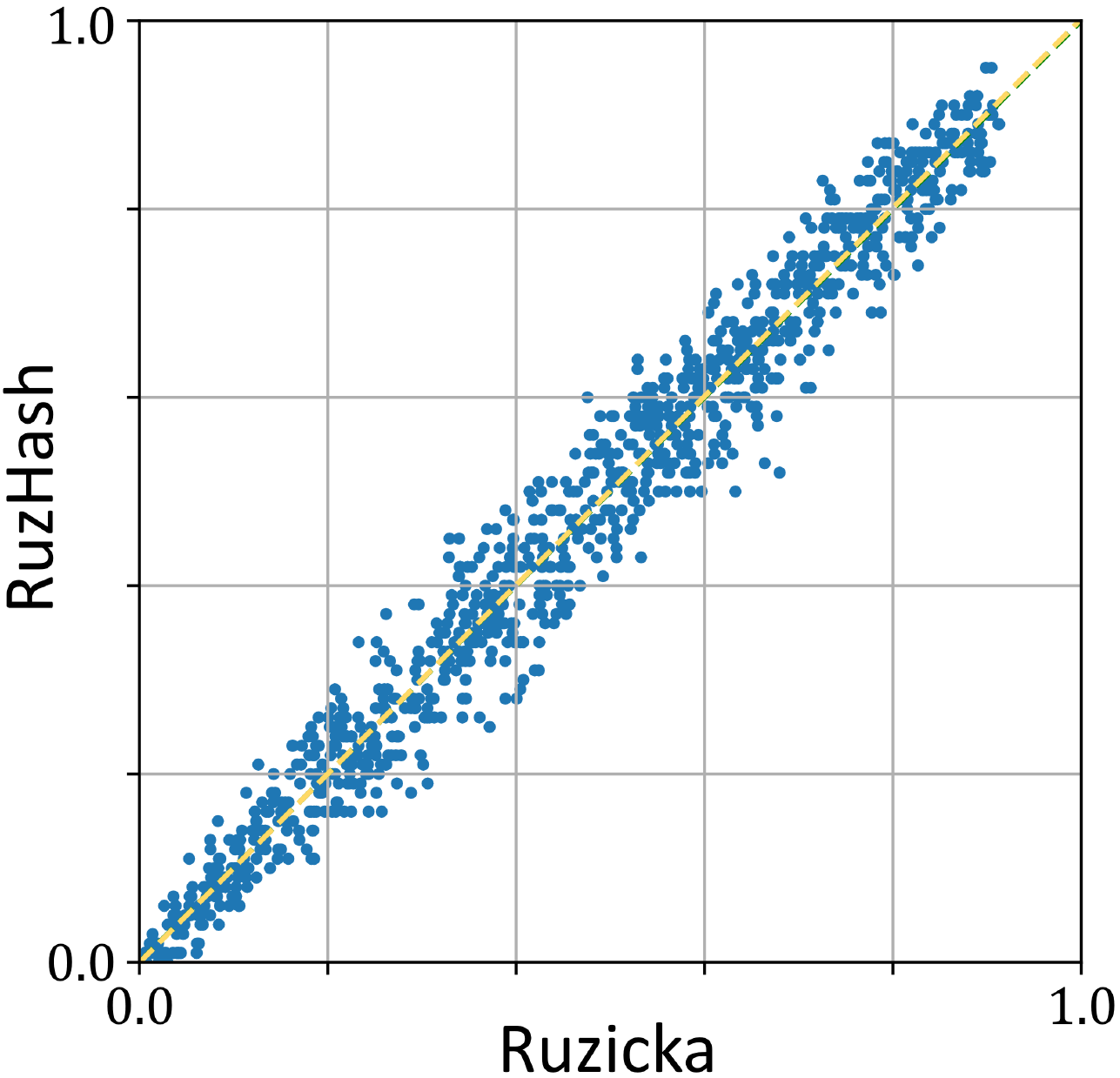}
                    \caption{}
                    \label{fig:ruzickacorr}
                \end{subfigure}
                \hfill
                \begin{subfigure}[t]{.45\linewidth}
                    \centering
                    \includegraphics[width=\linewidth]{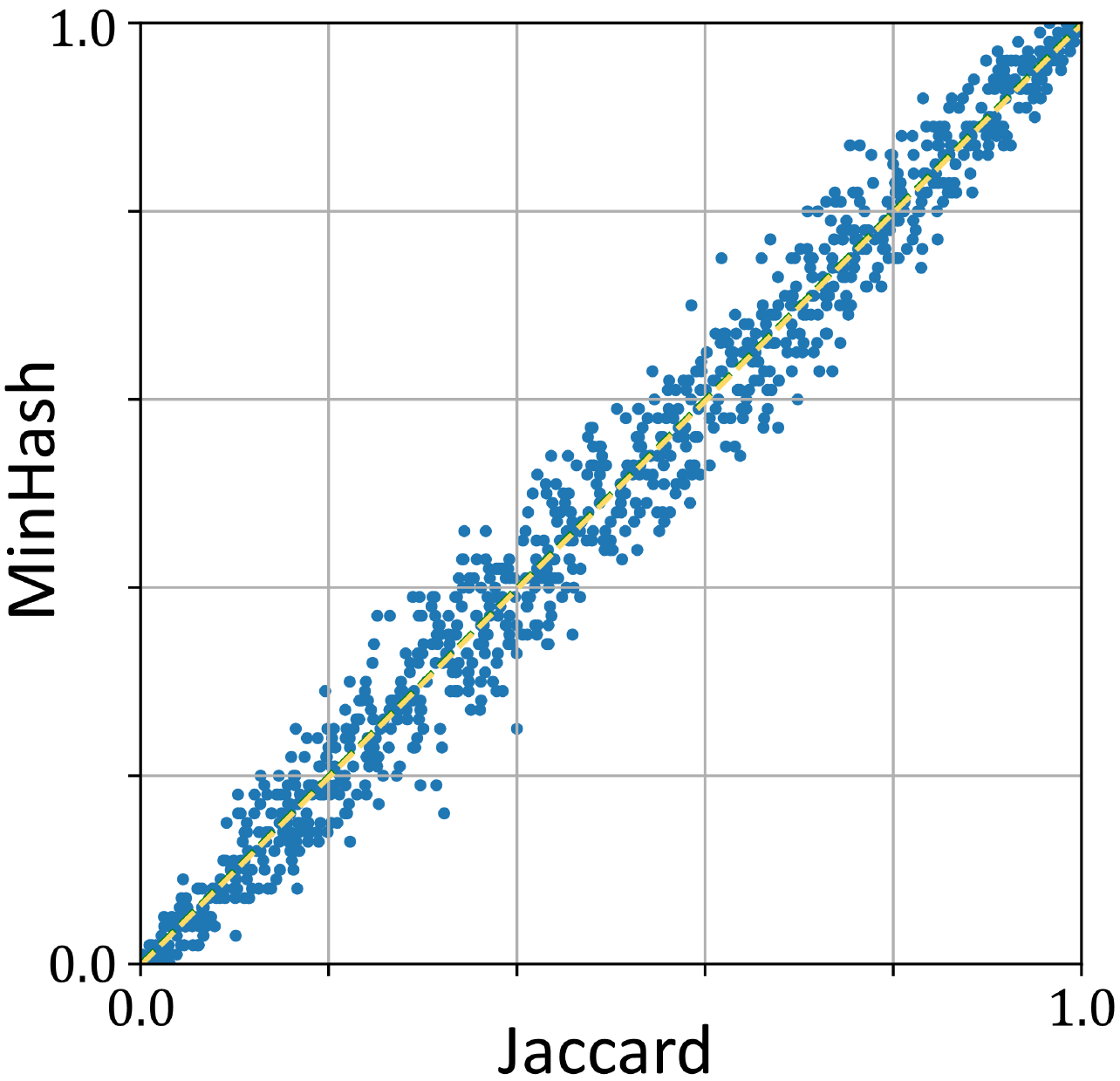}
                    \caption{}
                    \label{fig:minhashcorr}
                \end{subfigure}
                \caption{Correlation between (a) Ruzicka and \rzhash, proven in~\cref{thm:ruzhash}, and (b) Jaccard and \mhash, stated in~\cref{lem:minhash}.}
                \label{fig:correlation}
            \end{figure}
            
            \begin{figure}
                \centering
                \begin{subfigure}[t]{.45\linewidth}
                    \centering
                    \includegraphics[width=\linewidth]{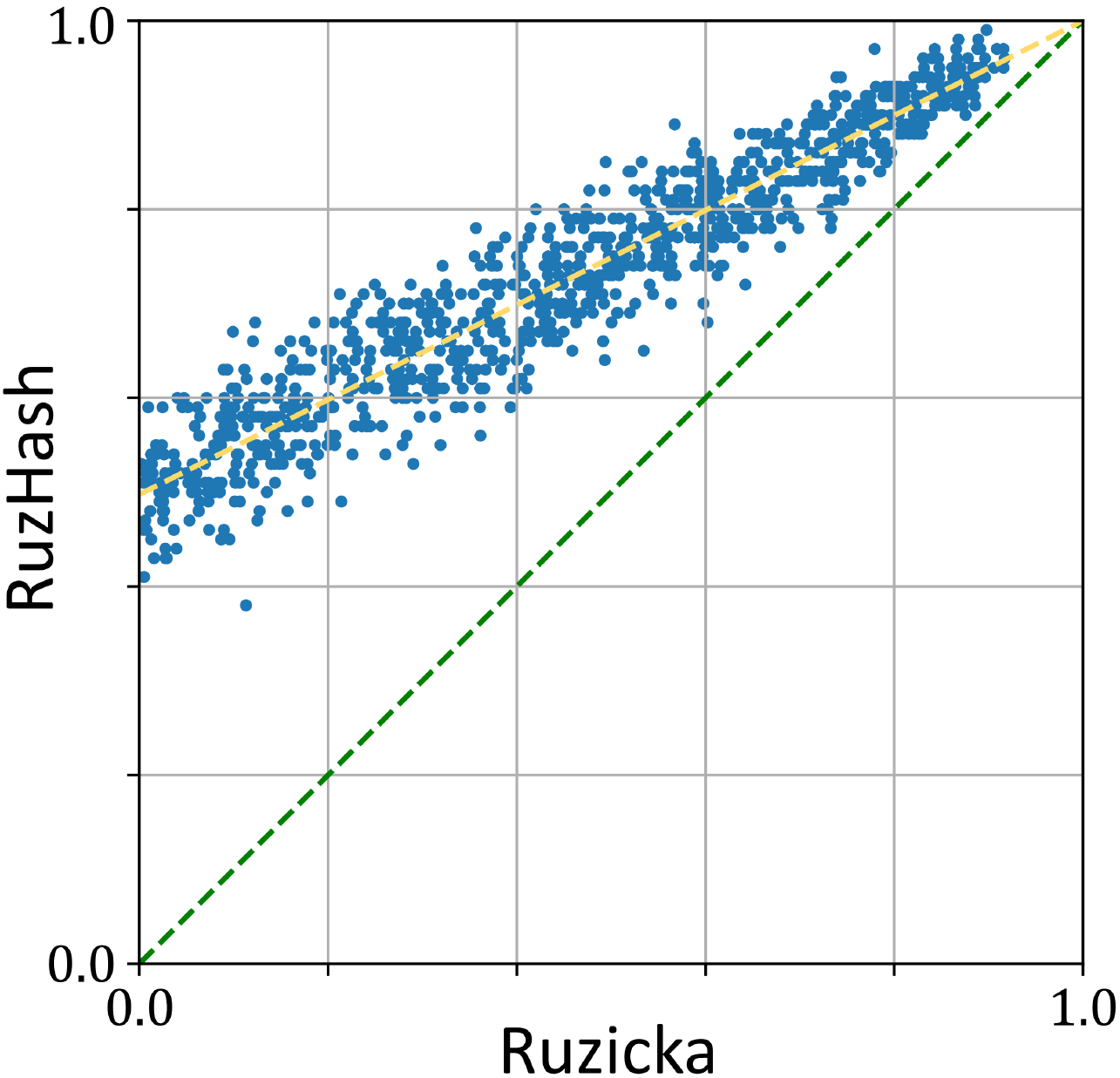}
                    \caption{$b = 2$}
                \end{subfigure}
                \hfill
                \begin{subfigure}[t]{.45\linewidth}
                    \centering
                    \includegraphics[width=\linewidth]{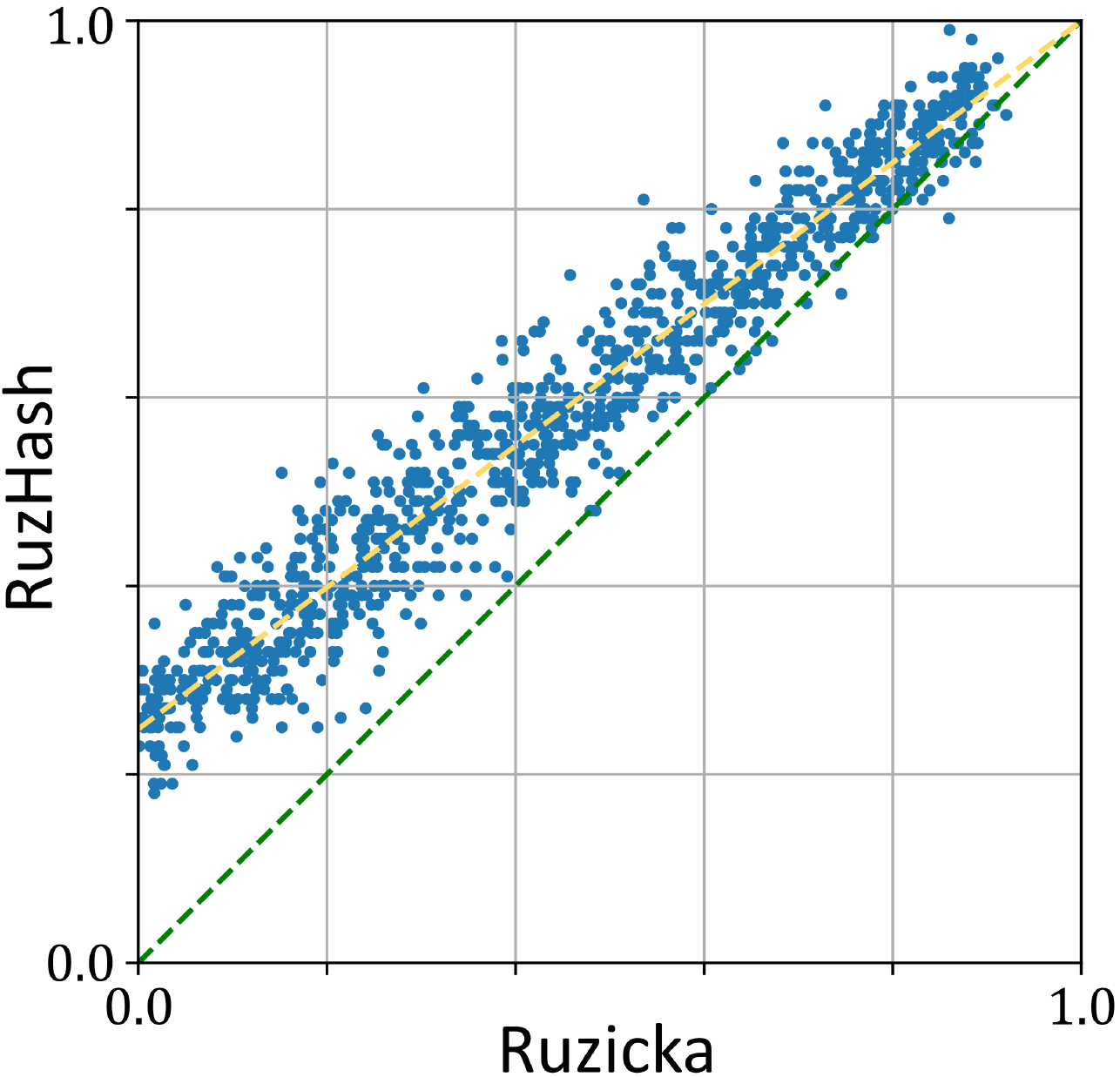}
                    \caption{$b = 4$}
                \end{subfigure}
                \\
                \begin{subfigure}[t]{.45\linewidth}
                    \centering
                    \includegraphics[width=\linewidth]{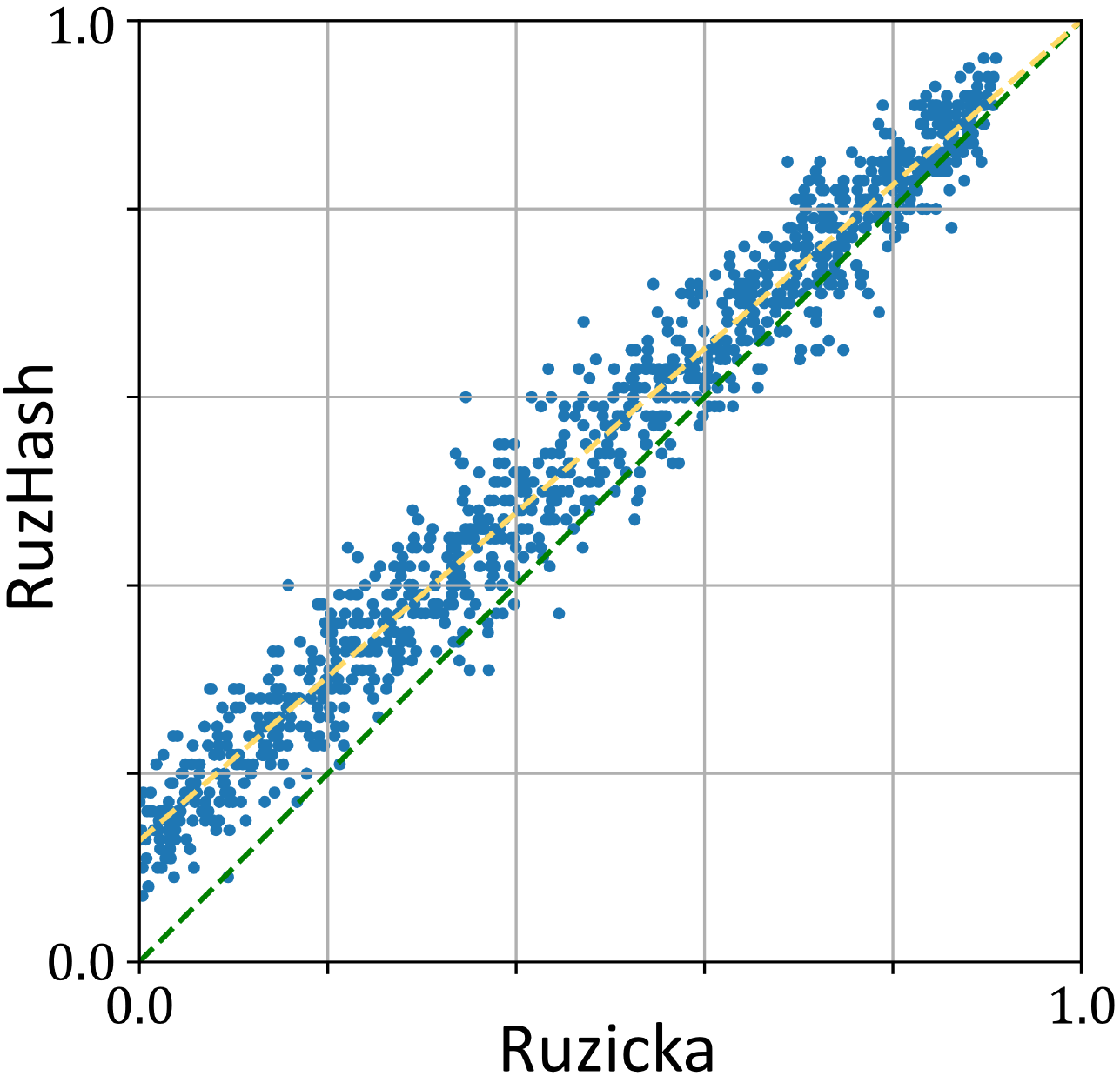}
                    \caption{$b = 8$}
                \end{subfigure}
                \caption{Linear correlation existing between $1 - d_R(\vect{p}, \vect{q})$ and the variable branching factor version of \rzhash introduced in~\cref{subsubsec:ruzhash_itree}.}
                \label{fig:ruzicka_variable}
            \end{figure}

    \cleardoublepage
    \pagestyle{empty}
    \mbox{}
    \cleardoublepage
    \pagestyle{fancy}
    \small
    \addcontentsline{toc}{chapter}{\bibname}
    \bibliographystyle{ieeetr}
    \bibliography{references}
\end{document}